\pdfoutput=1
\documentclass[11pt]{article}
\usepackage[final]{acl}

\usepackage{times}
\usepackage{latexsym}

\usepackage[T1]{fontenc}
\usepackage[utf8]{inputenc}

\usepackage{microtype}

\usepackage{inconsolata}

\newcommand{\ctt}[2]{\colorbox{#1}{{\texttt{#2}}}}

\title{Evaluating Large Language Models on Time Series Feature Understanding: A Comprehensive Taxonomy and Benchmark}

\usepackage{listings}
\usepackage{microtype}
\usepackage{graphicx}
\usepackage{booktabs} % for professional tables
\usepackage[most]{tcolorbox}
\usepackage{longtable}

\usepackage{amsmath}
\usepackage{amssymb}
\usepackage{multirow}
\usepackage{mathtools}
\usepackage{amsthm}
\usepackage{array}
\usepackage{caption}
\usepackage{subcaption}

\usepackage{hyperref}

\usepackage{enumitem}
\setlist[itemize]{align=parleft,left=5pt..10pt}

\author{Elizabeth Fons \quad Rachneet Kaur \quad Soham Palande \quad Zhen Zeng \\ 
        \textbf{Tucker Balch} \quad \textbf{Manuela Veloso} \quad \textbf{Svitlana Vyetrenko}\\
  \texttt{\{first\_name\}.\{last\_name\}@jpmchase.com} \\
        JP. Morgan AI Research}

\definecolor{efc}{rgb}{0.87, 0.19, 0.3}
\definecolor{sp}{rgb}{0.1, 0.5, 0.1}

\definecolor{headercolor}{gray}{0.85}
\definecolor{boxcolor}{gray}{0.95}

\definecolor{lightgray}{gray}{0.95}
\definecolor{darkgray}{gray}{0.85}
\definecolor{highlight}{gray}{0.9}
\definecolor{questioncolor}{gray}{0.9} % Light gray for question
\definecolor{answercolor}{gray}{0.95}   % Slightly darker gray for answers

\colorlet{Mycolor1}{red!30}
\usepackage{etoolbox}
\AtBeginEnvironment{tcolorbox}{\small}

\begin{document}
\maketitle

\begin{abstract}

Large Language Models (LLMs) offer the potential for automatic time series analysis and reporting, which is a critical task across many domains, spanning healthcare, finance, climate, energy, and many more. In this paper, we propose a framework for rigorously evaluating the capabilities of LLMs on time series understanding, encompassing both univariate and multivariate forms. We introduce a comprehensive taxonomy of time series features, a critical framework that delineates various characteristics inherent in time series data. Leveraging this taxonomy, we have systematically designed and synthesized a diverse dataset of time series, embodying the different outlined features, each accompanied by textual descriptions. This dataset acts as a solid foundation for assessing the proficiency of LLMs in comprehending time series. 
Our experiments shed light on the strengths and limitations of state-of-the-art LLMs in time series understanding, revealing which features these models readily comprehend effectively and where they falter. In addition, we uncover the sensitivity of LLMs to factors including the formatting of the data, the position of points queried within a series and the overall time series length.
\end{abstract}

\section{Introduction}
\label{sec:intro}
Time series analysis and reporting are crucial in diverse fields like healthcare, finance, and climate \cite{biosignal}. The recent progress in Large Language Models (LLMs) opens exciting possibilities for automating these processes. While recent studies have explored adapting LLMs for specific time series tasks, such as seizure localization in EEG time series~\cite{chen2024eegformer}, cardiovascular disease diagnosis in ECG time series~\cite{qiu2023transfer}, weather and climate data understanding~\cite{chen2023foundation}, and explainable financial time series forecasting~\cite{Yu2023TemporalDM}, a systematic evaluation of general-purpose LLMs' inherent capabilities in understanding time series data is lacking. 
One notable example of domain-specific application is the BioSignal Copilot framework presented by \cite{biosignal}, which focuses on leveraging LLMs for clinical report generation from biomedical signals. 

This paper aims to fill this gap by uncovering the strengths and weaknesses of general-purpose LLMs in time series understanding, without any domain-specific fine-tuning. Our focus is on assessing their potential for a key downstream task: time series annotation and summarization. By understanding the baseline capabilities of LLMs, practitioners can identify areas where these models are readily applicable and areas where targeted fine-tuning efforts may be necessary to improve performance.

To systematically evaluate the performance of general-purpose LLMs on generic time series understanding, we propose a taxonomy of time series features for both univariate and multivariate time series.
This taxonomy serves as a structured framework for evaluating LLM performance and provides a foundation for future research in this domain. 
Based on this taxonomy, we have created a diverse synthetic dataset of time series that covers a wide range of features, each accompanied by qualitative and quantitative textual descriptions. 

Our evaluations focus on tasks directly relevant to time series annotation and summarization, such as feature detection, classification, and data retrieval as well as arithmetic reasoning. Additionally, we assess the LLMs' ability to match textual descriptions to their corresponding time series, leveraging the textual descriptions in our dataset. These findings will be instrumental for developing LLM-powered tools for automated time series annotation and summarization, ultimately enhancing data analysis and reporting workflows across diverse domains.
\clearpage
Our contributions are three-fold:
\begin{itemize}
    \item \textbf{Taxonomy} - we introduce a comprehensive taxonomy that provides a systematic categorization of important time series features, an essential tool for standardizing the evaluation of LLMs in time series understanding.
    \item \textbf{Diverse Time Series Dataset} - we synthesize a diverse time series dataset with train/validation/test splits, ensuring a broad representation of various time series types, encompassing the spectrum of features identified in our taxonomy, each with accompanying textual descriptions.
    \item \textbf{Evaluations of LLMs} - our evaluations provide insights into LLMs' strengths and weaknesses in understanding time series. We analyze how LLMs handle data format, query location, and time series length, providing a nuanced understanding of their capabilities in this domain.
\end{itemize}

\section{Related Work}
\label{sec:relwork}

\paragraph{Large Language Models} Large Language Models (LLMs), such as Llama2 \cite{touvron2023llama}, PaLM \cite{chowdhery2023palm}, GPT-3 \cite{brown2020language}, GPT4 \cite{achiam2023gpt}, and Vicuna-13B \cite{vicuna2023}, have demonstrated remarkable capabilities in various language-related tasks and have recently been explored for their potential in time series analysis.

\paragraph{Language Models for Time Series} 
Recent progress in time series forecasting has capitalized on the versatile and comprehensive abilities of LLMs, merging their language expertise with time series data analysis. This collaboration marks a significant methodological change, underscoring the capacity of LLMs to revolutionize conventional predictive methods with their advanced information processing skills. Notably, \cite{gruver2023l} have set benchmarks for pre-trained LLMs such as GPT-3 and Llama2 by assessing their capabilities for zero-shot forecasting. Similarly, \cite{xue2023promptcast} introduced Prompcast, adopting a novel approach by treating forecasting as a question-answering activity, utilizing strategic prompts. Further, \cite{Yu2023TemporalDM} delved into the potential of LLMs for generating explainable forecasts in financial time series, tackling inherent issues like cross-sequence reasoning, integration of multi-modal data, and interpretation of results, which pose challenges in conventional methodologies. Additionally, \cite{zhou2023one} demonstrated that leveraging frozen pre-trained language models, initially trained on vast corpora, for time series analysis could achieve comparable or even state-of-the-art performance across various principal tasks in time series analysis including imputation, classification and forecasting.

Recent advancements in the application of LLMs to biomedical time series data have also shown promise in the automated generation of clinical reports. \cite{biosignal} introduce BioSignal Copilot, a system that leverages LLMs for drafting reports from biomedical signals, such as electrocardiograms (ECGs) and electroencephalograms (EEGs). Their work highlights the importance of domain-specific feature extraction in facilitating LLM understanding of time series data, aligning with our work on developing a comprehensive taxonomy of time series features to enhance LLM interpretability and analysis in various applications. Notably, their focus on automatic report generation from the processed signals serves as a specific downstream task, further emphasizing the need for a systematic evaluation of LLMs' ability to understand and extract relevant features from time series data, such as the one presented in this work.

\paragraph{LLMs for arithmetic tasks} Despite their advanced capabilities, LLMs face challenges with basic arithmetic tasks, crucial for time series analysis involving quantitative data \citep{azerbayev2023llemma, liu2023goat}. Research has identified challenges such as inconsistent tokenization and token frequency as major barriers \citep{nogueira2021investigating, Kim2021HaveYS}. Innovative solutions, such as Llama2's approach to digit tokenization \citet{yuan2023large}, highlight ongoing efforts to refine LLMs' arithmetic abilities, enhancing their applicability in time series analysis.
\section{Time Series Data}

\colorlet{pale1b}{blue!20}
\colorlet{pale1}{blue!10}
\colorlet{pale2}{green!10}
\colorlet{pale2b}{green!20}
\colorlet{pale3}{red!10}
\colorlet{pale3b}{red!20}
\colorlet{pale4}{orange!10}
\colorlet{pale4b}{orange!20}
\colorlet{pale5}{cyan!10}
\colorlet{pale5b}{cyan!20}
\colorlet{pale6}{magenta!10}
\colorlet{pale6b}{magenta!20}
\colorlet{pale7}{gray!10}
\colorlet{pale7b}{gray!20}
\colorlet{pale8}{teal!10}
\colorlet{pale8b}{teal!20}
\colorlet{pale9}{purple!10}
\colorlet{pale9b}{purple!20}
\colorlet{pale10}{olive!10}
\colorlet{pale10b}{olive!20}

\begin{table*}[h!]
    \centering
    \caption{Taxonomy of time series characteristics.}
    % \begin{footnotsize}
    \resizebox{\linewidth}{!}{
    \small
    % {\sffamily
    \begin{tabular}{p{0.23\linewidth} p{0.38\linewidth} >{\raggedright\arraybackslash}p{0.38\linewidth}}
    \toprule
        Main Category & Description & Sub-category\\
    \midrule
    \textit{Univariate} & \\
    \midrule
        \ctt{pale1b}{Trend} & Directional movements over time. & \ctt{pale1}{Up}, \ctt{pale1}{Down}\\
        \ctt{pale2b}{Seasonality and} \ctt{pale2b}{Cyclical Patterns} & Patterns that repeat over a fixed or irregular period. & \ctt{pale2}{Fixed-period},
        \ctt{pale2}{Shifting period}, \ctt{pale2}{Multiple seasonality} \\
        \ctt{pale4b}{Anomalies} & Significant deviations from typical patterns. & \ctt{pale4}{Spikes}, \ctt{pale4}{level shifts}, \newline \ctt{pale4}{temporal disruptions} \\
        \ctt{pale3b}{Volatility} & Degree of dispersion of a series over time. & \ctt{pale3}{Constant}, \ctt{pale3}{Trending}, \ctt{pale3}{Clustered},\ctt{pale3}{Dynamic} \\
        \ctt{pale5b}{Structural Breaks} & Fundamental shifts in the series data, such as regime changes or parameter shifts. & \ctt{pale5}{Regime changes}, \ctt{pale5}{parameter shifts} \\
        \ctt{pale10b}{Stationarity Properties} & Stationarity versus non-stationarity. & \ctt{pale10}{Stationarity} \\
        \ctt{pale6b}{Distribution Properties} & Characteristics like fat tails & \ctt{pale6}{Fat tails} \\
    \midrule
    \textit{Multivariate} & \\
    \midrule
        \ctt{pale7b}{Correlation} & Measure the linear relationship between series. Useful for predicting one series from another if they are correlated. &  \ctt{pale7}{Positive} \ctt{pale7}{Negative}  \\
        
        \ctt{pale8b}{Cross-Correlation} & Measures the relationship between two series at different time lags, useful for identifying lead or lag relationships. & \ctt{pale8}{Positive - direct}, \ctt{pale8}{Positive - lagged}, \newline \ctt{pale8}{Negative - direct}, \ctt{pale8}{Negative - lagged}  \\
        \ctt{pale9b}{Dynamic Conditional} \ctt{pale9b}{Correlation} & Assesses situations where correlations between series change over time. & \ctt{pale9}{Correlated first half}  \ctt{pale9}{Correlated second half} \\
    \bottomrule
    \end{tabular}
    }
    \label{tab:characteristics}
\end{table*}

\subsection{Taxonomy of Time Series Features}
\label{sec:taxonomy}

Our study introduces a comprehensive taxonomy for evaluating the analytical capabilities of Large Language Models (LLMs) in the context of time series data. This taxonomy categorizes the intrinsic characteristics of time series, providing a structured basis for assessing the proficiency of LLMs in identifying and extracting these features. The proposed taxonomy encompasses critical aspects of time series data that are frequently analyzed for different applications and are commonly used in qualitative descriptions of time series data. These features are considered the most relevant for evaluating the ability of LLMs to generate and understand textual reports of time series data. 

The features are organized in increasing order of complexity, starting with trend, seasonality, volatility, anomalies, structural breaks, and distribution properties. Each main feature is further divided into sub-categories to provide a more nuanced evaluation of LLM capabilities. This hierarchical organization allows for a detailed assessment of LLM performance on both simple and complex time series characteristics. Table~\ref{tab:characteristics} presents the selected features in order of increasing complexity and their sub-features. While we have strived to define the features as distinctly as possible, it is important to note that some overlap may exist between certain categories.

\paragraph{Justification for the proposed taxonomy}
Our selection of features is based on extensive literature review and expert consultations. Trends and seasonality are fundamental components widely recognized in time series analysis across various domains, such as finance and climate science \citep{tsbookmonash, shumstof2000}. Volatility and anomalies are crucial for understanding dynamic behaviors and identifying significant deviations in data \citep{tsay2005, chandola2009anomaly}. Structural breaks and distribution properties are essential for capturing shifts in underlying data generation processes and understanding the statistical nature of the data \citep{perron2005, Cont2001}.
Table~\ref{tab:taxonomy-definitions} provides definitions of each sub-category along with domain examples where these features could be referenced.

\subsection{Synthetic Time Series Dataset}
\label{synthetic_data}

\begin{figure}[htb]
\centering
\includegraphics[width=0.49\linewidth]{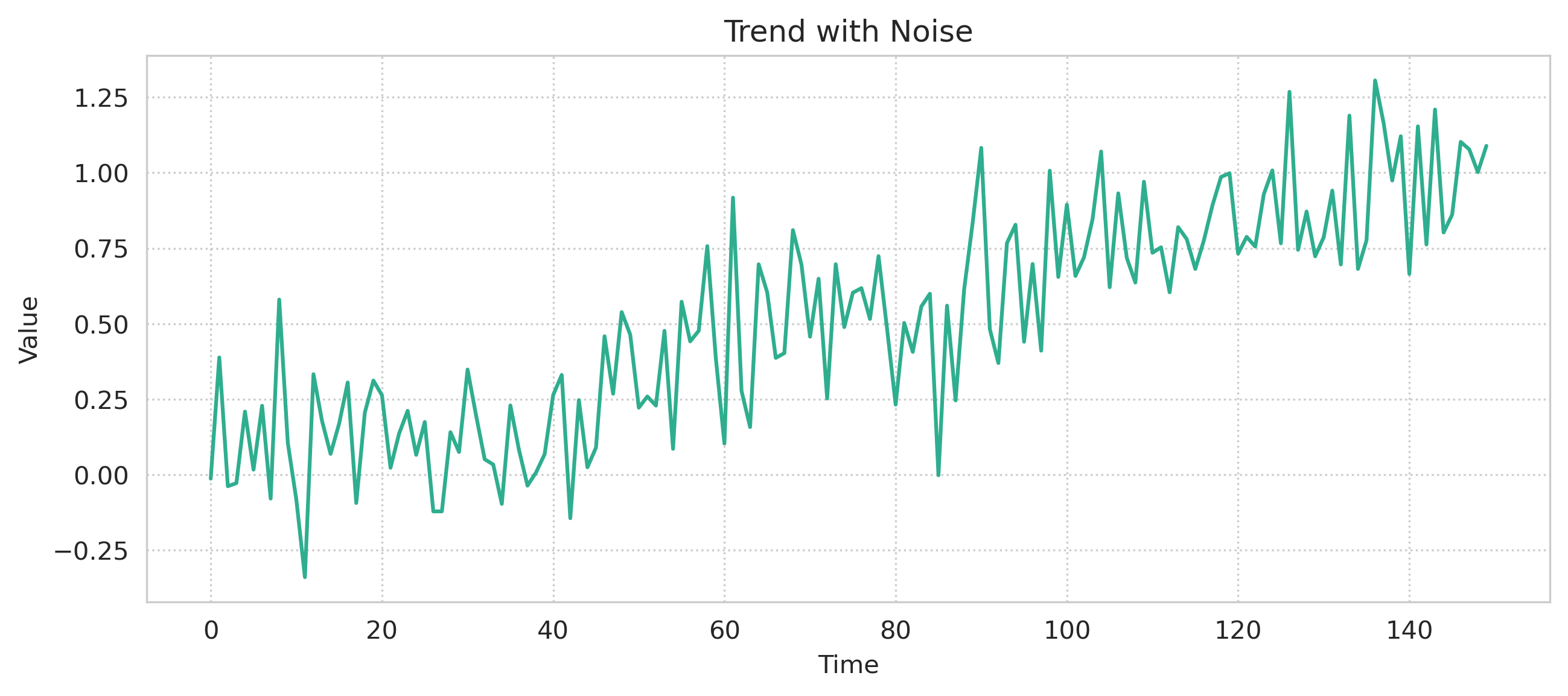}~\includegraphics[width=0.49\linewidth]{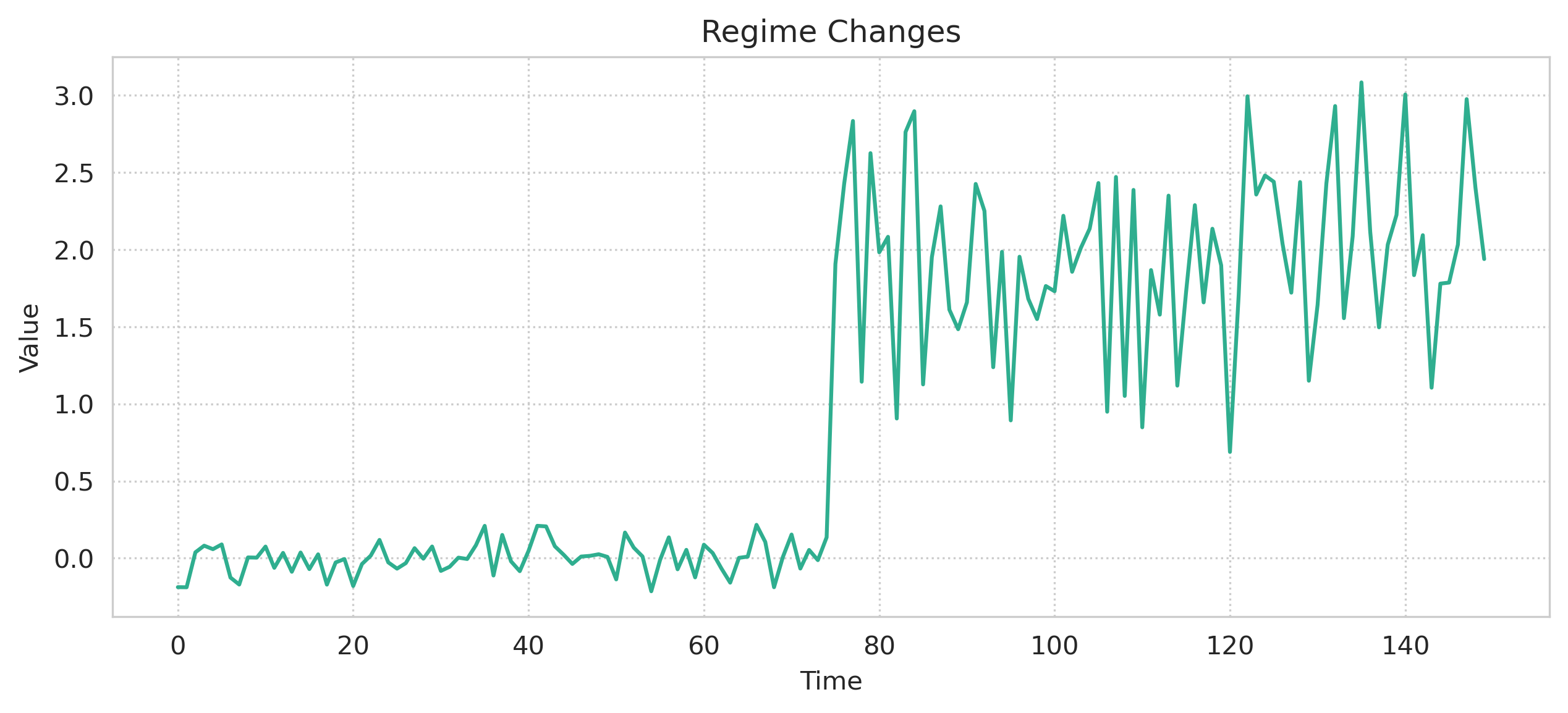}
\\
\includegraphics[width=0.49\linewidth]{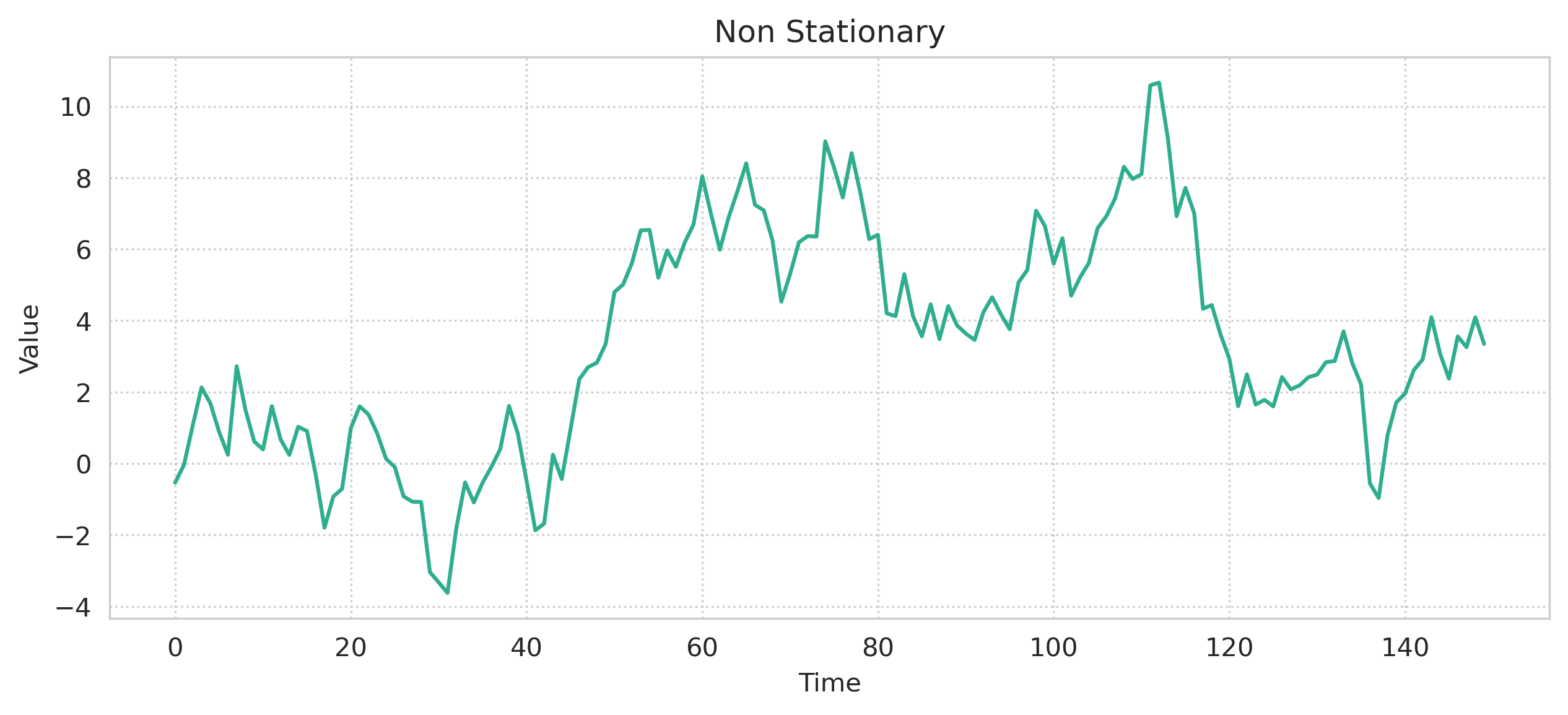}~\includegraphics[width=0.49\linewidth]{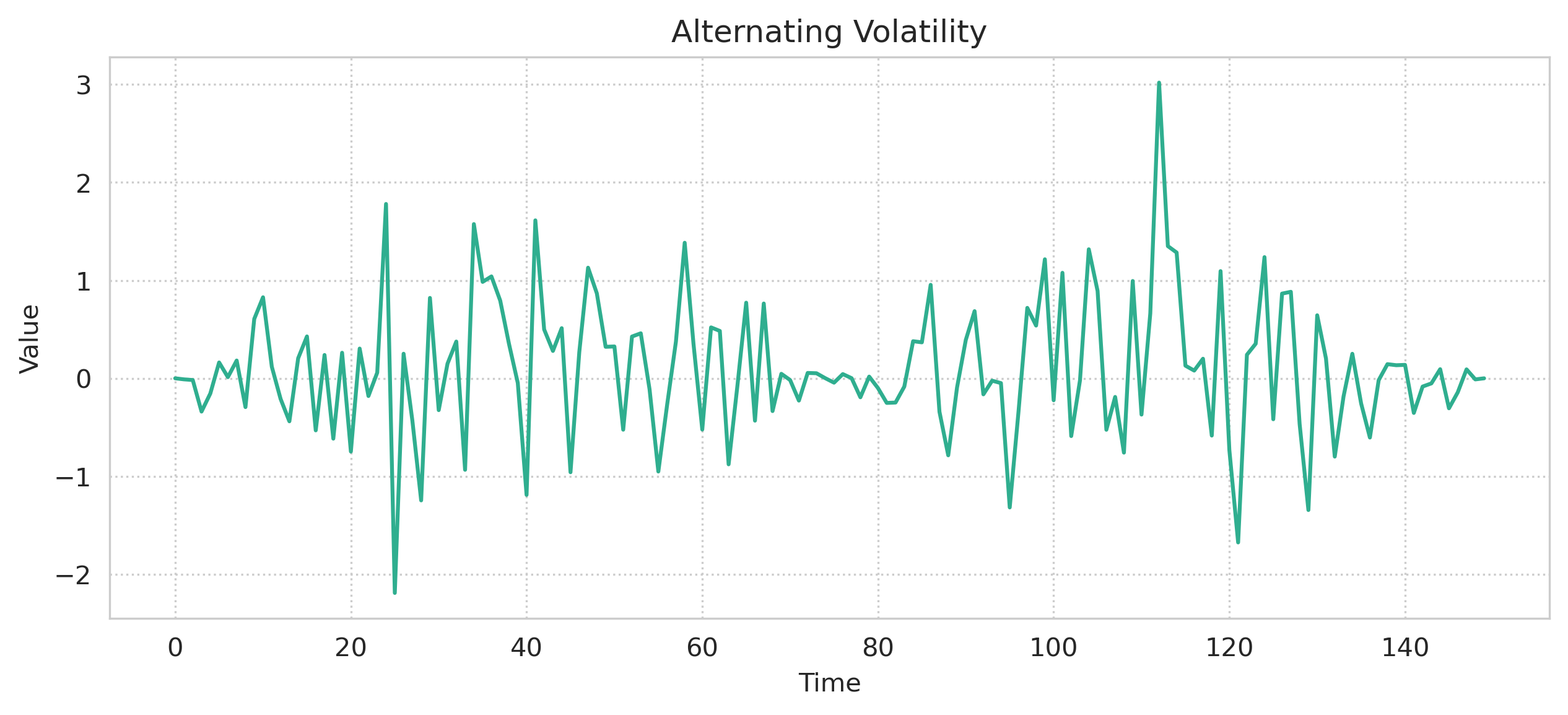}
\caption{Example synthetically generated time series.}
\label{fig:time_series_examples}
\end{figure}

Leveraging our taxonomy, we construct a diverse synthetic dataset of time series, covering the features outlined in the previous section. We generated in total 10 datasets, each with a training split (5000 samples), validation split (2000 samples), and test split (200 samples) to facilitate model development and evaluation. Within each dataset, the time series length is randomly chosen between 30 and 150 to encompass a variety of both short and long time series data. In order to make the time series more realistic, we add a time index, using predominantly daily frequency. Each time series in the dataset is accompanied by a qualitative description, a textual summary of the main features present in the time series (e.g., "This time series exhibits a downward quadratic trend, commencing with higher figures and falling gradually."), and a quantitative description, which includes the minimum and maximum values, the date range, and a textual description of the specific features present (e.g., "This daily time series covers the period from 2024-01-01 to 2024-05-04. It exhibits multiple seasonal patterns with monthly seasonality, with 5 peaks and 4 troughs, and an average amplitude of 24.25."). Fig.~\ref{fig:time_series_examples} showcases examples of our generated univariate time series.
Each univariate dataset showcases a unique single-dimensional pattern, whereas multivariate data explore series interrelations to reveal underlying patterns. See Table~\ref{fig:univariate_data} and Table~\ref{fig:multivariate_data} in the appendix for visual examples of each dataset.
For a detailed description of the generation of each dataset, refer to Appendix. \ref{sec:data}.

\section{Time Series Benchmark Tasks}
\label{sec:tasks}
Our evaluation framework is designed to assess the LLMs' capabilities in analyzing time series across the dimensions in our taxonomy (Sec.~\ref{sec:taxonomy}).
The evaluation includes four primary tasks:

\paragraph{Feature Detection}
This task evaluates the LLMs' ability to identify the presence of specific features within a time series, such as trend, seasonality, or anomalies. For instance, given a time series dataset with an upward trend, the LLM is queried to determine if a trend exists. 
Queries are structured as yes/no questions to assess the LLMs' ability to recognize the presence of specific time series features, such as "Is a trend present in the time series?"

\paragraph{Feature Classification}
Once a feature is detected, this task assesses the LLMs' ability to classify the feature accurately. For example, if a trend is present, the LLM must determine whether it is upward, downward, or non-linear.
This task involves a QA setup where LLMs are provided with definitions of sub-features within the prompt. Performance is evaluated based on the correct identification of sub-features, using the F1 score to balance precision and recall. This task evaluates the models' depth of understanding and ability to distinguish between similar but distinct phenomena.

\paragraph{Information Retrieval} Evaluates the LLMs' accuracy in retrieving specific data points, such as values on a given date.

\paragraph{Arithmetic Reasoning} Focuses on quantitative analysis tasks, such as identifying minimum or maximum values. Accuracy and Mean Absolute Percentage Error (MAPE) are used to measure performance, with MAPE offering a precise evaluation of the LLMs' numerical accuracy.

Additionally, to account for nuanced aspects of time series analysis, we propose in Sec.~\ref{sec:performance_factors} to study the influence of multiple factors, including time series formatting, location of query data point in the time series and time series length.
\paragraph{Time Series Description}
To evaluate the ability of LLMs to match time series to their corresponding descriptions, even in the presence of distractors, we introduce two new tasks: \textit{(1) Text Matching (inter-dataset):} the LLM is presented with a time series and four different descriptions from the same dataset, one of which is the correct description for the given time series. The descriptions include both qualitative commentaries and quantitative information about the time series. The LLM is asked to select the description that is closest to the time series. This task assesses the LLM’s ability to match a time series to its corresponding description, even in the case where the qualitative description is similar;
\textit{(2) Text Matching (cross-dataset):} the LLM is presented with a time series and four different qualitative descriptions, each from a different dataset. This task assesses the LLM’s ability to match a time series to its corresponding description based only on qualitative features, without relying on any quantitative information.

\begin{table*}[h!]
        % \centering
        \caption{Performances across all reasoning tasks (Bold indicates best performance).}
        \label{tab:main_res}
        \begin{center}
        \begin{small}
        \begin{sc}
        \resizebox{\textwidth}{!}{
        \begin{tabular}{llcccccccccc}
        \toprule
         & Metric & \multicolumn{2}{c}{GPT4} & \multicolumn{2}{c}{GPT3.5} & \multicolumn{2}{c}{Llama2} & \multicolumn{2}{c}{Vicuna} & \multicolumn{2}{c}{Phi3} \\
         &        & Zero-shot & CoT & Zero-shot & CoT & Zero-shot & CoT & Zero-shot & CoT & Zero-shot & CoT \\
        \midrule
        \multicolumn{3}{l}{\bf Univariate time series characteristics} & & & & & & & & & \\
        \midrule
        \multicolumn{3}{l}{\it Feature detection} & & & & & & & & & \\
        \midrule
        Trend & F1score & 0.79 & \bf 0.89 & 0.45 & 0.66 & 0.51 & 0.56 & 0.58 & 0.58 & 0.72 & 0.78  \\
        Seasonality & F1score & 0.94 & \bf 0.98 & 0.43 & 0.55 & 0.64 & 0.35 & 0.49 & 0.48 & 0.82 & 0.83\\
        Anomalies & F1score & \bf 0.84 & 0.81 & 0.57 & 0.47 & 0.47 & 0.51 & 0.49 & 0.52 & 0.43 & 0.71\\
        Volatility & F1score & 0.68 & \bf 0.73 & 0.43 & 0.43 & 0.42 & 0.53 & 0.45 & 0.48 & \bf 0.73 & 0.69 \\
        Struct. break & F1score & 0.59 & 0.61 & 0.57 & 0.48 & 0.39 & 0.44 & 0.48 & 0.52 & 0.44 & \bf 0.67\\
        Stationarity & F1score & 0.33 & \bf 0.59 & 0.33 & 0.40 & 0.33 & 0.39 & 0.44 & 0.42 & 0.33 & 0.46\\
        Fat Tails & F1score & 0.39 & -- & 0.44 & 0.36 & 0.34 & 0.39 & 0.44 & \bf 0.48 & 0.47 & 0.45\\
        \midrule
        \multicolumn{3}{l}{\it Feature classification} & & & & & & & & & \\
        \midrule
        Trend & F1score & \bf 0.98 & \bf 0.98 & 0.78 & 0.95 & 0.43 & 0.70 & 0.53 & 0.48 & 0.48 & 0.95 \\
        Seasonality & F1score & 0.17 & 0.21 & 0.17 & 0.16 & 0.31 & 0.27 & 0.23 & 0.18 & \bf 0.48 & 0.24 \\
        Anomalies & F1score & 0.87 & \bf 0.95 & 0.20 & 0.40 & 0.30 & 0.37 & 0.37 & 0.44 & 0.53 & 0.48\\
        Volatility & F1score & 0.18 & \bf 0.25 & 0.07 & 0.16 & 0.12 & 0.10 & 0.15 & 0.17 & 0.08 & 0.15 \\
        Struct. break & F1score & 0.42 & 0.41 & 0.56 & \bf 0.57 & 0.30 & 0.43 & 0.41 & 0.35 & 0.51 & 0.47 \\
        % stat-properties-stationary & NaN & 0.31 & 0.31 & 0.23 & 0.11 & NaN & 0.18 & 0.24 & 0.11 \\
        \midrule
        \multicolumn{3}{l}{\bf Multivariate time series characteristics} & & & & & & & & & \\
        \midrule
        Fixed Corr. & F1score & \bf 0.48 & -- & 0.39 & 0.43 & 0.38 & 0.43 & 0.40 & 0.46 & 0.43 & 0.57 \\
        Lagged Corr. & F1score & \bf 0.54 & -- & 0.52 & 0.46 & 0.45 & 0.42 & 0.42 & 0.45 & 0.41 & 0.40 \\
        Changing Corr. & F1score & 0.48 & -- & 0.43 & 0.44 & \bf 0.52 & 0.43 & 0.50 & 0.45 & 0.48  & 0.65\\ 
        \midrule
        \multicolumn{3}{l}{\bf Information Retrieval} & & & & & & & & & \\
        \midrule
        Value on Date & Acc & \bf 1.00 & \bf 1.00 & 0.99 & 0.99 & 0.54 & 0.49 & 0.61 & 0.62 & 0.93 & 0.89 \\
        Value on Date & MAPE & \bf 0.00 & \bf 0.00 & 0.03 & 0.03 & 1.06 & 0.73 & 0.75 & 0.76 & 0.19 & 0.17 \\
        \midrule
        \multicolumn{3}{l}{\bf Arithmetic Reasoning} & & & & & & & & & \\
        \midrule
        Min Value & Acc & \bf 1.00 & 0.99 & 0.99 & 0.98 & 0.63 & 0.55 & 0.63 & 0.72 & 0.94 & 0.91  \\
        & MAPE & \bf 0.00 & \bf 0.00 & 0.01 & 0.01 & 3.89 & 7.42 & 3.96 & 4.70 & 0.10 & 0.41 \\
        Min Date  & Acc & \bf  0.98 & 0.94 & 0.93 & 0.93 & 0.40 & 0.32 & 0.42 & 0.49 & 0.85 & 0.82 \\
        Max Value & Acc & \bf 1.00 & \bf 1.00 & 0.96 & 0.94 & 0.53 & 0.54 & 0.47 & 0.57 & 0.87 & 0.78 \\
        & MAPE & \bf 0.00 & \bf 0.00 & 3.66 & 3.96 & 3.23 & 1.09 & 3.12 & 2.27 & 0.11 & 0.26 \\
        Max Date  & Acc  & \bf 0.99 & 0.93 & 0.91 & 0.90 & 0.32 & 0.34 & 0.29 & 0.37 & 0.77 & 0.70 \\
        \bottomrule
        \end{tabular}
        }
        \end{sc}
        \end{small}
        \end{center}                
\end{table*}

\section{Performance Metrics and Factors}
\subsection{Performance Metrics}
We employ the following metrics to report the performance of LLMs on various tasks.
\paragraph{F1 Score} Applied to feature detection and classification, reflecting the balance between precision and recall.
\paragraph{Accuracy} Used for assessing the information retrieval and arithmetic reasoning tasks.
\paragraph{Mean Absolute Percentage Error (MAPE)} Employed for numerical responses in the information retrieval and arithmetic reasoning tasks, providing a measure of precision in quantitative analysis.

\subsection{Performance Factors}
\label{sec:performance_factors}
We identified various factors that could affect the performance of LLMs on time series understanding, for each we designed deep-dive experiments to reveal the impacts.

\paragraph{Time Series Formatting} 

Extracting useful information from raw sequential data as in the case of numerical time series is a challenging task for LLMs. The tokenization directly influences how the patterns are encoded within tokenized sequences \cite{gruver2023l}, and methods such as BPE separate a single number into tokens that are not aligned.
On the contrary, Llama2 has a consistent tokenization of numbers, where it splits each digit into an individual token, which ensures consistent tokenization of numbers \citep{liu2023goat}. 
We study different time series formatting approaches to determine if they influence the LLMs performance to capture the time series information. In total we propose 9 formats, ranging from simple CSV to enriched formats with additional information.

\paragraph{Time Series Length}

We study the impact that the length of the time series has in the retrieval task. Transformer-based models use attention mechanisms to weigh the importance of different parts of the input sequence. Longer sequences can dilute the attention mechanism's effectiveness, potentially making it harder for the model to focus on the most relevant parts of the text~\cite{vaswani2017}.

\paragraph{Position Bias}
Given a retrieval question, the position of where the queried data point occurs in the time series might impact the retrieval accuracy. Studies have discovered \textit{recency bias}~\cite{zhao2021calibrate} in the task of few-shot classification, where the LLM tends to repeat the label at the end. Thus, it is important to investigate whether LLM exhibits similar bias on positions in the task of time series understanding.

\section{Experiments}

\subsection{Experimental setup}

\subsubsection{Models}
We evaluate the following LLMs on our proposed framework using the test split of our dataset: 1) GPT4. \citep{achiam2023gpt} 2) GPT3.5. 3) Llama2-13B \cite{touvron2023llama}, 4) Vicuna-13B \cite{vicuna2023}, and 5) Phi3-Medium (14B)\cite{2024phi3}. We selected three open-source models, Phi3, Llama2 and Vicuna, the first with 14B parameters and the remaining with 13 billion; the version of Vicuna is 1.5 and was trained by fine-tuning Llama2. Additionally we selected GPT4 and GPT3.5 where the number of parameters is unknown. In the execution of our experiments, we used an Amazon Web Services (AWS) g5.12xlarge instance, equipped with four NVIDIA A10G Tensor Core GPUs, each featuring 24 GB of GPU RAM.

\subsubsection{Prompts}
\label{subsec:prompt_design}
The design of prompts for interacting with LLMs is separated into two approaches: retrieval/arithmetic reasoning and detection/classification questioning. In addition to zero-shot prompting, we also use chain-of-thought (CoT) \cite{CoTpaper} prompting to enhance the reasoning capabilities of LLMs. We employ regular expressions to parse the responses for feature detection and classification tasks in the zero-shot setting. However, for chain-of-thought prompting, we utilize an LLM to parse the responses due to their increased complexity and length.

\paragraph{Time series characteristics}
To evaluate the LLM reasoning over time series features, we use a two-step prompt with an adaptive approach, dynamically tailoring the interaction based on the LLM's responses. The first step involves detection, where the model is queried to identify relevant features within the data. If the LLM successfully detects a feature, we proceed with a follow-up prompt, designed to classify the identified feature between multiple sub-categories. %accurately.
For this purpose, we enrich the prompts with definitions of each sub-feature (e.g. up or down trend), ensuring a clearer understanding and more accurate identification process. 
The full list of prompts can be found in Sec.~\ref{sec:app_prompts} of the supplementary.

\paragraph{Information Retrieval/Arithmetic Reasoning}
We test the LLM's comprehension of numerical data represented as text by querying it for information retrieval and numerical reasoning, as exemplified in Fig.~\ref{fig:ret_prompt} and detailed in the supplementary Sec.~\ref{sec:app_prompts}.

\subsection{Benchmark Results}

In Table~\ref{tab:main_res}, we display the main results for the feature detection, feature classification, information retrieval and arithmetic reasoning tasks outlined in Sec.~\ref{sec:tasks}. The results for univariate time series feature detection and classification tasks illustrate GPT4's robustness in trend and seasonality detection, substantially outperforming Llama2, Vicuna, and GPT3.5 in zero-shot settings. This performance is further enhanced when chain-of-thought prompting is used. However, the detection of structural breaks and volatility presents challenges across all models, with lower accuracy scores, even with chain-of-thought prompting. GPT4 tends to always answer no for stationarity and fat tail detection tasks, while in the case of chain-of-thought prompting it does not answer, clarifying that it is only an AI model and cannot perform the necessary statistical tests.

\begin{figure}[h!]
\centering
% \captionsetup[sub]{font=footnotesize}
\begin{subfigure}{.49\linewidth}
  \centering
    \includegraphics[width=\textwidth]{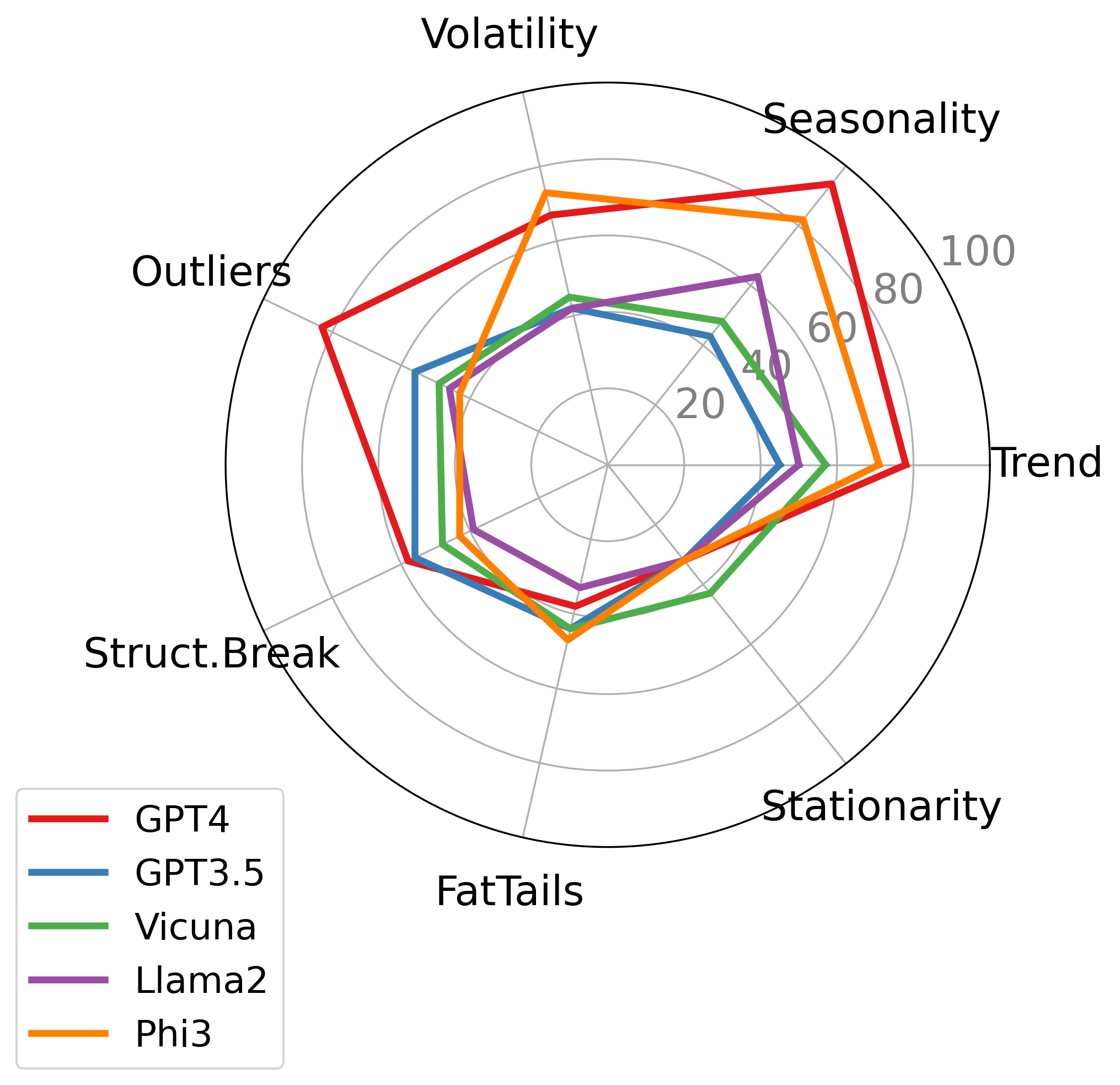}
  \caption{\small Feature detection}
  \label{fig:sub_radial_det}
\end{subfigure}~
\begin{subfigure}{.49\linewidth}
  \centering
  \includegraphics[width=\textwidth]{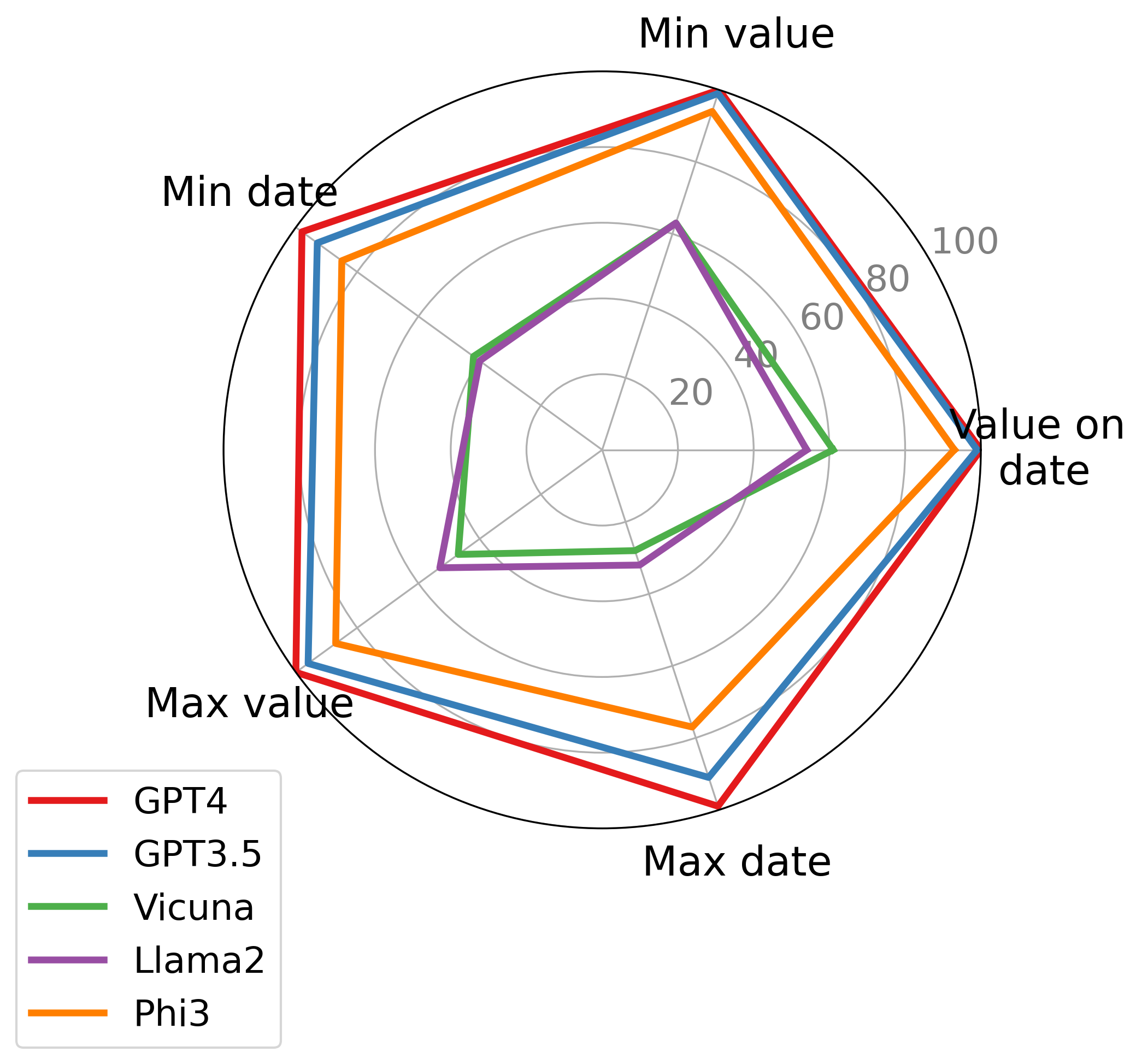}
  \caption{IR and math reasoning}
  \label{fig:sub_radial_ret}
\end{subfigure}%
\caption{Feature detection and arithmetic reasoning scores of GPT4, GPT3.5, Vicuna, Llama2 and Phi3.}
\label{fig:spider_feat_all}
\end{figure}

For trend classification, GPT4 excels in zero-shot and chain-of-thought prompting, demonstrating superior performance. Phi3 shows strong performance in zero-shot settings for trend classification, even surpassing GPT3.5 in zero-shot. In classifying seasonality, outliers, and structural breaks, Phi3 also demonstrates competitive performance, sometimes surpassing Llama2 and Vicuna, and outperforming GPT3.5 in seasonality classification, highlighting its distinct strengths. Additional plots of confusion matrices are provided in Appendix \ref{sec:app:cm} to better understand how the models select their choices, revealing potential biases such as consistently selecting the same label. Figure \ref{fig:spider_feat_all} (a) summarizes the F1 score for the feature detection task for all models, showing the strong performance on the four easier features, with Phi3 also being competitive in trend, seasonality and volatility detection. 

In multivariate time series feature detection and classification tasks, all models achieve moderate accuracy in zero-shot settings, suggesting potential for enhancement in intricate multivariate data analysis. Chain-of-thought prompting does not significantly improve performance in this context.

For \textit{information retrieval} tasks, GPT4 outperforms GPT3.5 and other models, achieving perfect accuracy in identifying the value on a given date. It also maintains a low Mean Absolute Percentage Error (MAPE), indicative of its precise value predictions. The \textit{arithmetic reasoning} results echo these findings, with GPT4 displaying superior accuracy, especially in determining minimum and maximum values within a series. Figure \ref{fig:spider_feat_all} summarizes the accuracy performance for the information retrieval and arithmetic reasoning tasks, where there are two clear groups with similar performance, GPT4, GPT3.5 and Phi3, and Llama2 and Vicuna.

In the \textit{text matching tasks}, Table \ref{tab:text-match} shows results intra-datasets, where GPT-4 significantly outperforms other models, achieving near-perfect accuracy across all datasets. This suggests that GPT-4 is capable of understanding the nuances of both qualitative and quantitative time series descriptions and effectively relating them to the underlying data. Table \ref{tab:text-match-cross} shows the results for the matching cross-datasets where GPT-4 outperforms other models on all datasets except two, showcasing its superior capability in understanding and matching qualitative descriptions even without explicit quantitative cues. The performance of GPT-3.5, Llama2, Vicuna, and Phi-3 is notably lower, indicating a greater reliance on quantitative information for accurate matching in these models. 
This overall decrease in performance, is in line with our overall findings that while numerical performance on simple arithmetic tasks is quite high, performance is generally lower for time series feature detection and classification.

\begin{table}[]
    \centering
    \caption{Accuracy of LLMs in matching time series to their corresponding textual descriptions, given four options. (Bold indicates best performance)}
\begin{subtable}{\linewidth}
\centering
    \resizebox{0.92\textwidth}{!}{
\begin{tabular}{lccccc}
\toprule
 & GPT-4 & GPT-3.5 & Llama2 & Vicuna & Phi3\\
\midrule
Trend & \bf 1.00 & 0.74 & 0.67 & 0.53 & 0.73\\
Seasonality & \bf 0.93 & 0.64 & 0.58 & 0.47 & 0.64\\
Anomalies & \bf 1.00 & 0.69 & 0.62 & 0.47 & 0.69 \\
Struct. break & \bf 0.99 & 0.63 & 0.57 & 0.39 & 0.63\\
Volatility & \bf 0.98 & 0.72 & 0.60 & 0.49 & 0.65 \\
Stationarity & \bf 0.99 & 0.72 & 0.64 & 0.52 & 0.69 \\
Fat Tails & \bf 0.99 & 0.69 & 0.61 & 0.43 & 0.68\\
% Fixed Corr. & \bf 1.00 & 0.70 & 0.64 & 0.43 & 0.68 \\
\bottomrule
\end{tabular}
}
\caption{Intra-dataset matching}
    \label{tab:text-match}
\end{subtable}
\begin{subtable}{0.92\linewidth}
\centering
    \resizebox{\textwidth}{!}{
    \begin{tabular}{lccccc}
\toprule
 & GPT-4 & GPT-3.5 & Llama2 & Vicuna & Phi3\\
\midrule
Trend & \bf 0.46 & 0.21 &  0.32 & 0.36 & 0.34  \\
Seasonality & 0.41 & \bf 0.50 &  0.32 & 0.35  & 0.31 \\
Anomalies & \bf 0.46 & 0.16  & 0.32 & 0.36 & 0.34  \\
Struct. break & \bf 0.28 & 0.1  & 0.26 & 0.27 & 0.24  \\
Volatility & 0.10  & 0.07 & \bf 0.15 & 0.12 & 0.14  \\
Stationarity & \bf 0.53 & \bf 0.53 & 0.36 & 0.42 & 0.35  \\
Fat Tails & \bf 0.10  & 0.04 & \bf 0.10 & \bf 0.10 & 0.09 \\
\bottomrule
\end{tabular}
}
\caption{Cross-dataset matching}
\label{tab:text-match-cross}
\end{subtable}
    \label{tab:my_label}
\end{table}

\subsection{Deep Dive on Performance Factors}

\paragraph{Time Series Formatting}

We present four formatting approaches in this section, \texttt{csv}, which is a common comma separated value, \texttt{plain} where the time series is formatted as \texttt{Date:YYYY-MM-DD,Value:num} for each pair date-value. We also use the formatting approach proposed by \citet{gruver2023l} which we denominate \texttt{spaces} that adds blank spaces between each digit of the time series, tokenizing each digit individually, and \texttt{symbol}, an enriched format where we add a column to the time series with arrows indicating if the value has moved up, down or remained unchanged. Examples of every approach can be found in Sec. \ref{sec:ts_format_extra} in the Appendix. 
% For the full results, please refer to Tables \ref{tab:perf_format_ret_acc} and \ref{tab:perf_format_reason_extra}. 

Table \ref{tab:prepro_short_acc} shows the results for the four time series formatting strategies.
For the information retrieval and arithmetic reasoning tasks, the \texttt{plain} formatting yields better results across all models. This approach provides more structure to the input, and outperforms other formats in a task where the connection between time and value is important. For the detection and classification tasks, the \texttt{plain} formatting does not yield better results. Interestingly the \texttt{symbol} formatting that adds an additional column to the time series yields better results in the trend classification task. 
This indicates that LLMs can effectively leverage symbolic representations of time series movements to enhance their understanding in trend classification. 

\definecolor{mypink1}{rgb}{0.858, 0.188, 0.478}
\definecolor{mygreen1}{rgb}{0.56, 0.74, 0.56}
\definecolor{myblue1}{rgb}{0.28, 0.24, 0.55}

\begin{table*}[h!]
    \centering
    \caption{Top: Time series feature detection and classification performance measured with F1 score. Bottom: Time series information retrieval and arithmetic reasoning performance measured by accuracy for different time series formats.  (Bold indicates best performance)}
        \resizebox{0.95\linewidth}{!}{
\begin{tabular}{l|cccc|cccc|cccc}
\toprule
 {} & \multicolumn{4}{c|}{\textcolor{myblue1}{GPT3.5}} & \multicolumn{4}{c|}{\textcolor{mypink1}{Llama2}} & \multicolumn{4}{c}{\textcolor{mygreen1}{Vicuna}}\\
 & csv & plain & spaces & symbol  & csv & plain & spaces &symbol   & csv & plain & spaces &symbol \\
\midrule
Min value & 0.98 & \textcolor{myblue1}{\bf 0.99} & 0.79 & 0.98  & 0.55 & \textcolor{mypink1}{\bf 0.58} & 0.20 &\textcolor{mypink1}{\bf 0.58}   & 0.63 & \textcolor{mygreen1}{\bf 0.67} & 0.17 &0.62 \\
Min date & 0.94 & \textcolor{myblue1}{\bf 0.95} & 0.69 & 0.93  & 0.28 & \textcolor{mypink1}{\bf 0.39} & 0.09 &0.29   & 0.50 & \textcolor{mygreen1}{\bf 0.55} & 0.13 &0.49 \\
Max value & 0.92 & 0.92 & 0.54 & \textcolor{myblue1}{\bf 0.94}  & 0.48 & \textcolor{mypink1}{\bf 0.56} & 0.05 &0.52   & 0.49 & 0.46 & 0.01 &\textcolor{mygreen1}{\bf 0.50} \\
Max date & 0.88 & 0.88 & 0.51 & \textcolor{myblue1}{\bf 0.89}  & 0.34 & \textcolor{mypink1}{\bf 0.46} & 0.04 &0.41   & 0.38 & \textcolor{mygreen1}{\bf 0.42} & 0.07 &0.41 \\
Value on date & \textcolor{myblue1}{\bf 0.94} & \textcolor{myblue1}{\bf 0.94} & 0.82 & \textcolor{myblue1}{\bf 0.94}  & \textcolor{mypink1}{\bf 0.39} & 0.38 & 0.07 &0.34   & 0.36 & \textcolor{mygreen1}{\bf 0.48} & 0.09 &0.41 \\
\midrule
Trend det   & \textcolor{myblue1}{\bf0.42} & 0.41 & \textcolor{myblue1}{\bf0.42} & \textcolor{myblue1}{\bf0.42} & \textcolor{mypink1}{\bf 0.51} & 0.44 & 0.34 & 0.40 & 0.51 & 0.49 & \textcolor{mygreen1}{\bf 0.54} & 0.45 \\
Trend class & 0.74 & 0.55 & 0.53 & \textcolor{myblue1}{\bf 0.92} & 0.41 & 0.48 & 0.43 & \textcolor{mypink1}{\bf 0.62} & 0.49 & 0.58 & 0.44 & \textcolor{mygreen1}{\bf 0.64} \\
Season det  & 0.61 & \textcolor{myblue1}{\bf 0.77} & 0.63 & 0.47 & \textcolor{mypink1}{\bf 0.55} & 0.24 & 0.40 & 0.50  & 0.47 & 0.47 & 0.53 & \textcolor{mygreen1}{\bf 0.54} \\
Season class & \textcolor{myblue1}{\bf 0.27} & 0.19 & 0.17 & 0.18 & 0.11 & \textcolor{mypink1}{\bf 0.13} & 0.08 & 0.10 & 0.14 & 0.14 & 0.14 & \textcolor{mygreen1}{\bf 0.15} \\
Outlier det  & 0.55 & 0.52 & 0.52 & \textcolor{myblue1}{\bf 0.62} & 0.44 & 0.35 & 0.41 & \textcolor{mypink1}{\bf 0.47} & 0.49 & 0.53 & \textcolor{mygreen1}{\bf 0.54} & 0.49 \\
Outlier class & \textcolor{myblue1}{\bfseries 0.17} & \textcolor{myblue1}{\bfseries 0.17} & \textcolor{myblue1}{\bfseries 0.17} & \textcolor{myblue1}{\bfseries 0.17} & 0.13 & \textcolor{mypink1}{\bf 0.14} & \textcolor{mypink1}{\bf 0.14} & 0.08 & \textcolor{mygreen1}{\bf 0.19} & 0.14 & 0.14 & 0.08 \\
\bottomrule
\end{tabular}
}
\label{tab:prepro_short_acc}
\end{table*}

\begin{figure*}[h!]
\centering
\begin{subfigure}{.3\textwidth}
  \centering
  \includegraphics[width=\textwidth]{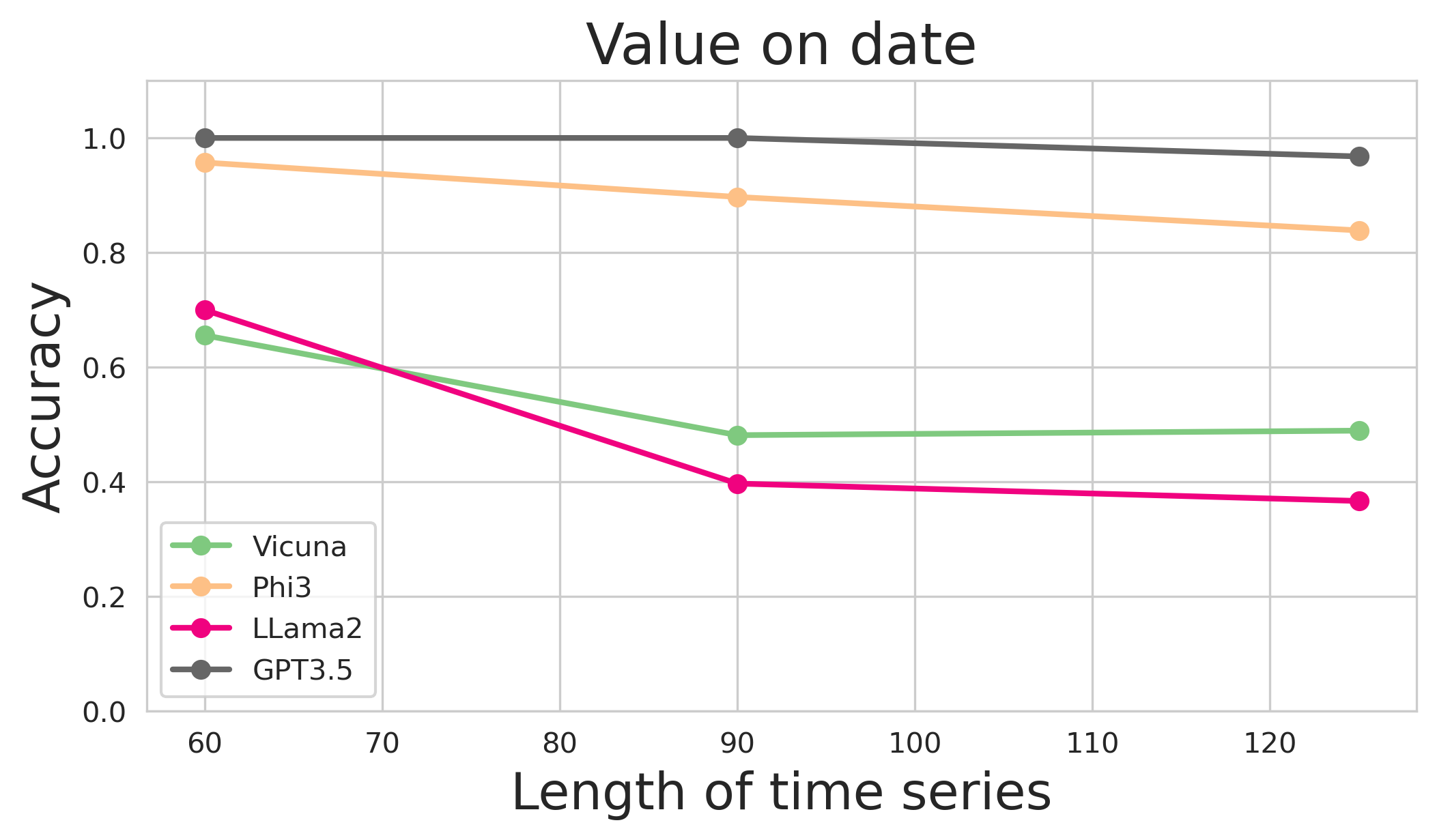}
  \caption{Trend}
  \label{fig:sub1}
\end{subfigure}%
\begin{subfigure}{.3\textwidth}
  \centering
    \includegraphics[width=\textwidth]{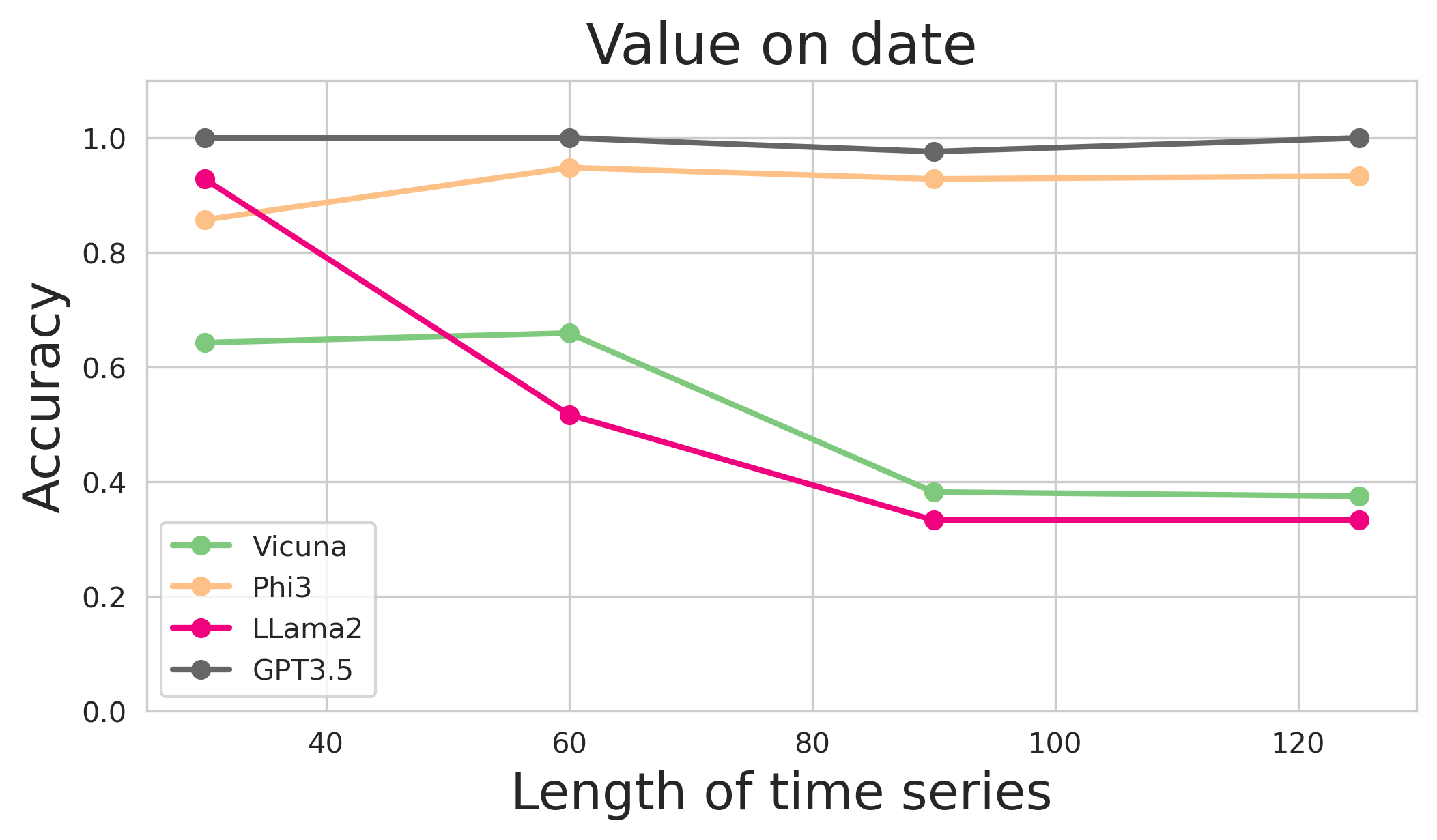}
  \caption{Seasonality}
  \label{fig:sub2}
\end{subfigure}
\begin{subfigure}{.3\textwidth}
  \centering
    \includegraphics[width=\textwidth]{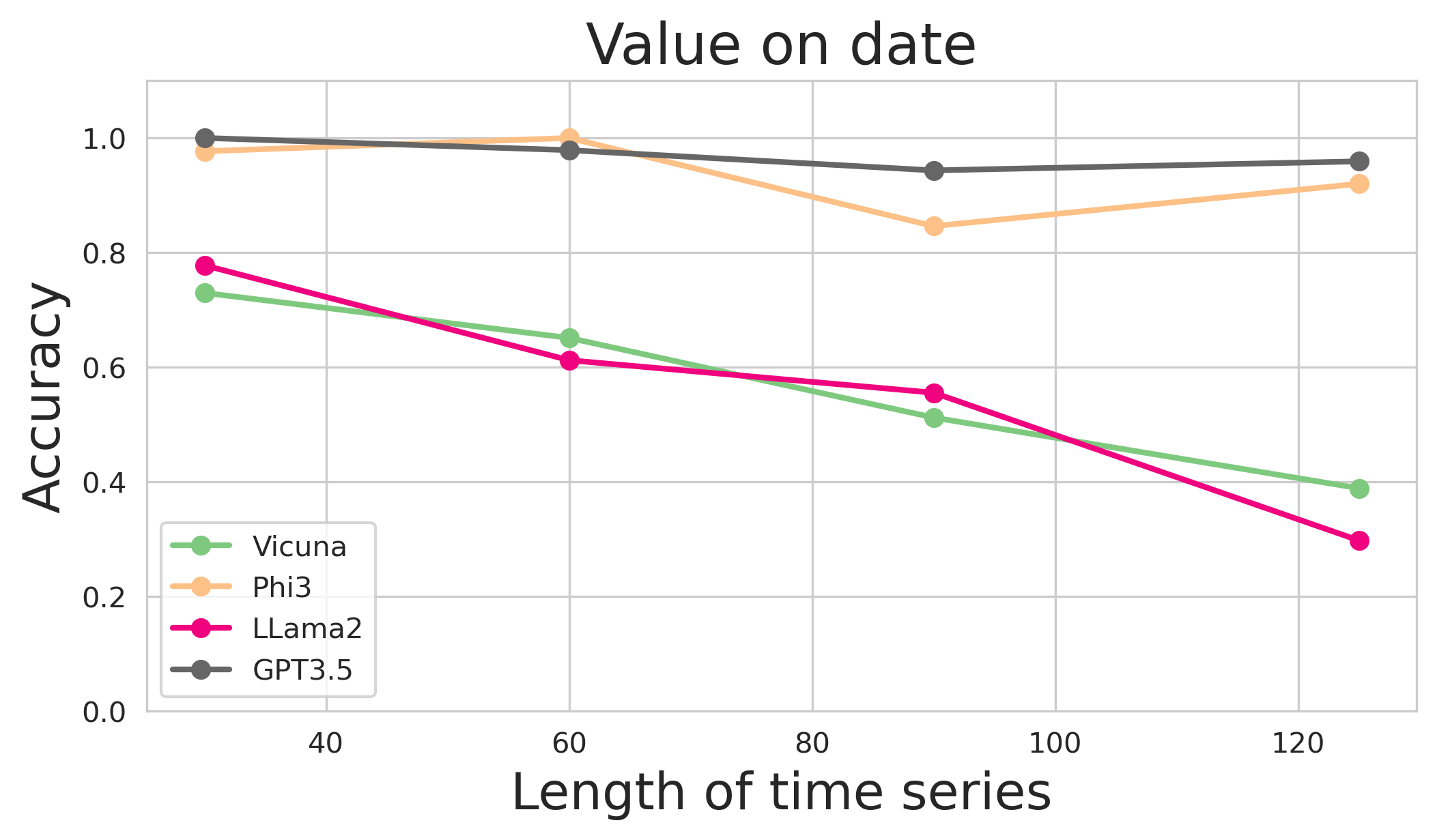}
  \caption{Outliers}
  \label{fig:sub3}
\end{subfigure}
\caption{Retrieval performance for different time series lengths.}
\label{fig:len_eval}
\end{figure*}
\paragraph{Time Series Length}

Figure \ref{fig:len_eval} shows the performance of GPT3.5, Phi3, Llama2 and Vicuna on three datasets, \texttt{trend}, \texttt{seasonality} and \texttt{outliers} which have time series with different lengths. We observe that GPT3.5 and Phi3 retrieval performance degrades slowly with increasing sequence length. Llama2 and and Vicuna suffer a more steep degradation especially from time series of length 30 steps to 60 steps.

\paragraph{Position Bias}
We carry out a series of experiments to determine how the  position of the target value affects task performance across various types of time series data. We address progressively more complex objectives: 1) identifying the presence of a value in a time series without a specified date (\ref{goal0_search}); 2) retrieving a value corresponding to a specific date (\ref{goal1_retrieval}); and 3) identifying the minimum and maximum values (\ref{goal2_min_max}). 
We cover a range of time series data, from monotonic series without noise to those with noise, sinusoidal patterns, data featuring outliers (spikes), and Brownian motion scenarios, each adding a layer of complexity. We examine how the position of the target value within the four quadrants — 1st, 2nd, 3rd, and 4th— affects the efficacy of these tasks across the varied time series landscapes. This approach helps reveal the influence of position on different LLMs (GPT3.5, Llama2, and Vicuna) in the task of time series understanding.

We consider the presence of position bias when the maximum performance gap between quadrants exceeds 10\%. 
Given this criterion, our analysis provides the following key takeaways on position bias impacting LLM performance across the defined tasks: (1) Pronounced position bias is observed across all tasks and LLMs: GPT models show significant bias exclusively in complex tasks that involve arithmetic reasoning. Both Llama2 and Vicuna demonstrate position biases across all tasks, from the simplest to the most complex ones. (2) The degree of complexity in the time series data tends to increase the extent of position bias observed within each task.
See Appendix \ref{position_experiments}, where we offer a detailed analysis of position bias across each task to further substantiate these conclusions.

\section{Conclusion}
In conclusion, we provide a critical examination of general-purpose Large Language Models (LLMs) in the context of time series understanding. Through the development of a comprehensive taxonomy of time series features and the synthesis of a diverse dataset that encapsulates these features, including qualitative and quantitative textual descriptions for each time series, we have laid a solid foundation for evaluating the capabilities of LLMs in understanding and interpreting time series data.
Our systematic evaluation sheds light on the inherent strengths and limitations of these models, offering valuable insights for practitioners aiming to leverage LLMs in time series understanding. Recognizing the areas of weakness and strength in general-purpose LLMs' current capabilities allows for targeted enhancements, ensuring that these powerful models can be more effectively adapted to specific domains.

In the future, we plan to study the performance of LLMs on real-world time series datasets to assess the generalizability of the proposed framework. This will involve testing LLMs on diverse datasets from various domains, such as finance, healthcare, and climate science. Additionally, future work should expand the analysis challenges LLMs face with multivariate time series data, including the ability to identify and interpret relationships between multiple series, such as correlation, cross-correlation, and dynamic conditional correlation. Understanding these challenges will be crucial for developing more effective LLMs for complex time series analysis. Finally, evaluating LLMs in few-shot settings is an important area for future work, as it can reveal the models' ability to learn and generalize from limited time series data. This can be particularly valuable in domains where labeled data is scarce or expensive to obtain.

\section{Limitations}
In this section, we detail the key limitations of our study and suggest pathways for future research.

Time series data frequently intersects with data from other domains. In the financial industry, for instance, analysis often combines time series data like stock prices and transaction volumes with supplementary data types such as news articles (text), economic indicators (tabular), and market sentiment analysis (textual and possibly visual). Our future work aims to delve into how LLMs can facilitate the integration of multimodal data, ensure cohesive data modality alignment within the embedding space, and accurately interpret the combined data insights.

Currently, our application of LLMs in time series analysis is primarily focused on comprehending time series features. However, the lack of interpretability mechanisms within our framework stands out as a significant shortcoming. Moving forward, we plan to focus on developing and integrating interpretability methodologies for LLMs specifically tailored to time series data analysis contexts.

\section*{Acknowledgements}

This paper was prepared for informational purposes by the Artificial Intelligence Research group of JPMorgan Chase $\&$ Co and its affiliates (“J.P. Morgan”) and is not a product of the Research Department of J.P. Morgan.  J.P. Morgan makes no representation and warranty whatsoever and disclaims all liability, for the completeness, accuracy or reliability of the information contained herein.  This document is not intended as investment research or investment advice, or a recommendation, offer or solicitation for the purchase or sale of any security, financial instrument, financial product or service, or to be used in any way for evaluating the merits of participating in any transaction, and shall not constitute a solicitation under any jurisdiction or to any person, if such solicitation under such jurisdiction or to such person would be unlawful.

\bibliography{main}

%%%%%%%%%%%%%%%%%%%%%%%%%%%%%%%%%%%%%%%%%%%%%%%%%%%%%%%%%%%%%%%%%%%%%%%%%%%%%%%
%%%%%%%%%%%%%%%%%%%%%%%%%%%%%%%%%%%%%%%%%%%%%%%%%%%%%%%%%%%%%%%%%%%%%%%%%%%%%%%
% APPENDIX
%%%%%%%%%%%%%%%%%%%%%%%%%%%%%%%%%%%%%%%%%%%%%%%%%%%%%%%%%%%%%%%%%%%%%%%%%%%%%%%
%%%%%%%%%%%%%%%%%%%%%%%%%%%%%%%%%%%%%%%%%%%%%%%%%%%%%%%%%%%%%%%%%%%%%%%%%%%%%%%
\newpage
\appendix
\onecolumn

\section{Additional details of Taxonomy}
\begin{table}[h!]
\centering
\begin{small}
\resizebox*{!}{0.85\textheight}{
% \begin{tabular}{|p{3.5cm}|p{6cm}|p{7cm}|}
\begin{tabular}{|p{0.23\linewidth}|p{0.35\linewidth}|p{0.41\linewidth}|}
% \begin{tabular}p{0.23\linewidth} p{0.38\linewidth} p{0.38\linewidth}
\toprule
\textbf{Feature} & \textbf{Description} & \textbf{Example Use Cases} \\ 
\midrule
\ctt{pale1b}{Trend} & The general direction of a time series, either increasing (upward) or decreasing (downward) over a long period. & \textbf{Finance:} Stock price trends, inflation rates. \textbf{Climate:} Global temperature trends. Energy: Long-term energy consumption trends. \\ 
\midrule
\ctt{pale2b}{Seasonality} & A repeating pattern in a time series that occurs at regular intervals, such as daily, weekly, monthly, or yearly. & \textbf{Energy:} Seasonal variations in electricity demand. \textbf{Retail:} Seasonal sales patterns (e.g., holiday shopping). \textbf{Tourism:} Seasonal fluctuations in visitor numbers. \\ \hline
\ctt{pale2}{Fixed-Period} & Seasonality with a constant, unchanging period (e.g., monthly seasonality). & \textbf{Energy:} Monthly variations in electricity usage. \textbf{Finance:} Quarterly earnings reports. \\ \hline
\ctt{pale2}{Shifting Period} & Seasonal patterns where the length of the period shifts over time. & \textbf{Climate:} Shifting seasonal temperature patterns due to climate change. \textbf{Retail:} Shifting sales patterns due to changing consumer behavior. \\ \hline
\ctt{pale2}{Multiple Seasonality} & Presence of multiple overlapping seasonal patterns (e.g., both weekly and monthly seasonality). & \textbf{Finance:} Weekly and monthly trading cycles. \textbf{Health:} Weekly and annual cycles in flu cases. \\ 
\midrule
\ctt{pale3b}{Volatility} & The degree of variation of a time series over time, often measured by the standard deviation or variance. & \textbf{Finance:} Stock market volatility, exchange rate fluctuations. \textbf{Energy:} Price volatility in commodity markets. \textbf{Weather:} Day-to-day fluctuations in temperature or precipitation. \\ \hline
\ctt{pale3}{Constant Volatility} & The degree of variation in the time series remains consistent and predictable over time. & \textbf{Finance:} Stable bond markets. \textbf{Energy:} Consistent electricity prices. \\ \hline
\ctt{pale3}{Trending Volatility} & The level of variation in the time series shows a clear increasing or decreasing trend over time. & \textbf{Finance:} Increasing volatility in emerging markets. \textbf{Climate:} Increasing variability in weather patterns. \\ \hline
\ctt{pale3}{Clustered Volatility} & The time series exhibits periods where volatility is significantly higher or lower, with these periods tending to cluster together. & \textbf{Finance:} Volatility clustering in financial markets during crises. \textbf{Economics:} Clustered periods of high inflation. \\ \hline
\ctt{pale3}{Dynamic Volatility} & The volatility of the time series changes over time in response to external factors (e.g., leverage effect where the volatility of the time series tends to increase when the series experiences negative returns). & \textbf{Finance:} Changing volatility due to market interventions. \textbf{Climate:} Volatility changes in response to natural disasters. \\ 
\midrule
\ctt{pale4b}{Anomalies} & Data points that deviate significantly from the expected pattern of a time series. & \textbf{Quality Control:} Detecting defective products in a manufacturing process. \textbf{Network Security:} Identifying unusual traffic patterns that may indicate cyberattacks. \textbf{Finance:} Detecting fraudulent transactions. \\ \hline
\ctt{pale4}{Spike} & A sudden and brief deviation from the overall pattern of the data. & \textbf{Finance:} Sudden stock price jumps. \textbf{Weather:} Temperature spikes during heatwaves. \\ \hline
\ctt{pale4}{Level Shift} & A sudden and lasting change in the average value of a time series. & \textbf{Economics:} Changes in consumer confidence or business sentiment. \textbf{Energy:} Shifts in energy consumption patterns due to technological advancements or policy changes. \textbf{Environmental Science:} Changes in water levels or pollutant concentrations due to natural or human-induced factors. \\ \hline
\ctt{pale4}{Temporal Disruption} & An interval where data is missing or not recorded. & \textbf{Network Security:} Periods of data loss in network traffic. \textbf{Health:} Missing data in patient records. \\ 
\midrule
\ctt{pale5b}{Structural Breaks} & Abrupt changes in the underlying structure of a time series, often caused by external events or policy changes. & \textbf{Economics:} Changes in economic policy or regulations. \textbf{Finance:} Market crashes or financial crises. \textbf{Epidemiology:} Changes in disease transmission patterns due to interventions. \\ 
\midrule
\ctt{pale6b}{Stationarity} & A time series is stationary if its statistical properties, such as mean and variance, do not change over time. & \textbf{Econometrics:} Assumption for many time series models. \textbf{Finance:} Assessing the stability of financial markets. \\
\midrule
\ctt{pale7b}{Fat Tails} & A distribution of a time series where extreme events are more likely than expected under a normal distribution. & \textbf{Finance:} Modeling extreme price movements in financial markets. \textbf{Insurance:} Pricing insurance policies for catastrophic events. \\ 
\bottomrule
\end{tabular}
}
\end{small}
\caption{Definitions and examples of time series analysis features and sub-categories.}
\label{tab:taxonomy-definitions}
\end{table}

\newpage

\section{Synthetic Time Series Dataset}
\label{sec:data}

\subsection{Univariate Time Series}
The primary characteristics considered in our univariate dataset include:
\begin{enumerate}
\item \textbf{Trend} We generated time series data to analyze the impact of trends on financial market behavior. This dataset encompasses linear and quadratic trends. For linear trends, each series follows a simple linear equation a * t + b, where a (the slope) varies between 0.1 and 1, multiplied by the direction of the trend, and b (the intercept) is randomly chosen between 100 and 110. This simulates scenarios of steadily increasing or decreasing trends. For quadratic trends, the series is defined by $a * t^2 + b * t + c$, with a varying between 0.01 and 0.05 (again adjusted for trend direction), b between 0 and 1, and c between 0 and 10, or adjusted to ensure non-negative values. The quadratic trend allows us to simulate scenarios where trends accelerate over time, either upwards or downwards, depending on the direction of the trend. This approach enables the exploration of different types of trend behaviors in financial time series, from gradual to more dynamic changes, providing a comprehensive view of trend impacts in market data.

\item \textbf{Seasonality} In our study, we meticulously crafted a synthetic dataset to explore and analyze the dynamics of various types of seasonality within time series data, aiming to closely mimic the complexity found in real-world scenarios. This dataset is designed to include four distinct types of seasonal patterns, offering a broad spectrum for analysis: (1) Fixed Seasonal Patterns, showcasing regular and predictable occurrences at set intervals such as daily, weekly, or monthly, providing a baseline for traditional seasonality; (2) Varying Amplitude, where the strength or magnitude of the seasonal effect fluctuates over time, reflecting phenomena where seasonal influence intensifies or diminishes; (3) Shifting Seasonal Pattern, characterized by the drift of seasonal peaks and troughs over the timeline, simulating scenarios where the timing of seasonal effects evolves; and (4) Multiple Seasonal Patterns, which presents a combination of different seasonal cycles within the same series, such as overlapping daily and weekly patterns, to capture the complexity of real-world data where multiple seasonalities interact. This diverse dataset serves as a foundation for testing the sensitivity and adaptability of analytical models to detect and quantify seasonality under varying and challenging conditions.

\item \textbf{Anomalies and outliers} refer to observations that significantly deviate from the typical pattern or trend observed in the dataset. The types of outliers included in our generated dataset are: 1) single sudden spike for isolated sharp increases, 2) double and triple sudden spikes for sequences of consecutive anomalies, 3) step spike and level shift for persistent changes, and 4) temporal disruption for sudden interruptions in the pattern. We also include a no outlier category as a control for comparative analysis. Parameters such as the location and magnitude of spikes, the duration and start of step spikes, the placement and size of level shifts, and the initiation and conclusion of temporal disruptions are randomly assigned to enhance the dataset's diversity and relevance.

\item \textbf{Structural breaks}
in time series data signify substantial changes in the model generating the data, leading to shifts in parameters like mean, variance, or correlation. These are broadly classified into two types: parameter shifts and regime shifts, with a third category for series without breaks. 
Parameter shifts involve changes in specific parameters such as mean or variance, including sub-types like mean shifts, variance shifts, combined mean-variance shifts, seasonality amplitude shifts, and autocorrelation shifts. 
Regime shifts represent deeper changes that affect the model's structure, including: distribution changes (e.g., normal to exponential), stationarity changes (stationary to non-stationary), linearity changes (linear to non-linear models), frequency changes, noise trend changes, error correlation changes, and variance type changes. 
The occurrence of these shifts is randomly determined within the time series.

\item \textbf{Volatility} 
We generated synthetic time series data to simulate various volatility patterns, specifically targeting clustered volatility, leverage effects, constant volatility, and increasing volatility, to mimic characteristics observed in financial markets.

For clustered volatility, we utilized a GARCH(1,1) model with parameters \(\omega=0.1\), \(\alpha=0.2\), and \(\beta=0.7\), ensuring the sum of \(\alpha\) and \(\beta\) remained below 1 for stationarity, thus capturing high volatility persistence. The GARCH(1,1) model is defined by the equations:
\[
\sigma_t^2 = \omega + \alpha r_{t-1}^2 + \beta \sigma_{t-1}^2
\]
\[
r_t = \sigma_t \epsilon_t
\]
where \(\sigma_t^2\) is the conditional variance, \(r_t\) is the return at time \(t\), and \(\epsilon_t\) is white noise.

To simulate the leverage effect, our model increased volatility in response to negative returns, reflecting typical market dynamics. The leverage effect model was designed with a base volatility of 0.1 and a leverage strength of 0.3, ensuring that volatility would significantly increase after negative returns while gradually reverting to the base level after positive returns. The model is defined by:
\[
r_t = \sigma_{t-1} \epsilon_t
\]
\[
\sigma_t = 
\begin{cases} 
\sigma_{t-1} (1 + \text{leverage\_strength}) & \text{if } r_t < 0 \\
\max(\sigma_{t-1} (1 - \text{leverage\_strength}), 0.01) & \text{if } r_t \geq 0 
\end{cases}
\]

Additionally, we created time series with constant volatility by adding normally distributed random noise (standard deviation of 1) to a cumulative sum of random values. This produced a time series with a consistent level of volatility throughout the period. Mathematically, this is represented as:
\[
r_t = \sum_{i=1}^t \epsilon_i + \eta_t
\]
where \(\epsilon_i\) is white noise and \(\eta_t \sim N(0, 1)\).

For increasing volatility, we scaled the noise in proportion to the increasing range of the series, with a scaling factor up to 5 towards the end of the series. This was achieved by multiplying the standard deviation of the random noise by a linearly increasing factor, resulting in a volatility profile that progressively intensified. This can be described by:
\[
\sigma_t = \sigma_0 \left(1 + \frac{t}{n} \cdot 5\right)
\]
\[
r_t = \epsilon_t \cdot \sigma_t
\]
where \(\sigma_0\) is the initial standard deviation and \(n\) is the total number of points.

To ensure non-negative volatility values across all simulations, we took the absolute values of the generated noise. These methodologies enabled us to comprehensively represent different volatility behaviors in financial time series, including constant, increasing, clustered, and leverage-induced volatilities. By using these varied approaches, we enriched our analysis with diverse market conditions, providing a robust dataset for evaluating the performance of models designed to handle different volatility patterns.

\item \textbf{Statistical properties}
Next, we constructed a dataset to delve into significant features of time series data, centering on fat tails and stationarity. The dataset sorts series into four categories: those exhibiting fat tails, characterized by a higher likelihood of extreme values than in a normal distribution; non-fat-tailed, where extreme values are less probable; stationary, with unchanging mean, variance, and autocorrelation; and non-stationary series. Non-stationary series are further divided based on: 1) changing mean: series with a mean that evolves over time, typically due to underlying trends.
2) changing variance: series where the variance, or data spread, alters over time, suggesting data volatility.
3) seasonality: series with consistent, cyclical patterns occurring at set intervals, like seasonal effects.
4) trend and seasonality: series blending both trend dynamics and seasonal fluctuations.
\end{enumerate}

\subsection{Multivariate Time Series}
For our analysis, we confined each multivariate series sample to include just 2 time series. The main features of our generated multivariate dataset encompass:
\begin{enumerate}
\item \textbf{Correlation} involves analyzing the linear relationships between series, which is crucial for forecasting one time series from another when a correlation exists. The randomly selected correlation coefficient quantifies the strength and direction of relationships as positive (direct relationship), negative (inverse relationship), or neutral (no linear relationship) between series.

\item \textbf{Cross-correlation} evaluates the relationship between two time series while considering various time lags, making it valuable for pinpointing leading or lagging relationships between series. For our data generation, the time lag and correlation coefficient are randomly chosen.

\item \textbf{Dynamic conditional correlation}  focuses on scenarios where correlations between series vary over time.  The points in the time series at which correlation shifts take place are selected randomly.
\end{enumerate}

\newpage
\subsection{Data Examples}
\begin{longtable}{c}
\hline
\\
\textbf{Trend} \\
\begin{minipage}{4cm}
\includegraphics[scale = 0.3]{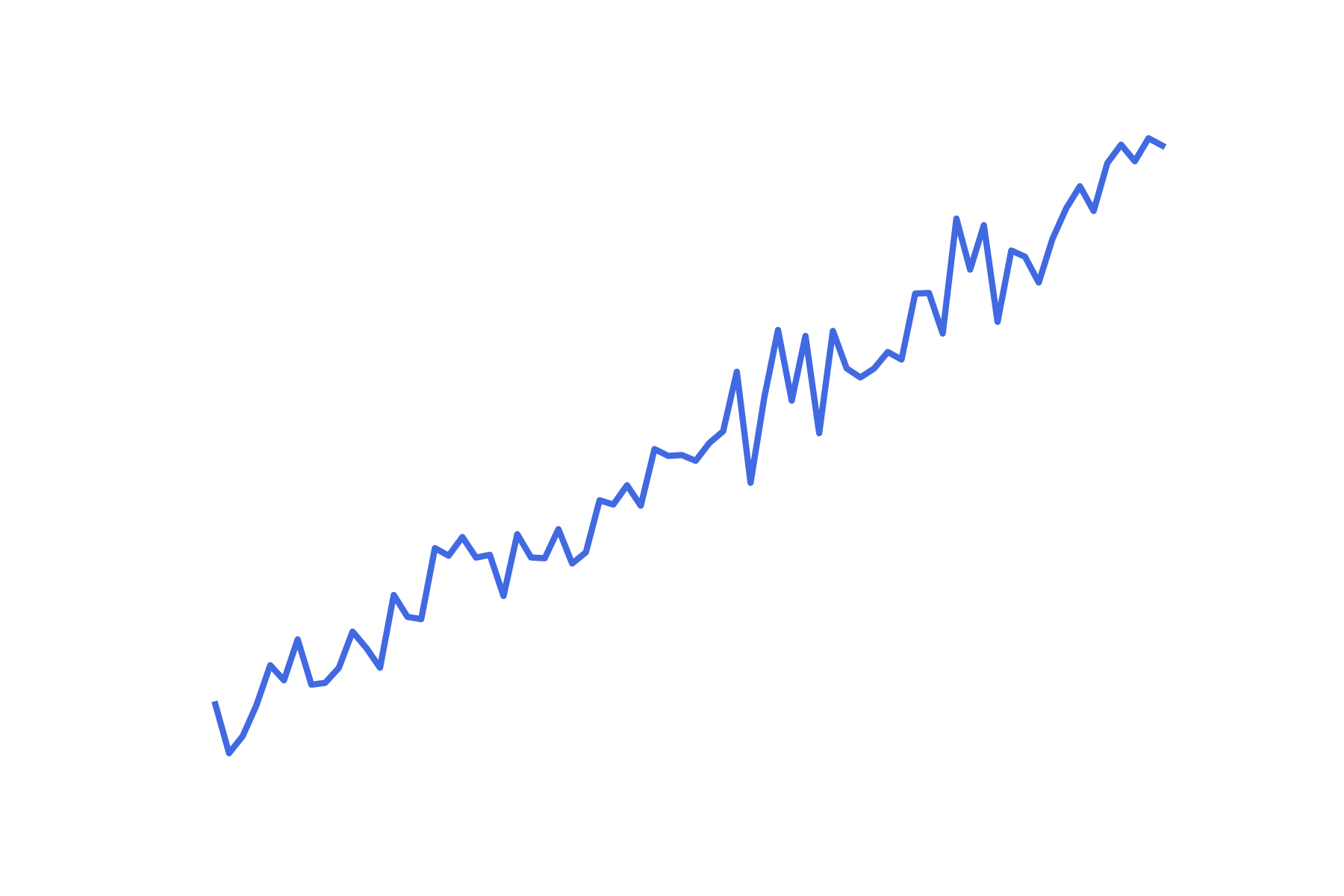}
\centering
(a) Positive trend
\end{minipage}
\begin{minipage}{4cm}
\includegraphics[scale = 0.3]{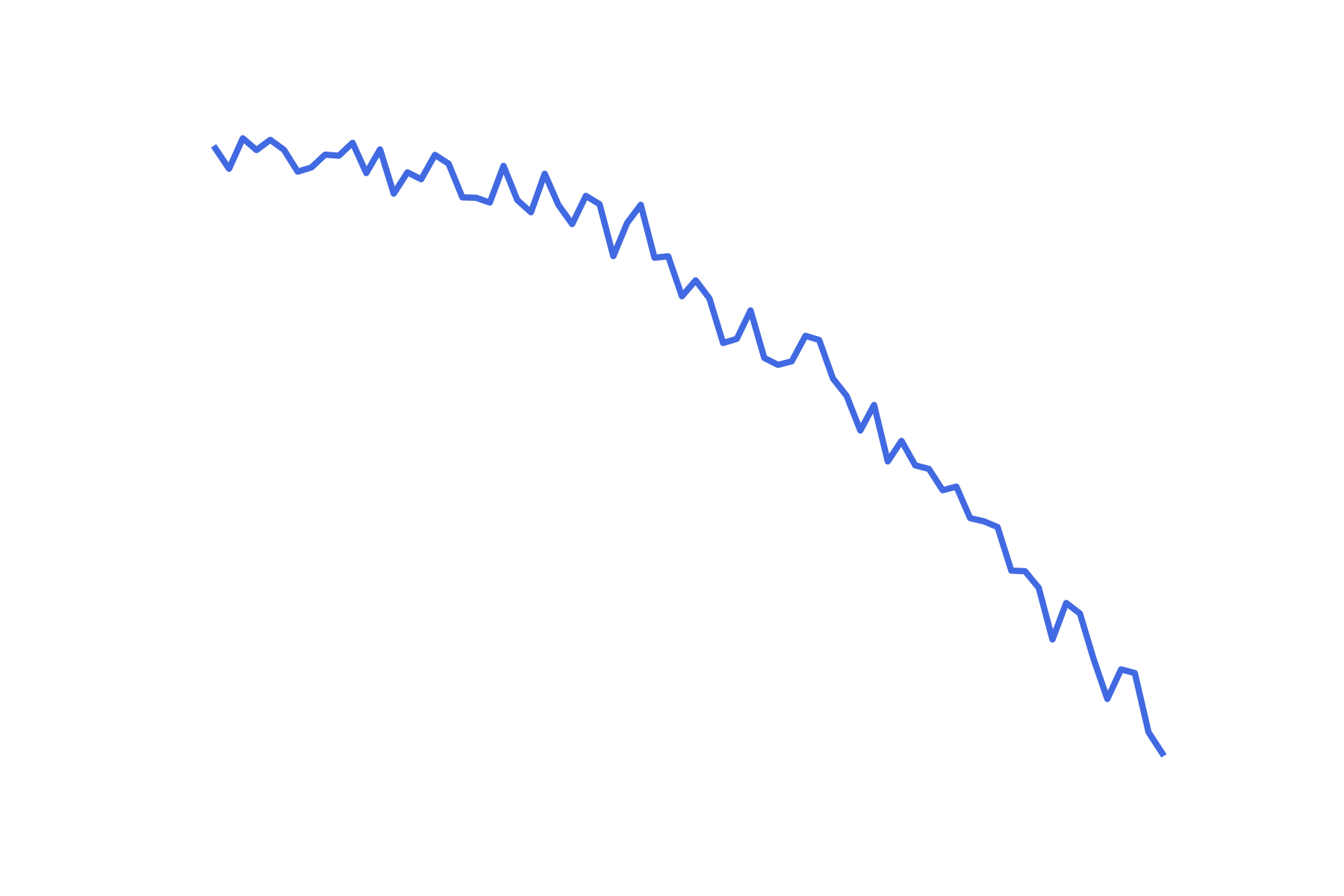}
\centering
(b) Negative trend
\end{minipage}
\begin{minipage}{4cm}
\includegraphics[scale = 0.3]{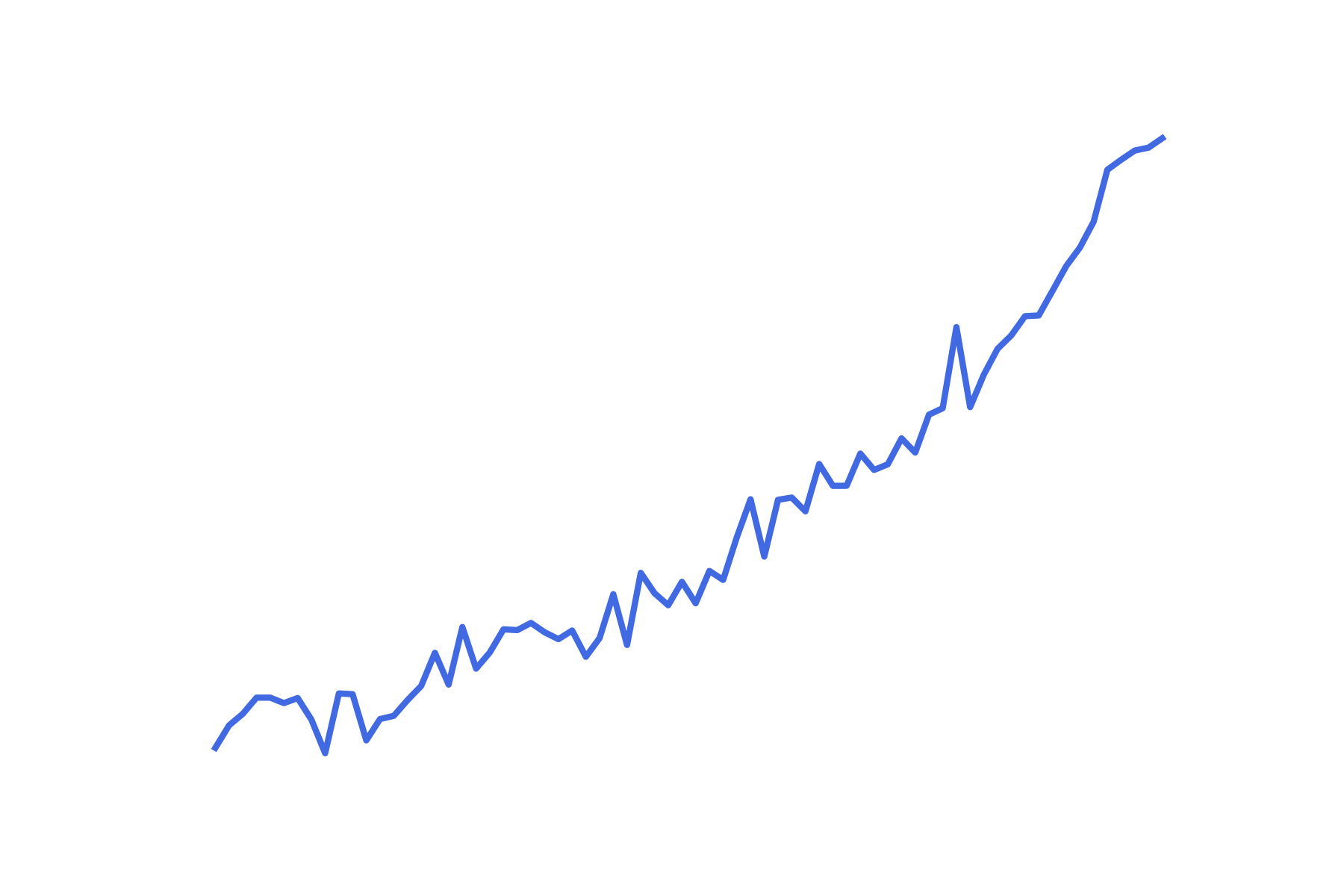}
\centering
(c) Positive trend
\end{minipage}
\begin{minipage}{4cm}
\includegraphics[scale = 0.3]{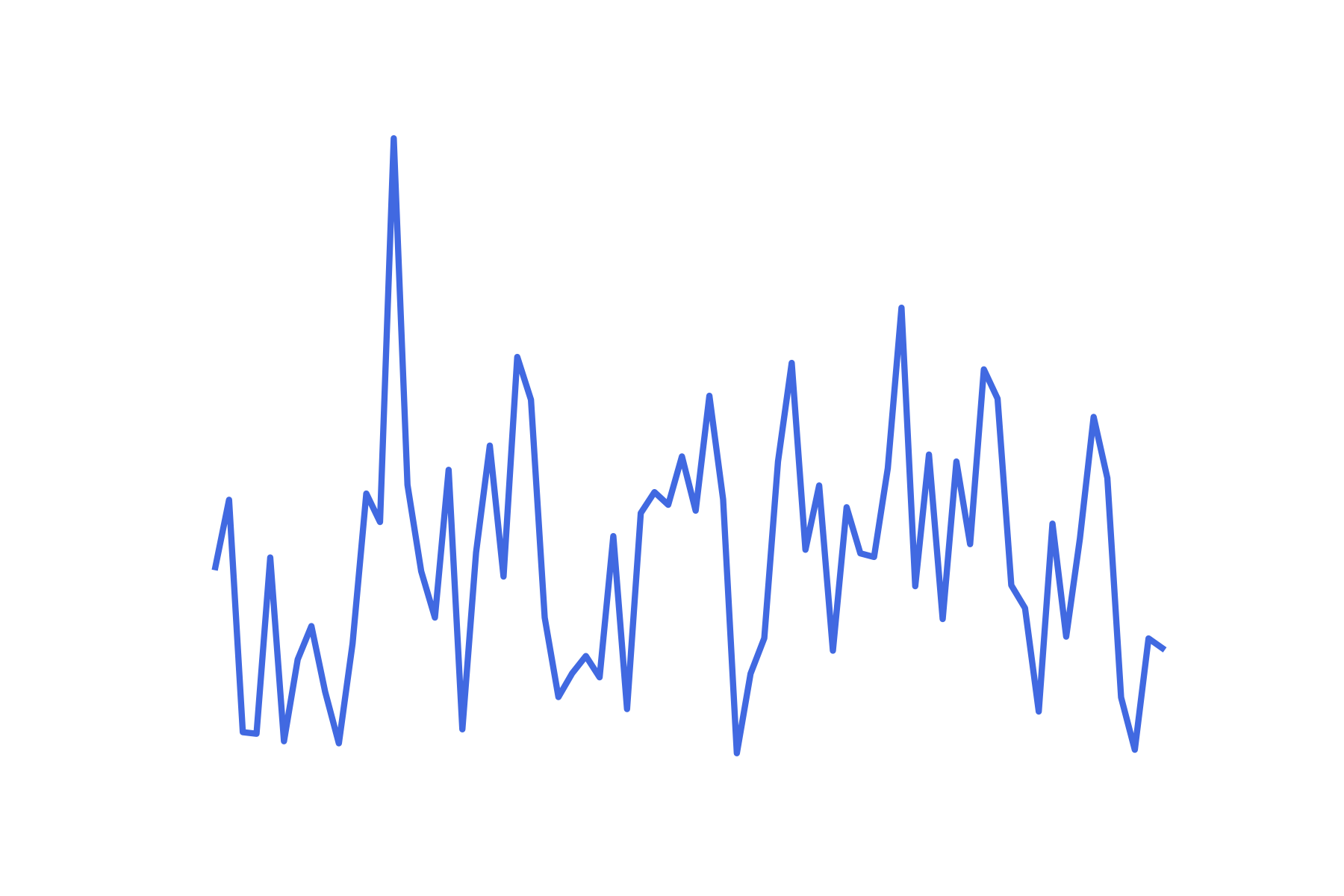}
\centering
(d) No clear trend
\end{minipage}
\\ % End of the first row of images and labels
\midrule
\\
\textbf{Seasonality} \\
\begin{minipage}{4cm}
\includegraphics[scale = 0.3]{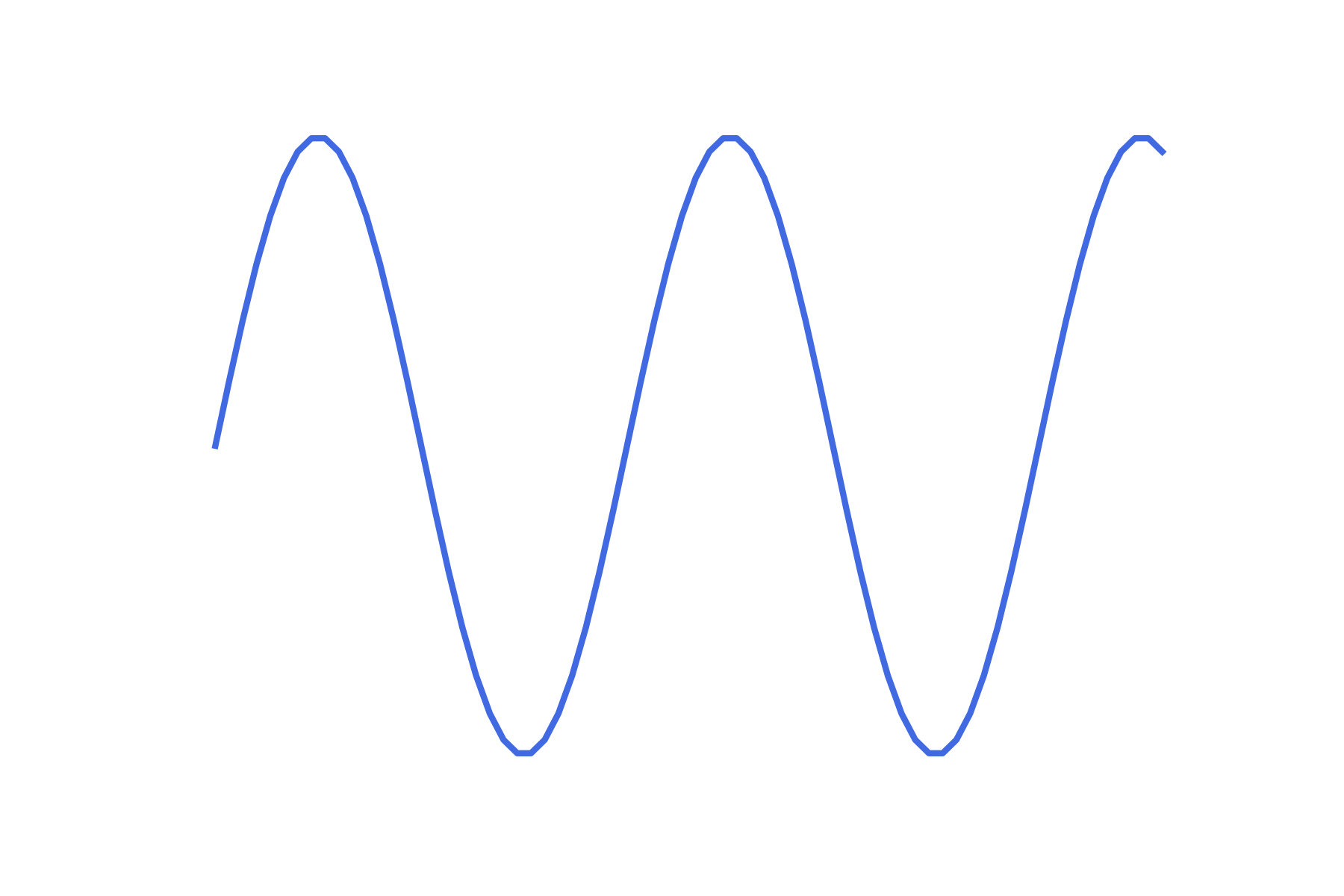}
\centering
(a) Fixed seasonality
\end{minipage}
\begin{minipage}{4cm}
\includegraphics[scale = 0.3]{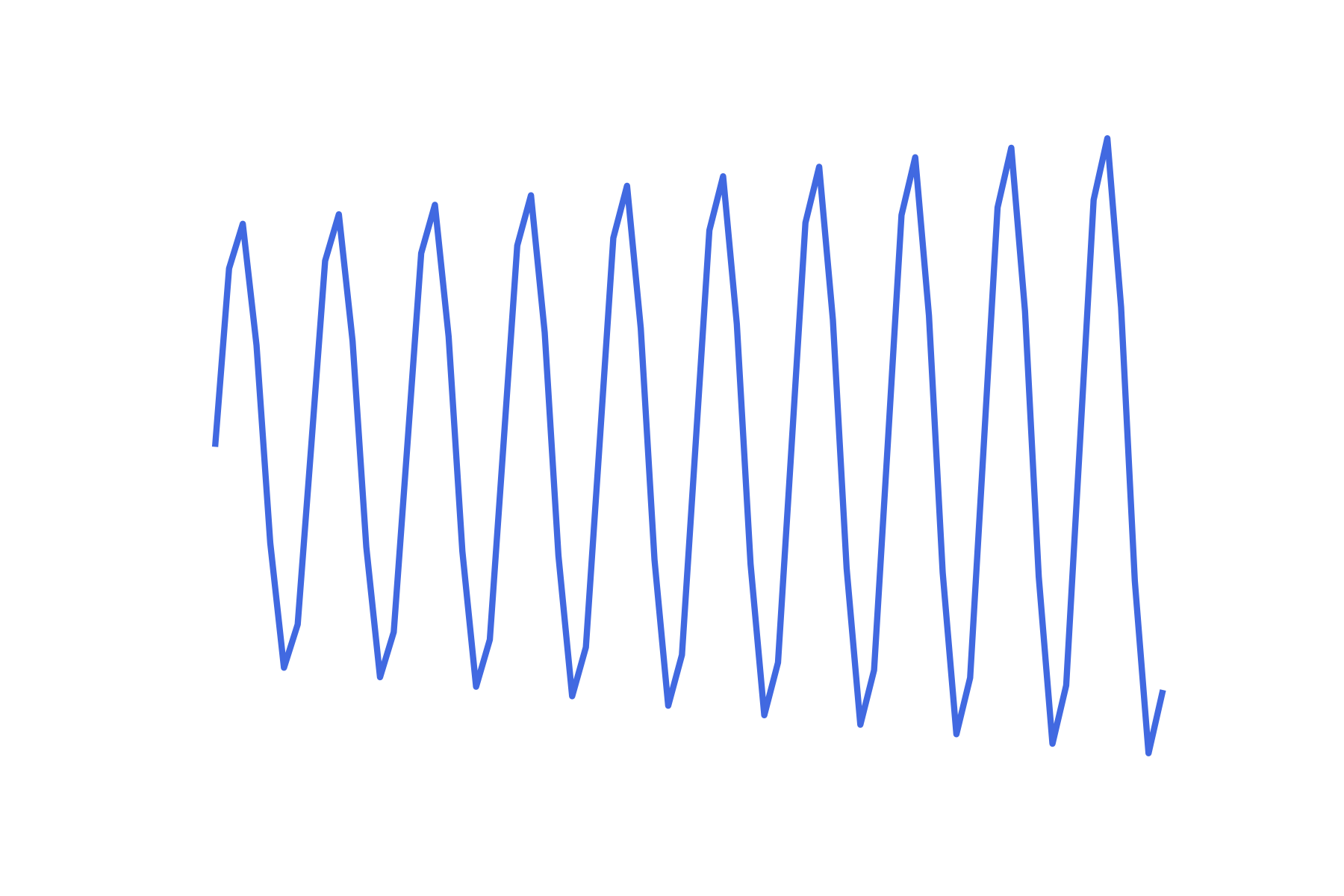}
\centering
(b) Fixed seasonality 
\end{minipage}
\begin{minipage}{4cm}
\includegraphics[scale = 0.3]{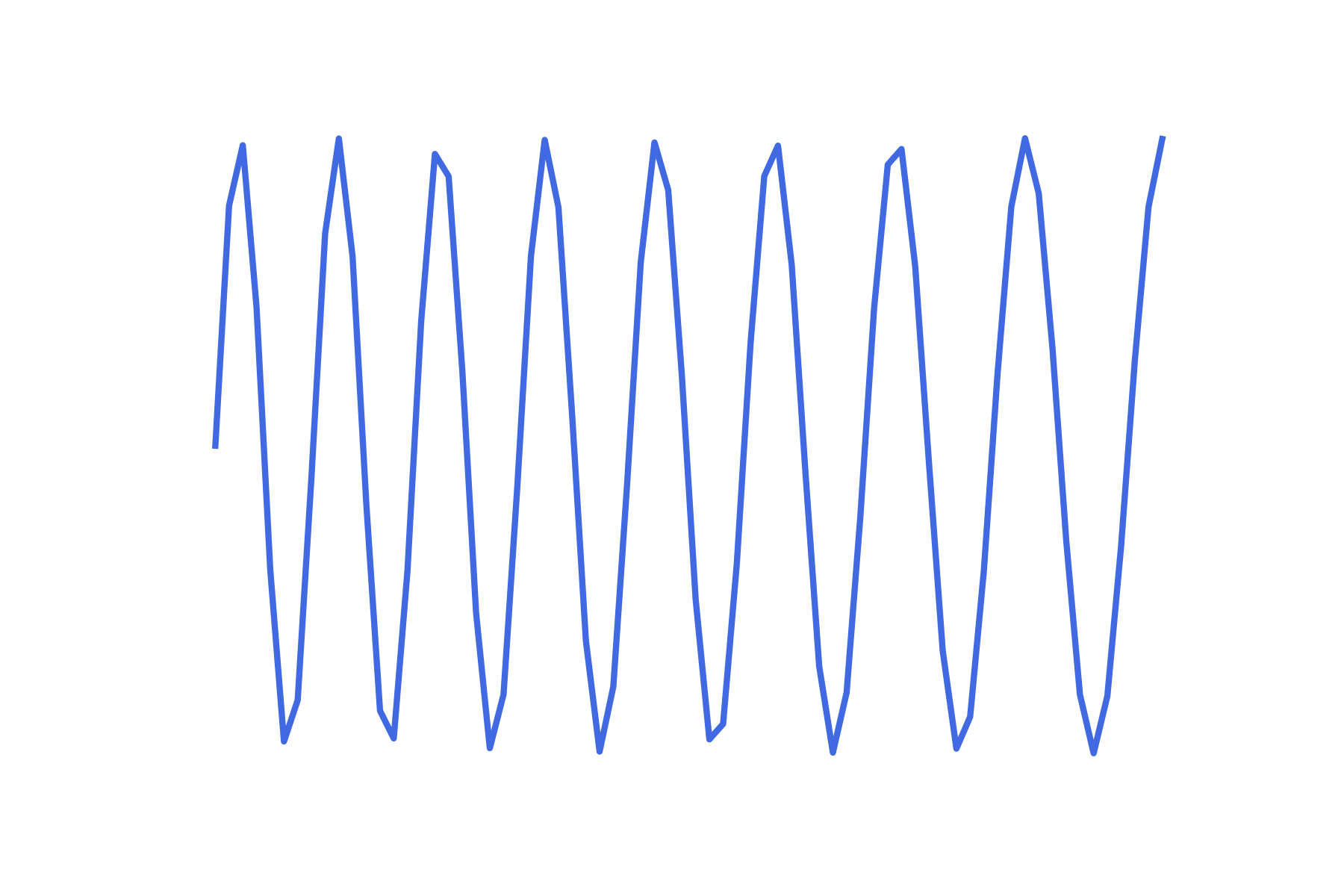}
\centering
(c) Shifting patterns 
\end{minipage}
\begin{minipage}{4cm}
\includegraphics[scale = 0.3]{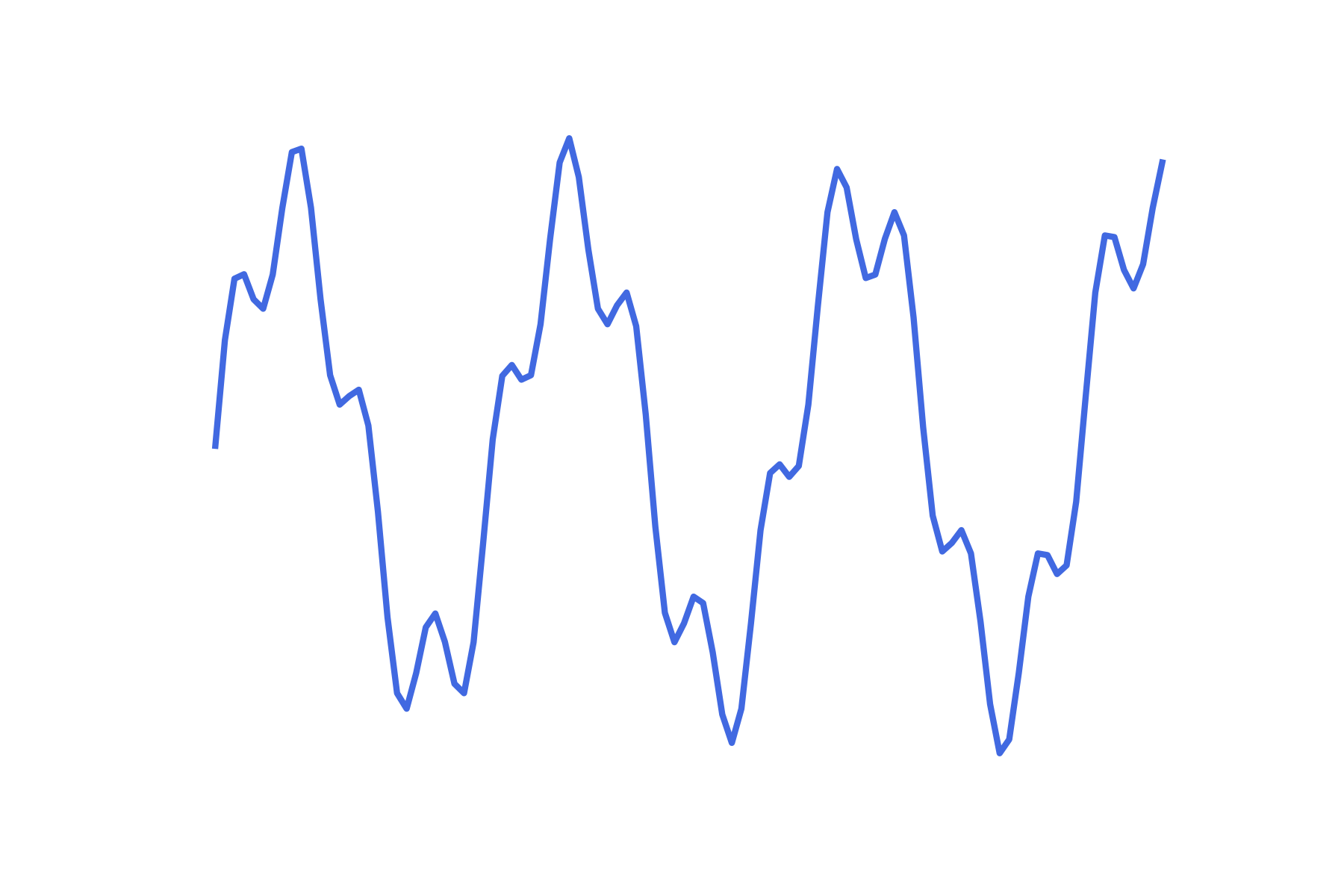}
\centering
(d) Multiple Seasonalities
\end{minipage}
\\ % End of the first row of images and labels
\midrule
\\
\textbf{Volatility} \\
\begin{minipage}{4cm}
\includegraphics[scale = 0.3]{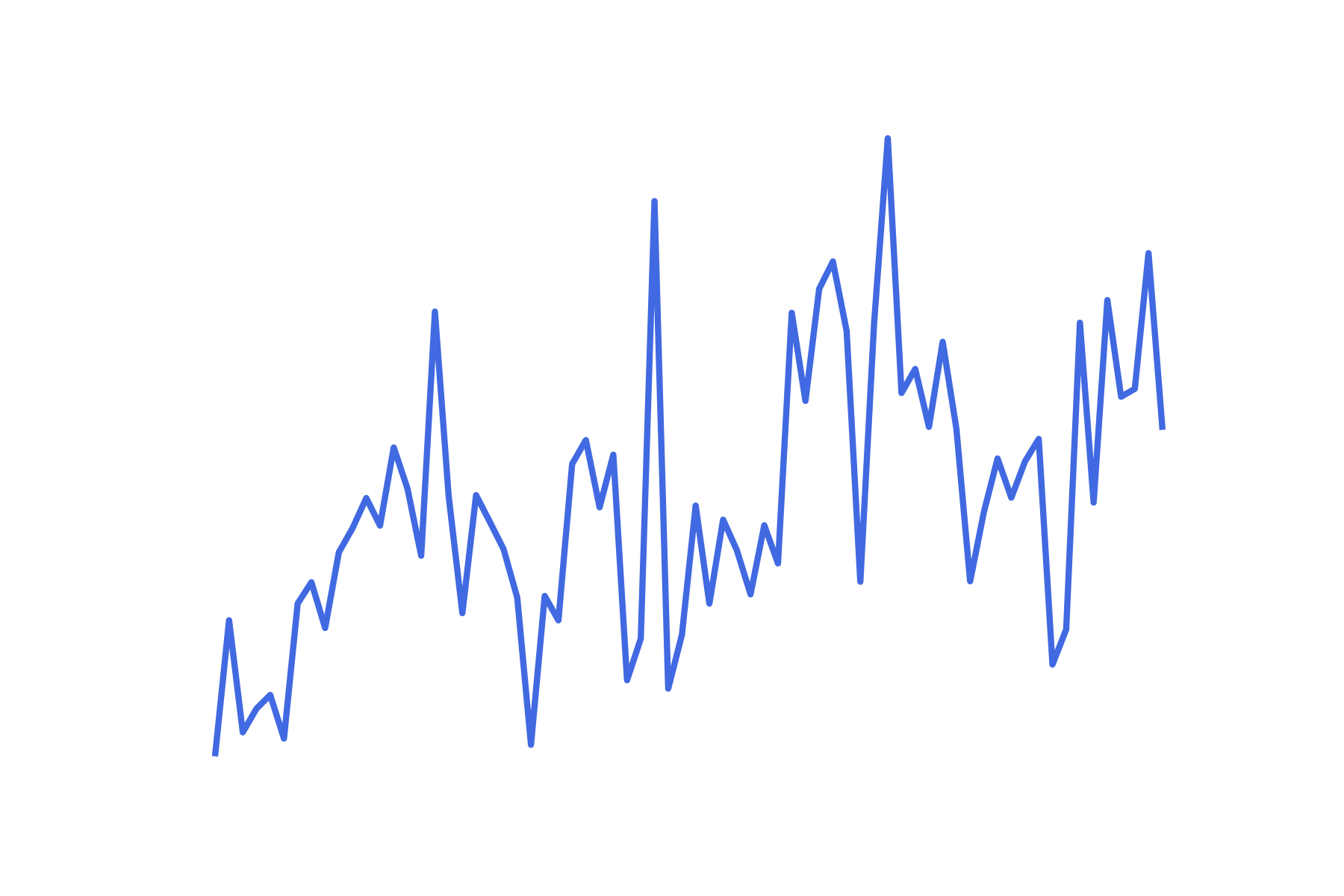}
\centering
(a) Constant volatility 
\end{minipage}
\begin{minipage}{4cm}
\includegraphics[scale = 0.3]{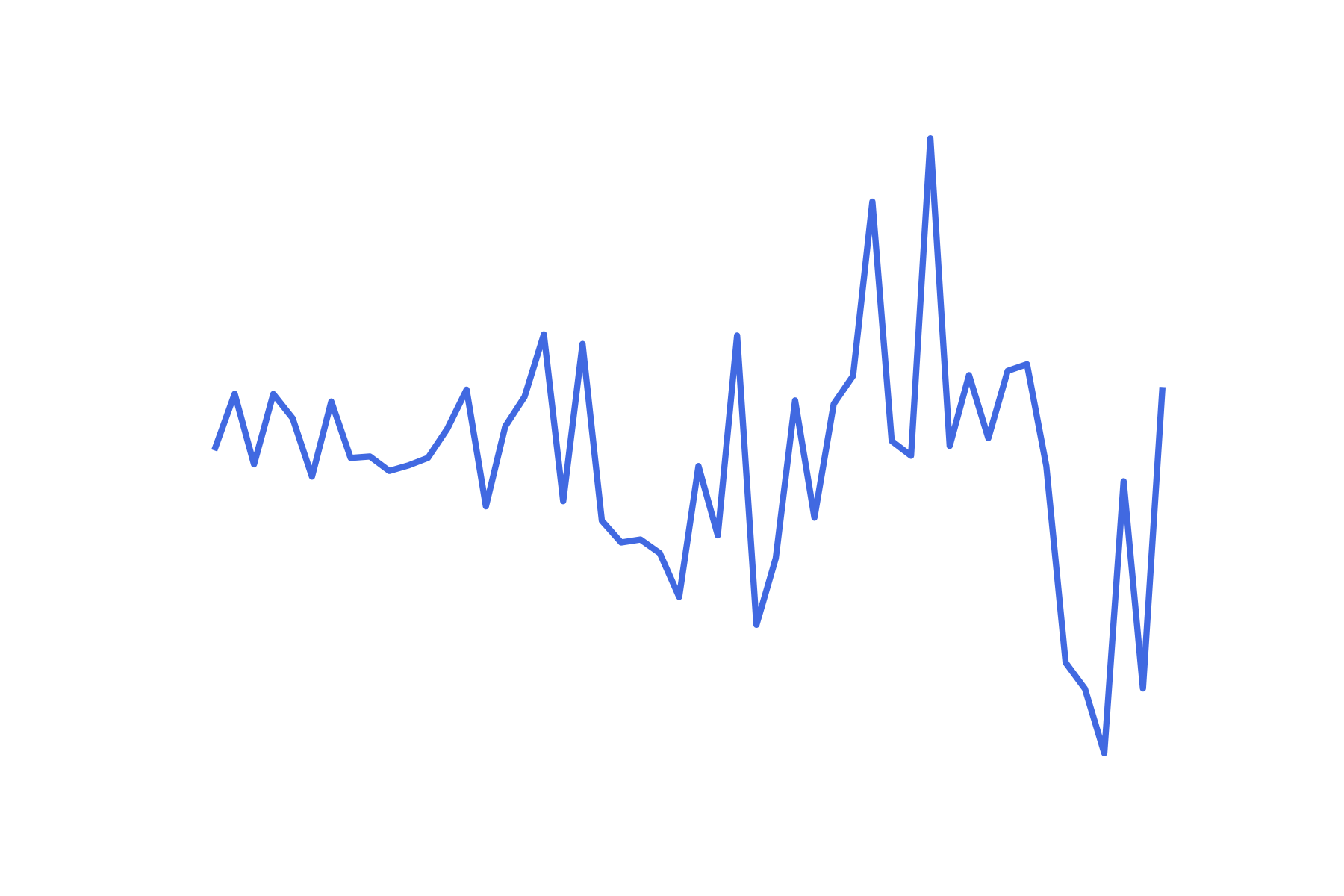}
\centering
(b) Increasing volatility 
\end{minipage}
\begin{minipage}{4cm}
\includegraphics[scale = 0.3]{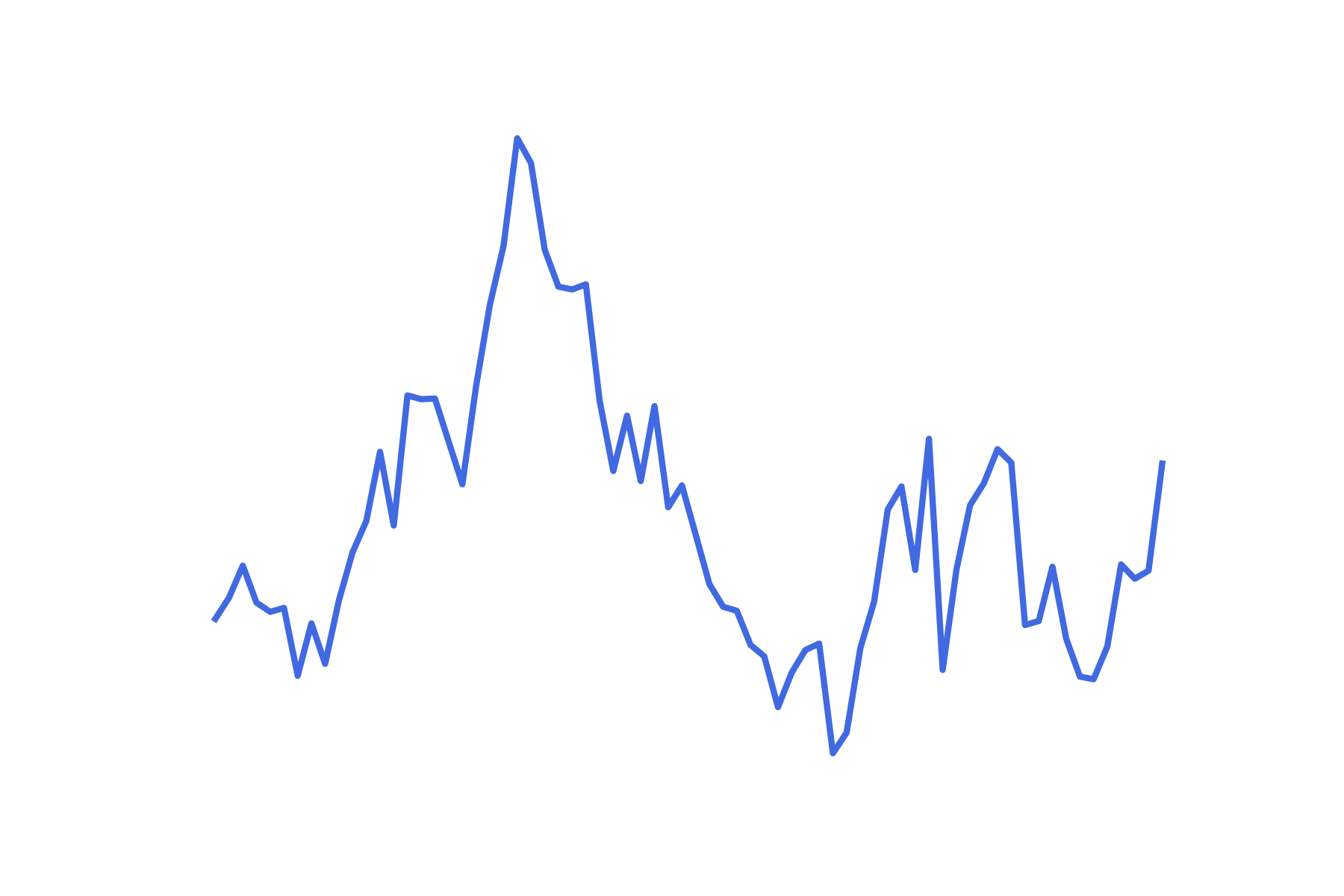}
\centering
(c) Clustered volatility
\end{minipage}
\begin{minipage}{4cm}
\includegraphics[scale = 0.3]{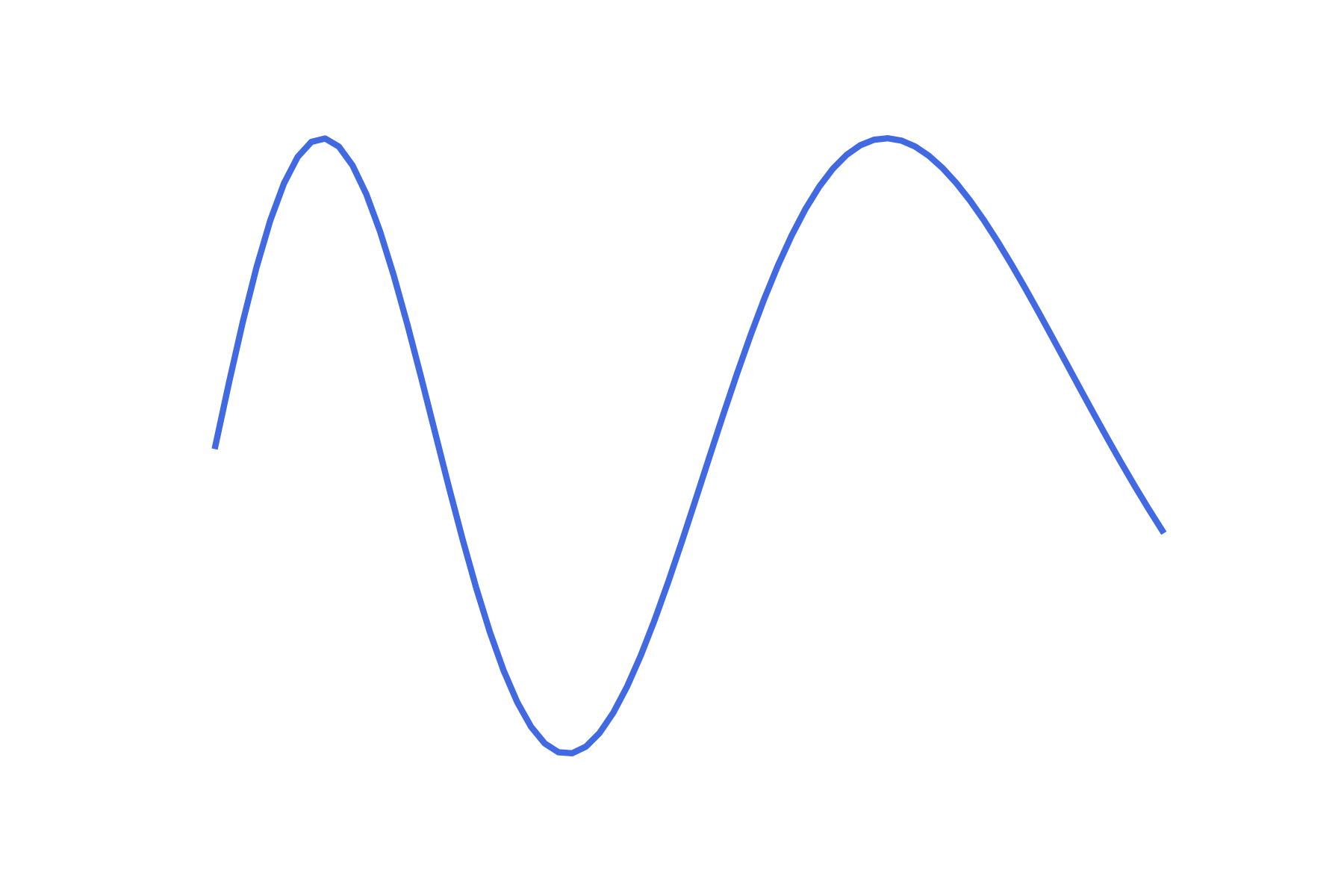}
\centering
(d) No volatility 
\end{minipage}
\\ % End of the first row of images and labels
\midrule
\\
\textbf{Anomalies and Outliers} \\
\begin{minipage}{4cm}
\includegraphics[scale = 0.3]{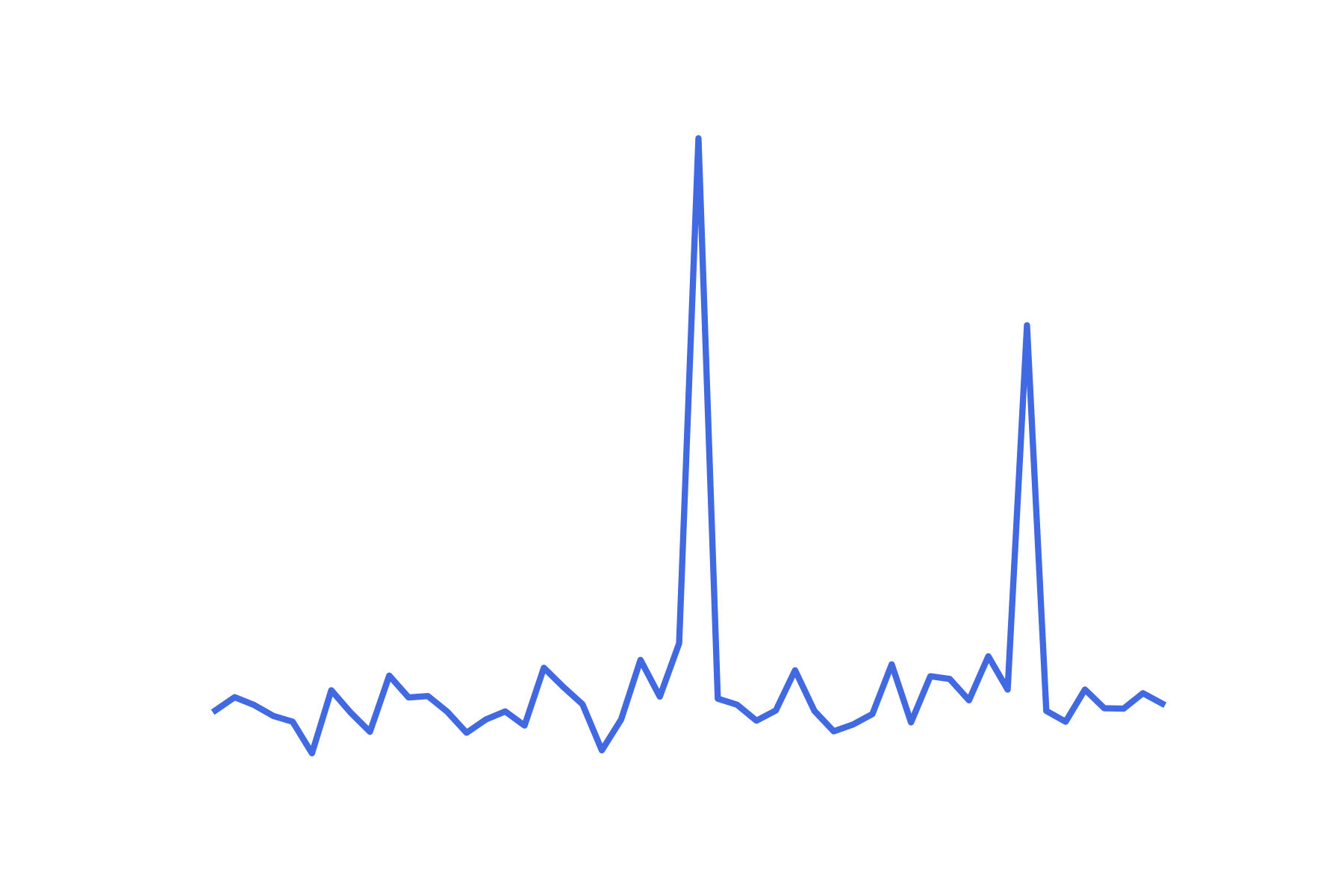}
\centering
(a) Double sudden spikes
\end{minipage}
\begin{minipage}{4cm}
\includegraphics[scale = 0.3]{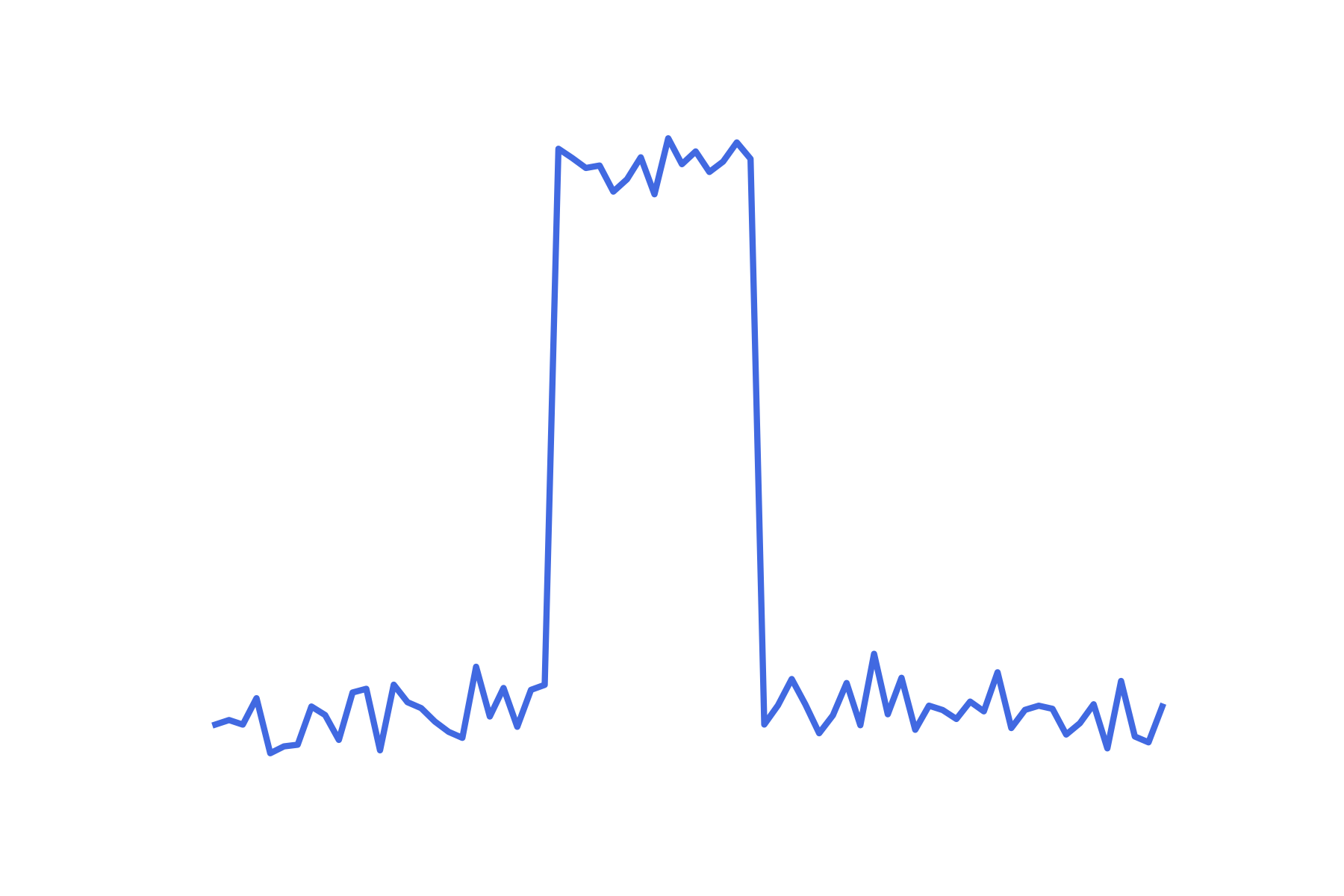}
\centering
(b) Step spike
\end{minipage}
\begin{minipage}{4cm}
\includegraphics[scale = 0.3]{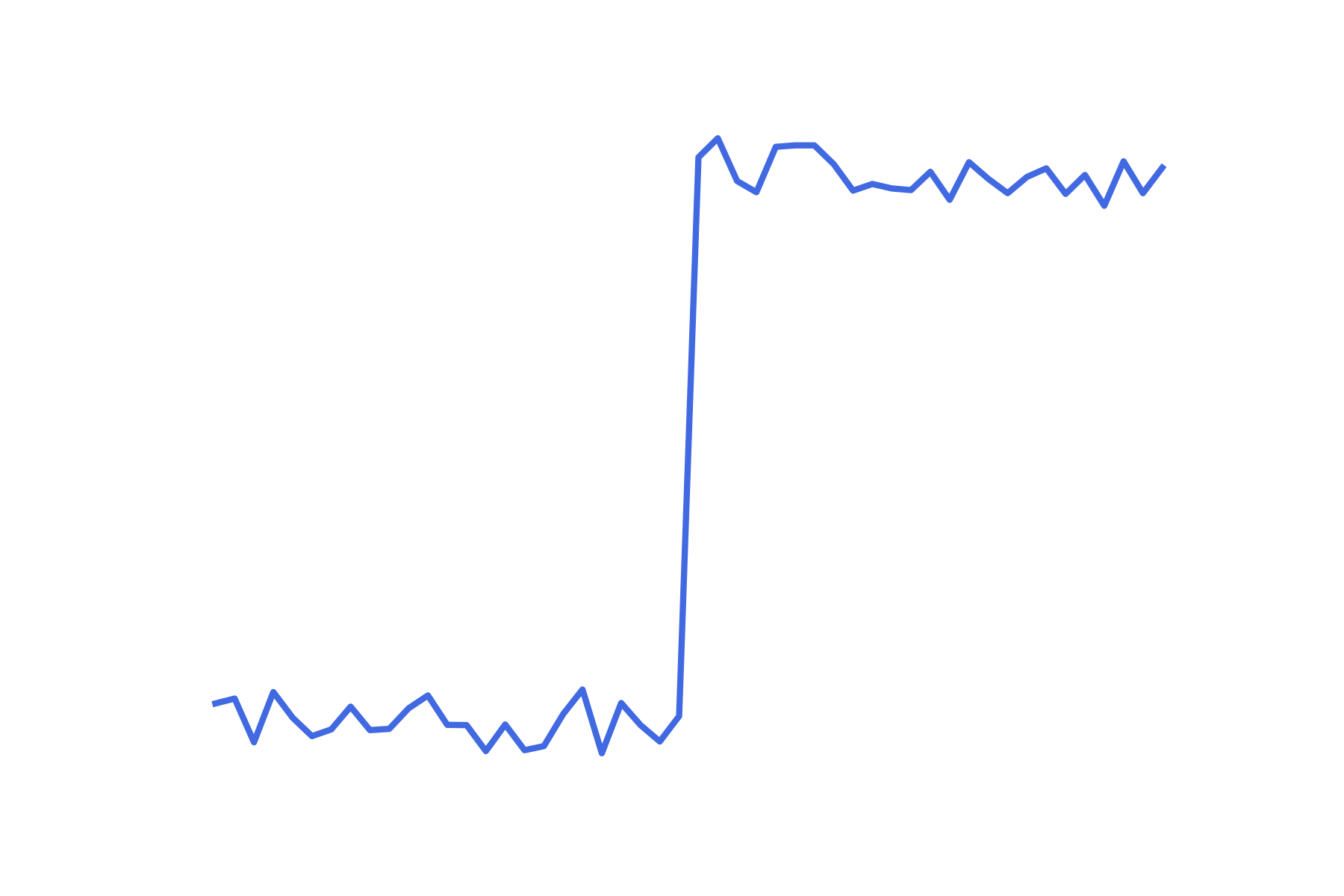}
\centering
(c) Level shift 
\end{minipage}
\begin{minipage}{4cm}
\includegraphics[scale = 0.3]{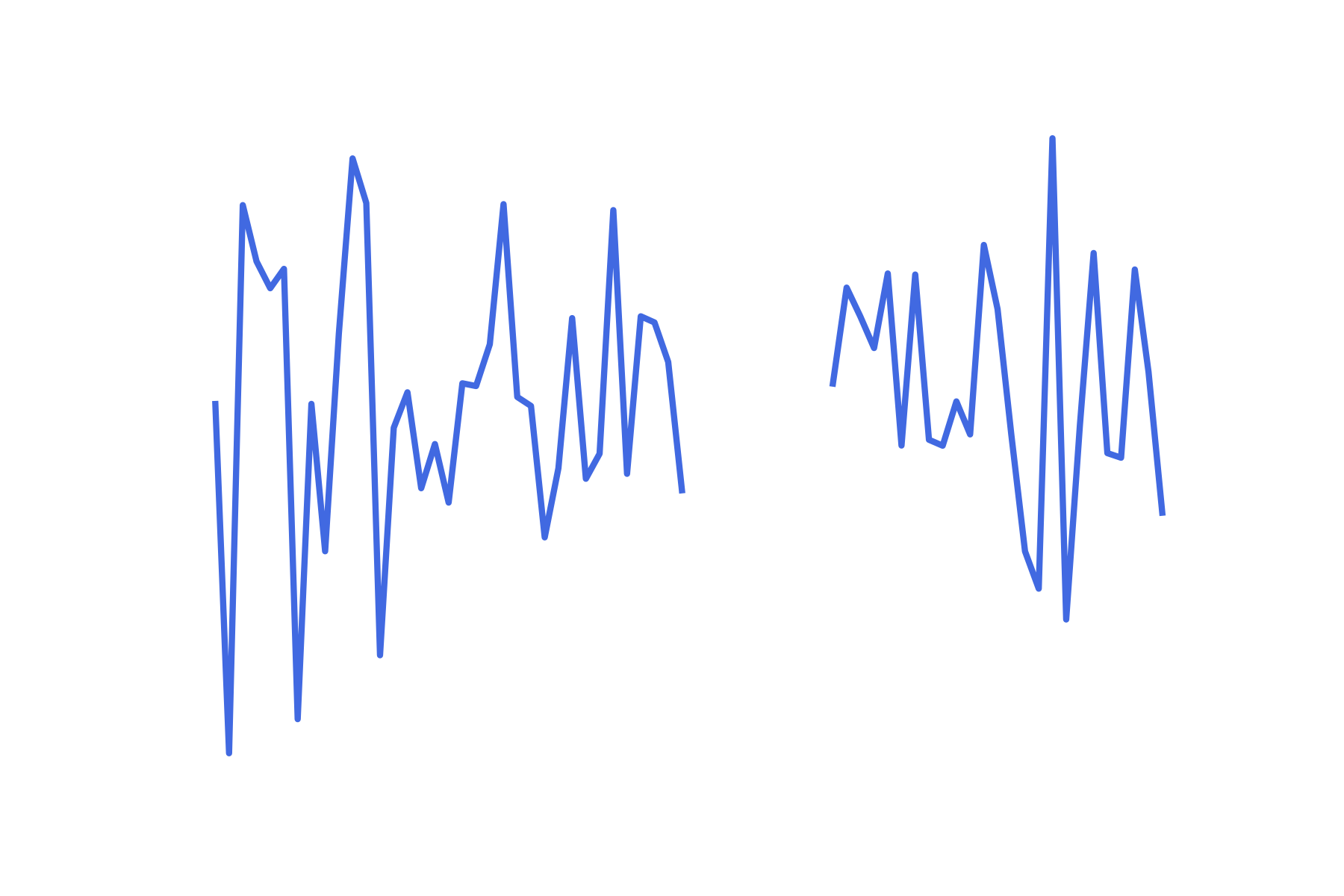}
\centering
(d) Temporal Disruption
\end{minipage}
\\ % End of the first row of images and labels
\midrule
\\
\newpage
\textbf{Structural breaks} \\
\begin{minipage}{4cm}
\includegraphics[scale = 0.3]{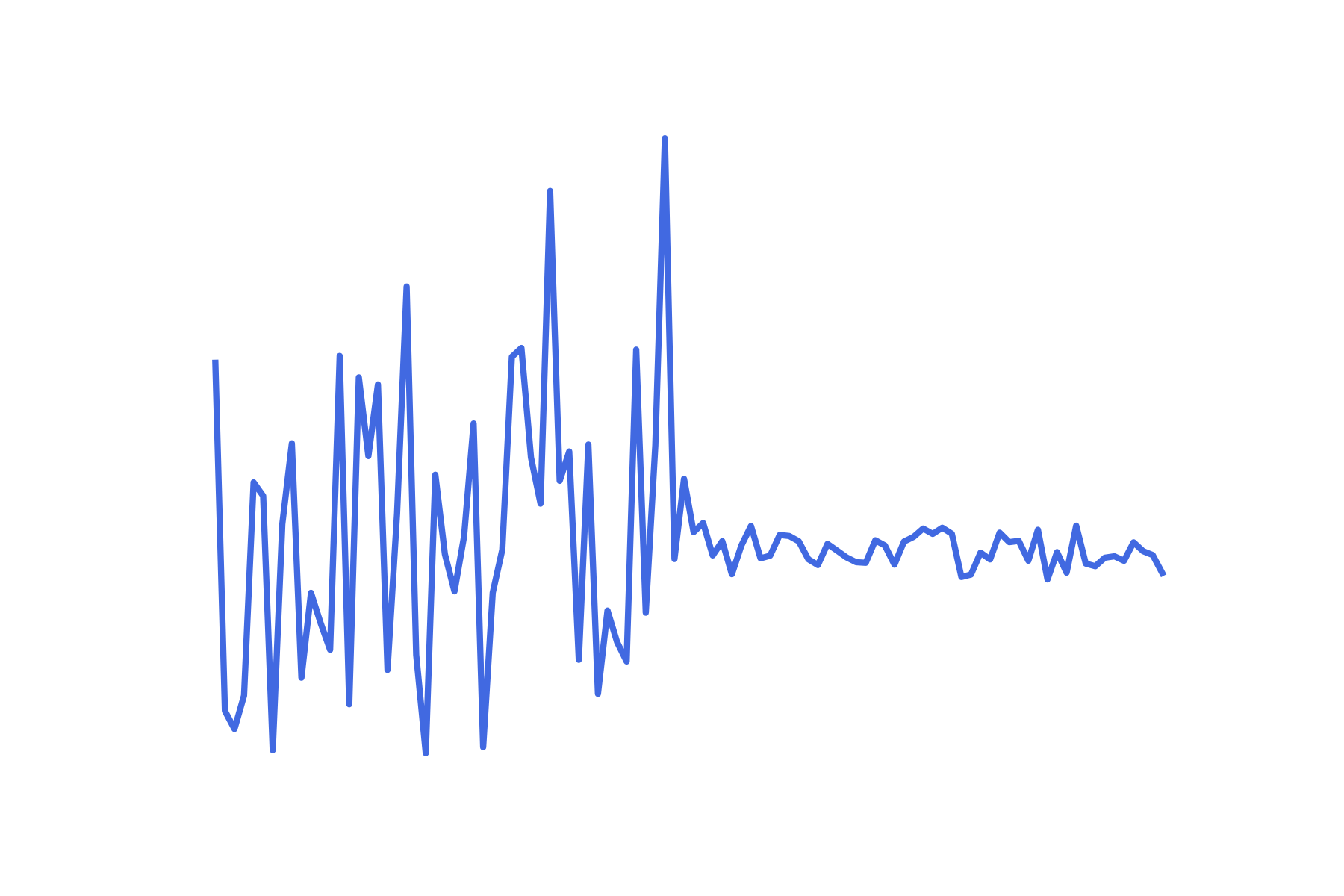}
\centering
(a) Parameter shift \\ (change in variance)
\end{minipage}
\begin{minipage}{4cm}
\includegraphics[scale = 0.3]{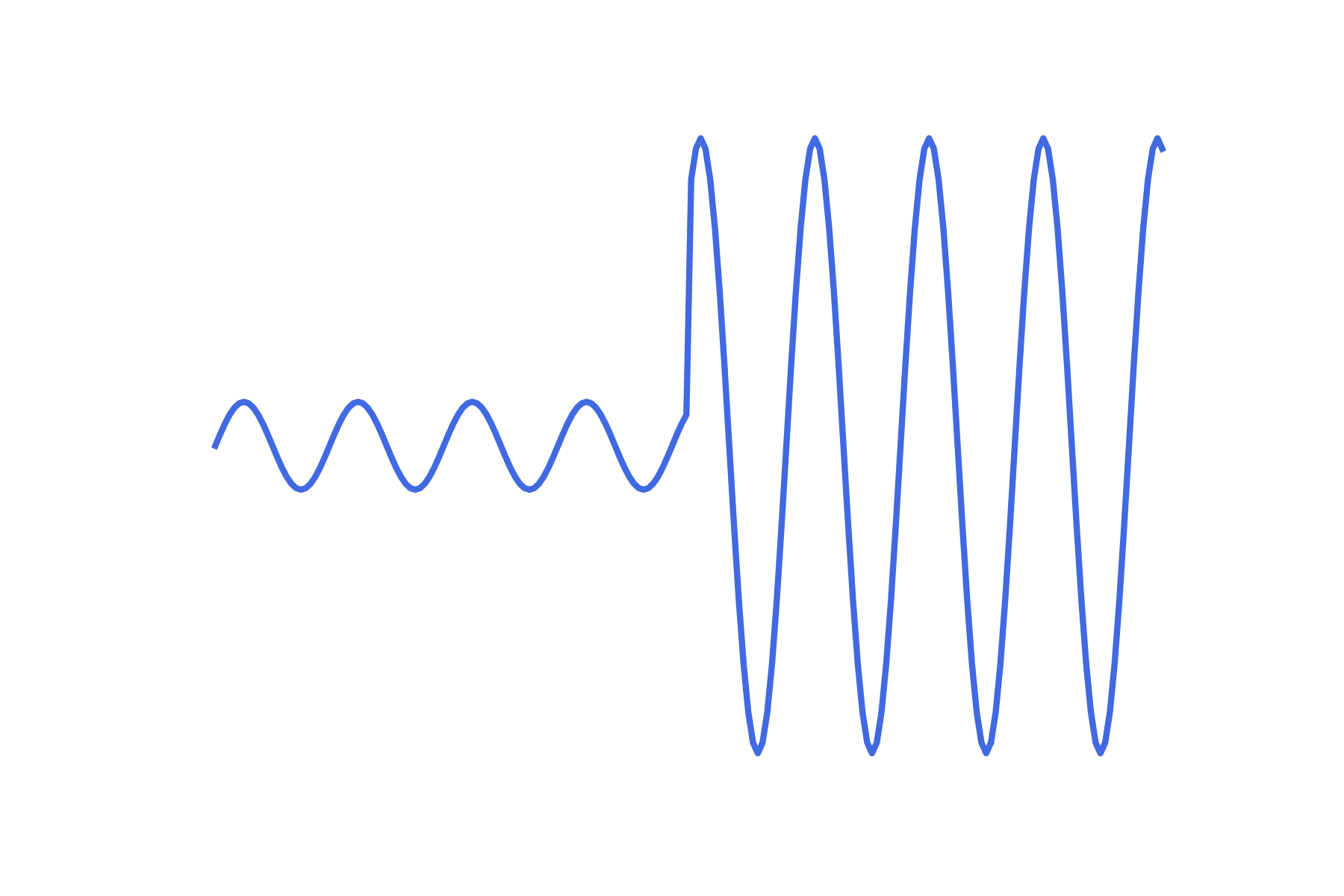}
\centering
(b) Parameter shift \\ (change in seasonality amplitude) 
\end{minipage}
\begin{minipage}{4cm}
\includegraphics[scale = 0.3]{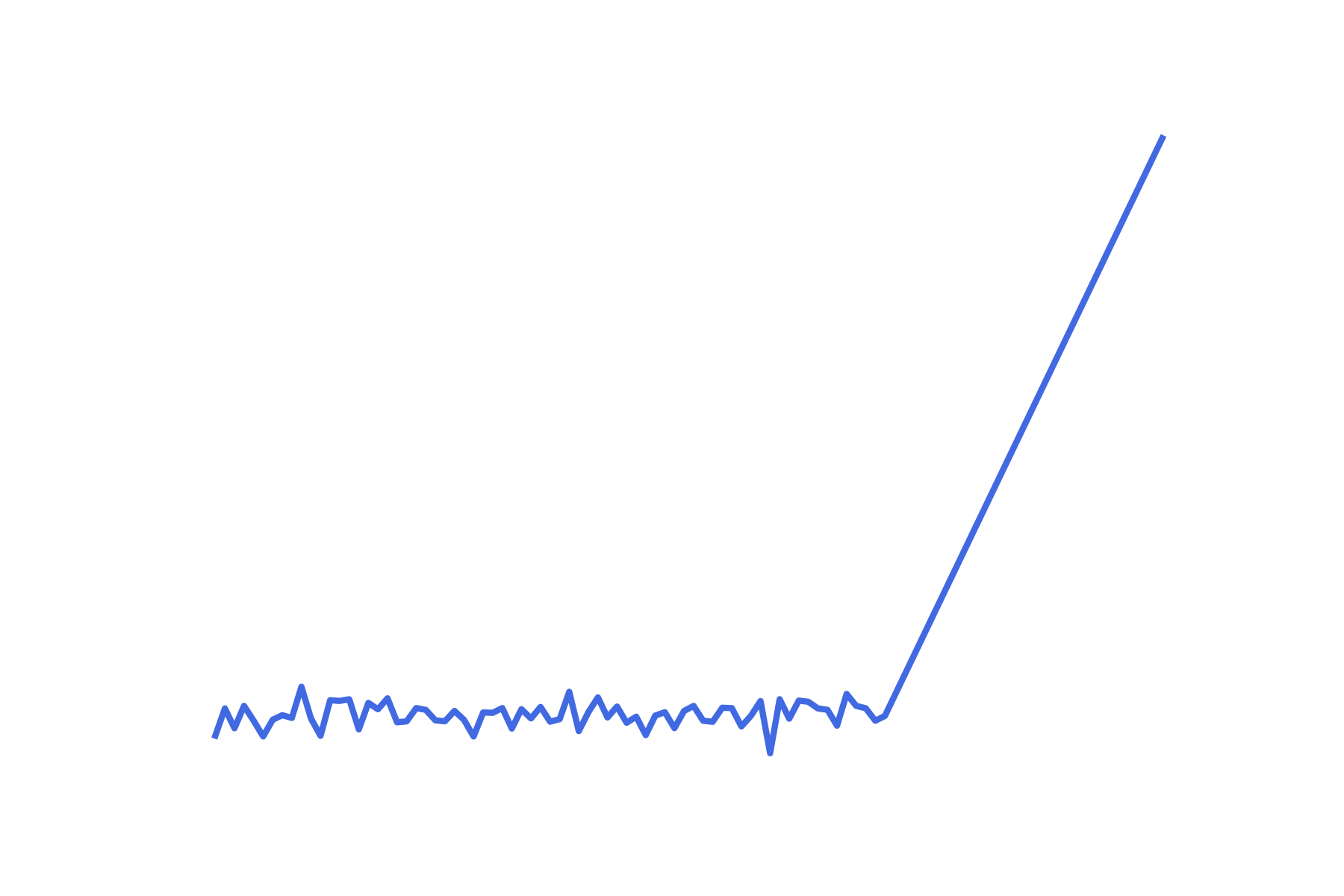}
\centering
(c) Regime shift \\ (noise trend change)
\end{minipage}
\begin{minipage}{4cm}
\includegraphics[scale = 0.3]{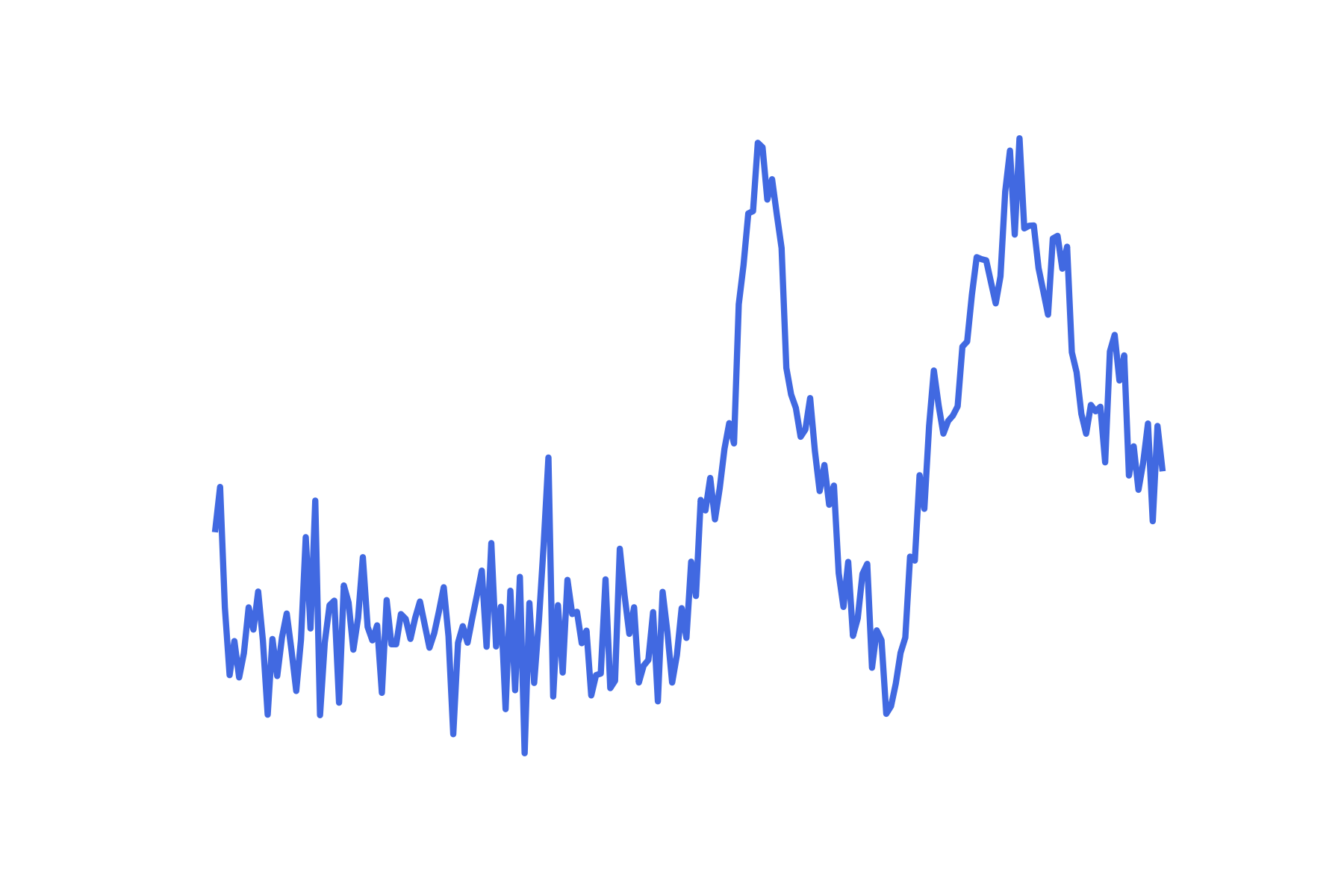}
\centering
(d) Regime shift \\ (stationarity change)
\end{minipage}
\\ % End of the first row of images and labels
\midrule
\\
\textbf{Fat Tails and Stationarity} \\ 
\begin{minipage}{4cm}
\includegraphics[scale = 0.3]{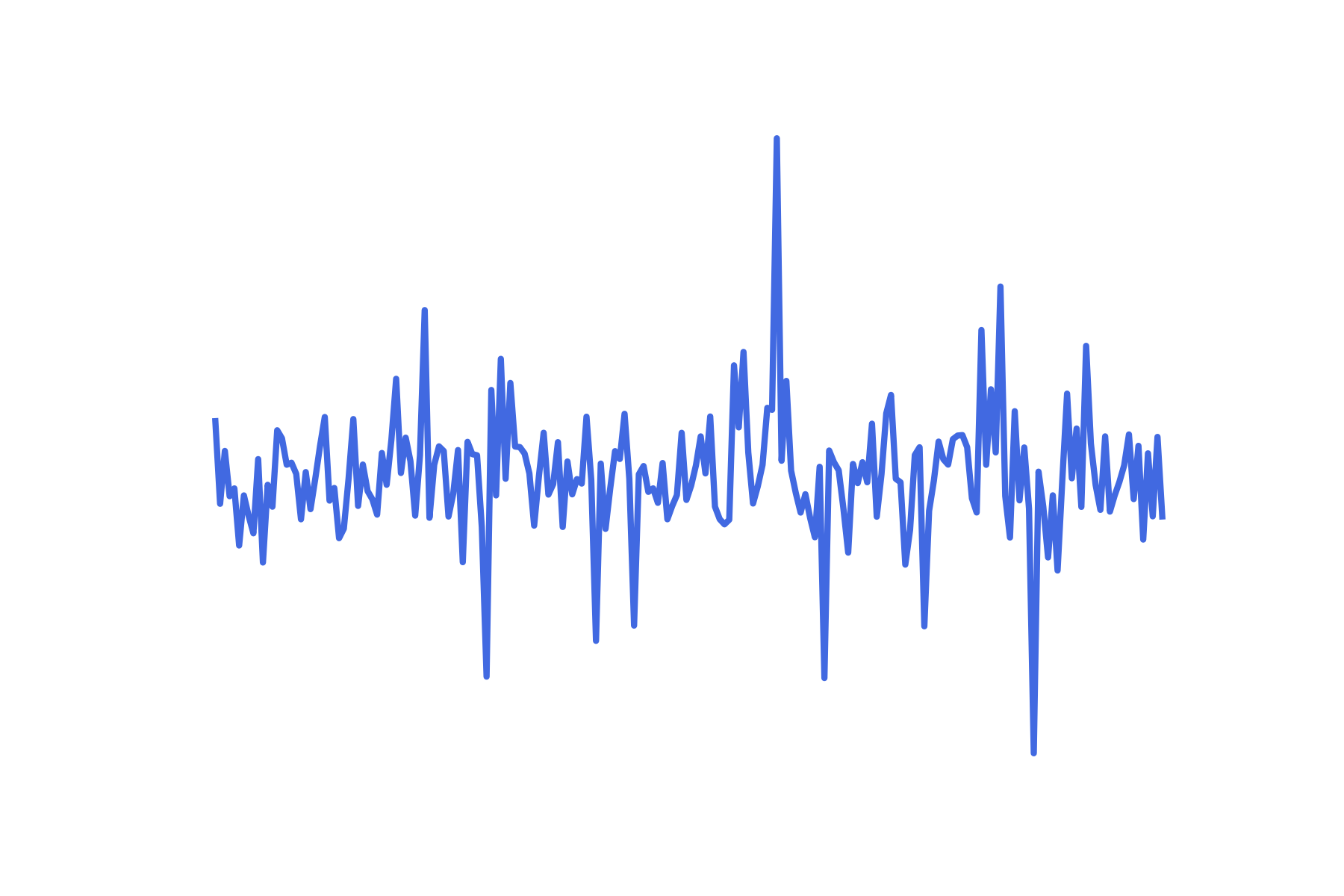}
\centering
(a) Fat tailed
\end{minipage}
\begin{minipage}{4cm}
\includegraphics[scale = 0.3]{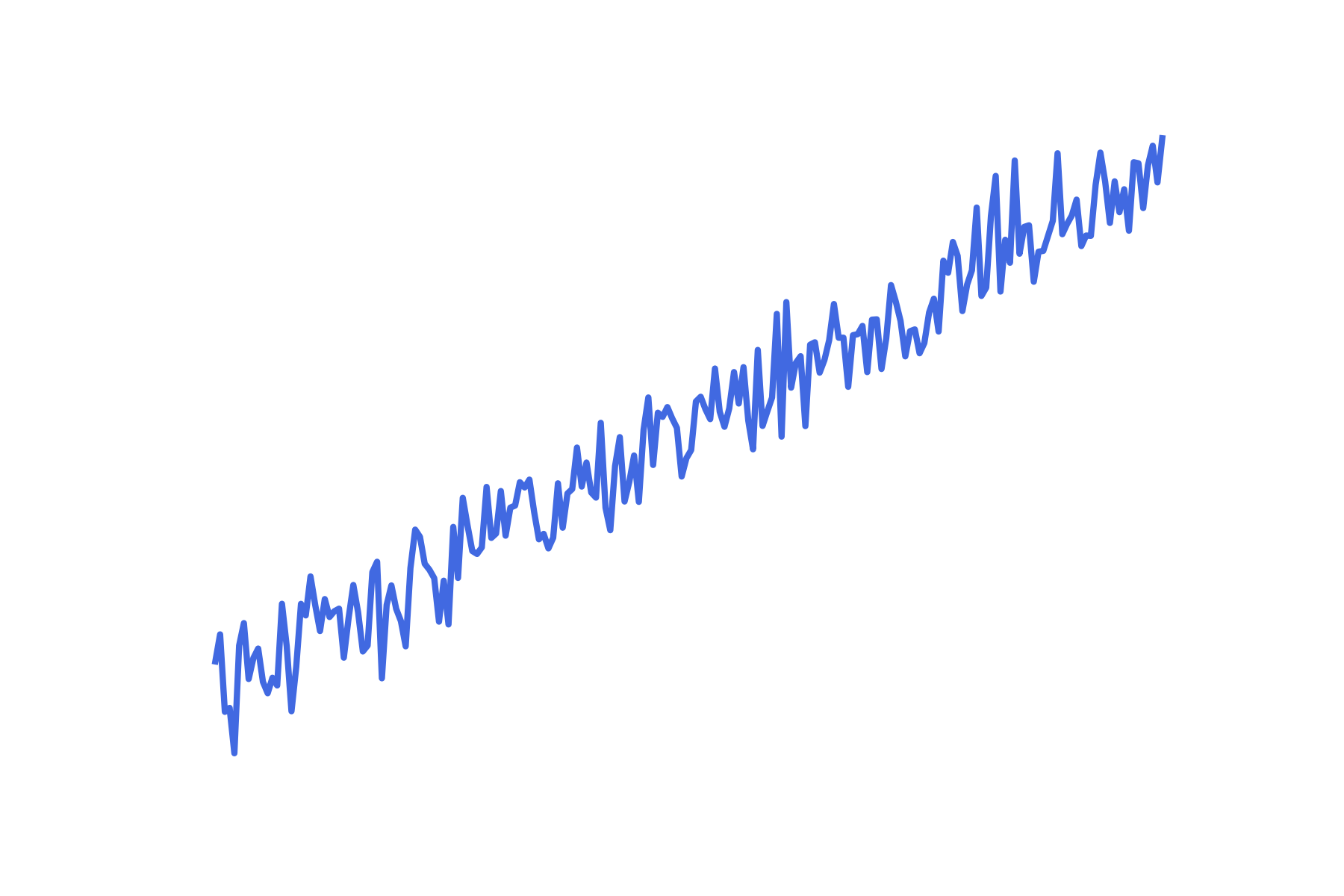}
\centering
(b) Non-stationary \\ (trend)
\end{minipage}
\begin{minipage}{4cm}
\includegraphics[scale = 0.3]{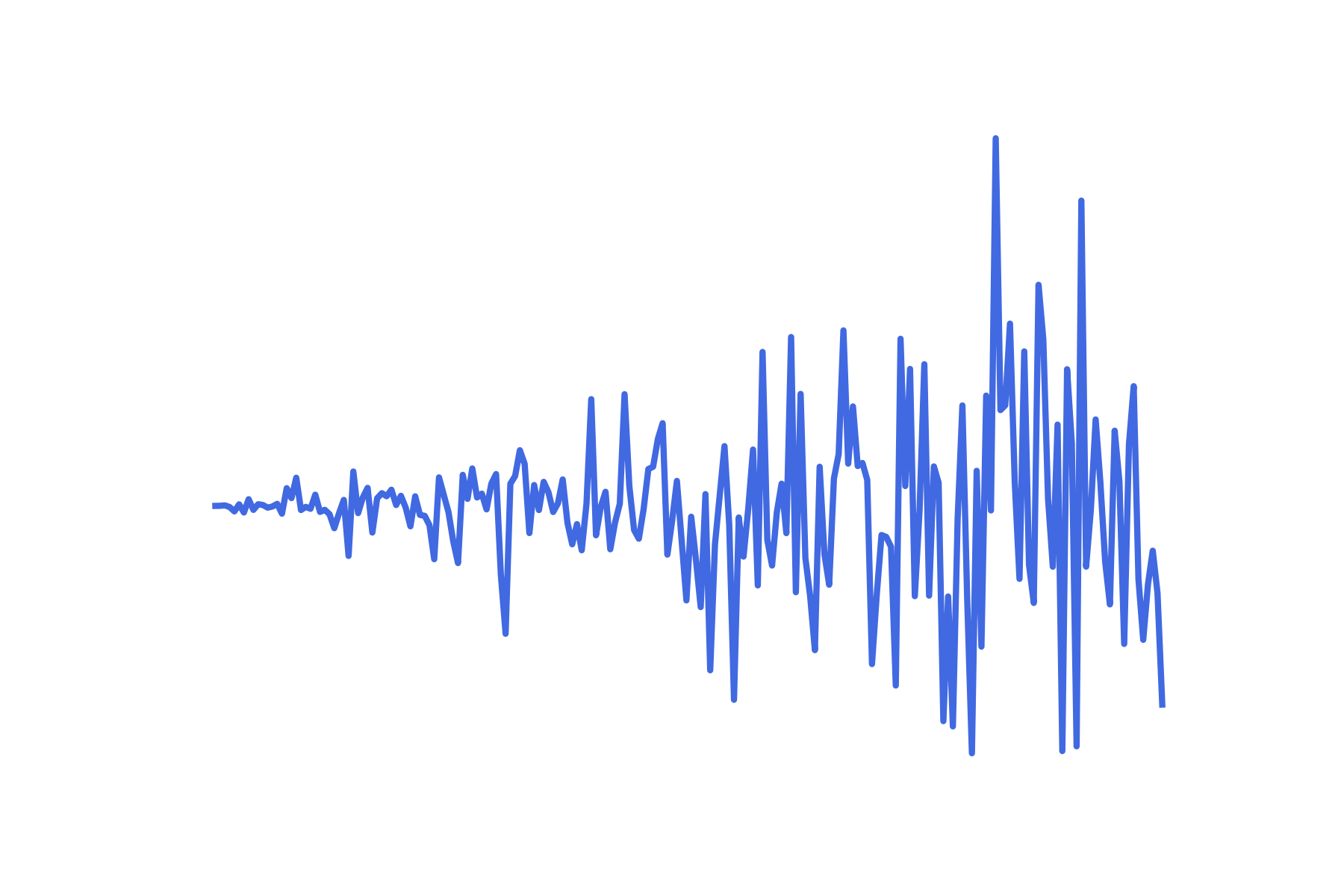}
\centering
(c) Non-stationary \\ (changing variance over time)
\end{minipage}
\begin{minipage}{4cm}
\includegraphics[scale = 0.3]{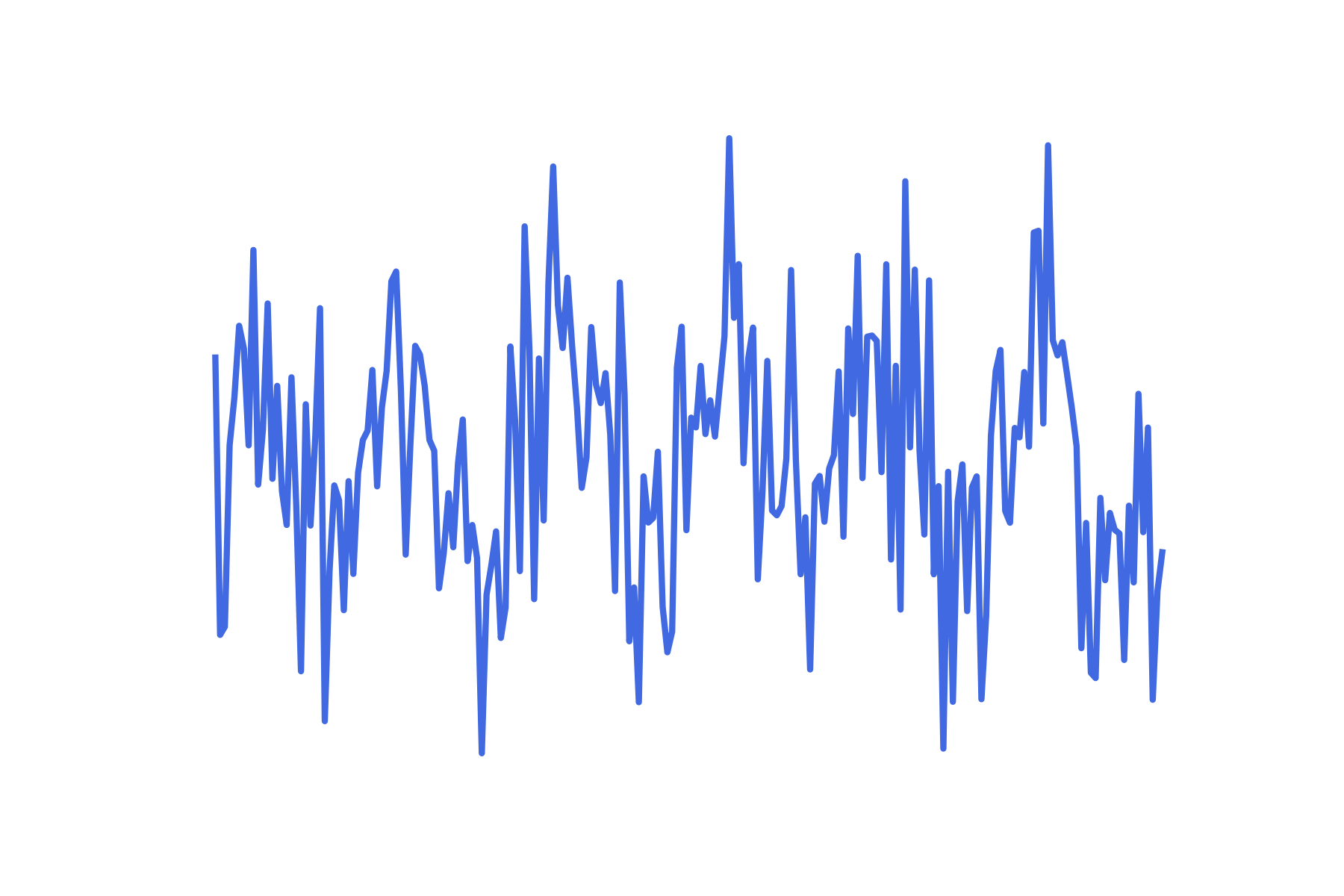}
\centering
(d) Non-stationary \\ (seasonality)
\end{minipage}
% \hline
\\
% \bottomrule
% \end{tabular}
\caption{Examples of the generated univariate time series. The x- and y-axis are intentionally omitted to focus exclusively on the shape and characteristics of the time series.}
\label{fig:univariate_data}
\end{longtable}

%%%% Multivariate Time Series Examples 
\begin{longtable}{c}
% \centering
% \begin{tabular}{c} % Two columns taking up equal space
\toprule
% \\
 \textbf{Correlation}
 \\
\begin{minipage}{5cm}
\includegraphics[scale = 0.4]{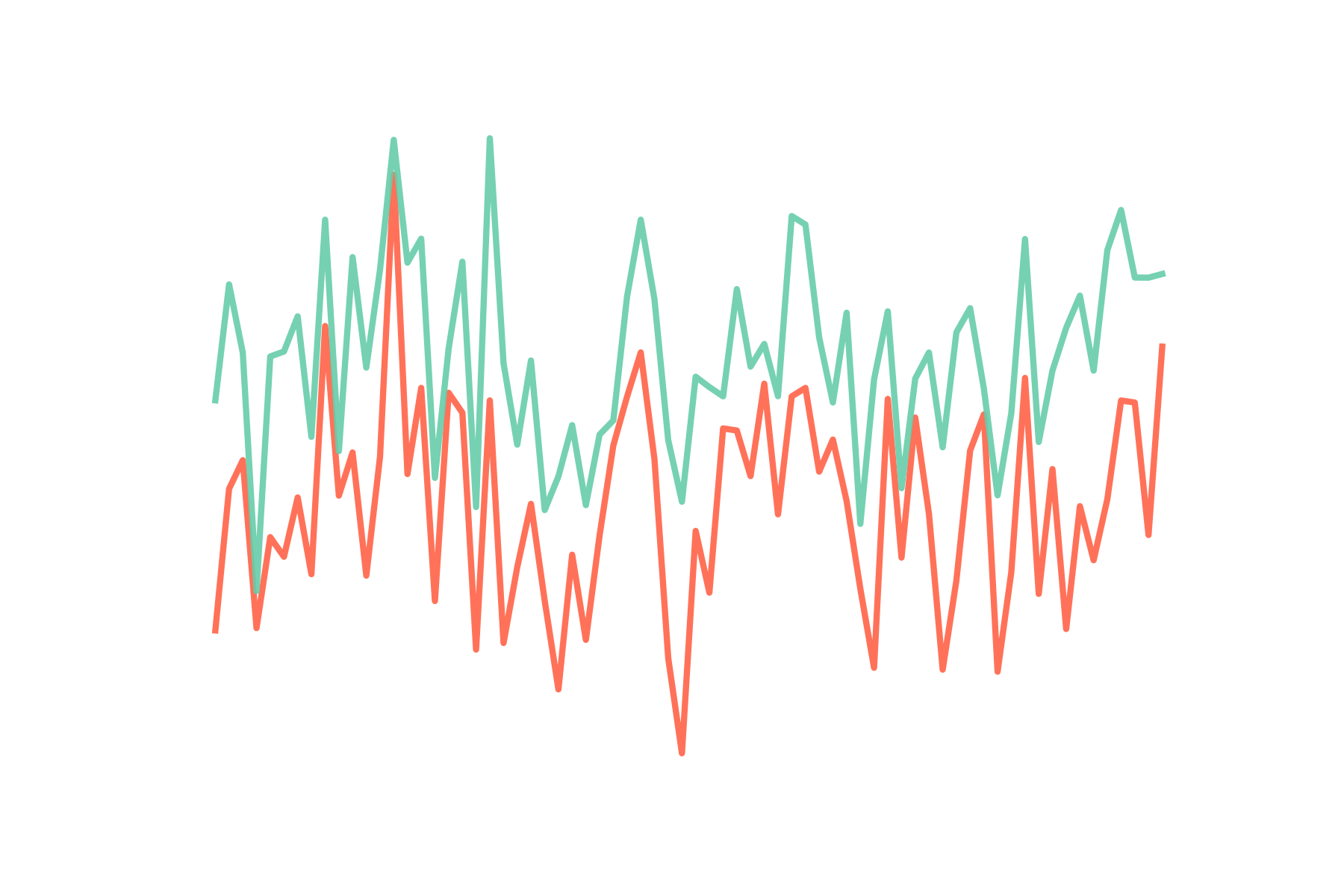}
\centering
(a) Positive correlation
\end{minipage}
\begin{minipage}{5cm}
\includegraphics[scale = 0.4]{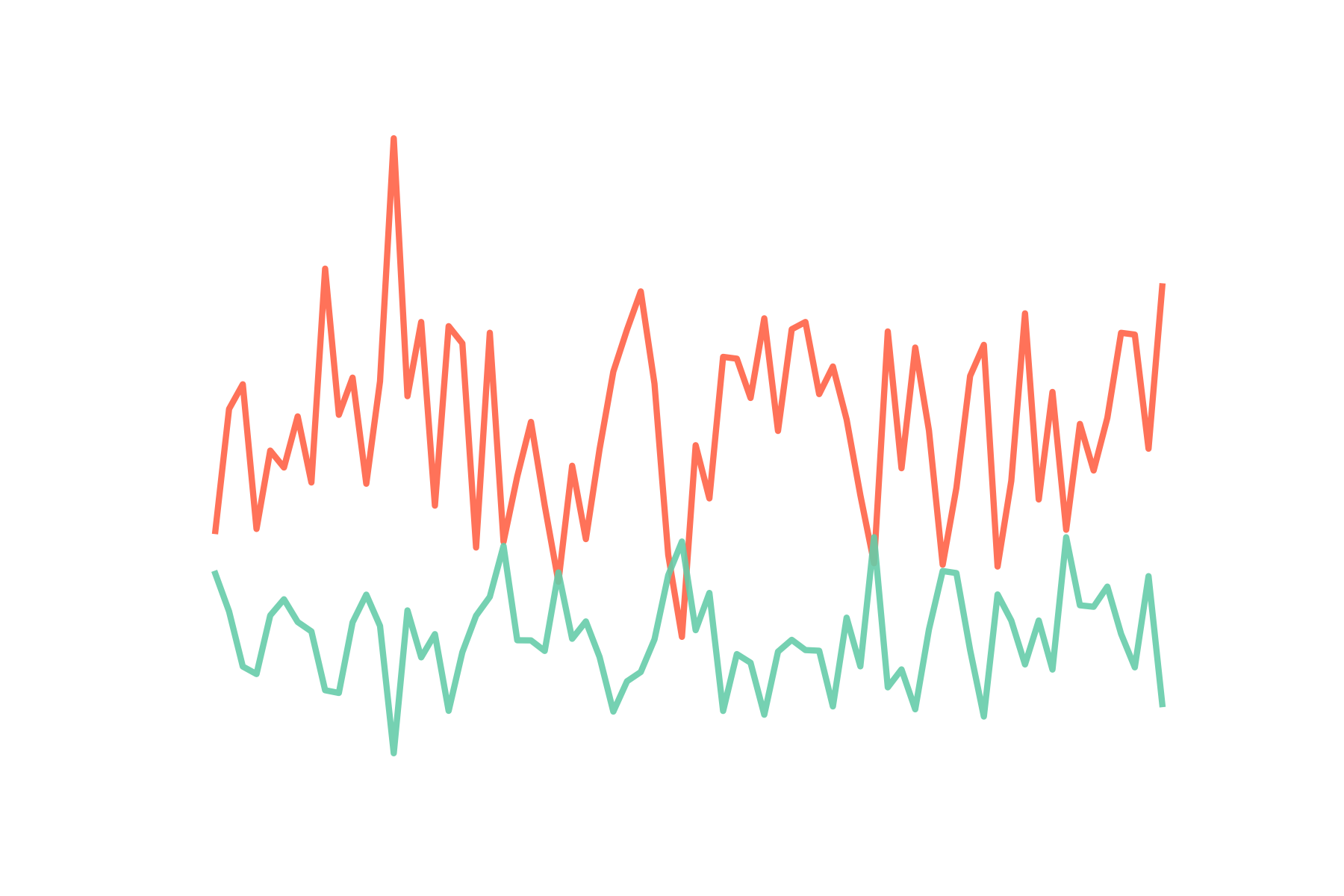}
\centering
(b) Negative correlation
\end{minipage}
\begin{minipage}{5cm}
\includegraphics[scale = 0.4]{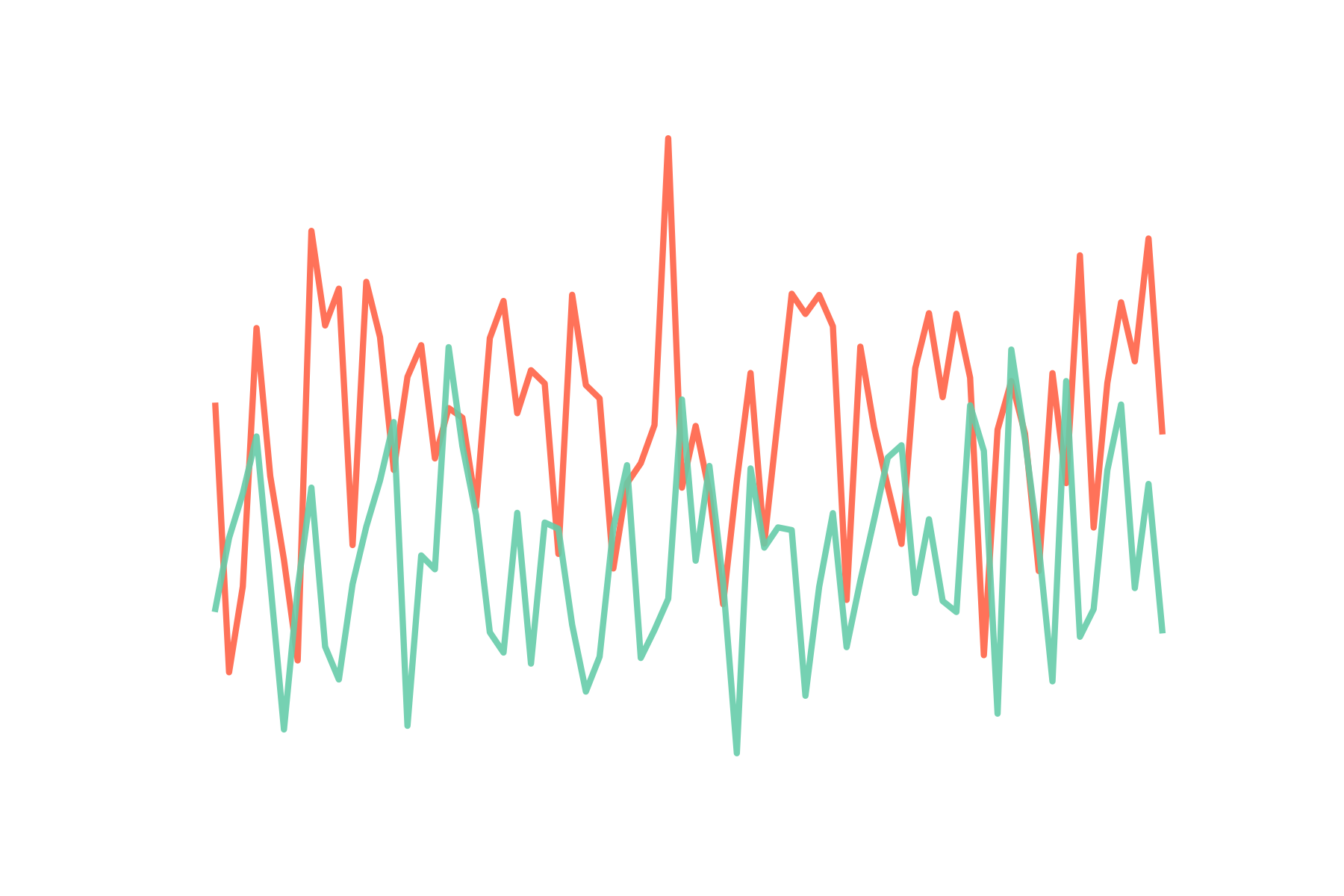}
\centering
(c) No correlation
\end{minipage}
\\ % End of the first row of images and labels
\midrule
\\
 \textbf{Cross-correlation} 
 \\
\begin{minipage}{5cm}
\includegraphics[scale = 0.4]{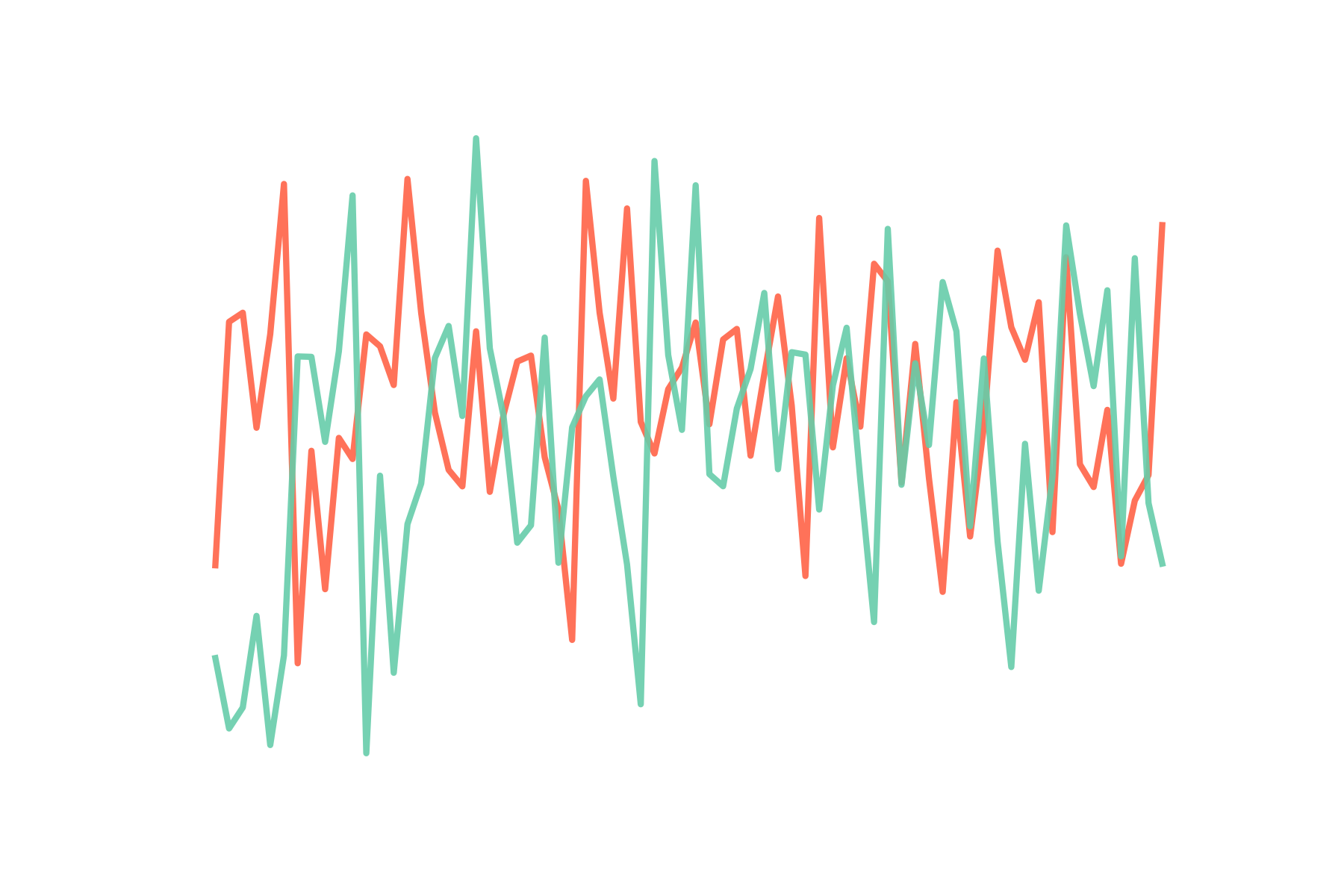}
\\
\centering
(a) Lagged positive correlation
\end{minipage}
\begin{minipage}{5cm}
\includegraphics[scale = 0.4]{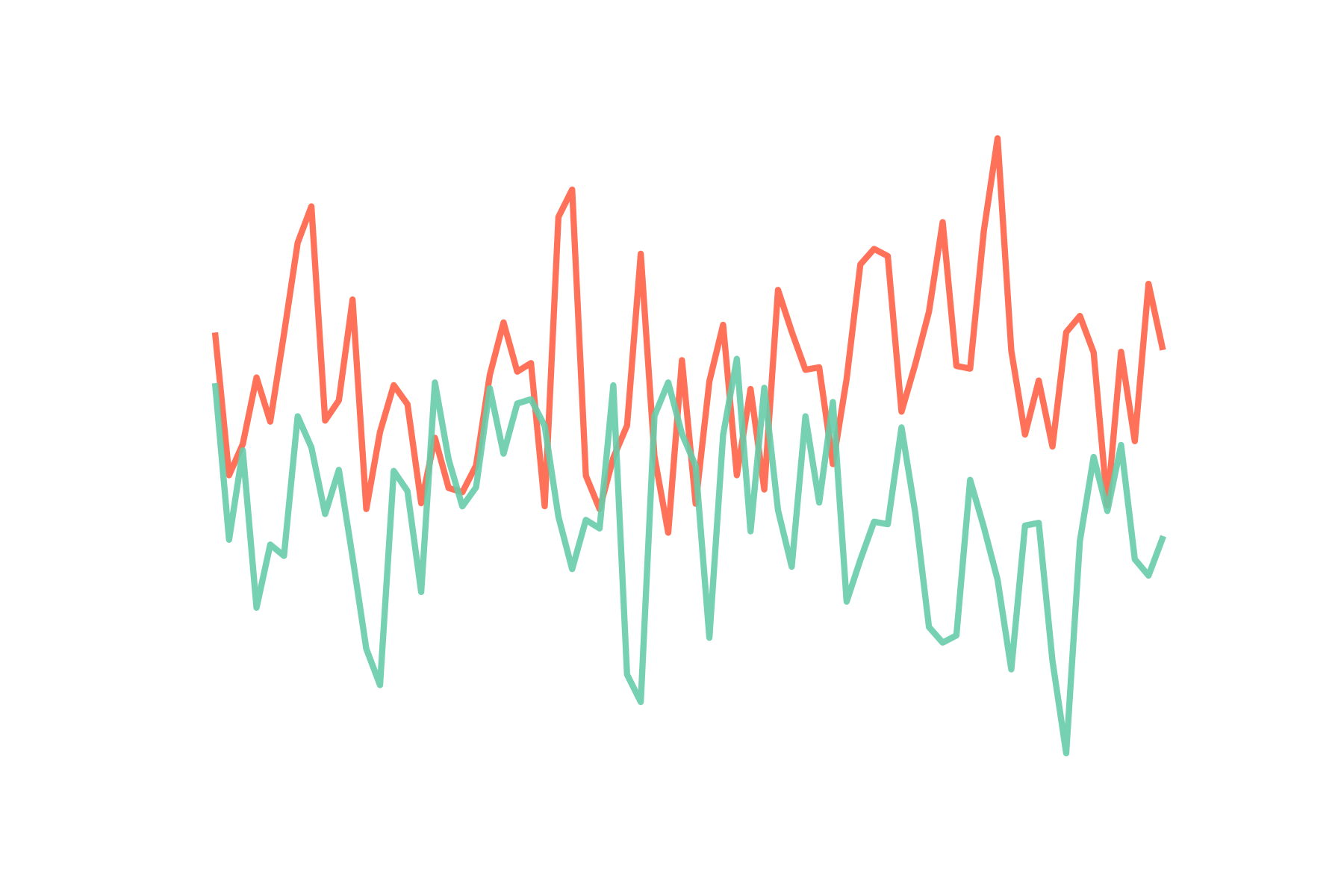}
\\
\centering
(b) Lagged negative correlation
\end{minipage}
\\ % End of the first row of images and labels
% \newline
% Text below last image in Block 1 \\
% Text below image in Block 2 \\
\midrule
\\
\textbf{Dynamic conditional correlation} 
\\ 
\begin{minipage}{5cm}
\includegraphics[scale = 0.4]{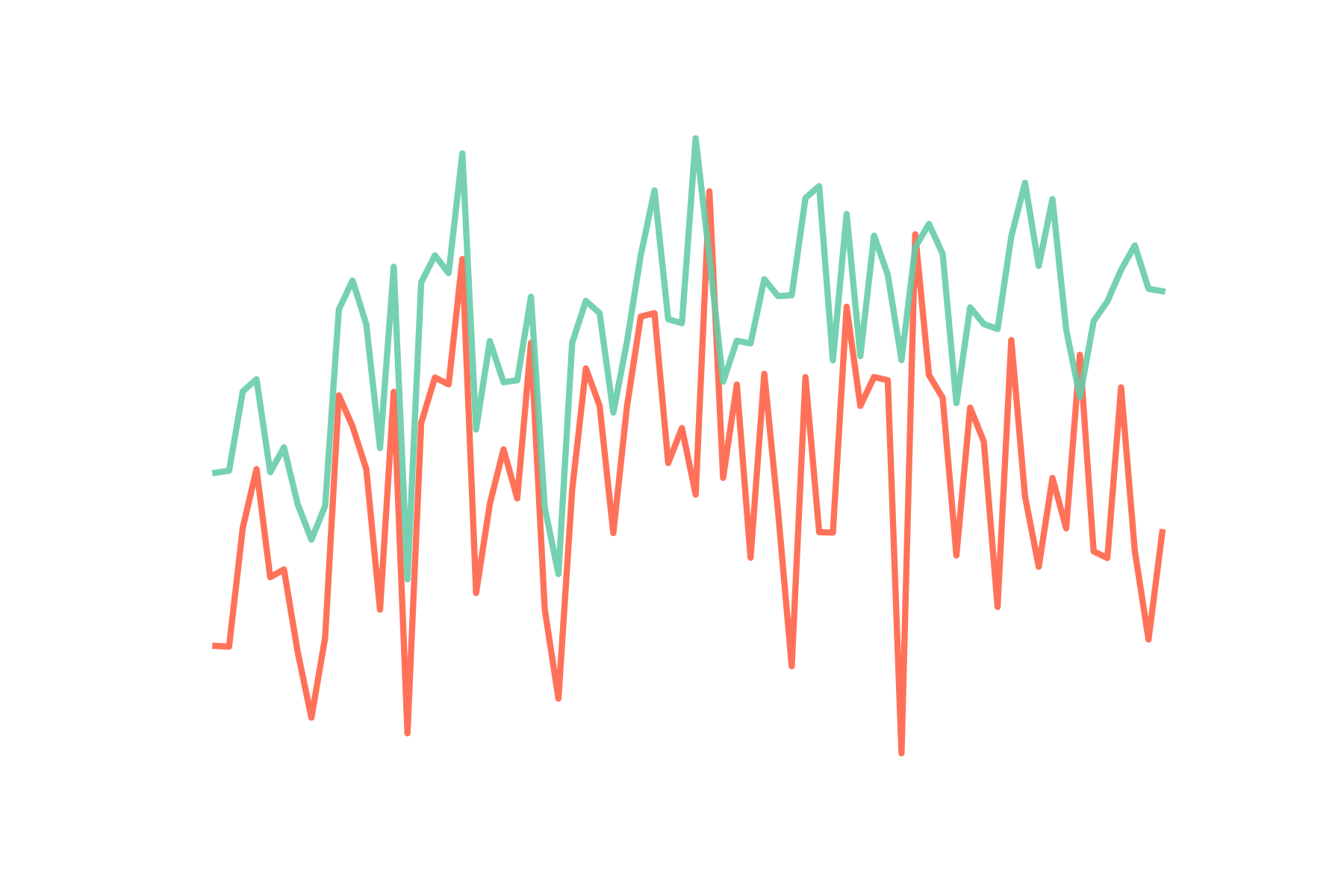}
\\
\centering
(a) Positive correlation \\ (first half)
\end{minipage}
\begin{minipage}{5cm}
\includegraphics[scale = 0.4]{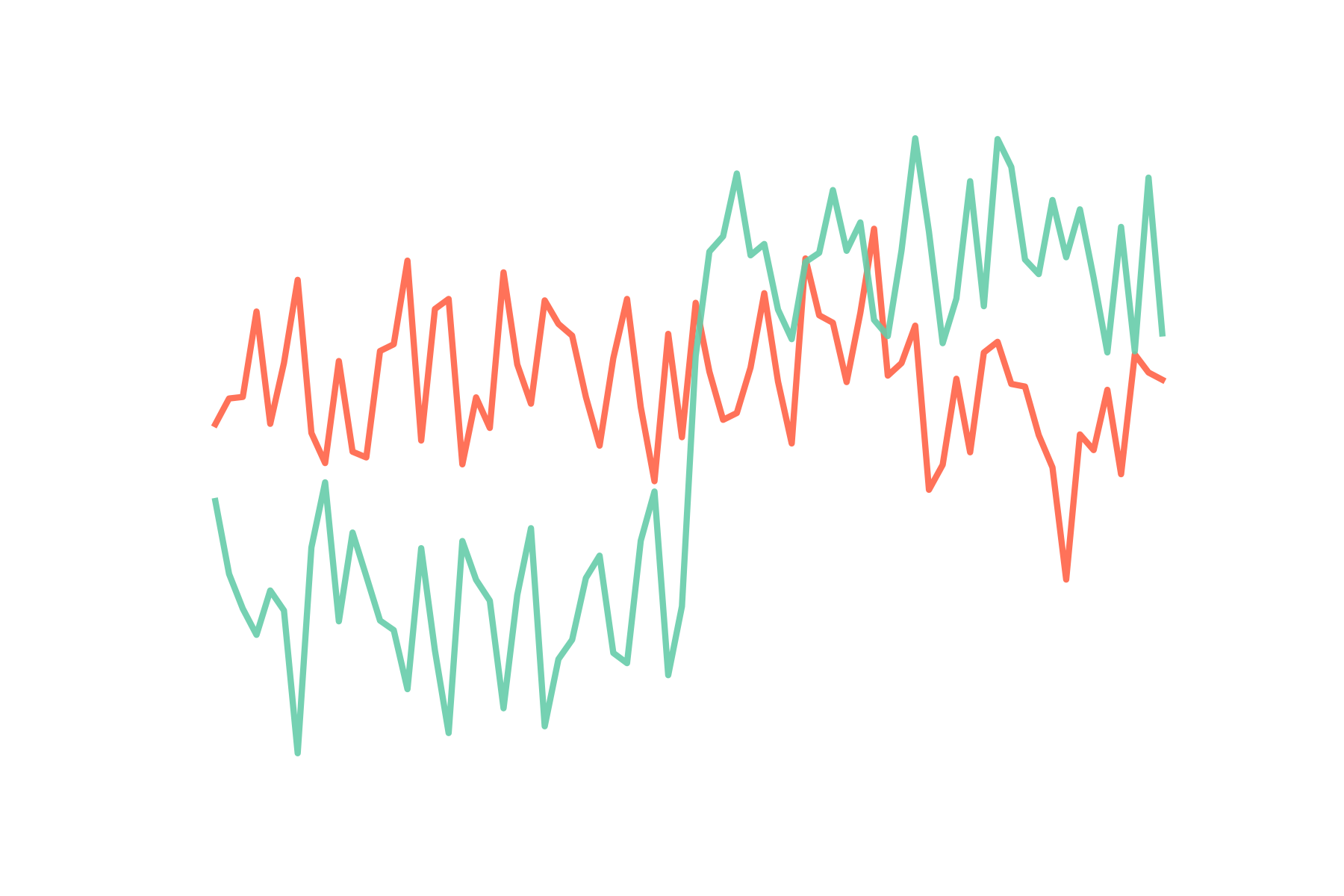}
\\
\centering
(b) Negative correlation \\ (first half)
\end{minipage}
\begin{minipage}{5cm}
\includegraphics[scale = 0.4]{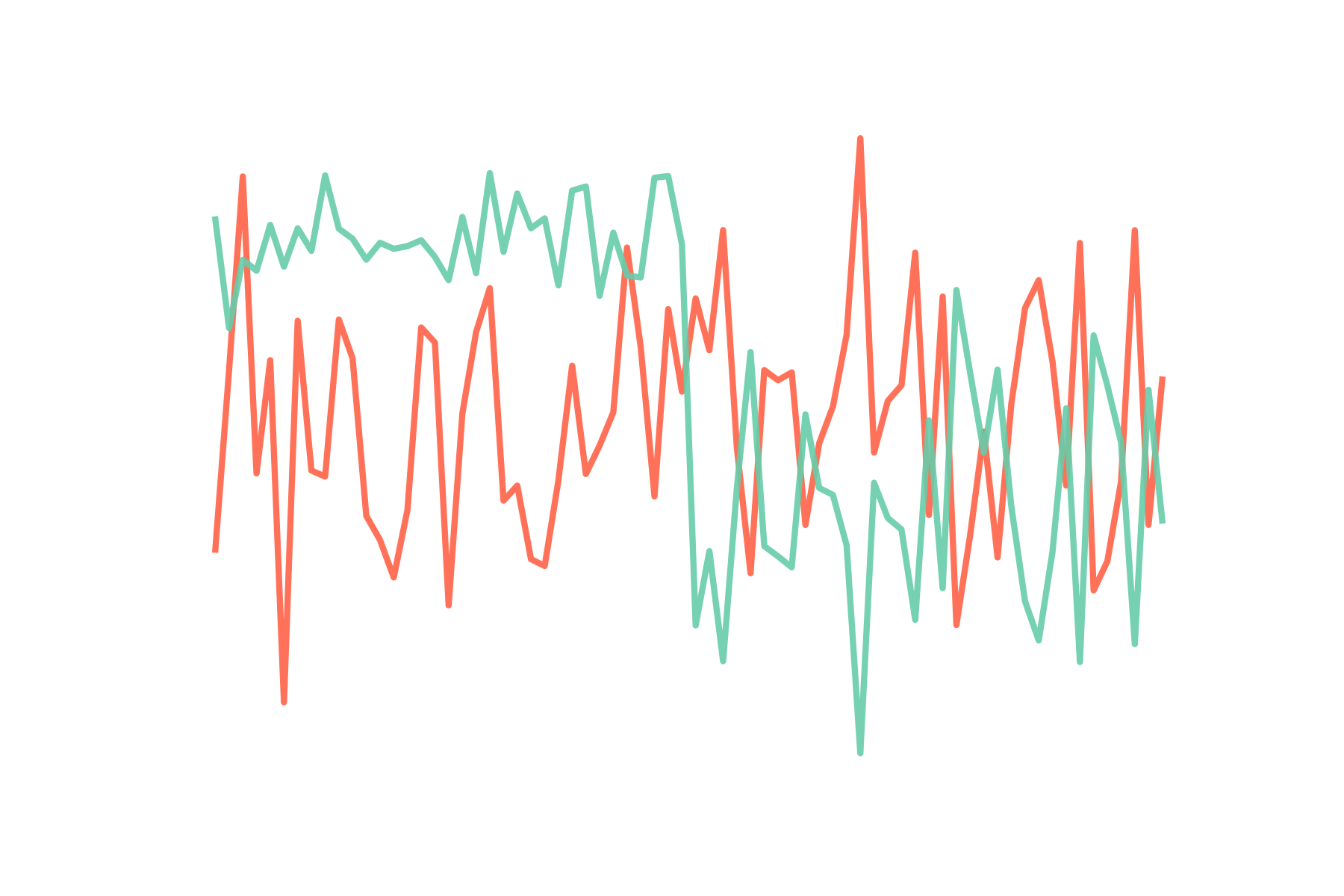}
\\
\centering
(d) Negative correlation \\ (second half)
\end{minipage}
\\
\bottomrule
% \end{tabular}
\caption{Examples of the generated multivariate
time series. The x- and y-axis are intentionally omitted to focus exclusively on the shape and characteristics of the time series.}
\label{fig:multivariate_data}
\end{longtable}

%%%%%%%%%%%%%%%%%%%%%%%%%%%%%%%%%%%%%%%%%%%%%%%%

\section{Additional datasets}

\textbf{Brownian Data}: We generate a synthetic time series dataset exhibiting brownian motion. The data consists of 400 samples where each time series has a length of 175. We control for the quadrant in the which the maximum and minimum values appear using rejection sampling i.e. there are 50 samples for which the maximum value in the time series occurs in the first quadrant, 50 samples for which the maximum value appears in the second quadrant, and so on, upto the fourth quadrant. In a similar manner we control for presence of the minimum value in each quadrant. 
\newline
\textbf{Outlier Data}: We generate a synthetic time series dataset where each time series contains a single outlier which is the either the minimum or maximum values in the time series. The data consists of 400 samples where each time series has a length of 175. We control for the quadrant in the which the maximum and minimum (outlier) values appear using rejection sampling i.e. there are 50 samples for which the maximum value in the time series occurs in the first quadrant, 50 samples for which the maximum value appears in the second quadrant, and so on, upto the fourth quadrant. In a similar manner we control for presence of the minimum value in each quadrant. 
\newline
\textbf{Monotone Data}: We generate a synthetic time series dataset where each time series is monotonically increasing or decreasing. The data consists of 400 samples (200 each for increasing/decreasing) where each time series has a length of 175.  
\newline
\textbf{Monotone (with Noise) Data}: We generate a synthetic time series dataset where each time series is increasing or decreasing. The data consists of 400 samples (200 each for increasing/decreasing) where each time series has a length of 175. Note that dataset is different from the Monotone data as the time series samples are not strictly increasing/decreasing.

\clearpage
\section{Additional results}
\label{sec:app:cm}

\subsection{Trend}

\begin{figure}[h!]
    \centering
    \includegraphics[width=0.20\textwidth]{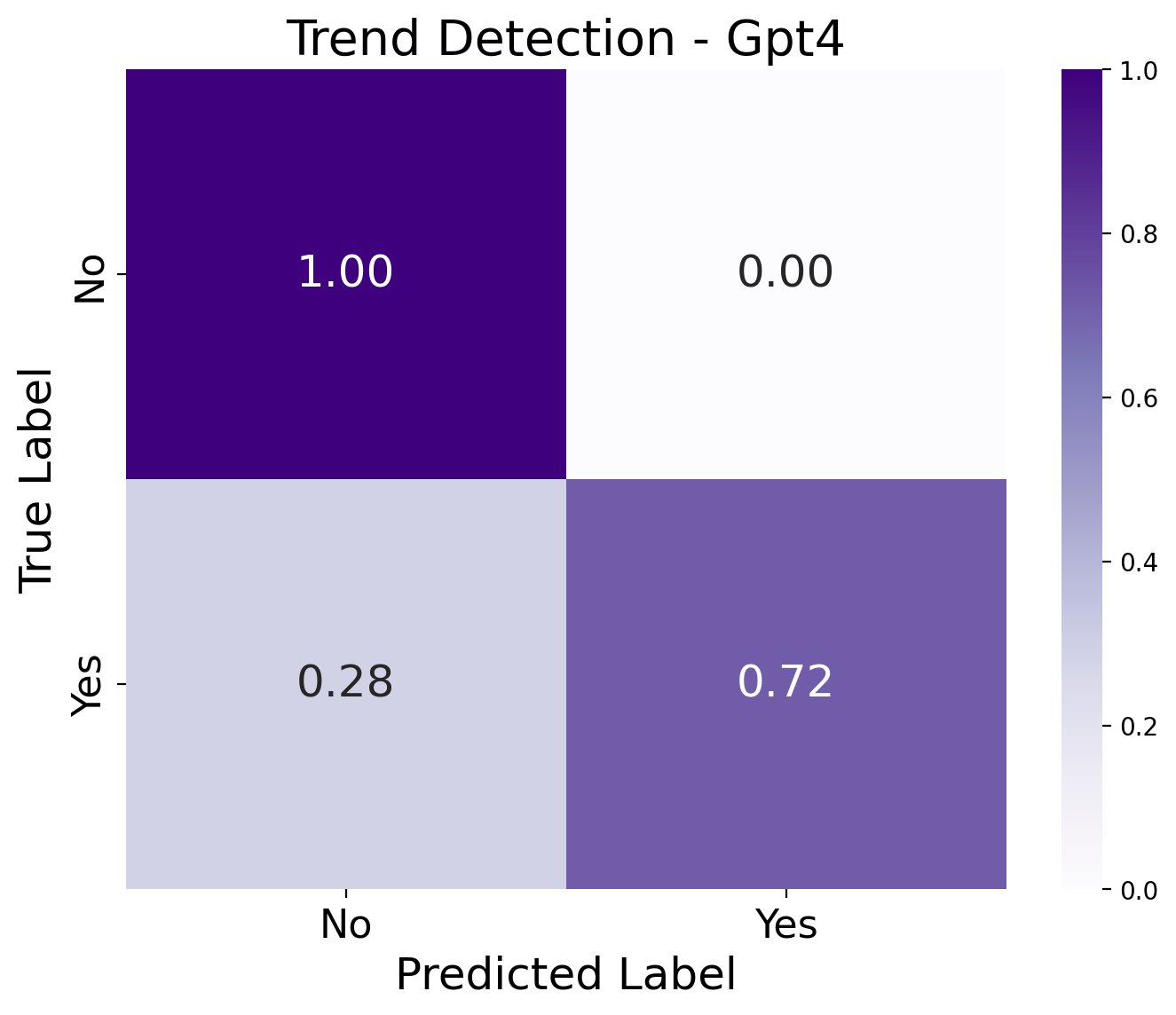}~
    \includegraphics[width=0.20\textwidth]{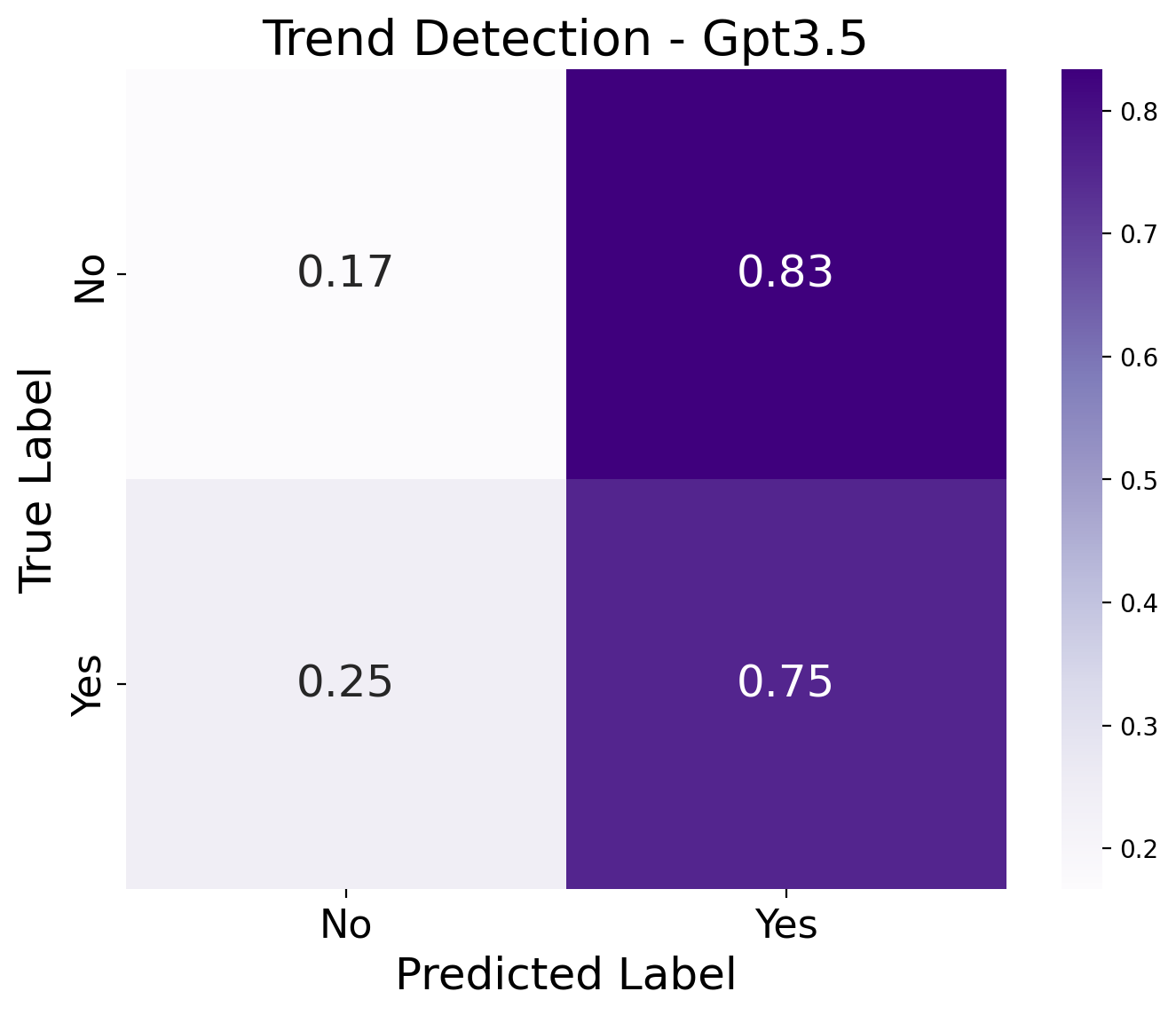}~
    \includegraphics[width=0.20\textwidth]{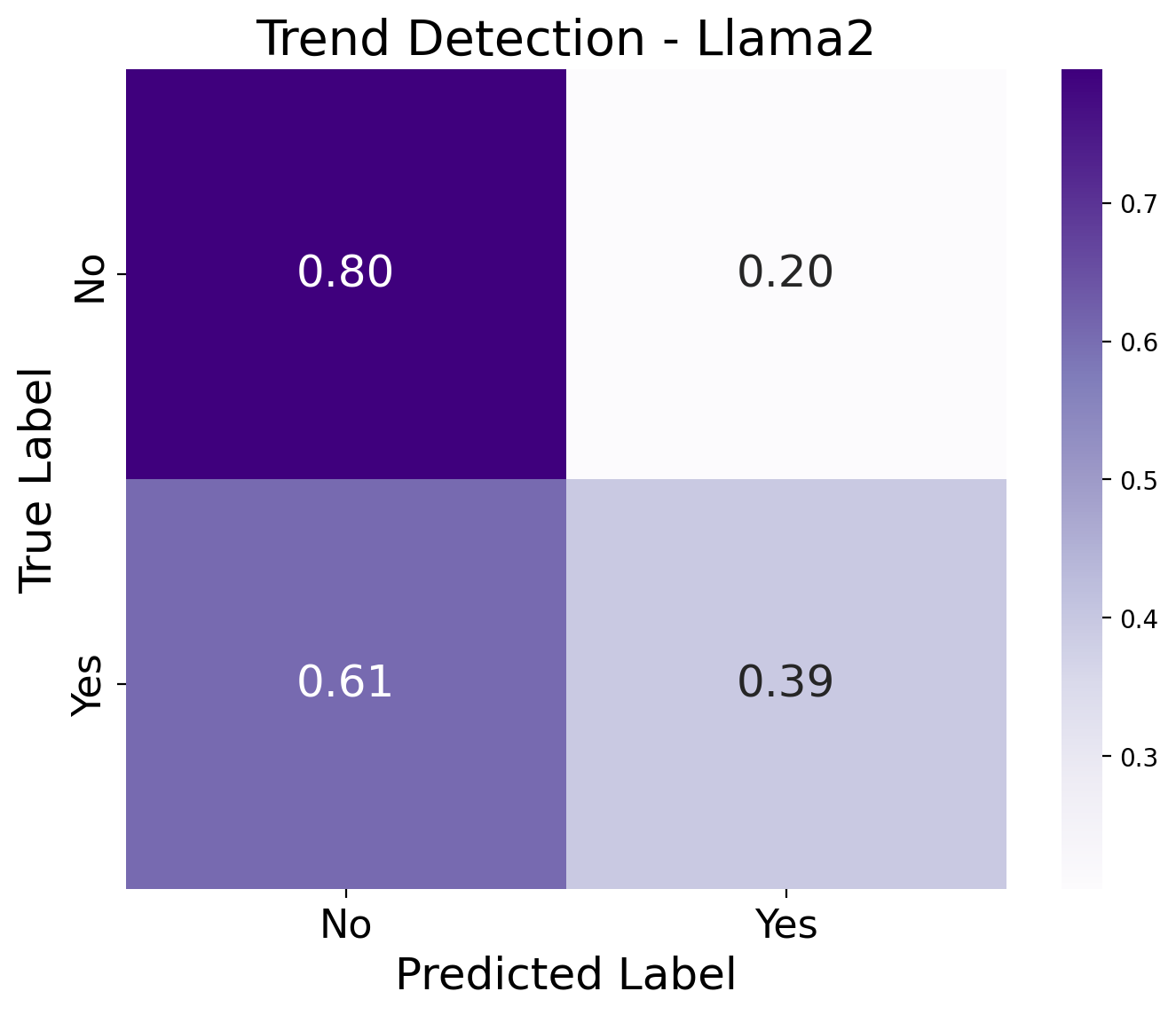}~
    \includegraphics[width=0.20\textwidth]{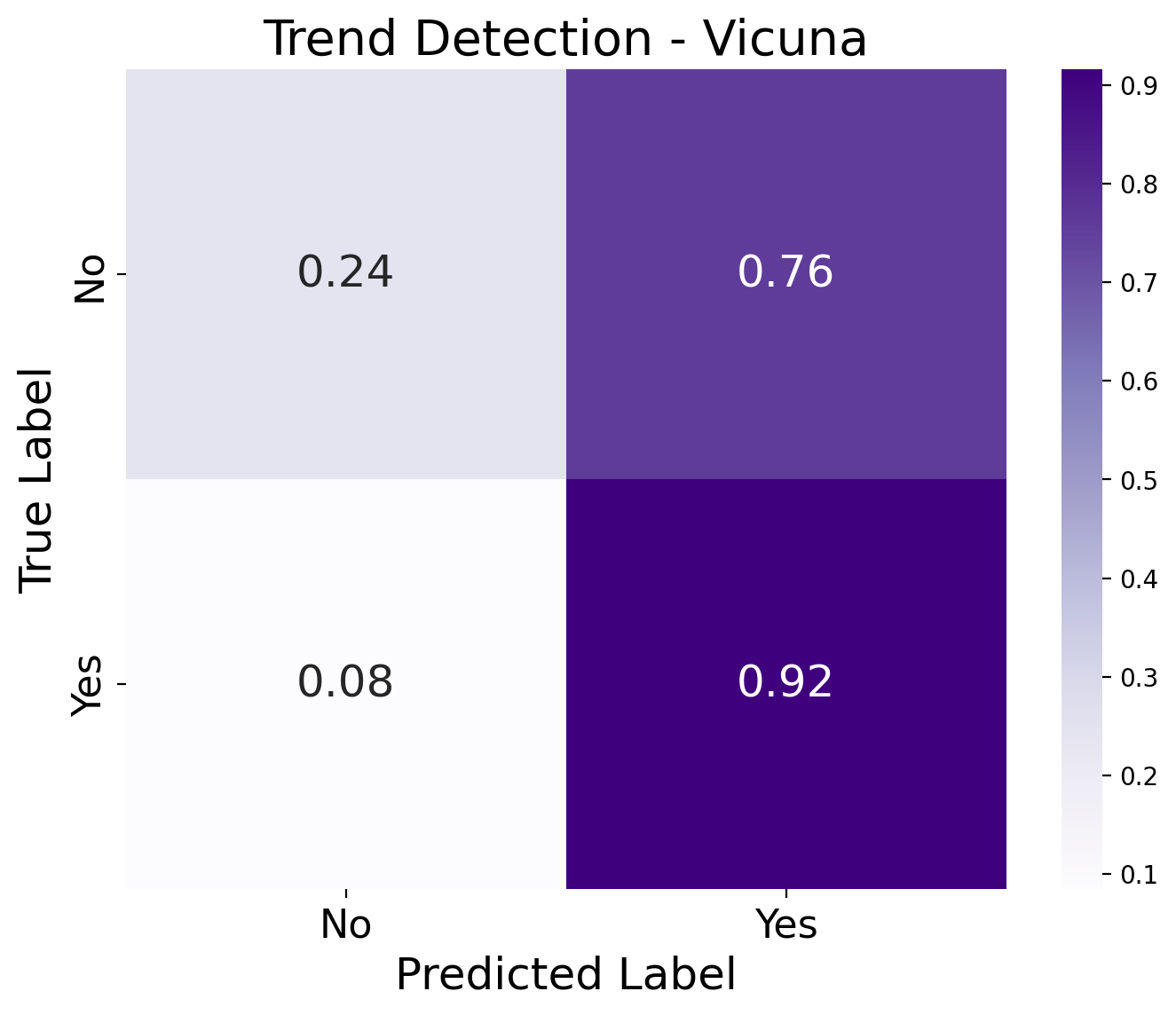}~
    \includegraphics[width=0.20\textwidth]{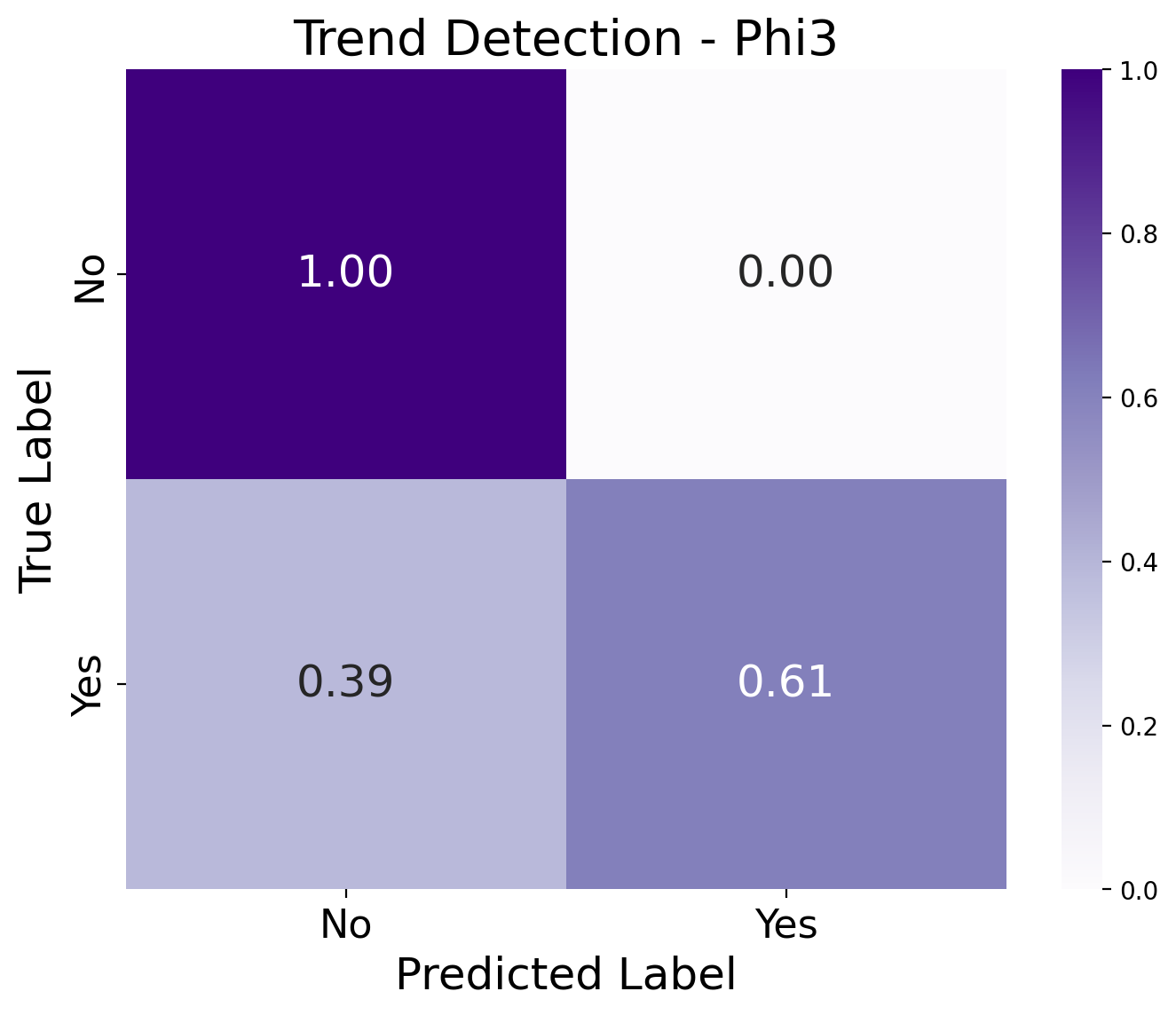}~
    \caption{Trend detection}
    \label{fig:enter-label}
\end{figure}

\begin{figure}[h!]
    \centering
    \includegraphics[width=0.20\textwidth]{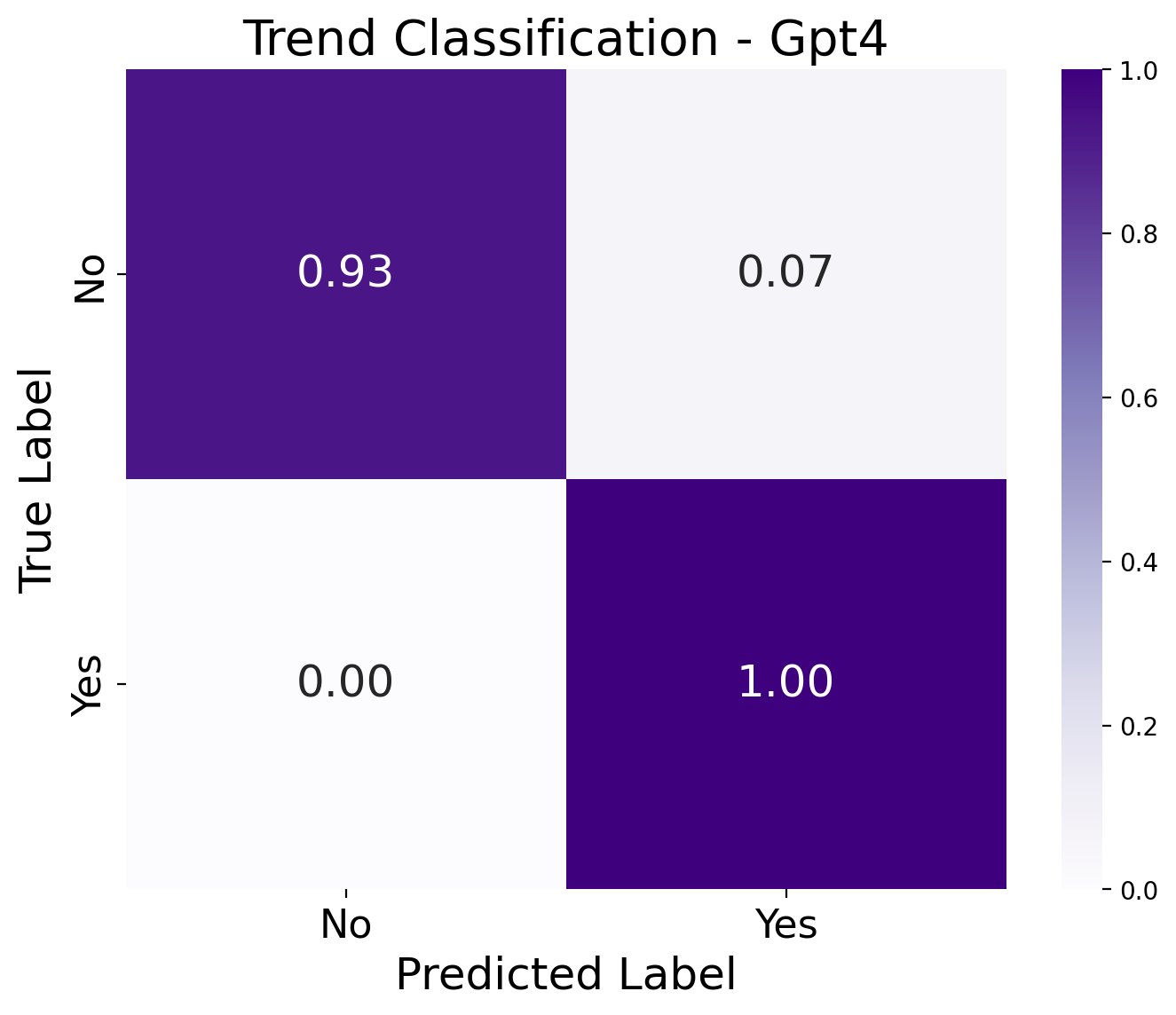}~
    \includegraphics[width=0.20\textwidth]{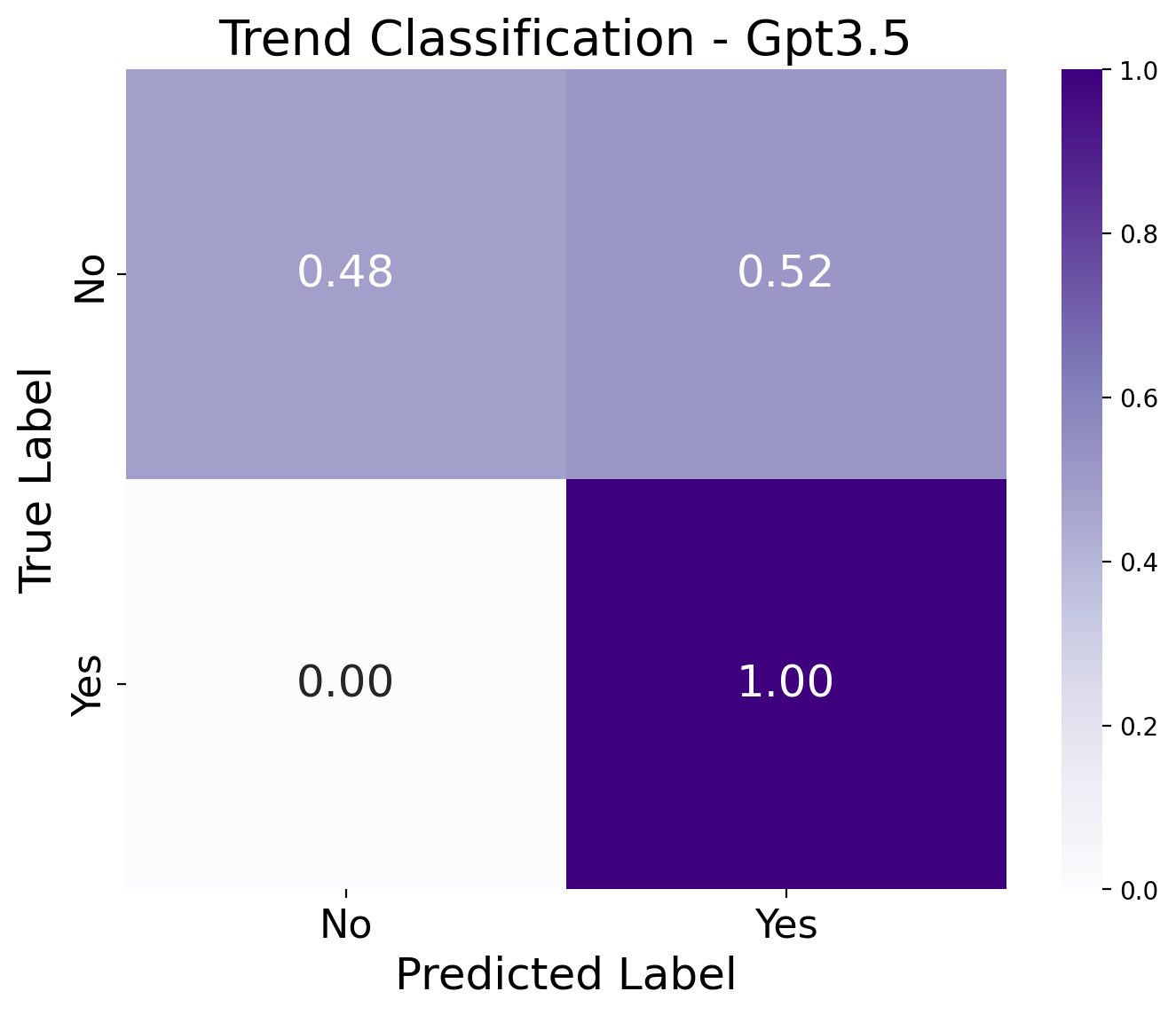}~
    \includegraphics[width=0.20\textwidth]{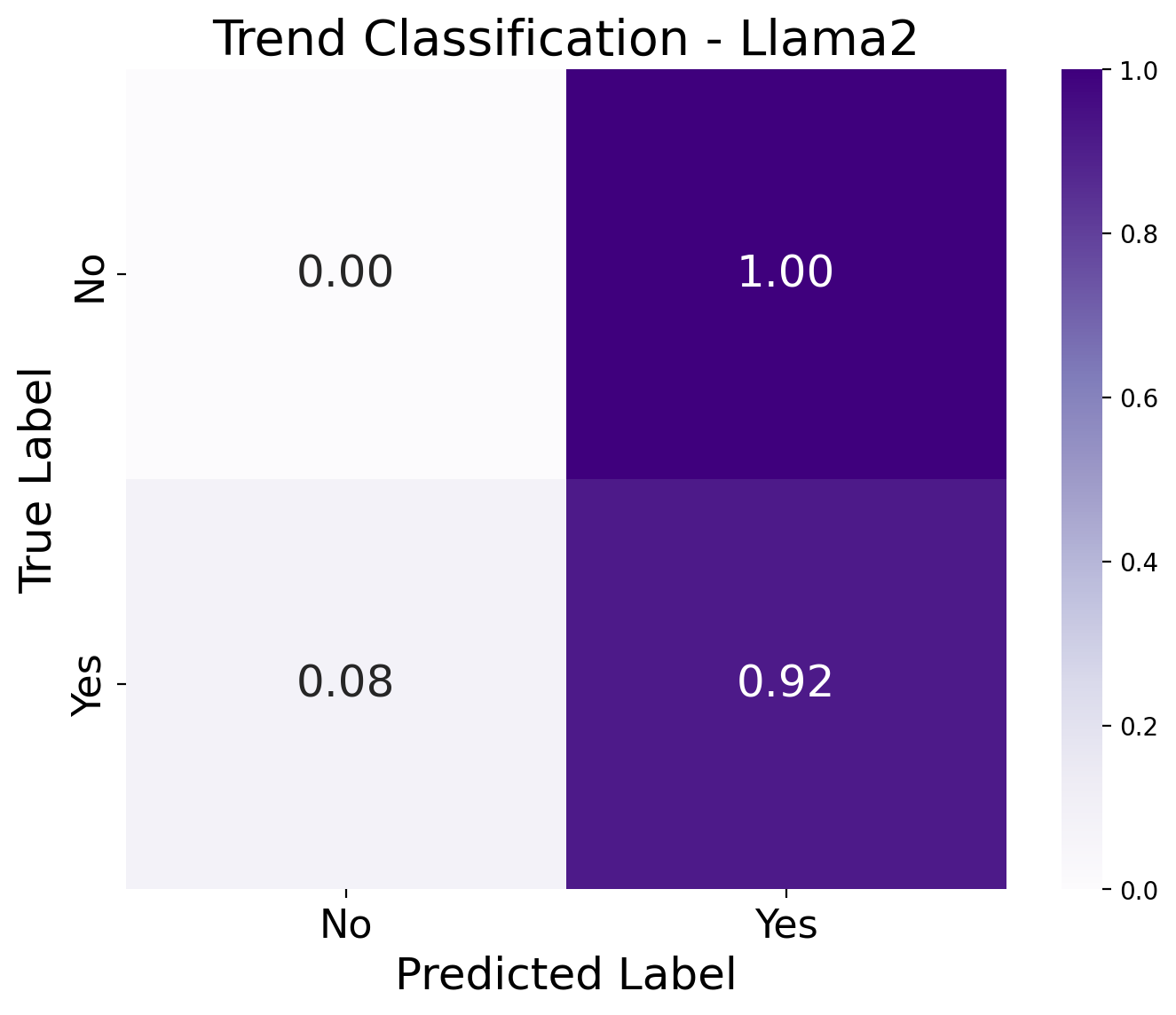}~
    \includegraphics[width=0.20\textwidth]{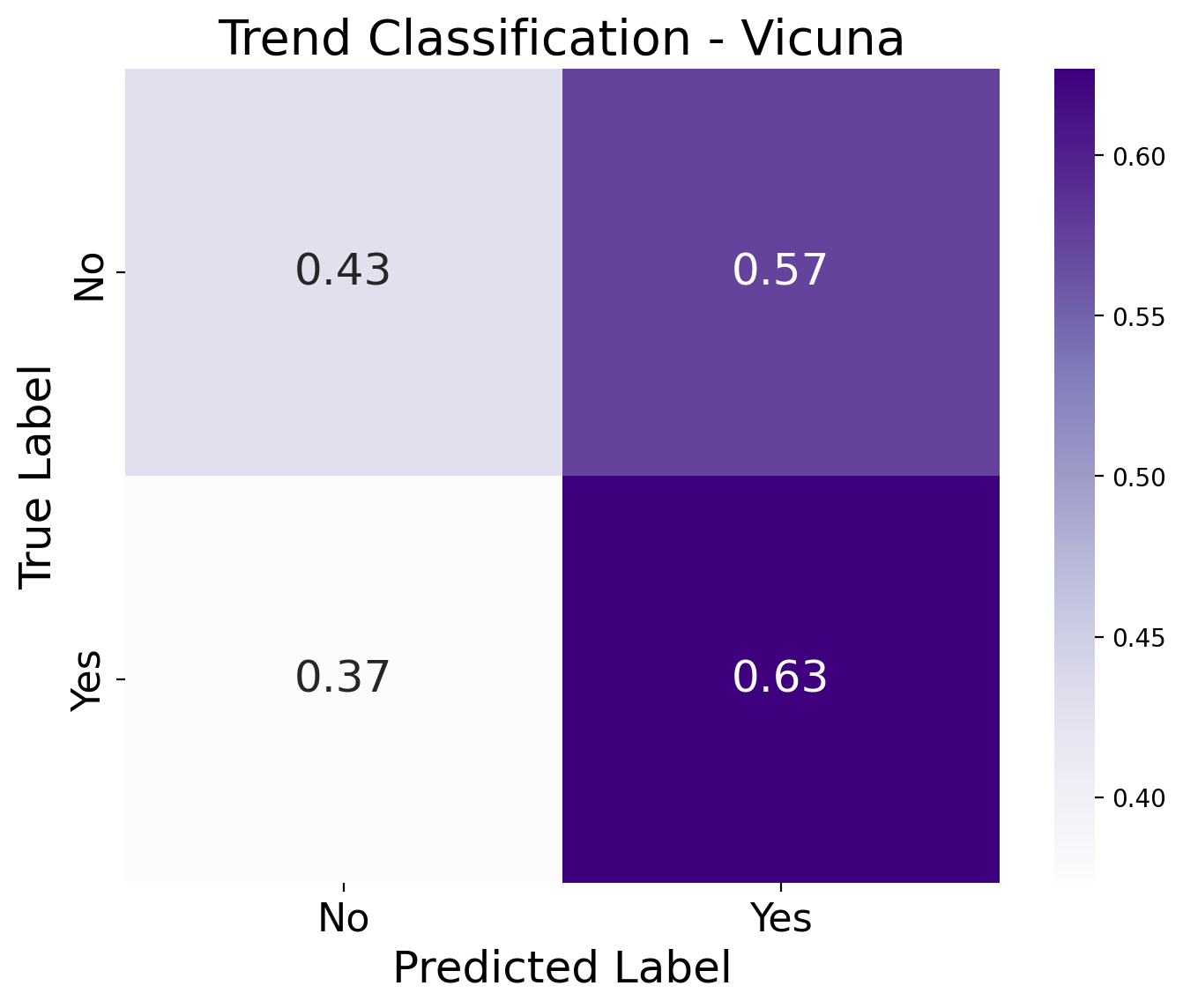}~
    \includegraphics[width=0.20\textwidth]{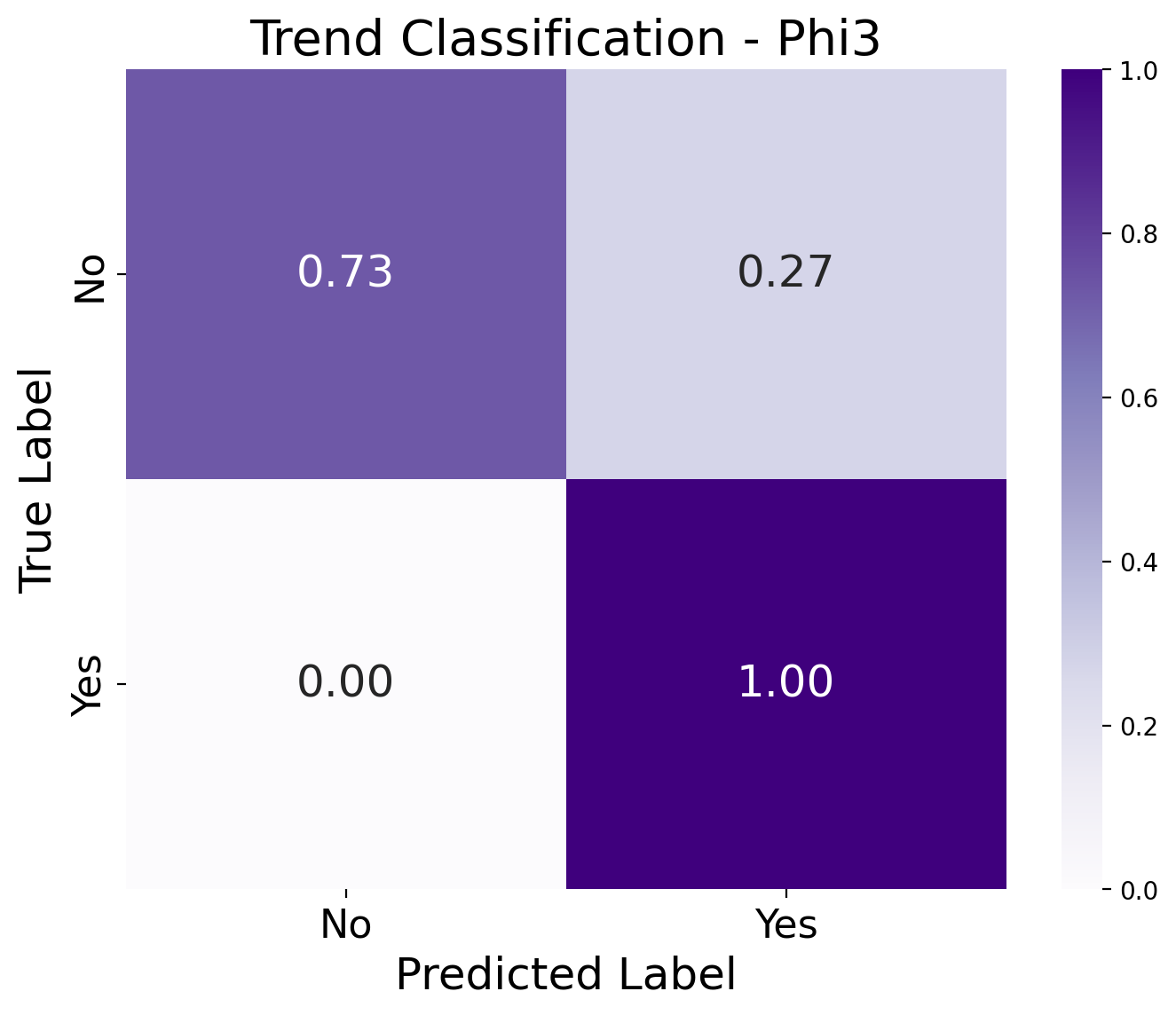}~
    \caption{Trend classification}
    \label{fig:enter-label}
\end{figure}

\subsection{Seasonality}
\begin{figure}[h!]
    \centering
    \includegraphics[width=0.20\textwidth]{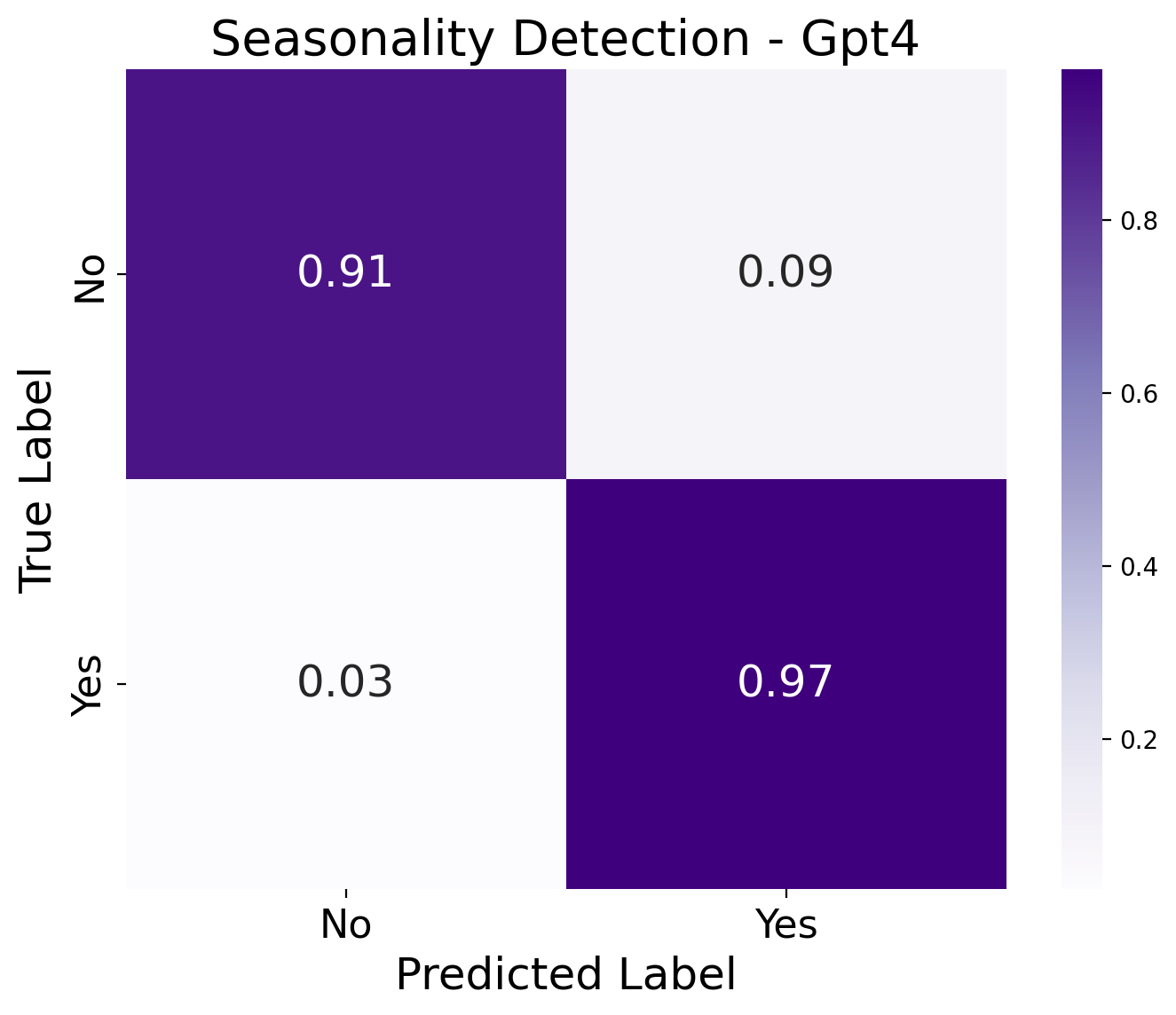}~
    \includegraphics[width=0.20\textwidth]{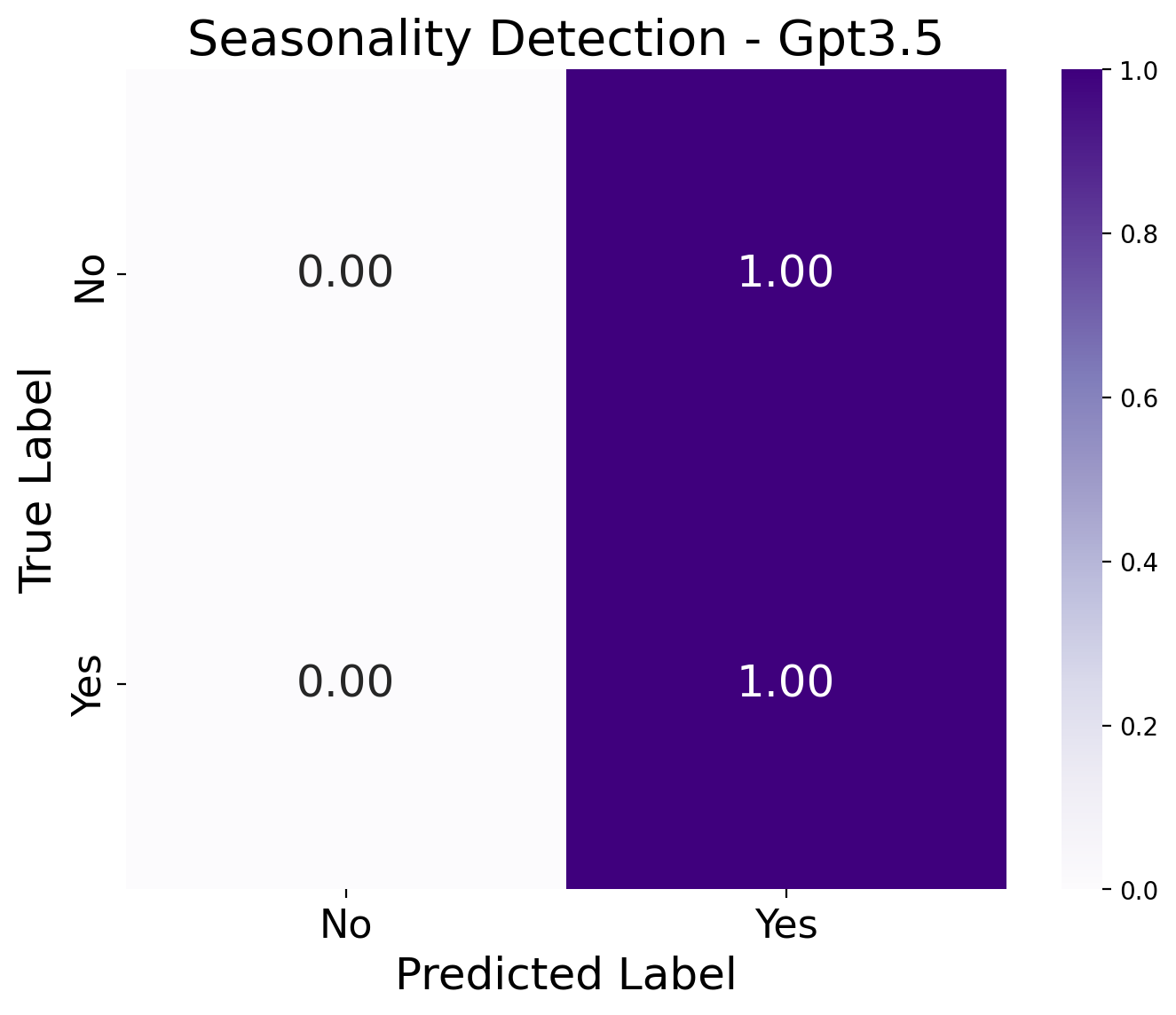}~
    \includegraphics[width=0.20\textwidth]{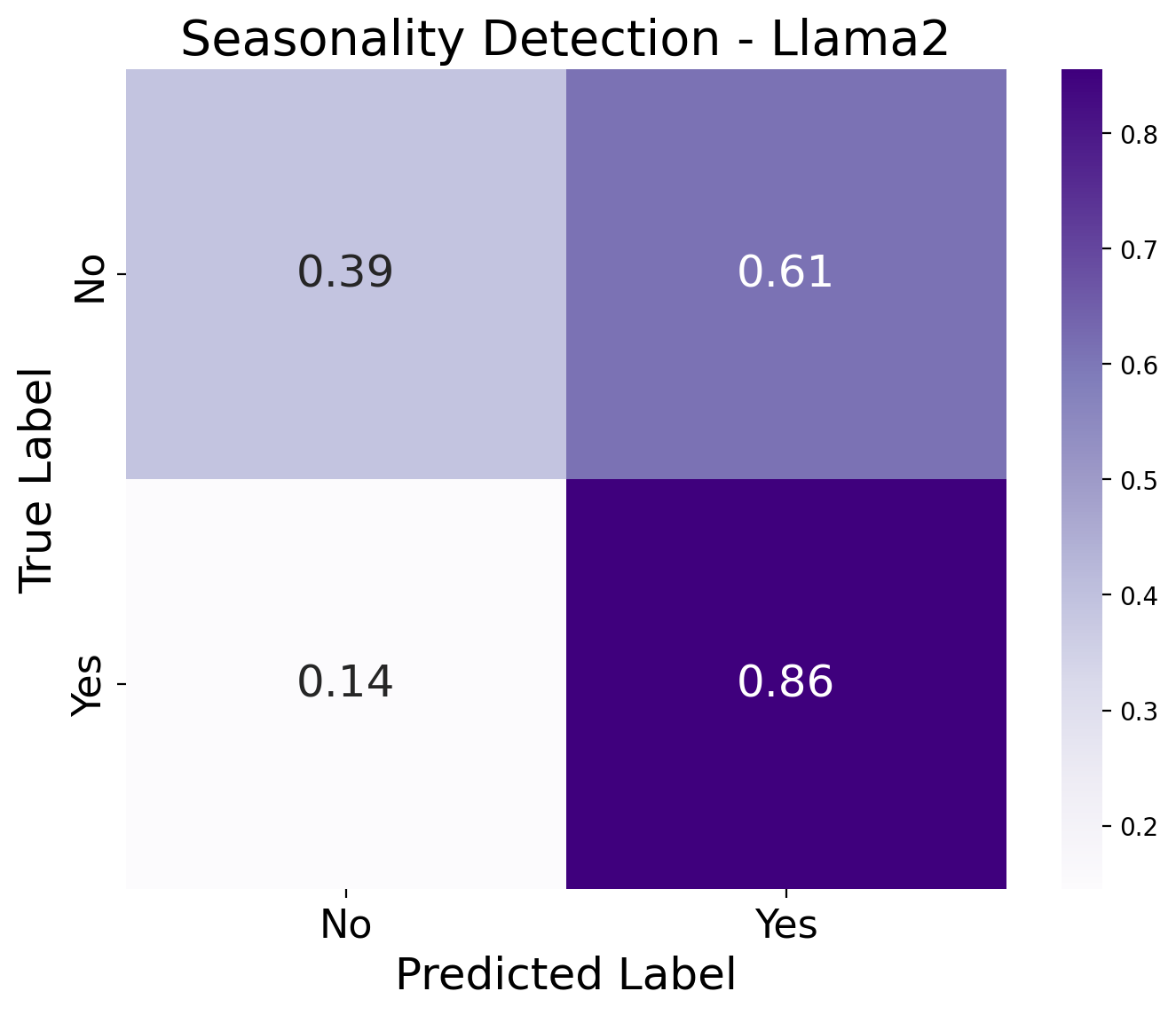}~
    \includegraphics[width=0.20\textwidth]{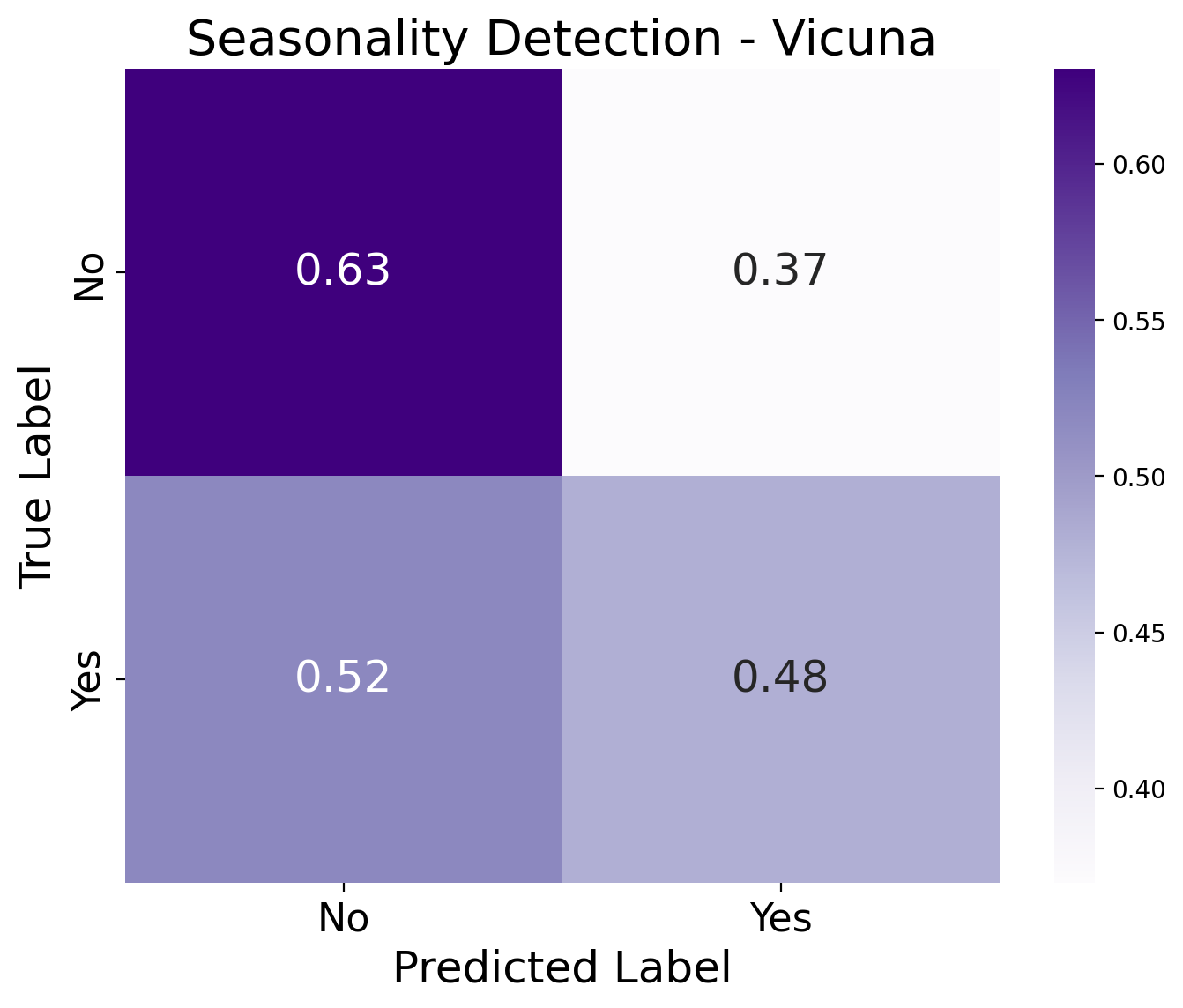}~
    \includegraphics[width=0.20\textwidth]{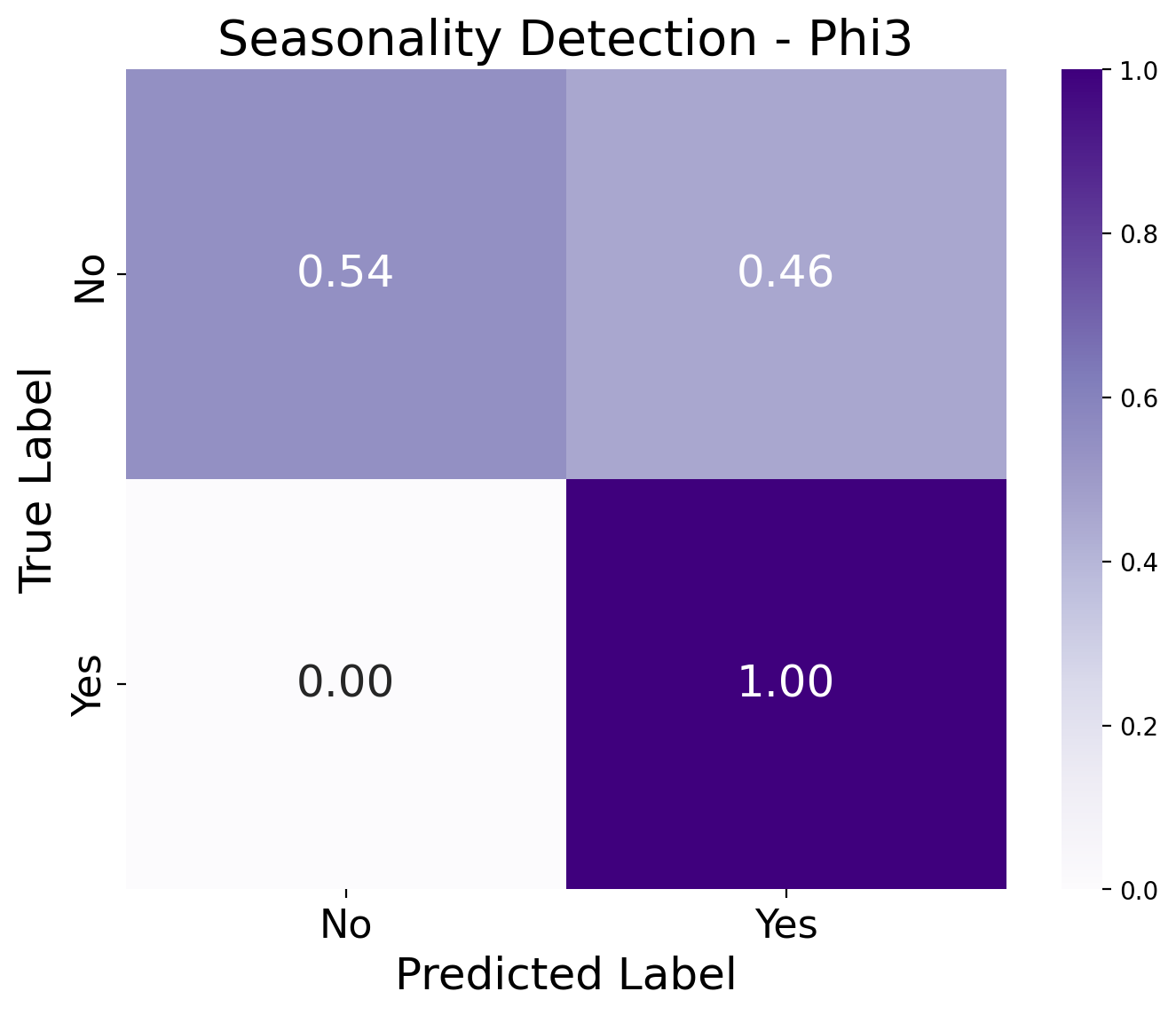}
    \caption{Seasonality detection}
    \label{fig:enter-label}
\end{figure}
\begin{figure}[h!]
    \centering
    \includegraphics[width=0.20\textwidth]{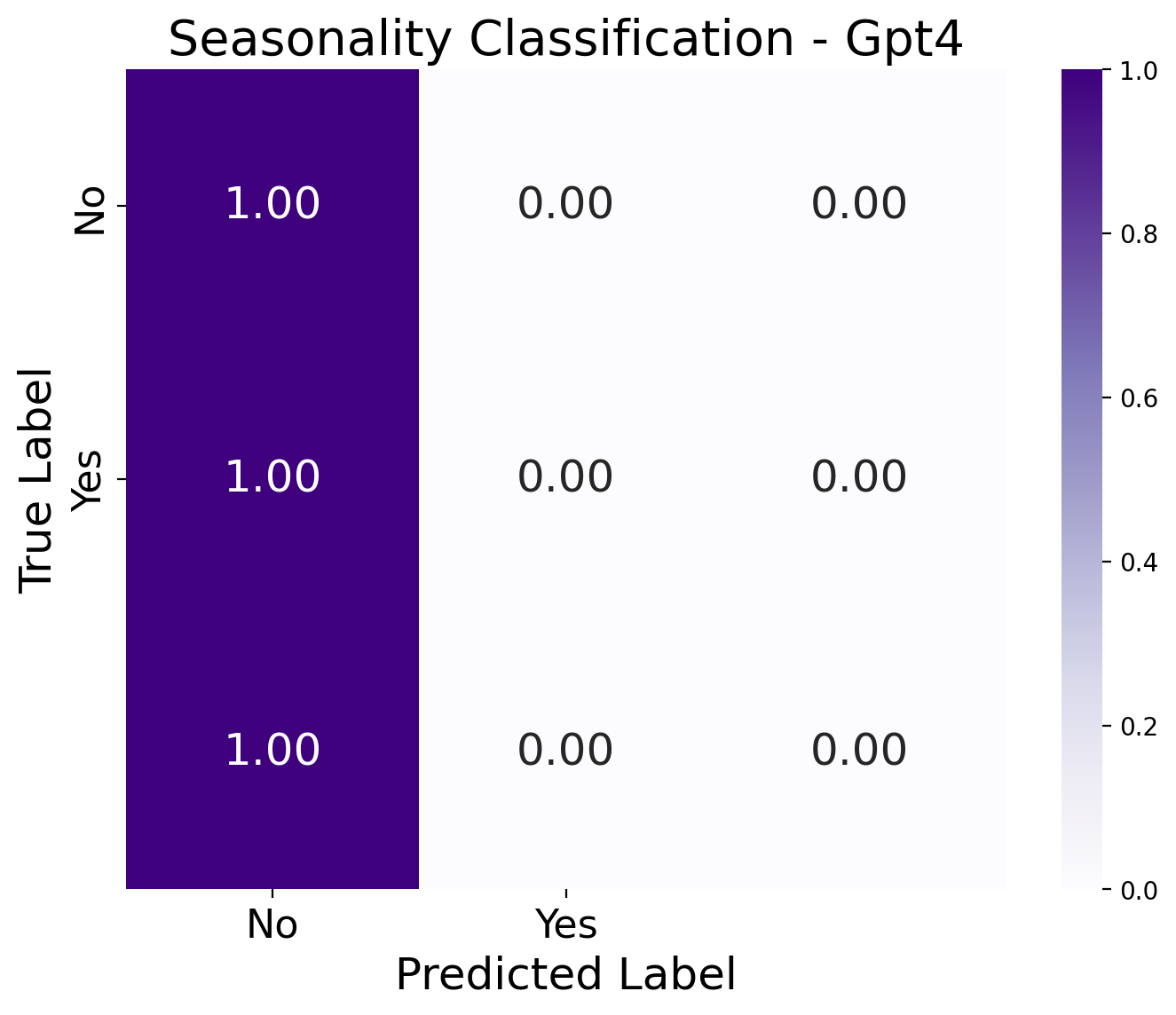}~
    \includegraphics[width=0.20\textwidth]{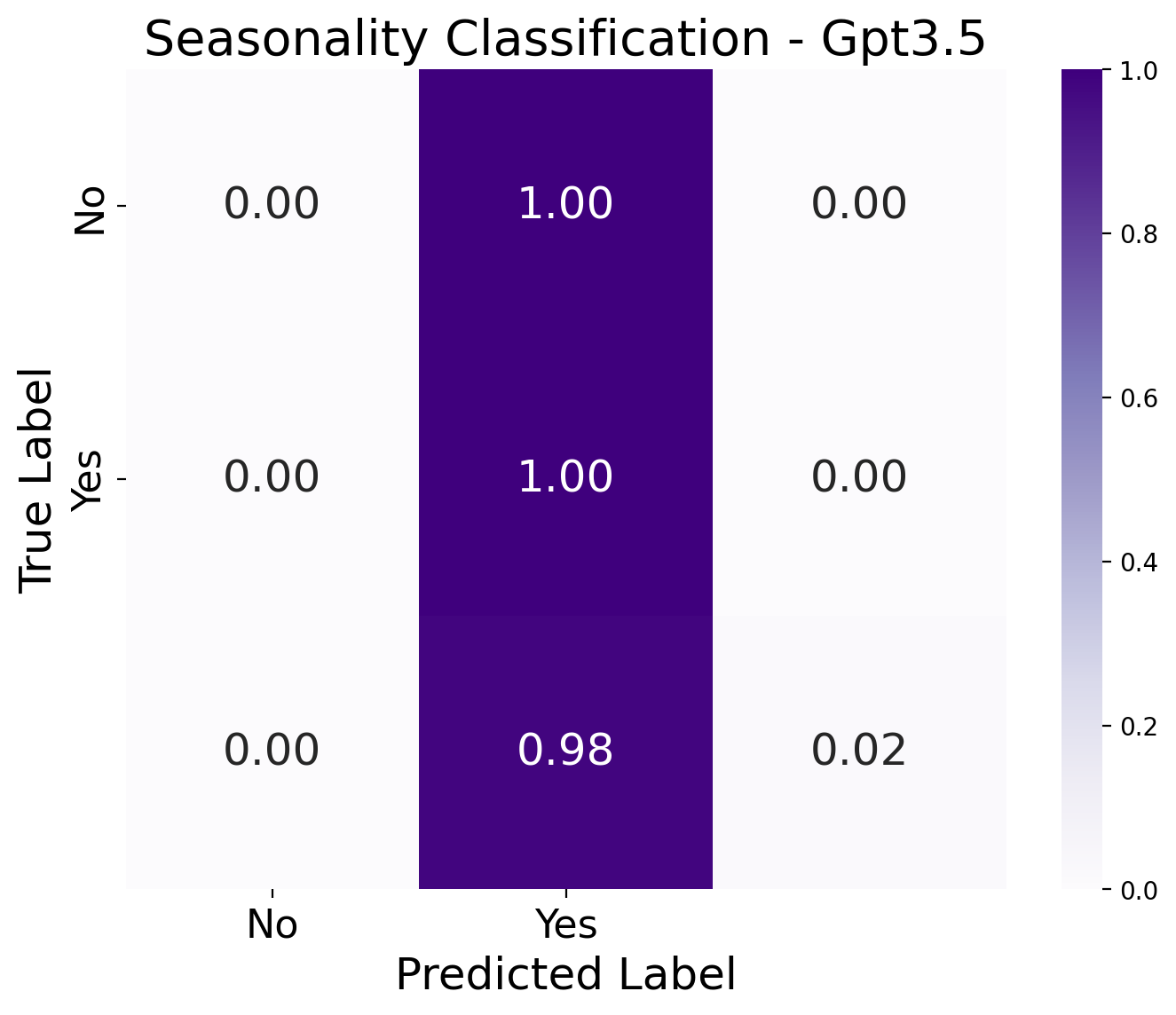}~
    \includegraphics[width=0.20\textwidth]{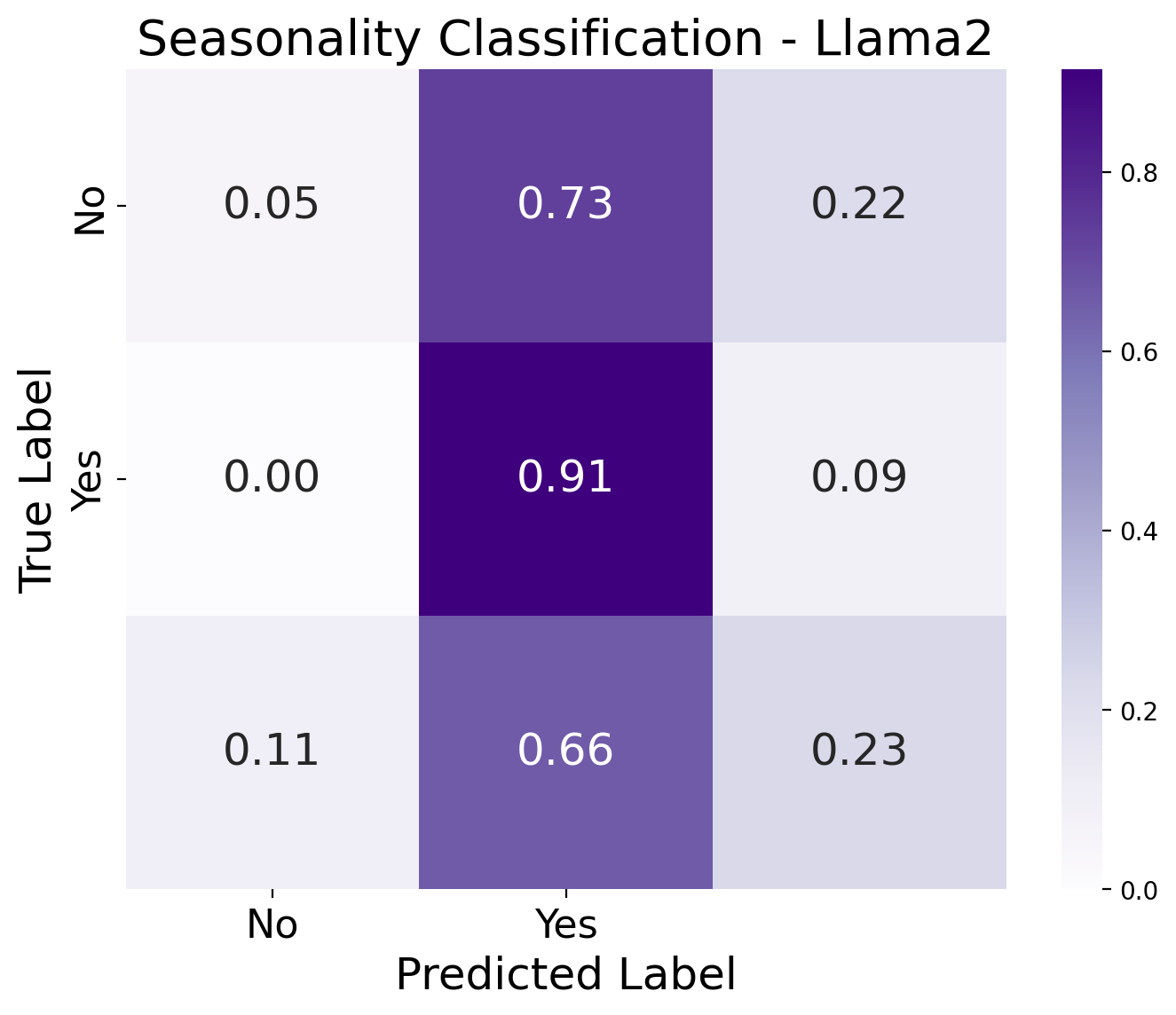}~
    \includegraphics[width=0.20\textwidth]{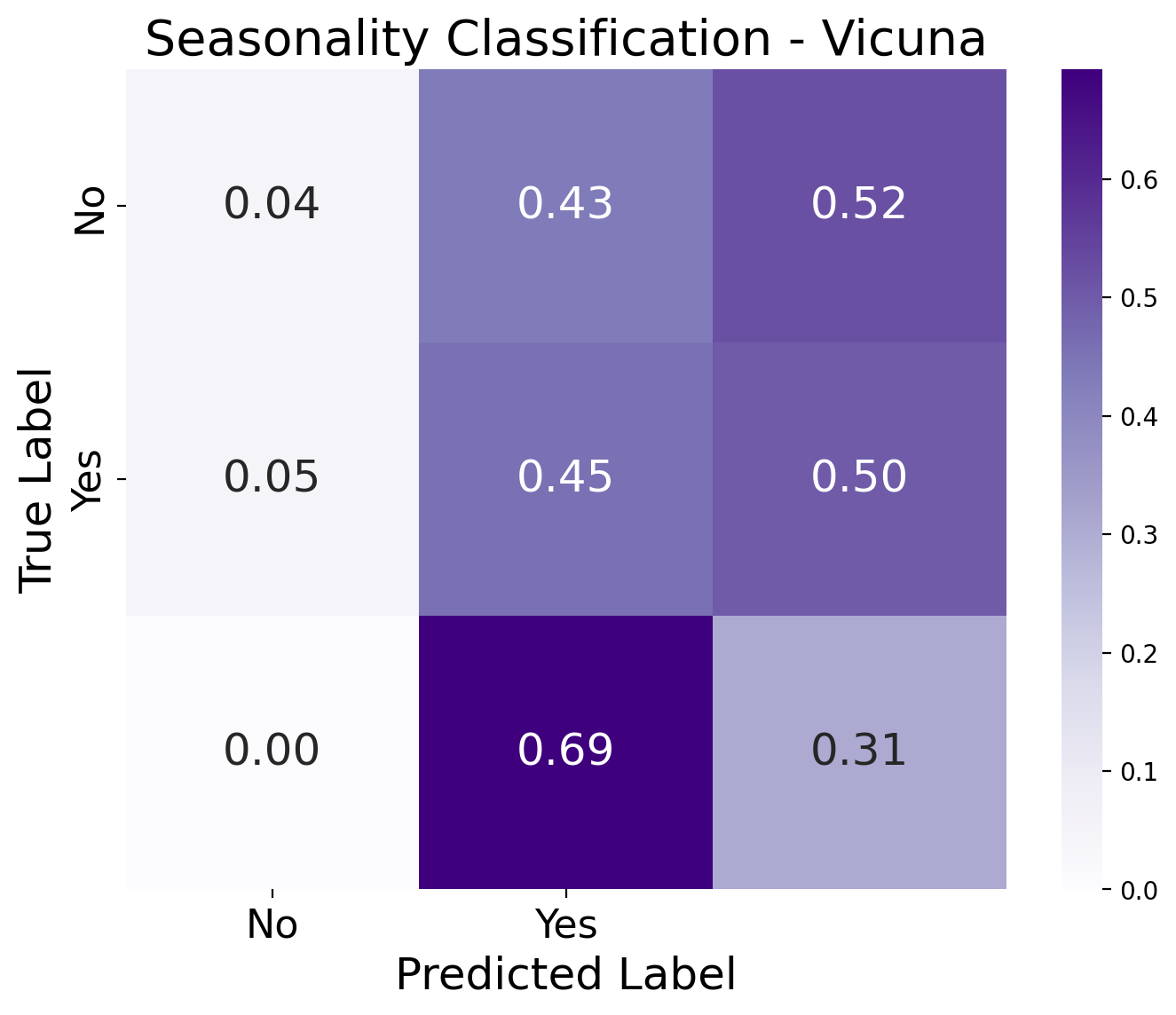}~
    \includegraphics[width=0.20\textwidth]{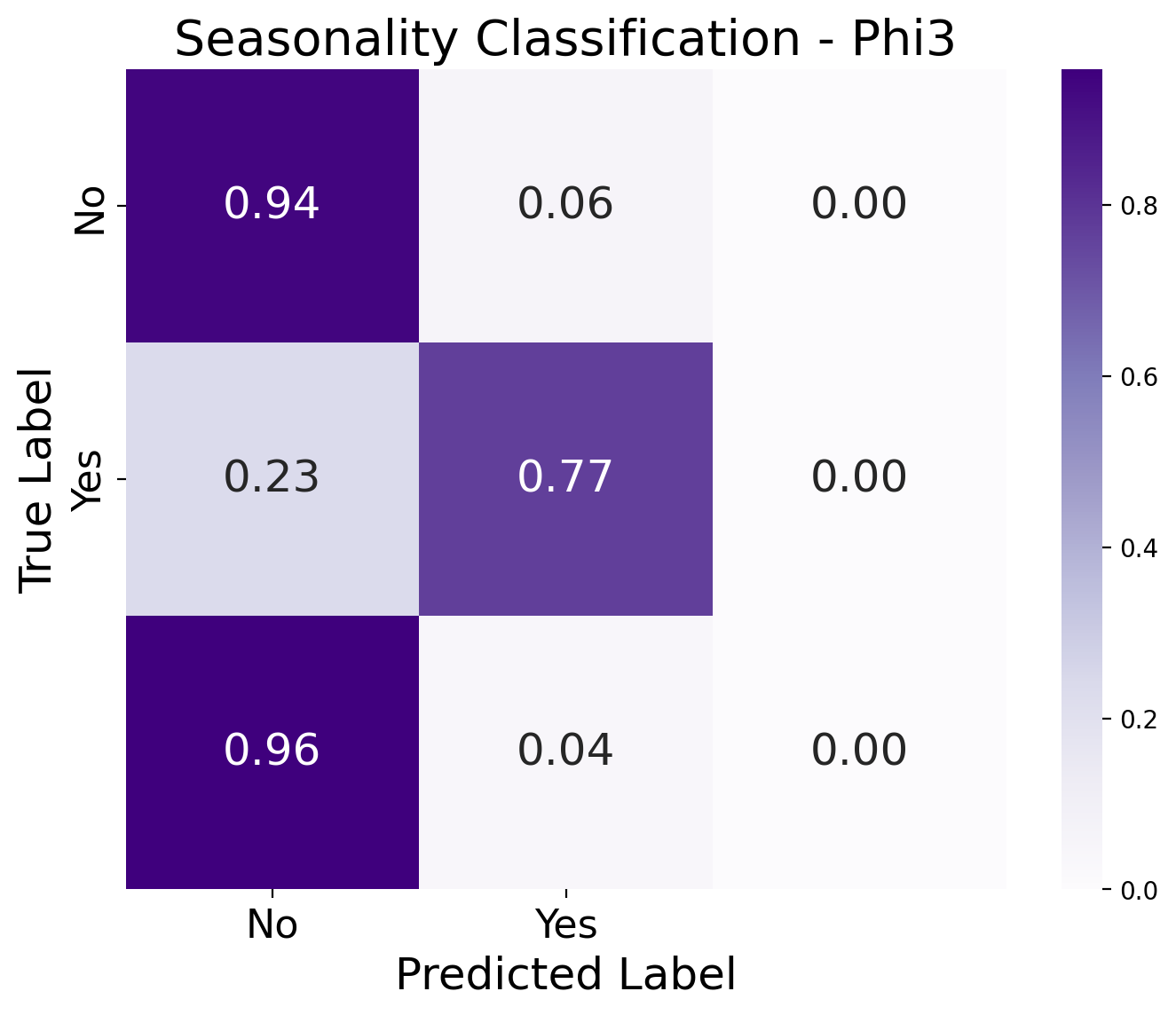}
    \label{fig:enter-label}
\end{figure}

\subsection{Anomalies}
\begin{figure}[h!]
    \centering
    \includegraphics[width=0.20\textwidth]{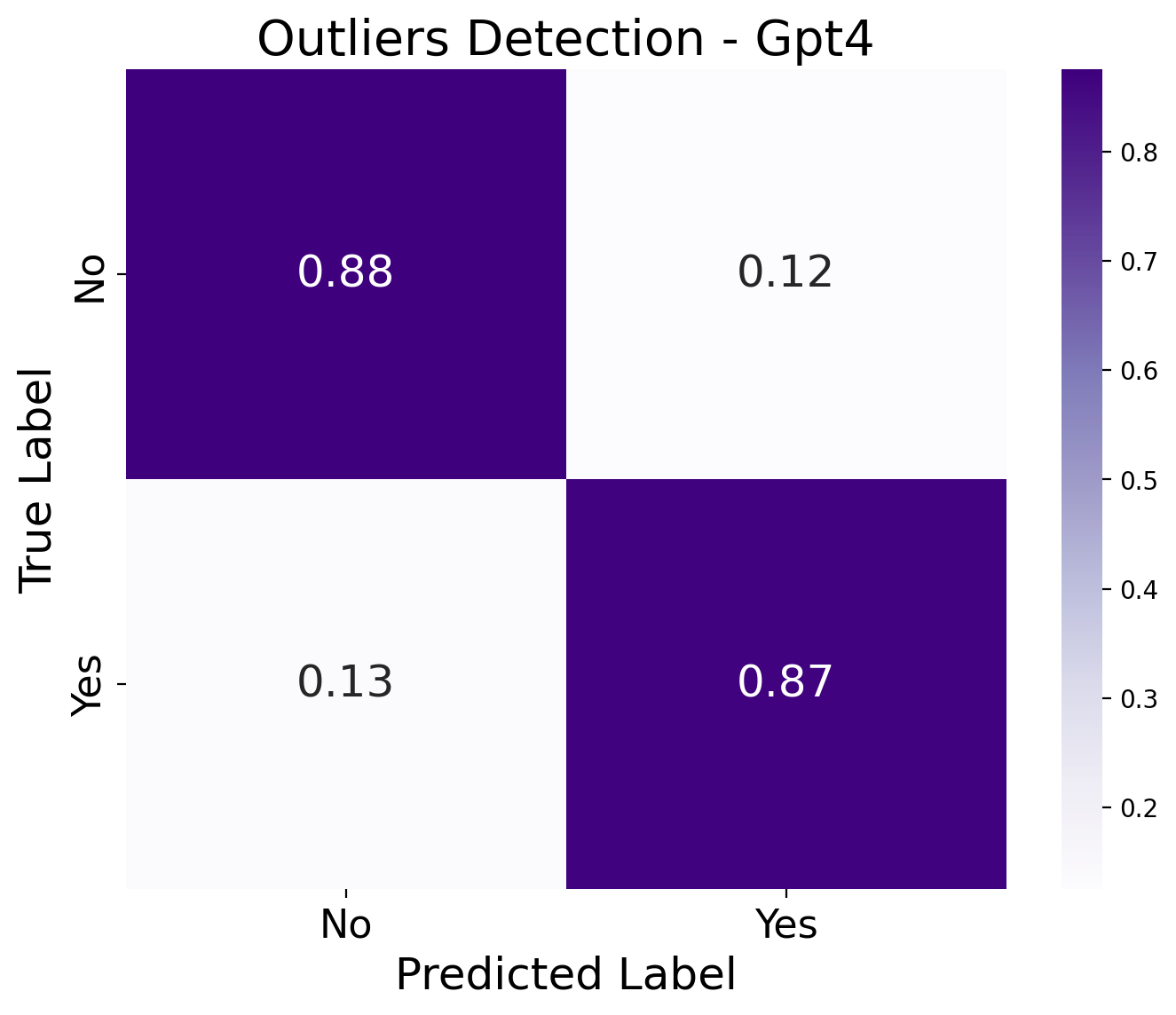}~
    \includegraphics[width=0.20\textwidth]{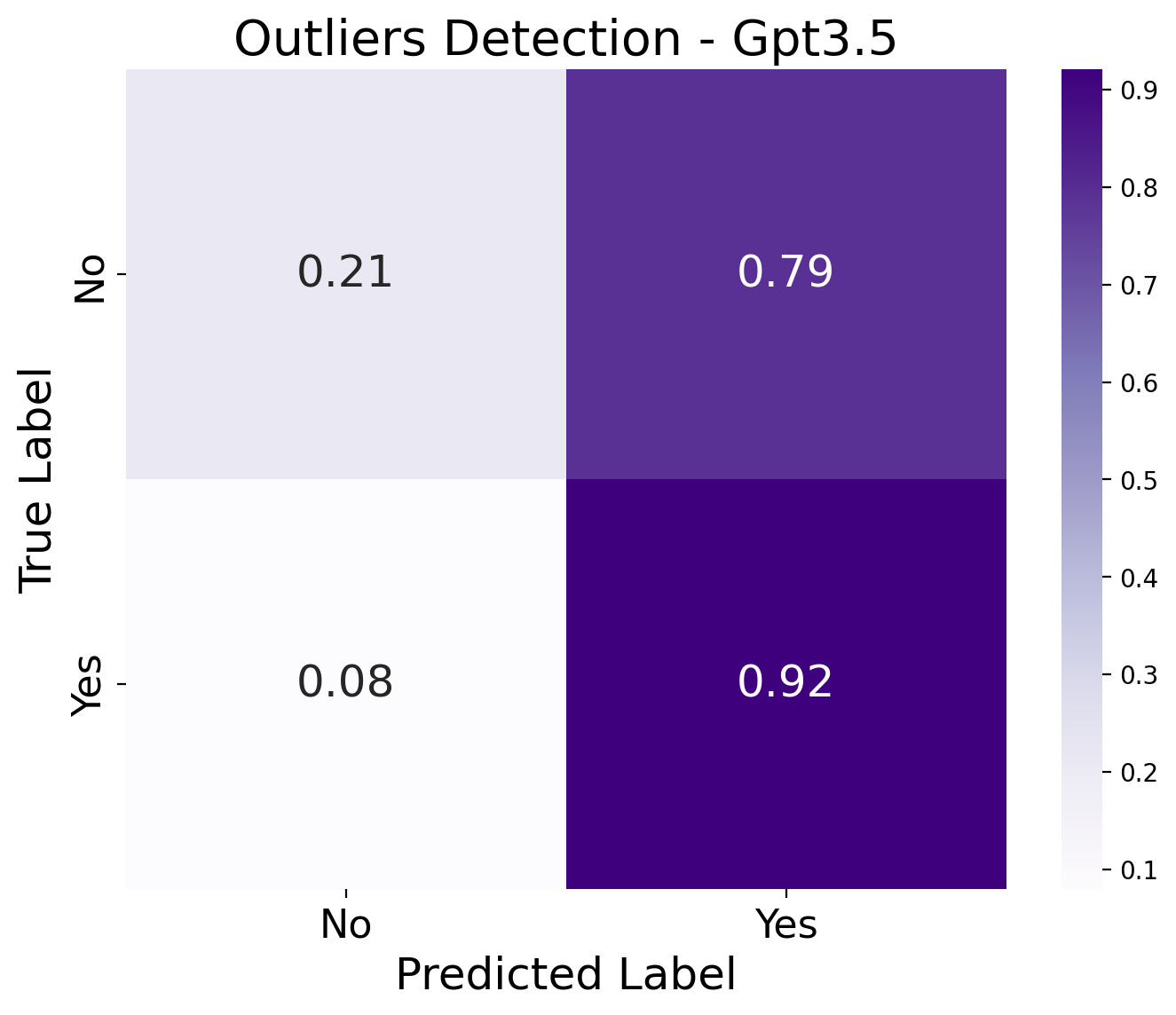}~
    \includegraphics[width=0.20\textwidth]{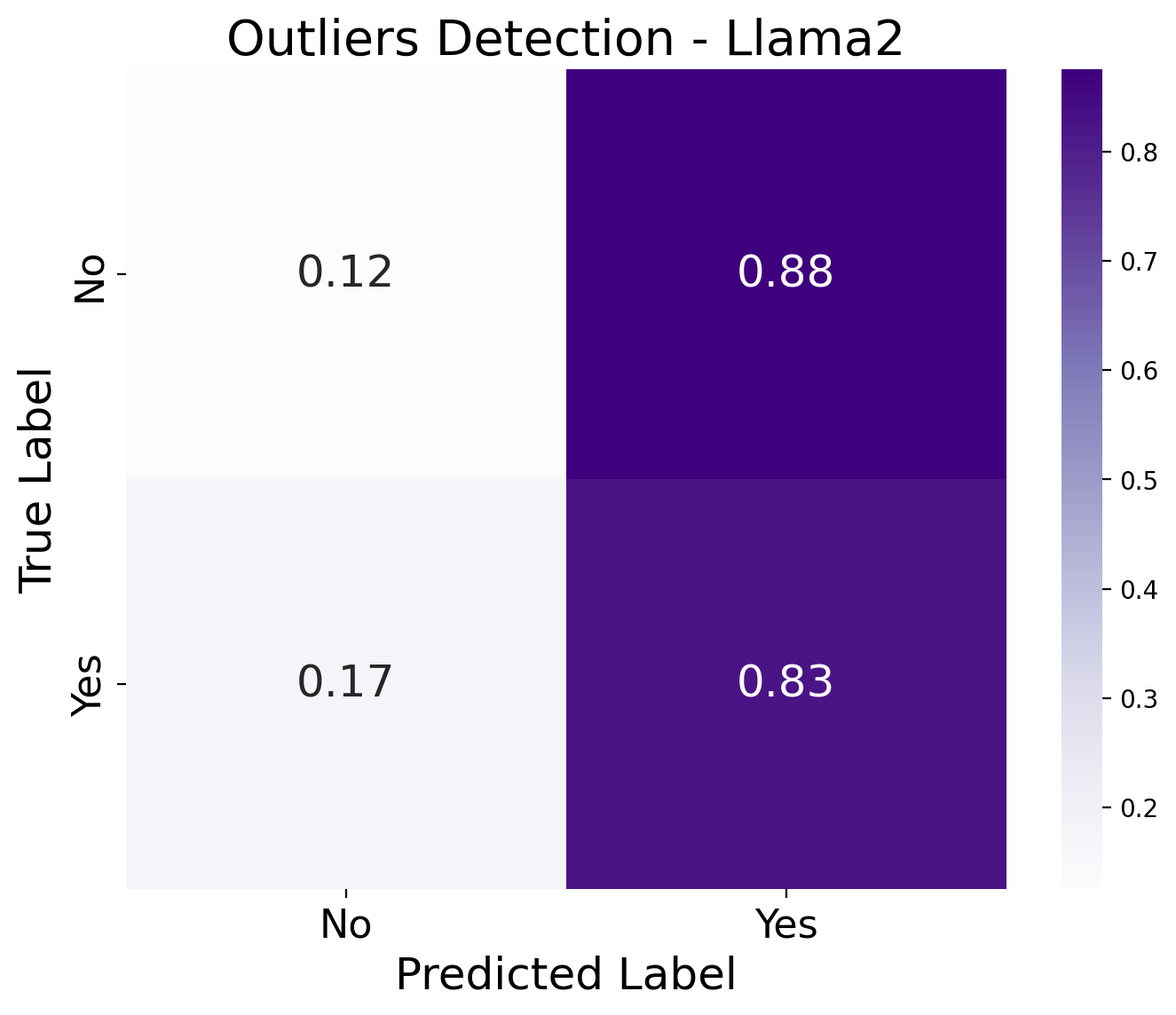}~
    \includegraphics[width=0.20\textwidth]{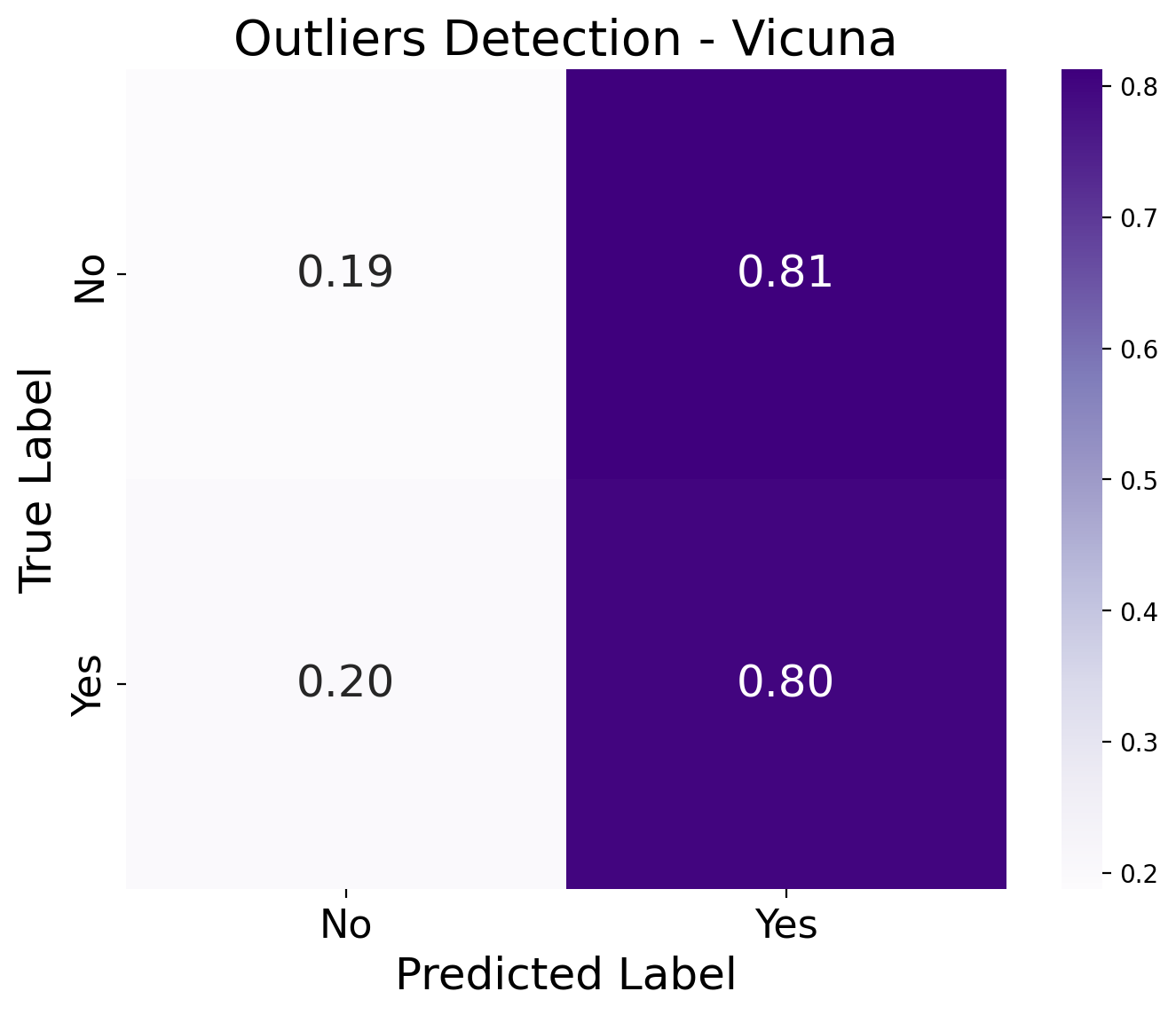}~
    \includegraphics[width=0.20\textwidth]{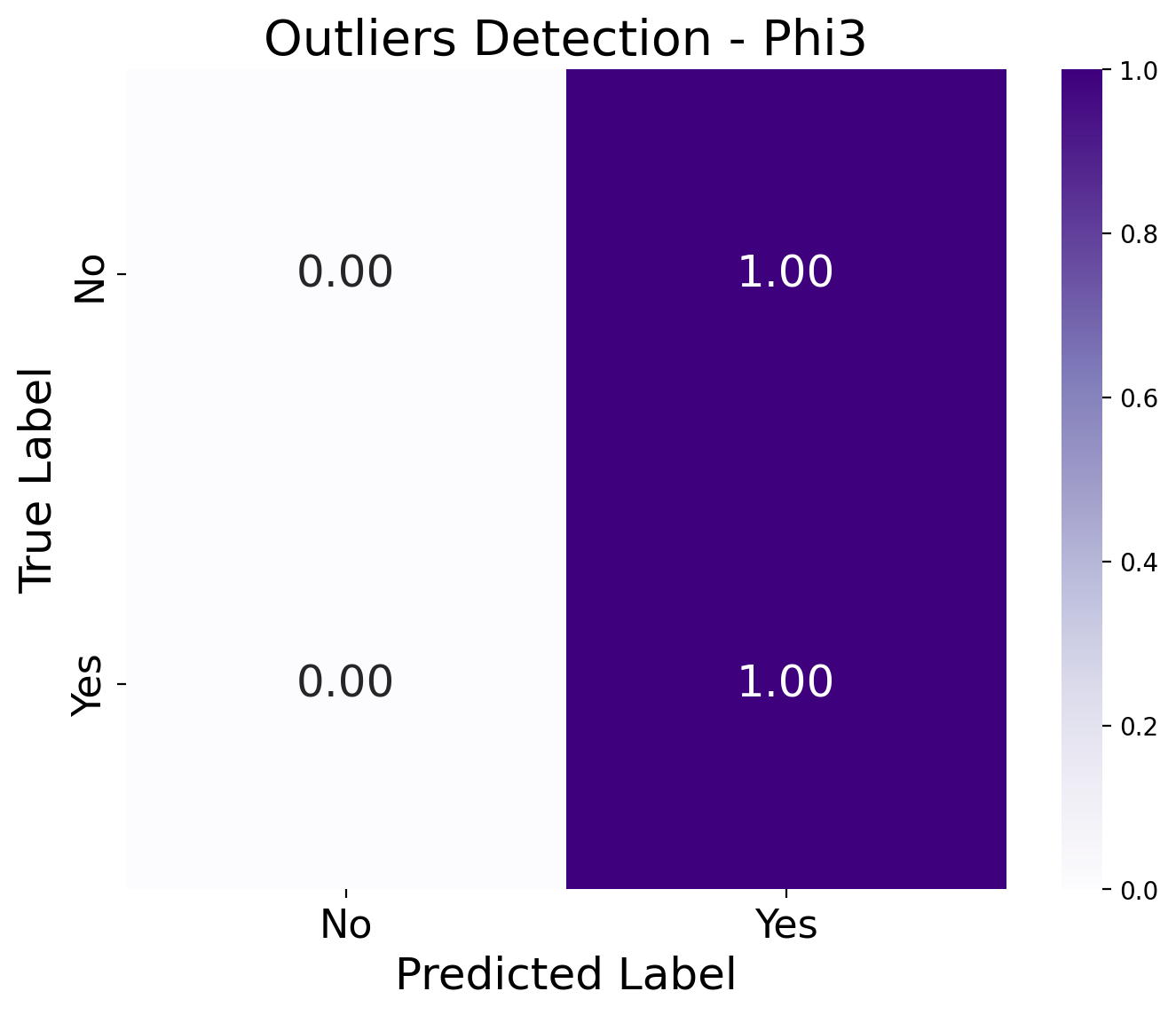}
    \caption{Anomaly detection}
    \label{fig:enter-label}
\end{figure}
\begin{figure}[h!]
    \centering
    \includegraphics[width=0.20\textwidth]{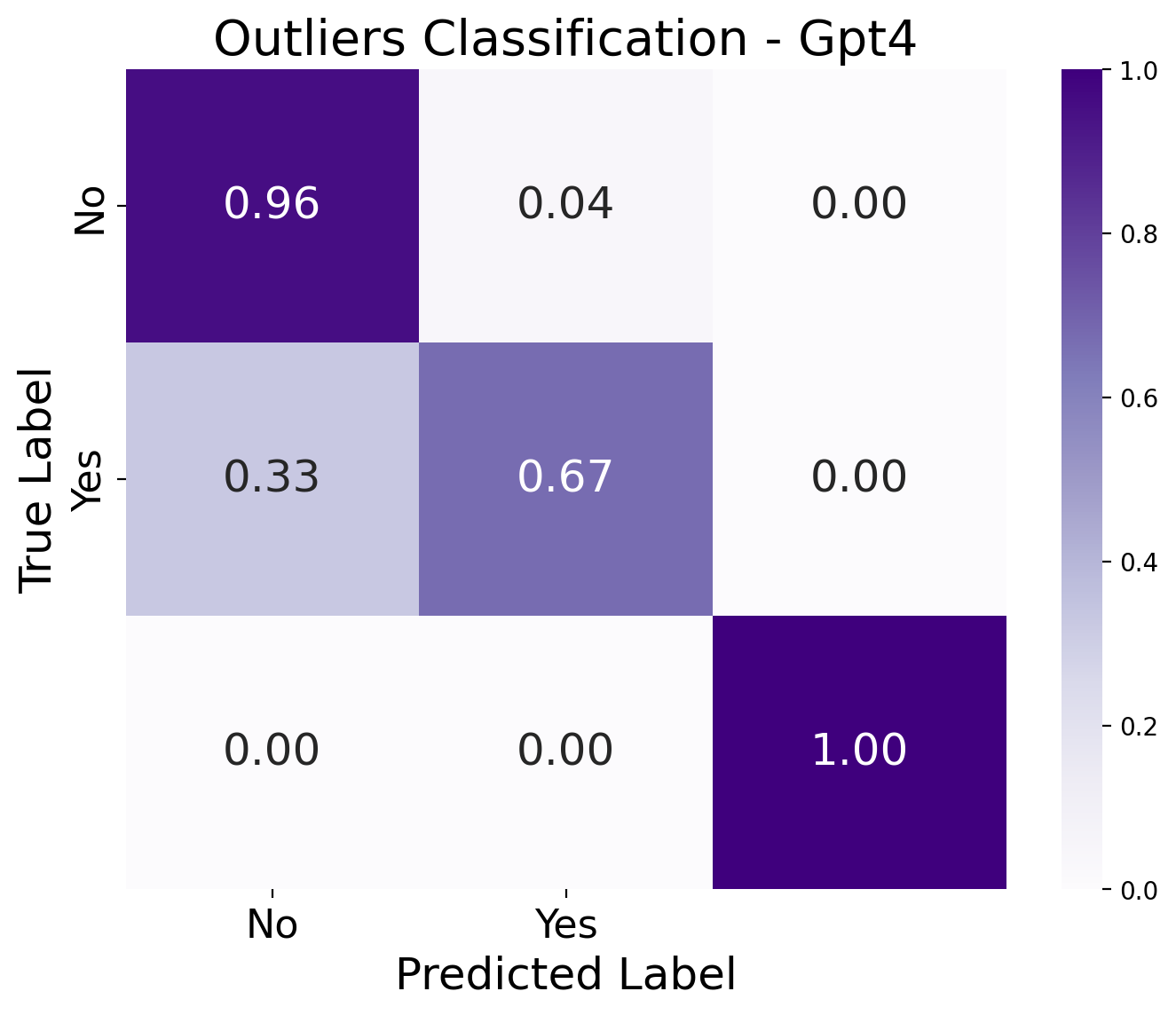}~
    \includegraphics[width=0.20\textwidth]{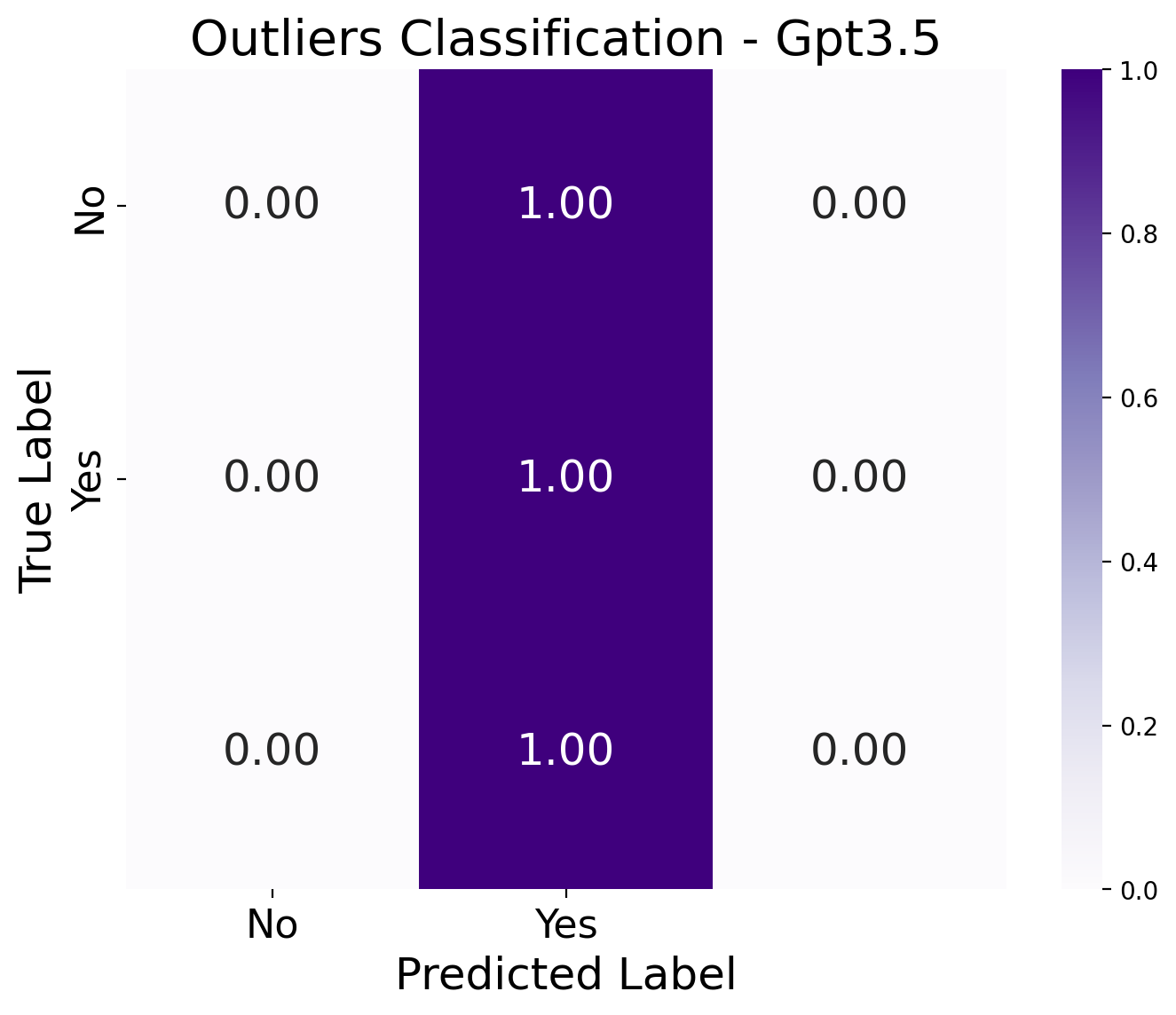}~
    \includegraphics[width=0.20\textwidth]{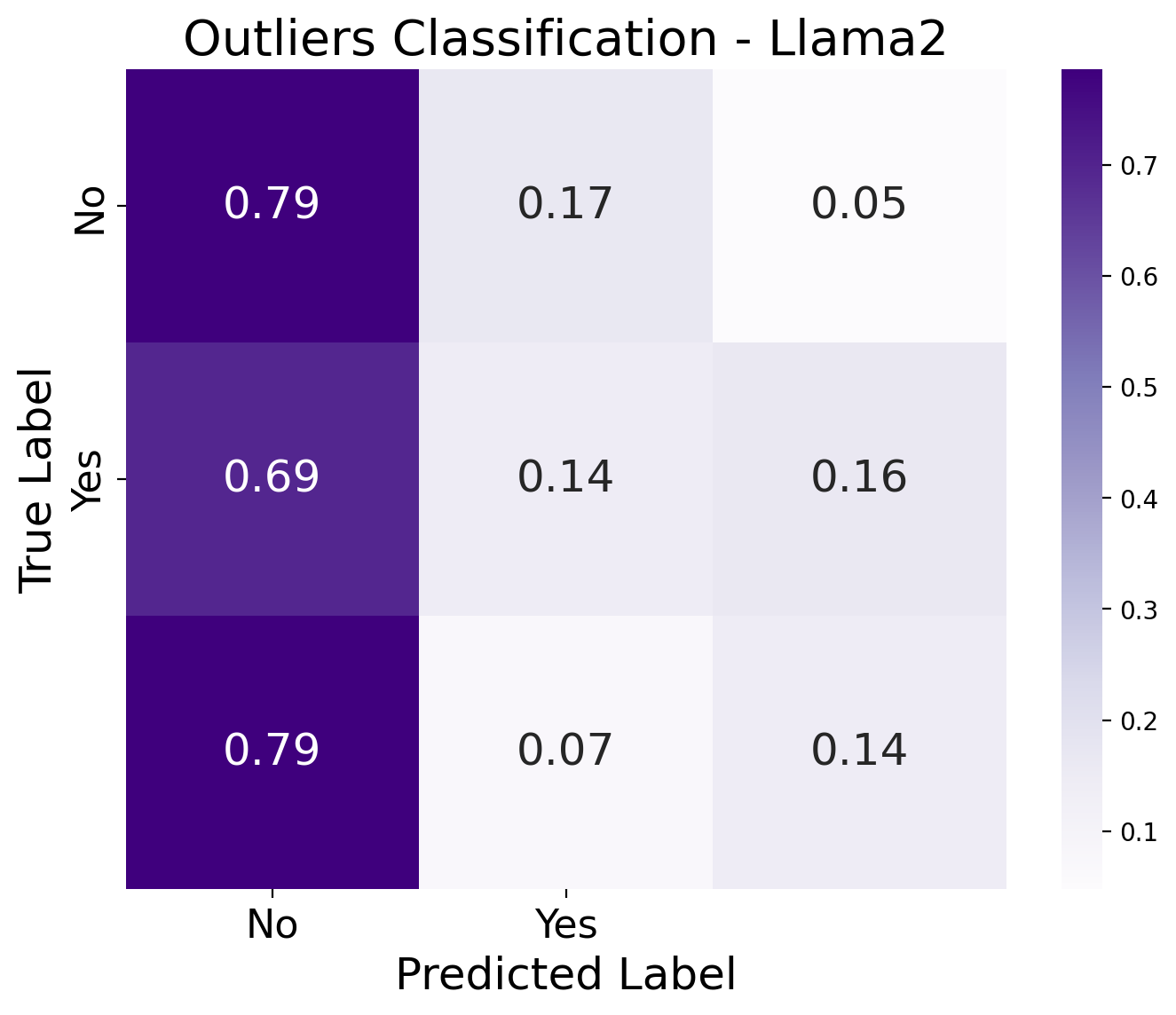}~
    \includegraphics[width=0.20\textwidth]{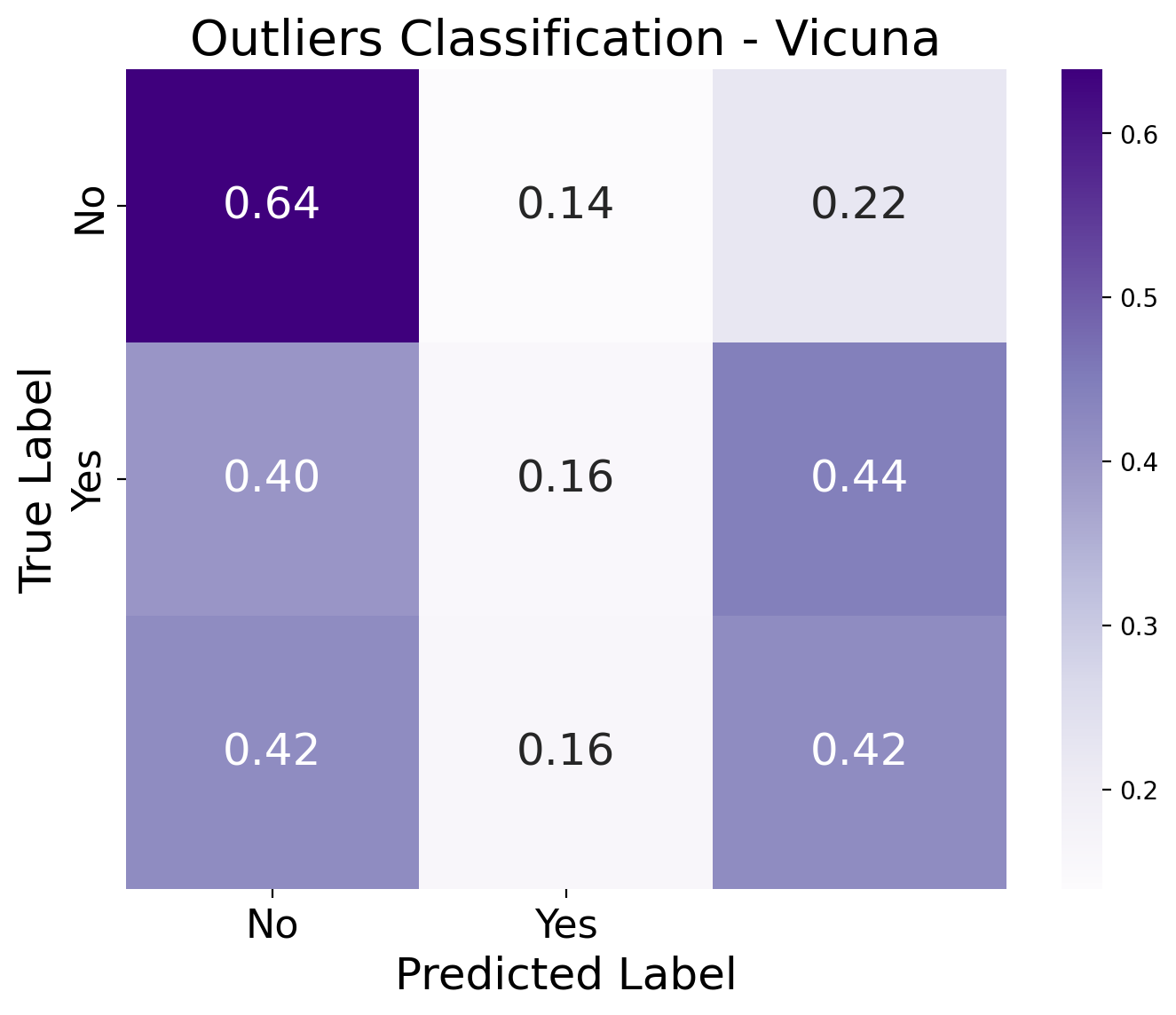}~
    \includegraphics[width=0.20\textwidth]{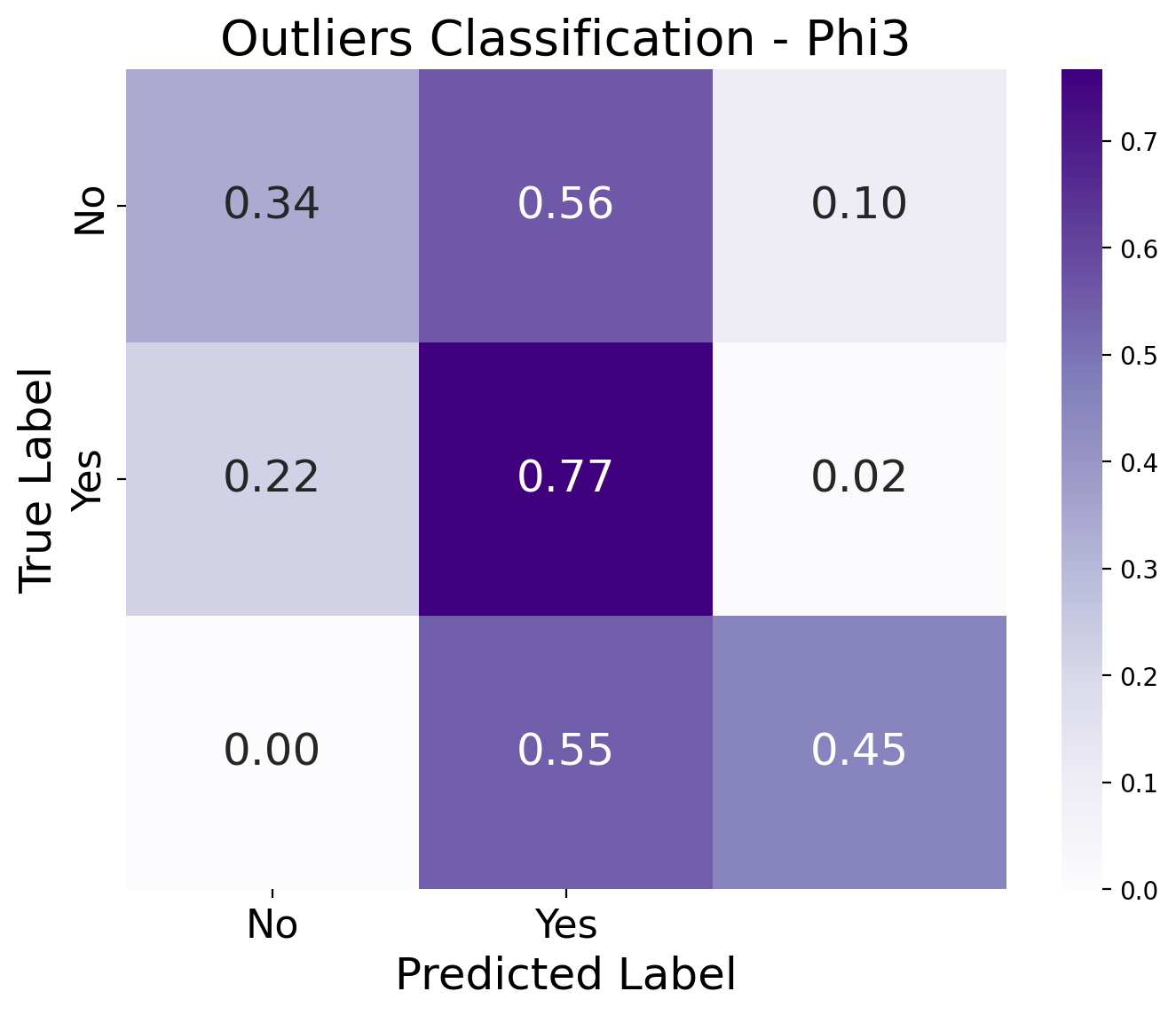}~
    \caption{Anomaly classification}
    \label{fig:enter-label}
\end{figure}

\subsection{Volatility}
\begin{figure}[h!]
    \centering
    \includegraphics[width=0.20\textwidth]{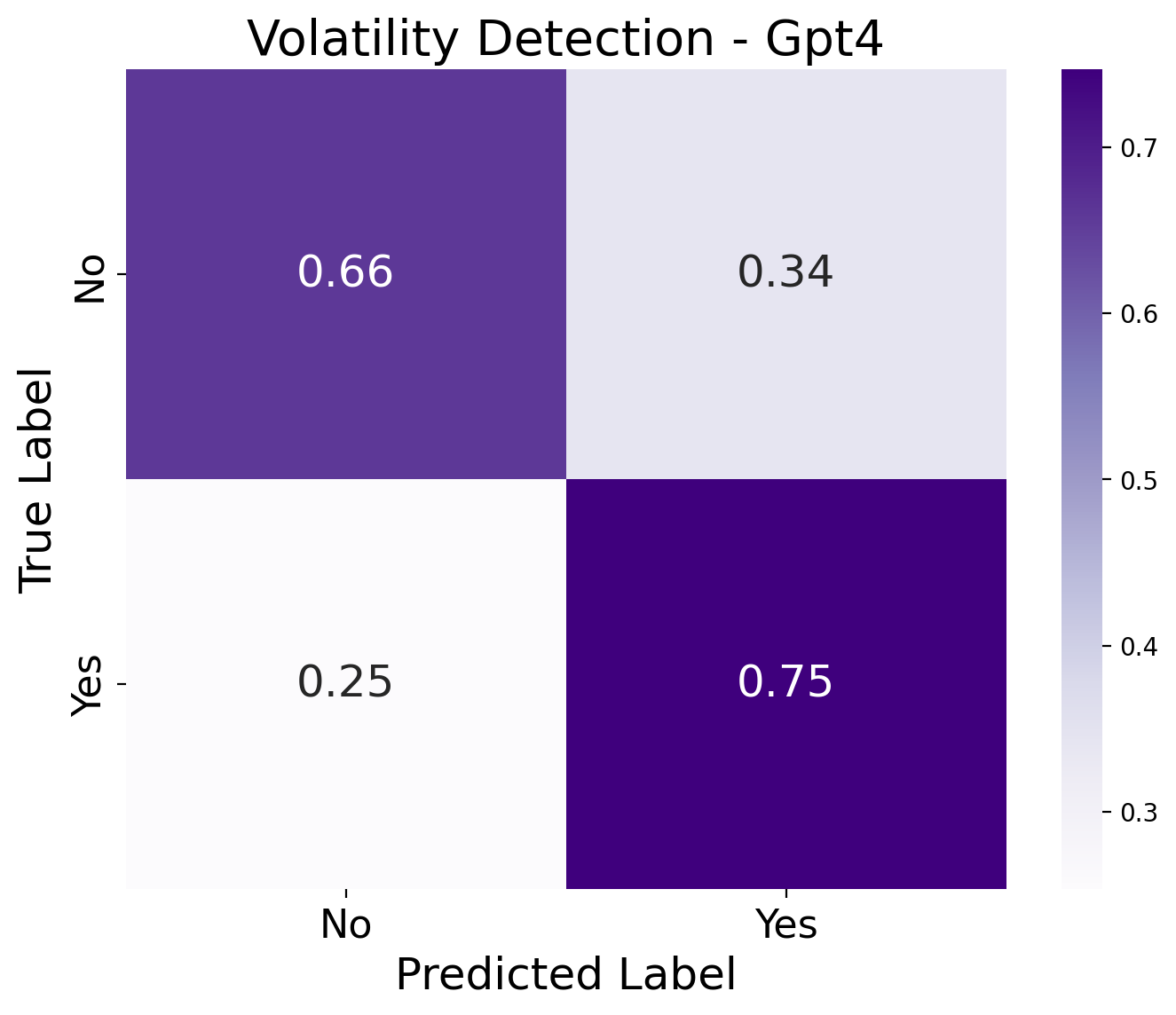}~
    \includegraphics[width=0.20\textwidth]{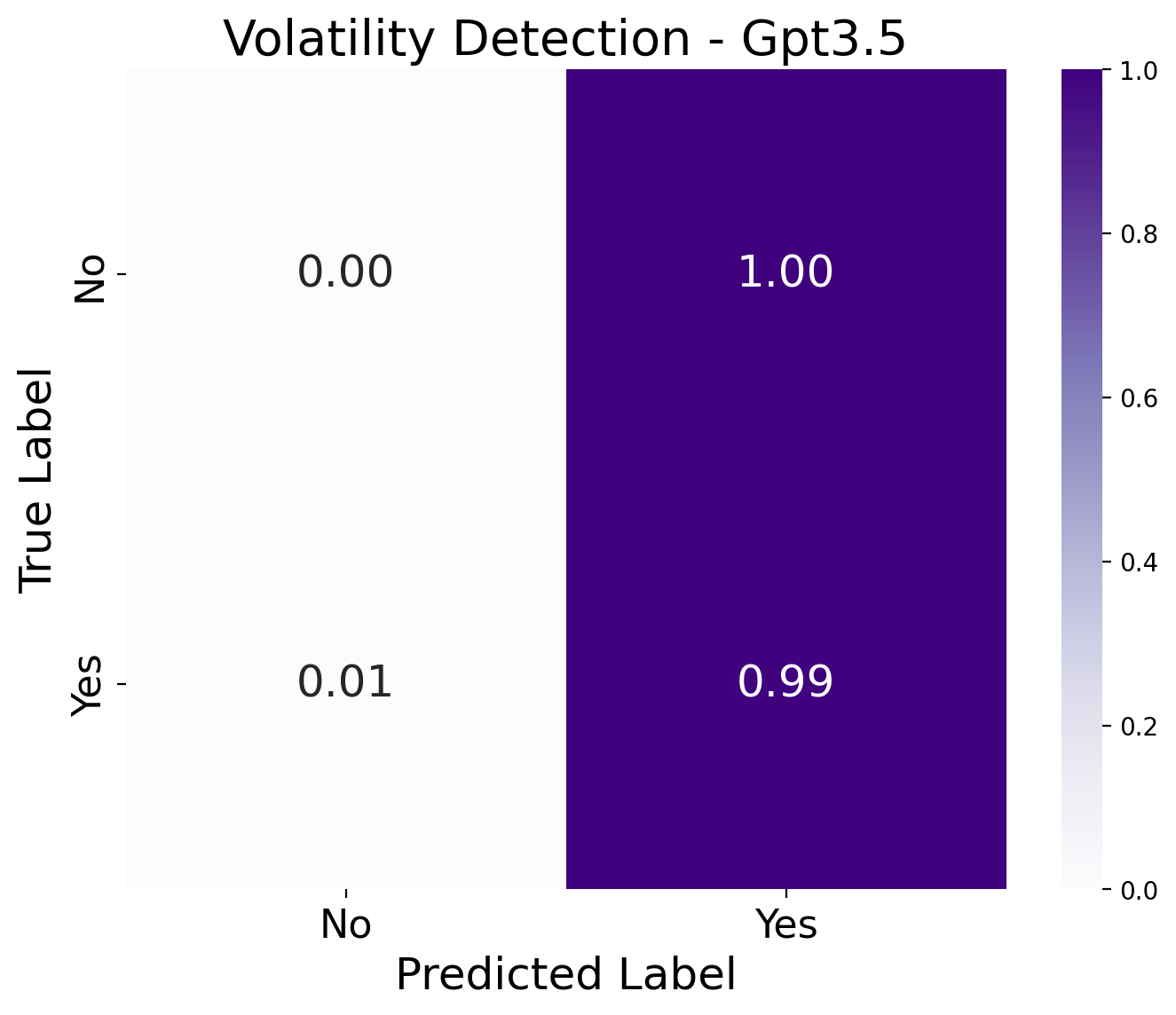}~
    \includegraphics[width=0.20\textwidth]{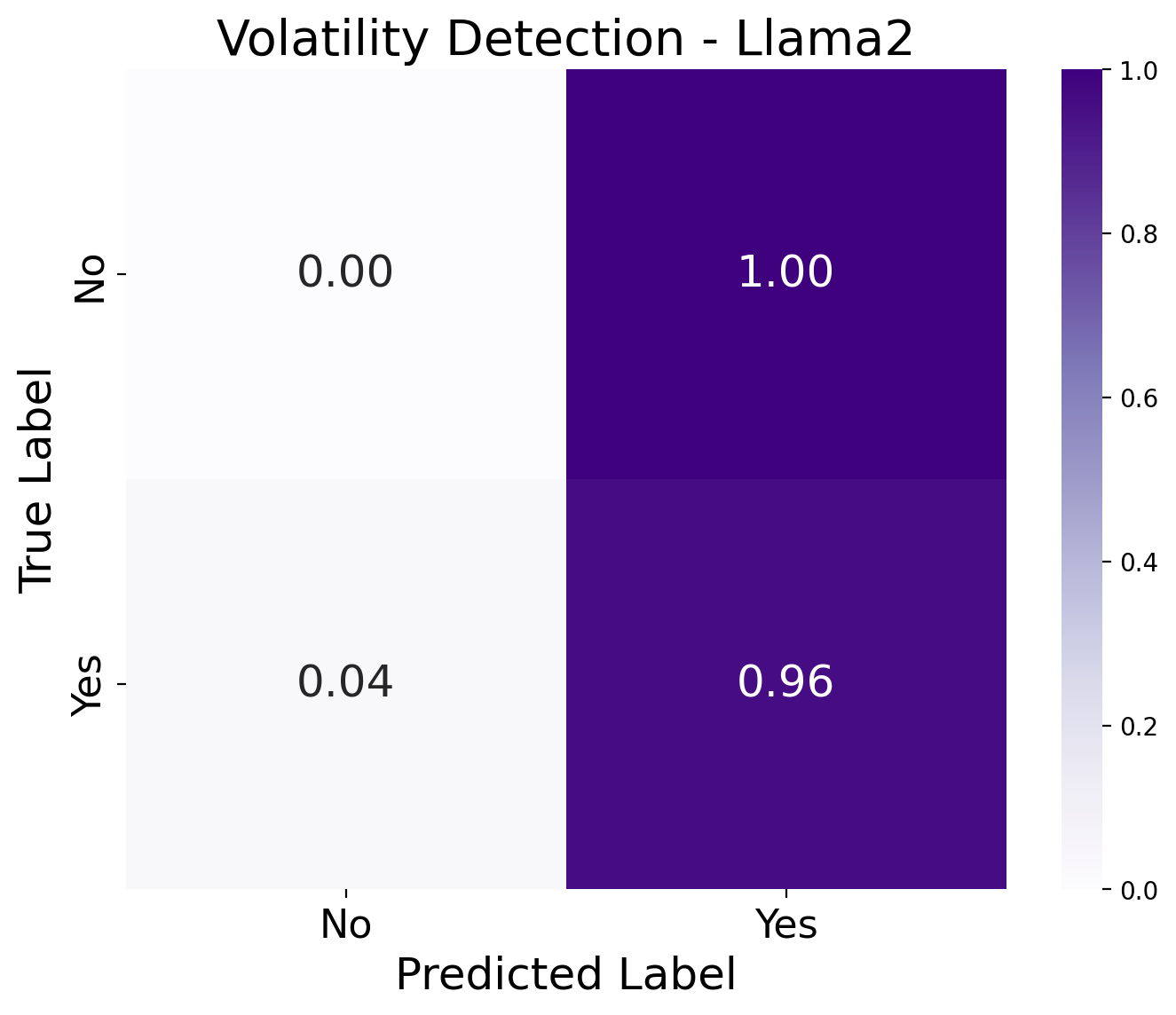}~
    \includegraphics[width=0.20\textwidth]{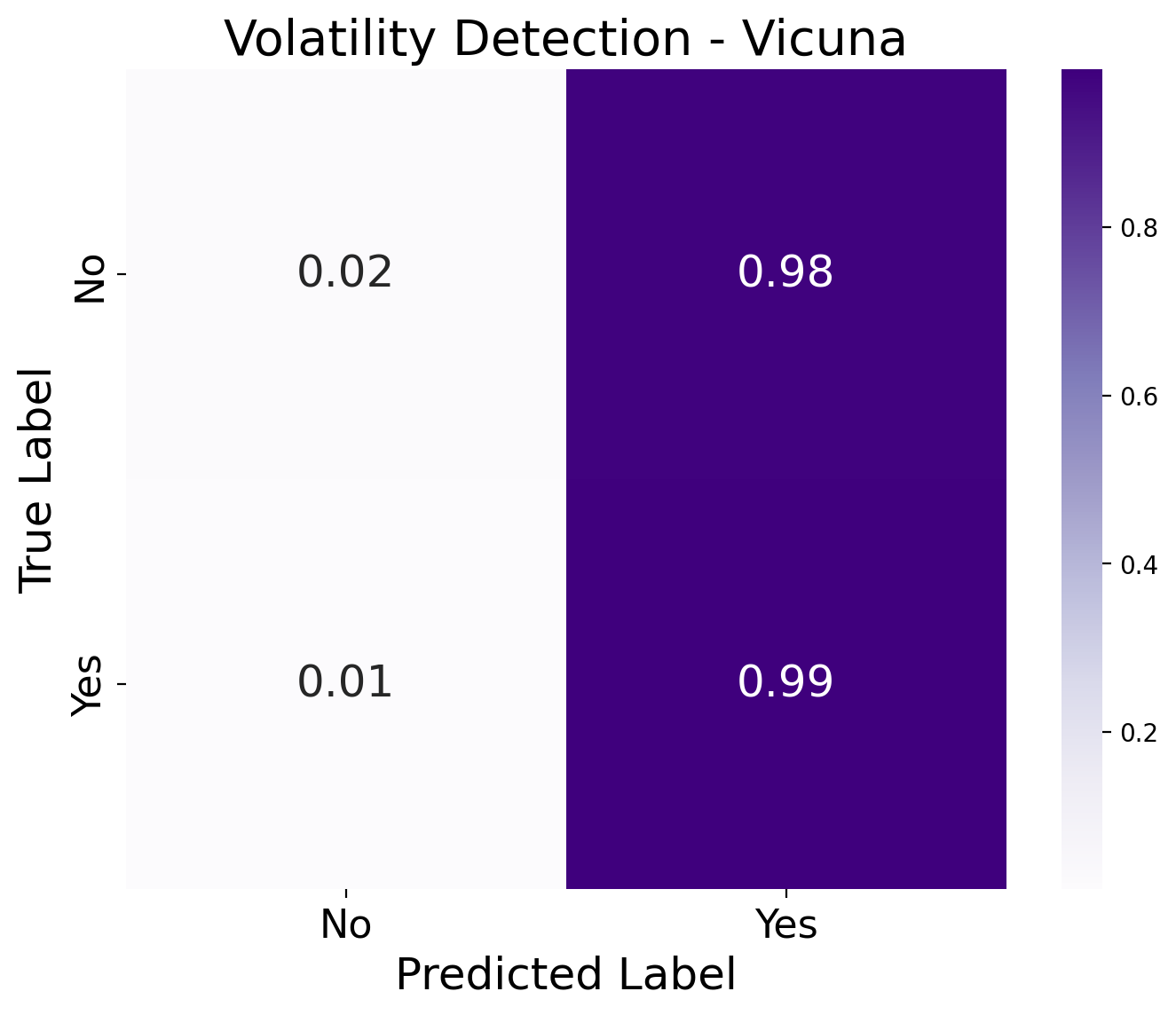}~
    \includegraphics[width=0.20\textwidth]{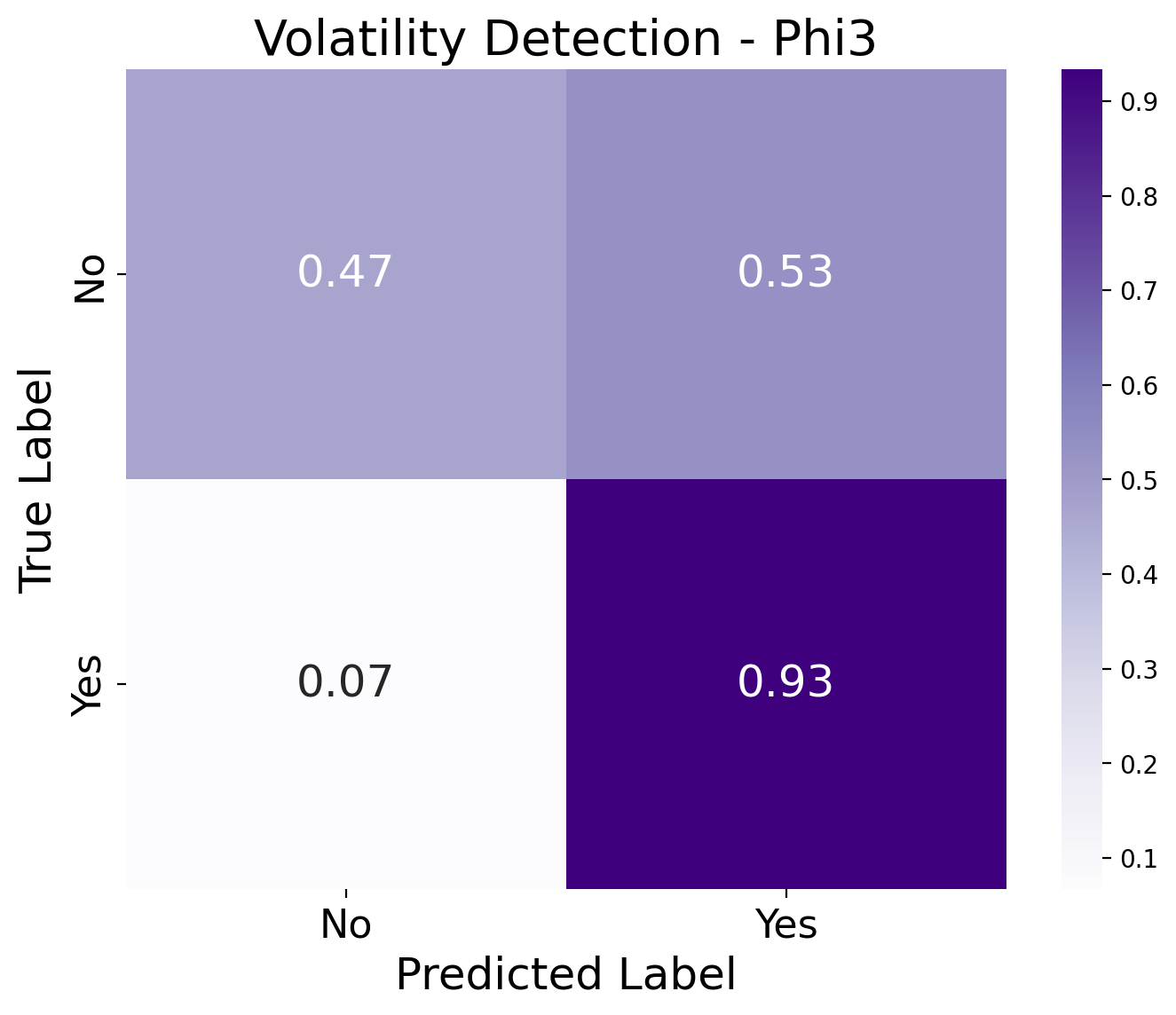}
    \caption{Volatility detection}
    \label{fig:enter-label}
\end{figure}
\begin{figure}[h!]
    \centering
    \includegraphics[width=0.20\textwidth]{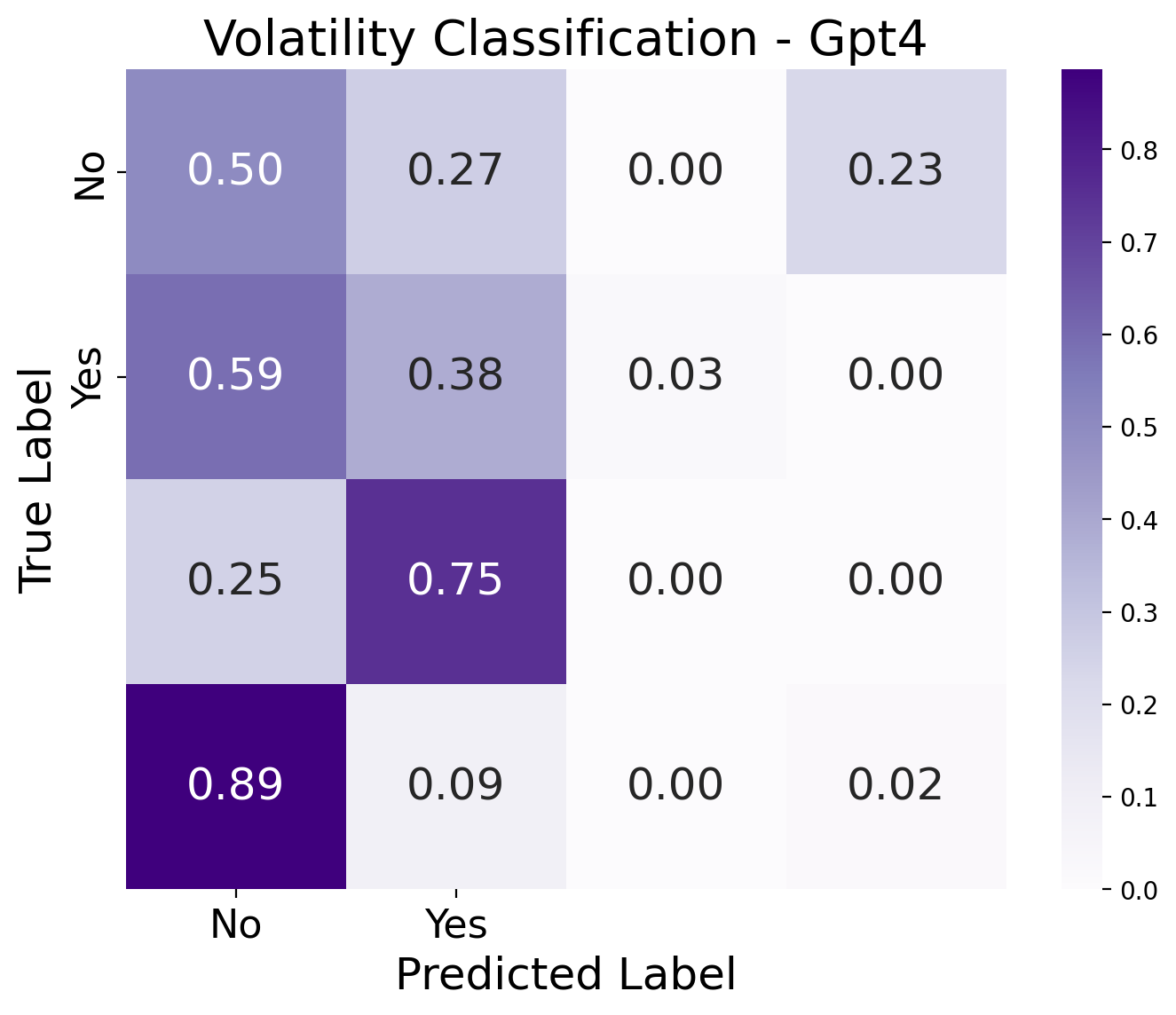}~
    \includegraphics[width=0.20\textwidth]{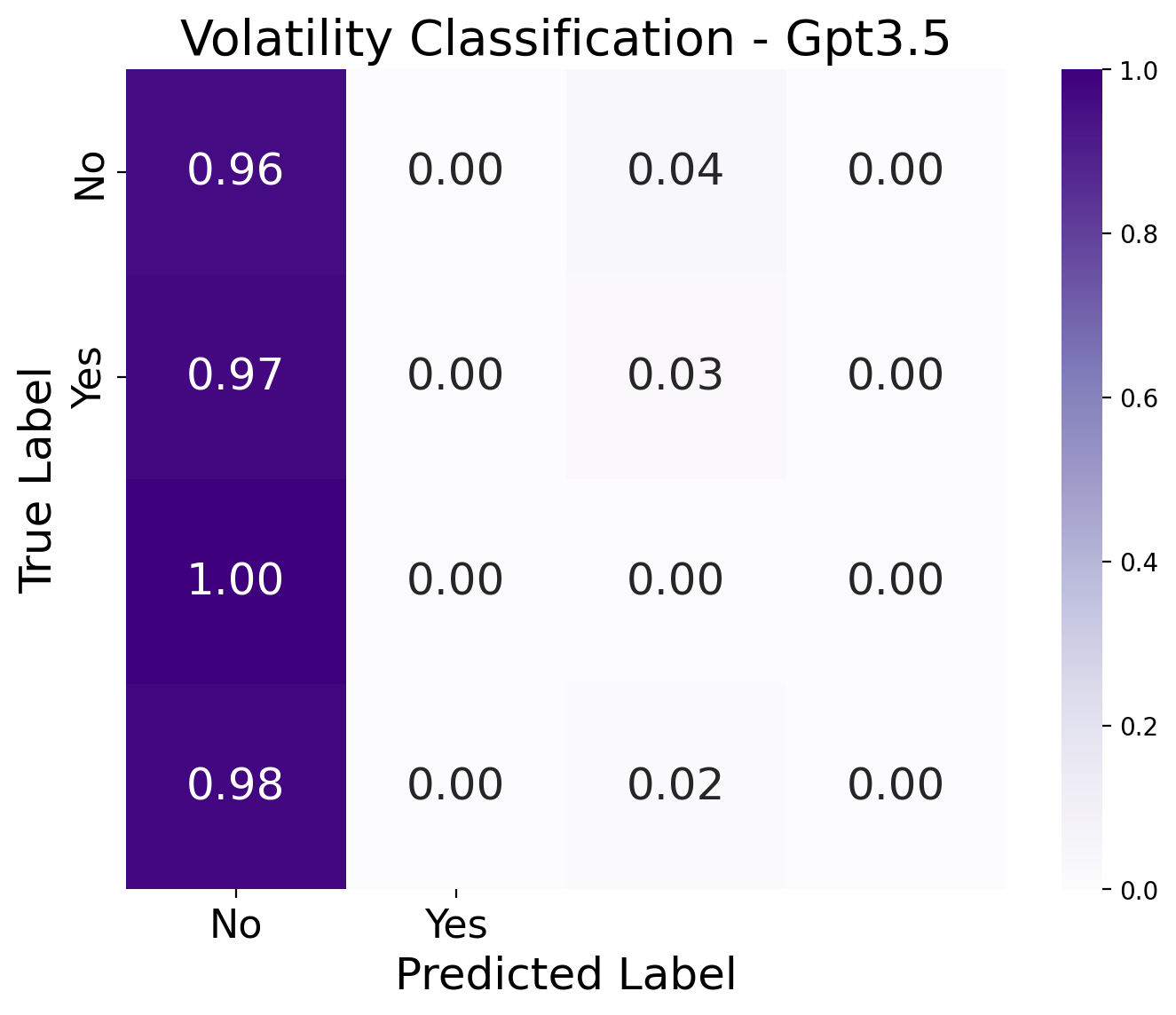}~
    \includegraphics[width=0.20\textwidth]{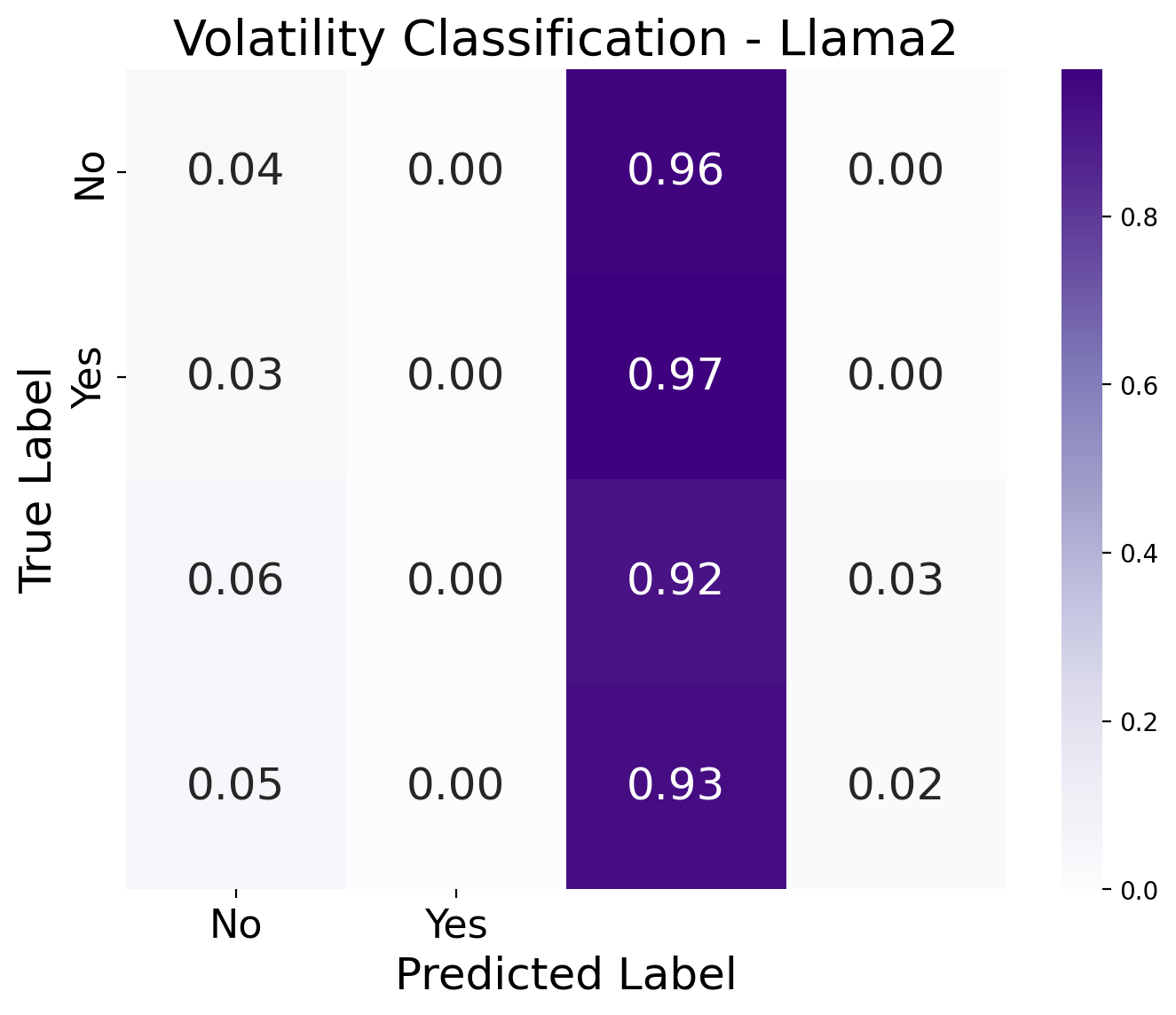}~
    \includegraphics[width=0.20\textwidth]{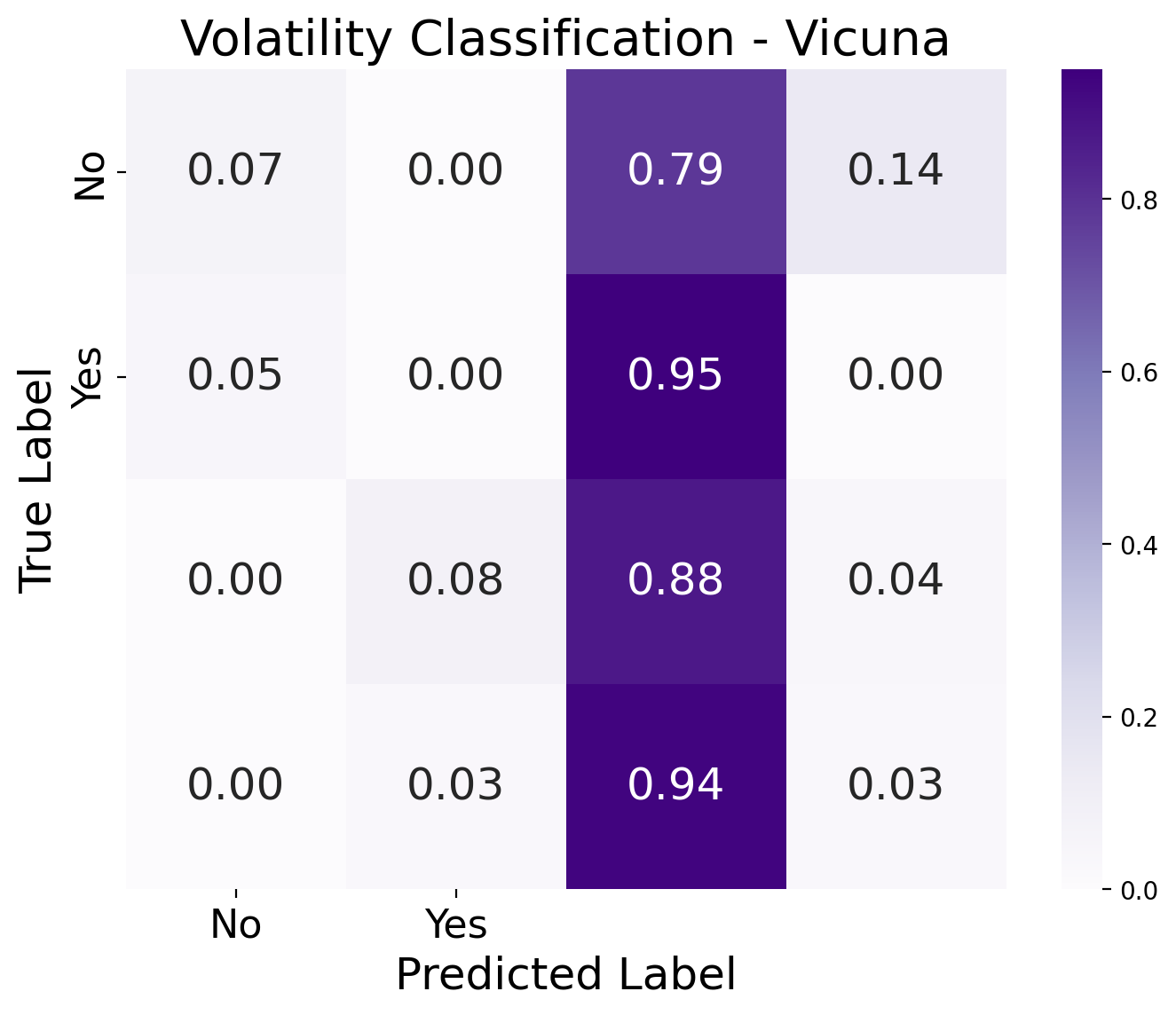}~
    \includegraphics[width=0.20\textwidth]{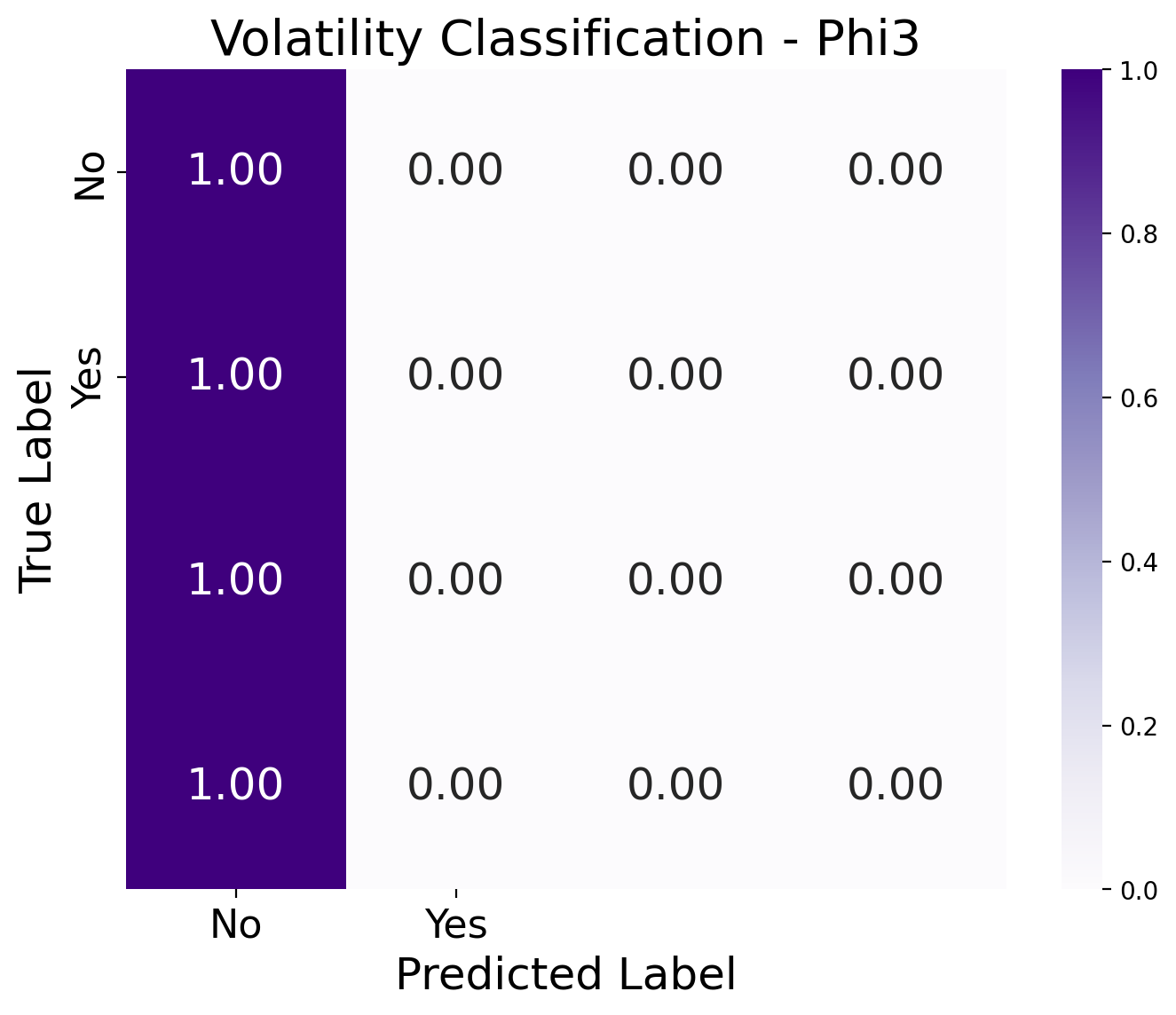}
    \caption{Volatility classification}
    \label{fig:enter-label}
\end{figure}

\newpage
\pagebreak

\begin{section}{Position Bias} \label{position_experiments}

\begin{subsection} {Does the position of the target value affect the performance of identifying its presence in various types of time series data?} \label{goal0_search}
Refer to Figure \ref{fig:position_experiments_G0_search}, which includes a confusion matrix (with `1: yes' indicating presence of the number in the series and `0: no' indicating its absence) and bar plot showing the accuracy in each quadrant for each LLM and type of time series data.

GPT achieves nearly perfect performance across all quadrants and time series types, indicating an absence of position bias in detecting the presence of a number within the time series. 
Llama2 does not exhibit position bias in monotonic series without noise but begins to show position bias as the complexity of the time series increases, such as in monotonic series with noise and sinusoidal series. We believe this bias is also present in Brownian series; however, due to the higher complexity of the dataset, Llama2's performance is poor across all quadrants, making the impact of the bias less discernible.
Vicuna displays superior performance compared to Llama2 across all datasets but continues to exhibit position bias. Notably, this bias appears in most datasets, such as monotonic series without noise, sinusoidal series, and Brownian motion series.

\begin{longtable}{c}
% \centering
% \begin{tabular}{c} % Two columns taking up equal space
% \toprule
\hline
\\
\textbf{GPT 3.5} \\
\begin{minipage}{4cm}
\includegraphics[scale = 0.2]{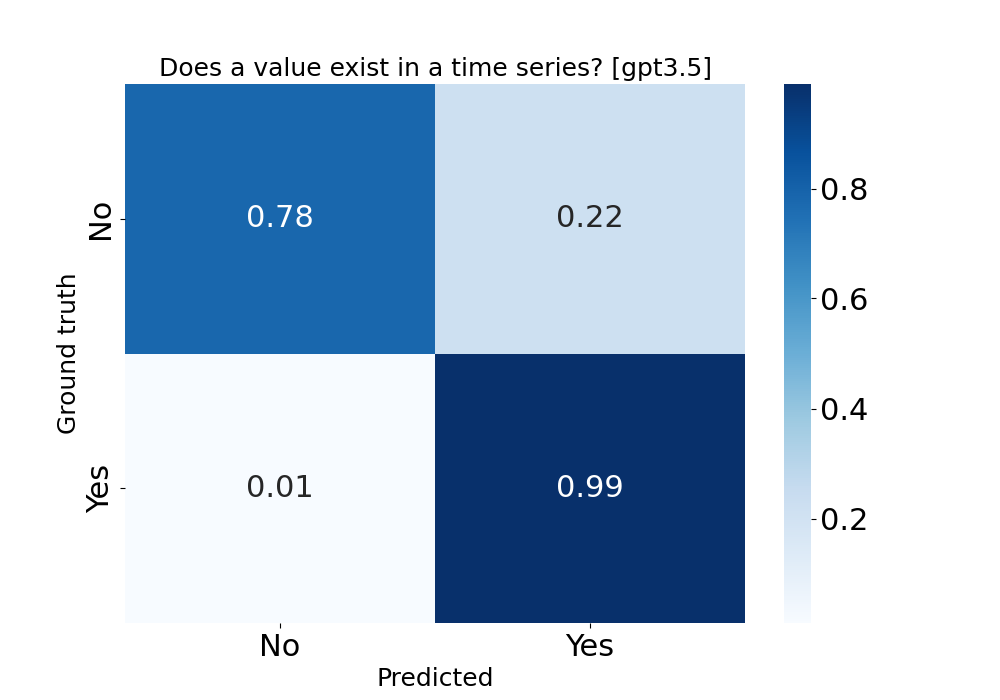}
\centering
\end{minipage}
\begin{minipage}{4cm}
\includegraphics[scale = 0.2]{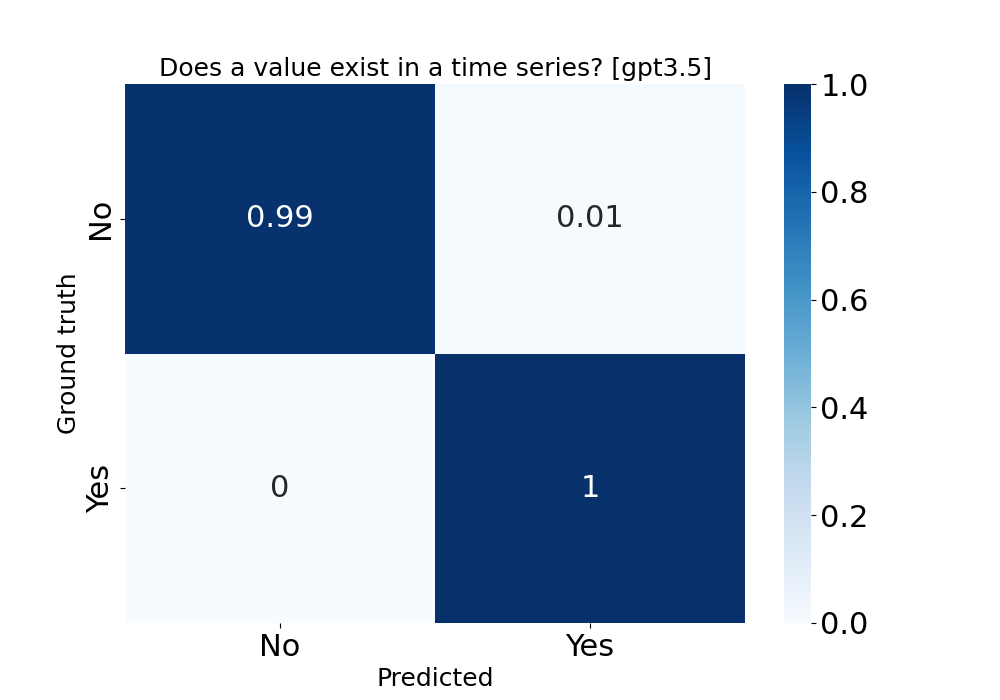}
\centering
\end{minipage}
\begin{minipage}{4cm}
\includegraphics[scale = 0.2]
{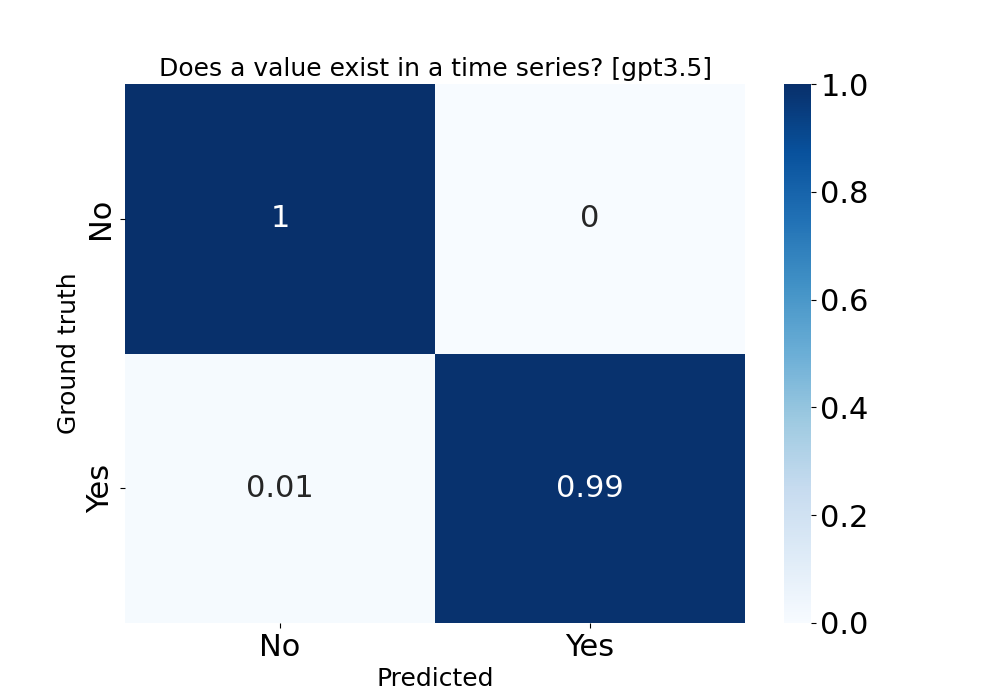}
\centering
\end{minipage}
\begin{minipage}{4cm}
\includegraphics[scale = 0.2]{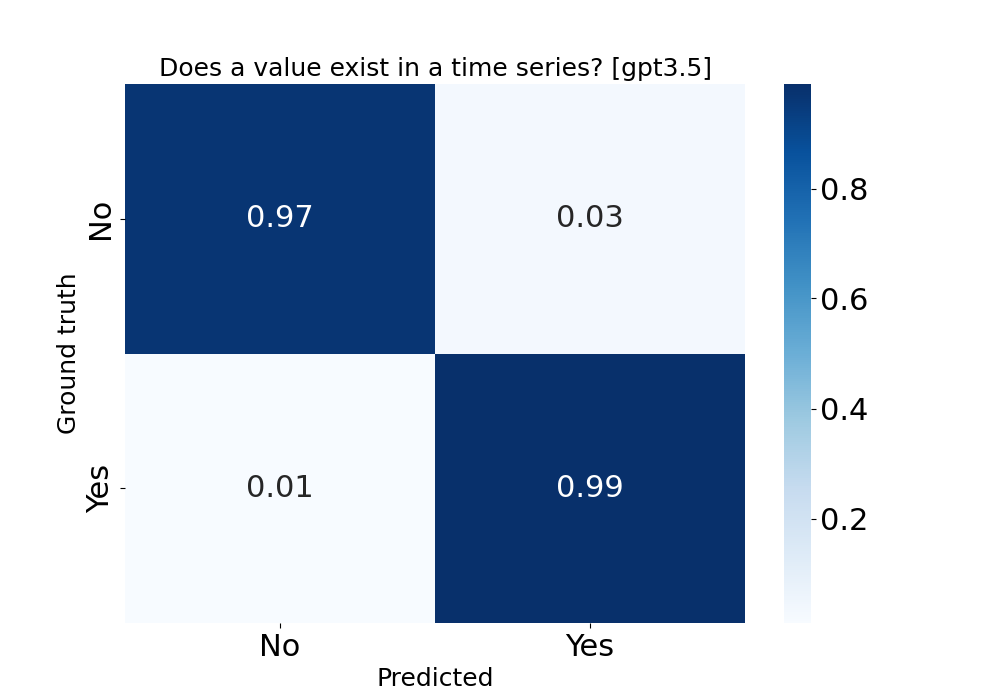}
\centering
\end{minipage}
\\ % End of the first row of images and labels
% \newline
% Text below last image in Block 1 \\
% Text below image in Block 2 \\
\\
\begin{minipage}{4cm}
\includegraphics[scale = 0.25]{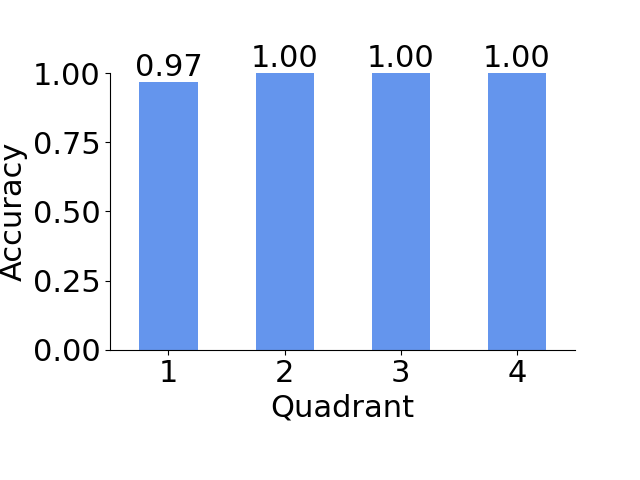}
\centering
\\
(a) Monotonic (no noise)
\end{minipage}
\begin{minipage}{4cm}
\includegraphics[scale = 0.25]
{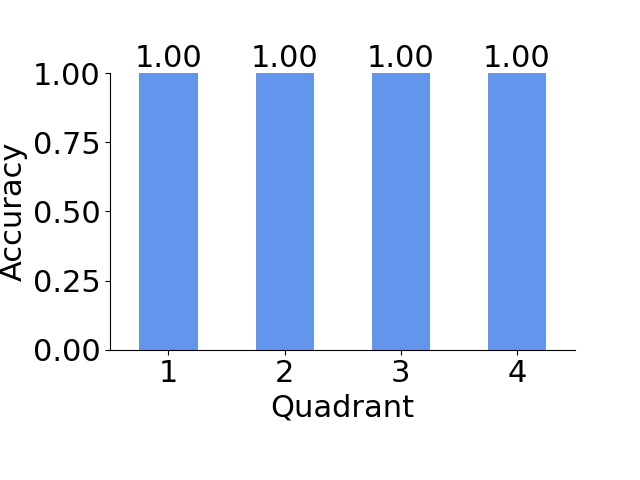}
(b) Monotonic with noise
\end{minipage}
\begin{minipage}{4cm}
\includegraphics[scale = 0.25]{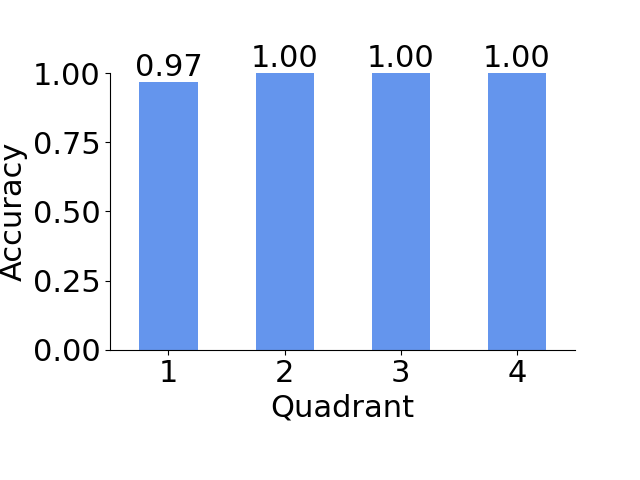}
\centering
(c) Sinusoidal
\end{minipage}
\begin{minipage}{4cm}
\includegraphics[scale = 0.25]{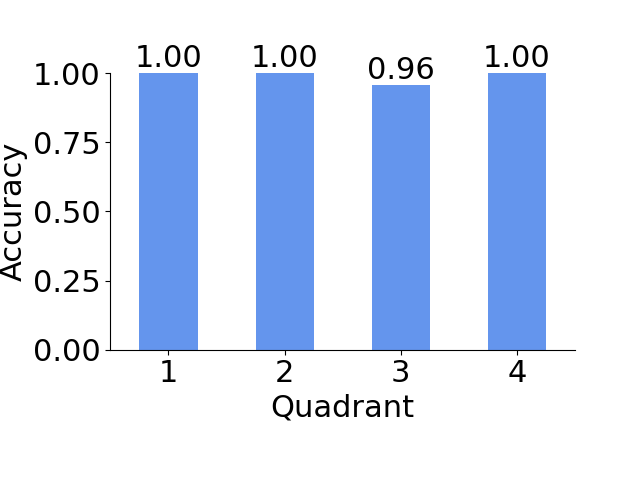}
\centering
(d) Brownian motion
\end{minipage}
\\ % End of the first row of images and labels
% \newline
% Text below last image in Block 1 \\
% Text below image in Block 2 \\
\midrule
\\
\textbf{Llama2} \\
\begin{minipage}{4cm}
\includegraphics[scale = 0.2]{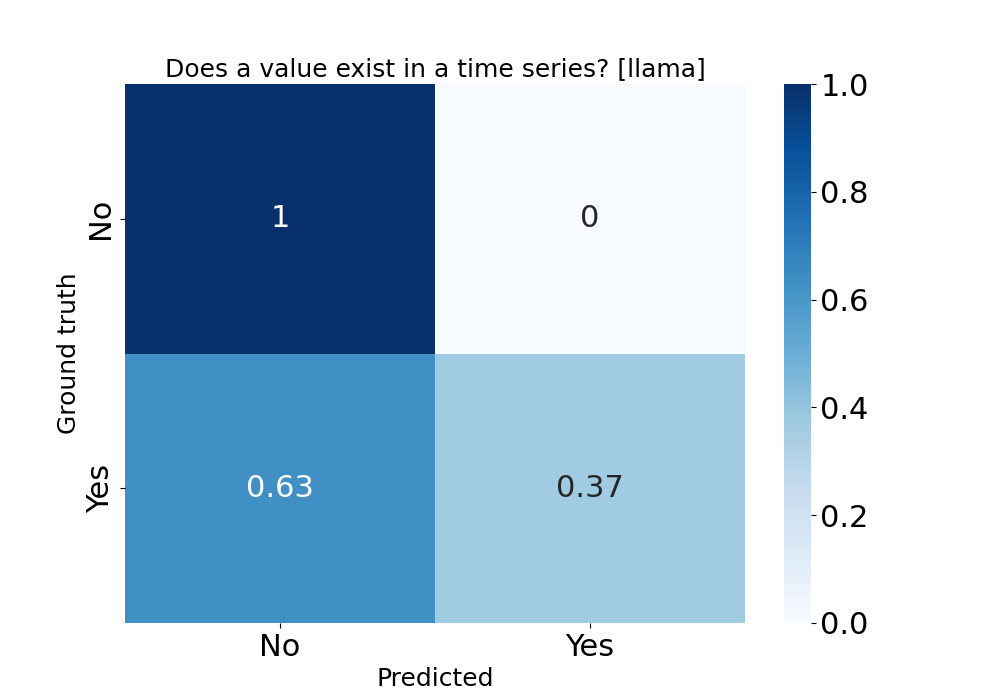}
\centering
\end{minipage}
\begin{minipage}{4cm}
\includegraphics[scale = 0.2]{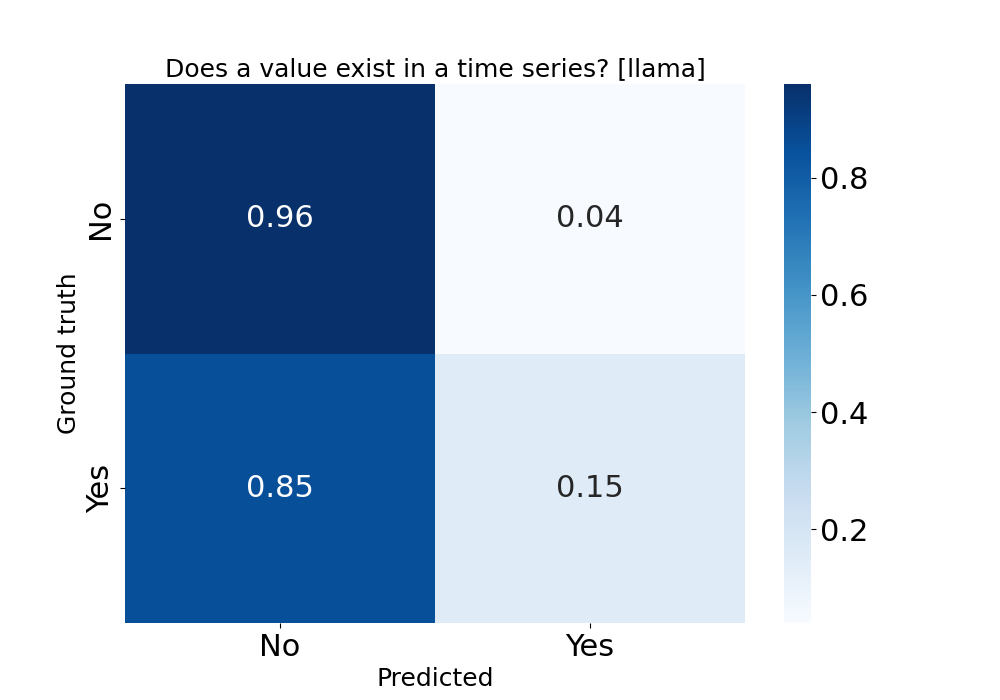}
\centering
\end{minipage}
\begin{minipage}{4cm}
\includegraphics[scale = 0.2]{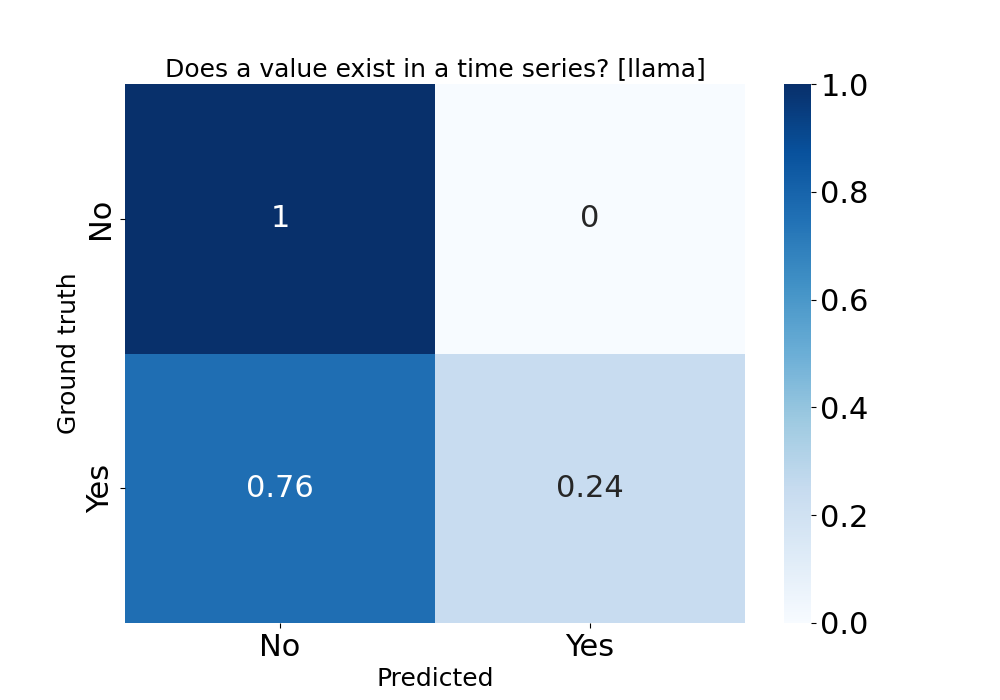}
\centering
\end{minipage}
\begin{minipage}{4cm}
\includegraphics[scale = 0.2]{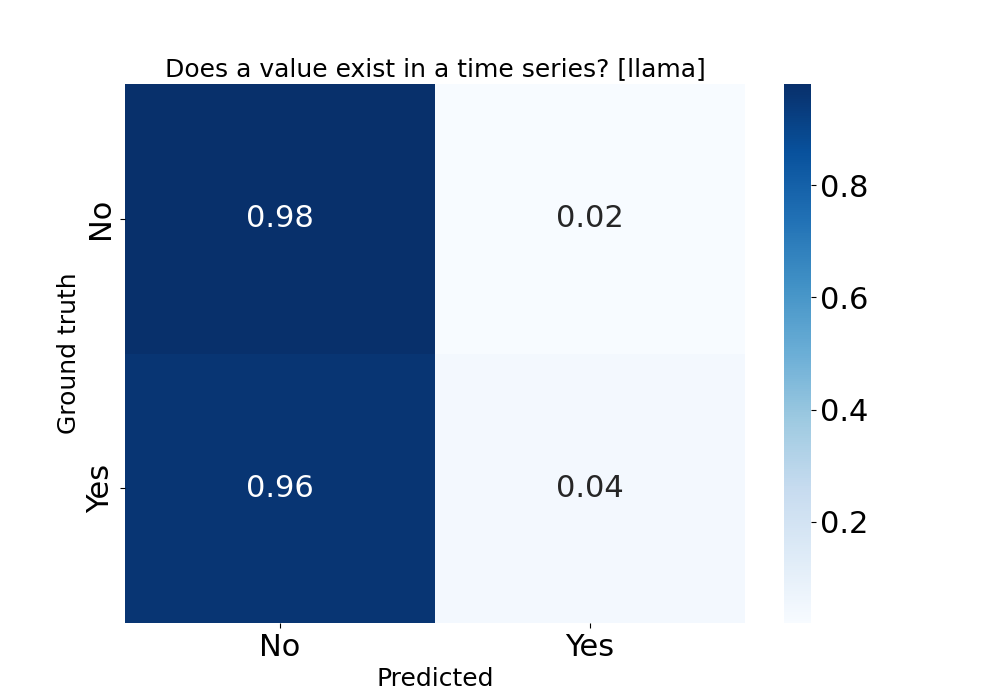}
\centering
\end{minipage}
\\ % End of the first row of images and labels
% \newline
% Text below last image in Block 1 \\
% Text below image in Block 2 
\begin{minipage}{4cm}
\includegraphics[scale = 0.25]{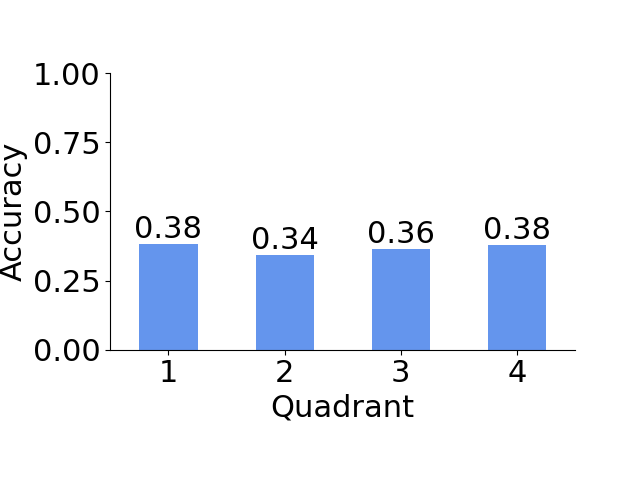}
\centering
(a) Monotonic (no noise)
\end{minipage}
\begin{minipage}{4cm}
\includegraphics[scale = 0.25]{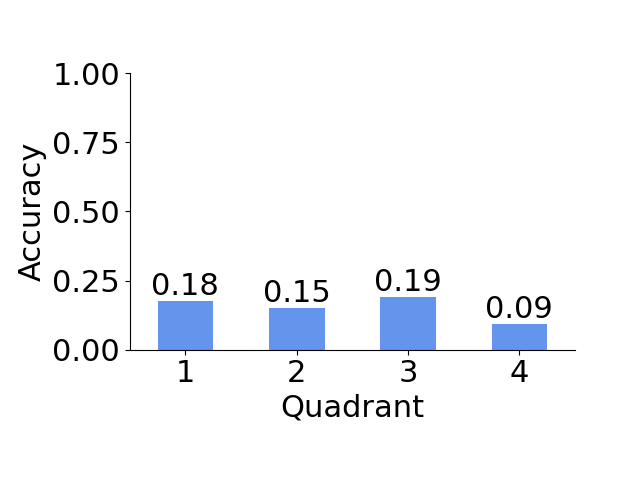}
\centering
(b) Monotonic with noise
\end{minipage}
\begin{minipage}{4cm}
\includegraphics[scale = 0.25]{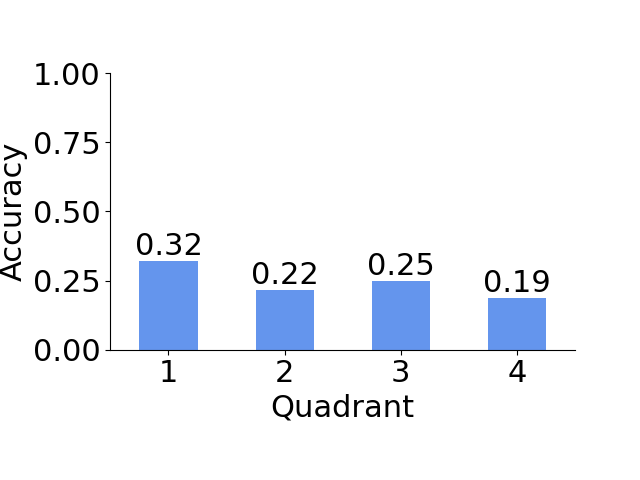}
\centering
(c) Sinusoidal
\end{minipage}
\begin{minipage}{4cm}
\includegraphics[scale = 0.25]{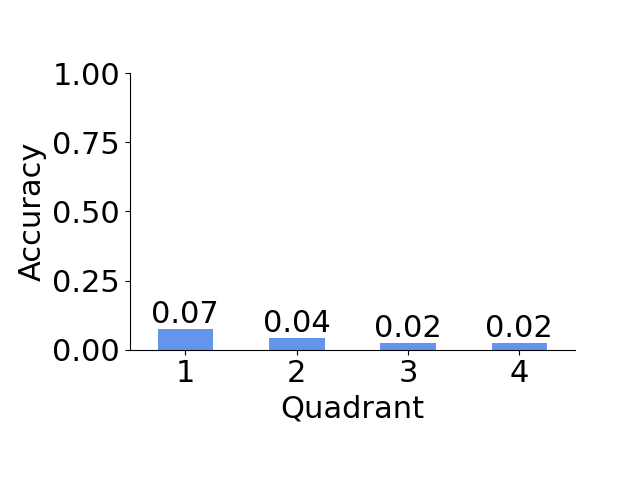}
\centering
(d) Brownian motion
\end{minipage}
\\ % End of the first row of images and labels
% \newline
% Text below last image in Block 1 \\
% Text below image in Block 2 \\
\midrule
\\
\textbf{Vicuna} \\
\begin{minipage}{4cm}
\includegraphics[scale = 0.2]{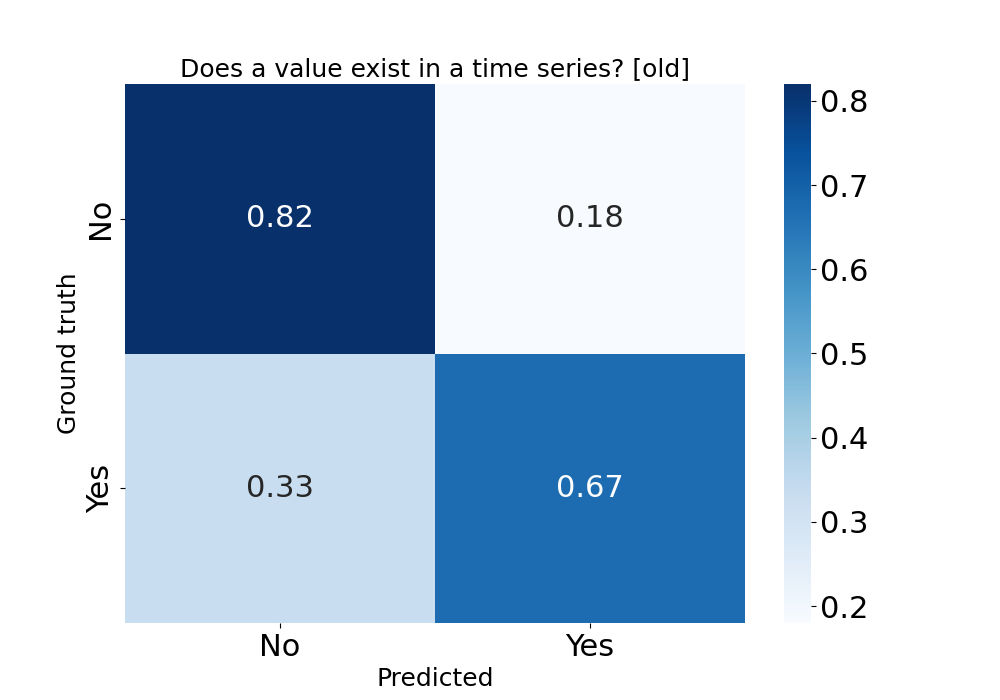}
\centering
\end{minipage}
\begin{minipage}{4cm}
\includegraphics[scale = 0.2]{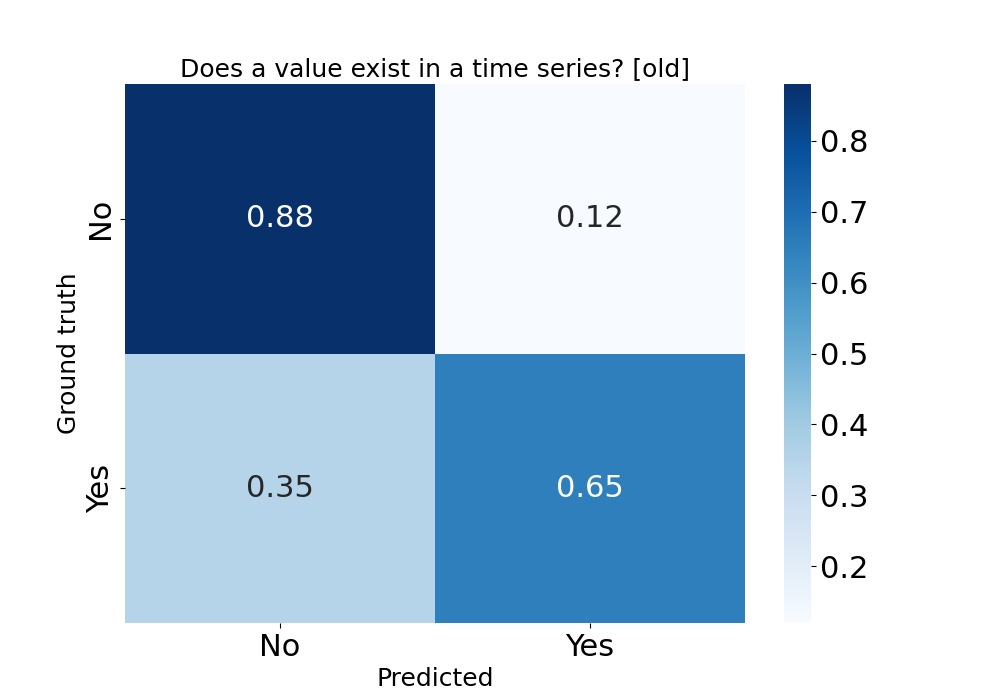}
\centering
\end{minipage}
\begin{minipage}{4cm}
\includegraphics[scale = 0.2]{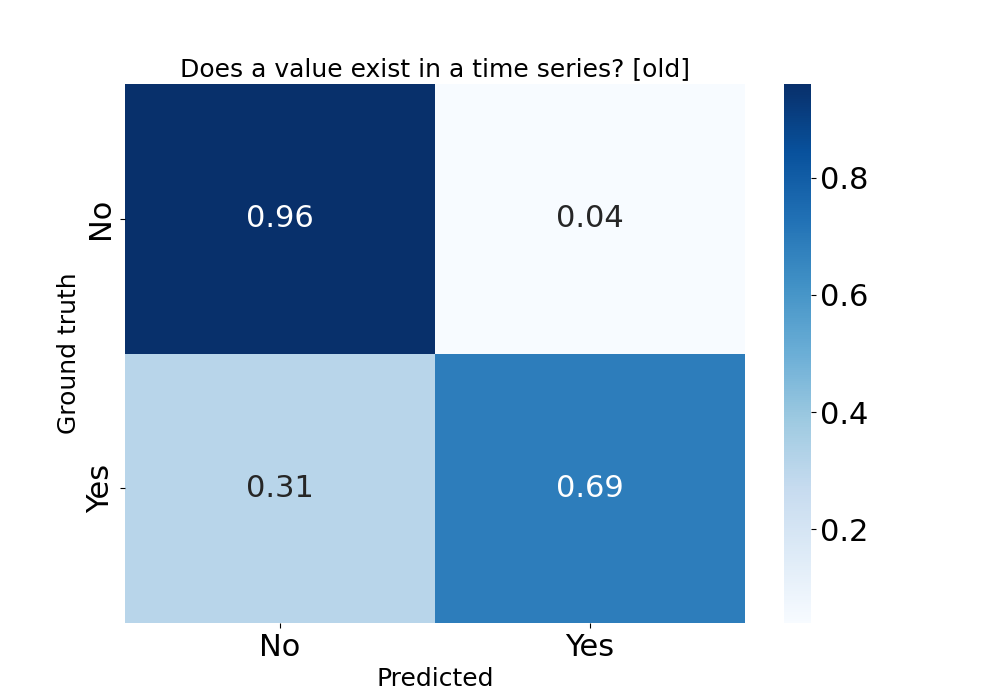}
\centering
\end{minipage}
\begin{minipage}{4cm}
\includegraphics[scale = 0.2]{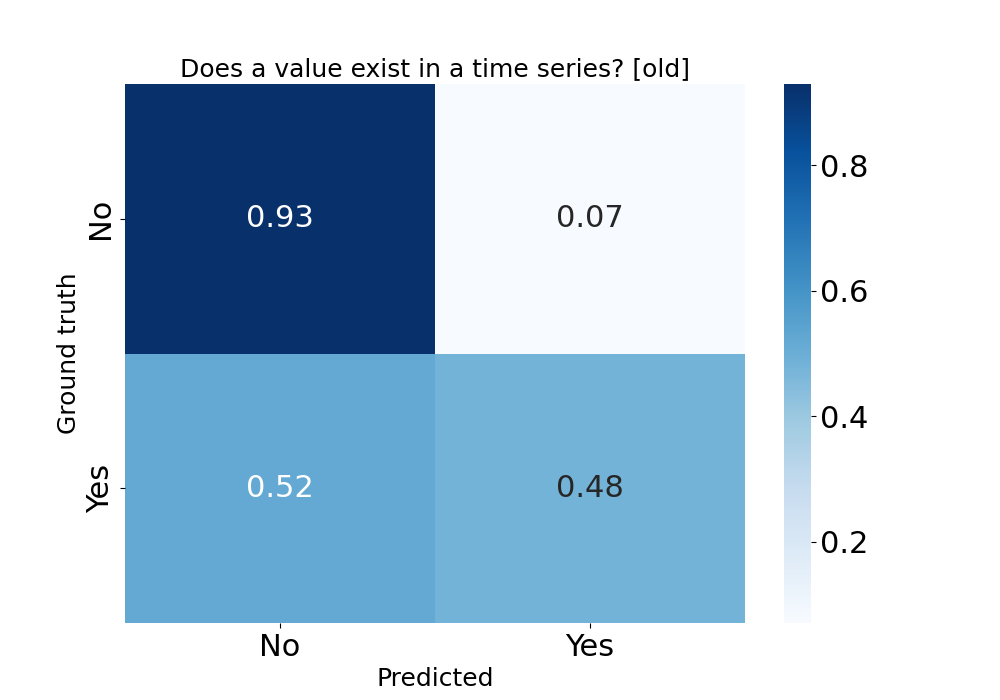}
\centering
\end{minipage}
\\ % End of the first row of images and labels
% \newline
% Text below last image in Block 1 \\
% Text below image in Block 2 \\
\begin{minipage}{4cm}
\includegraphics[scale = 0.25]{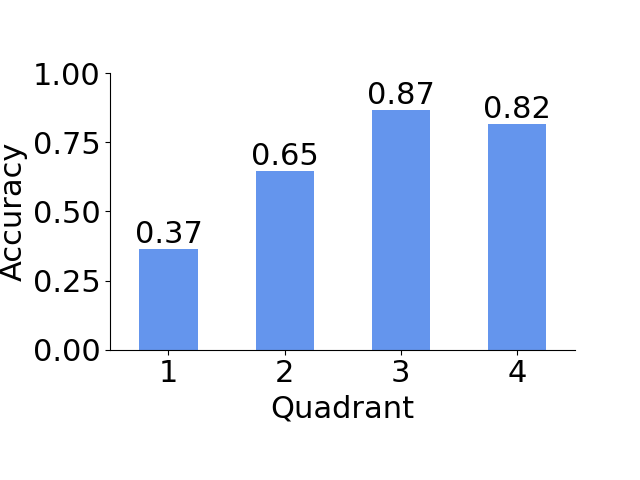}
\centering
(a) Monotonic (no noise)
\end{minipage}
\begin{minipage}{4cm}
\includegraphics[scale = 0.25]{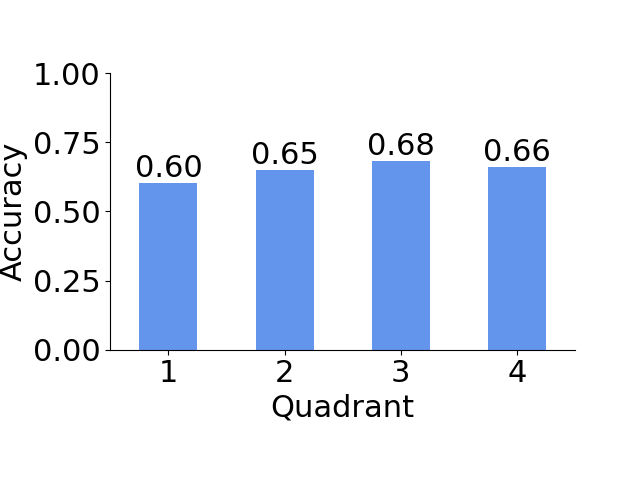}
\centering
(b) Monotonic with noise
\end{minipage}
\begin{minipage}{4cm}
\includegraphics[scale = 0.25]{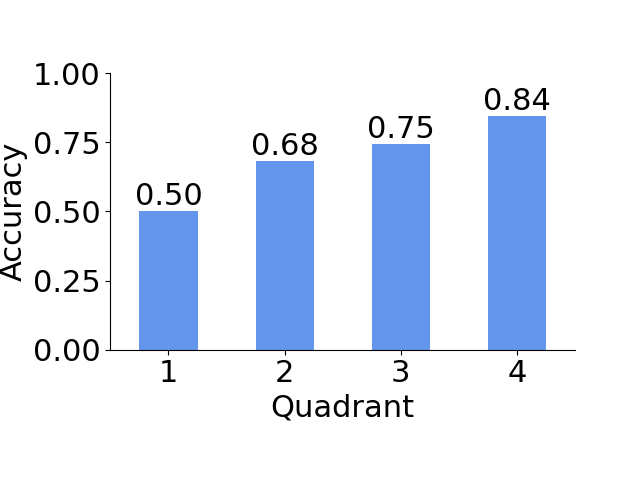}
\centering
(c) Sinusoidal
\end{minipage}
\begin{minipage}{4cm}
\includegraphics[scale = 0.25]{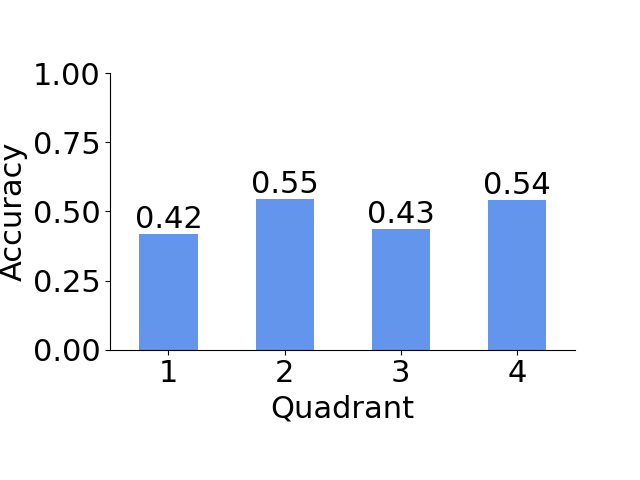}
\centering
(d) Brownian motion
\end{minipage}
% \hline
\\
\bottomrule
% \end{tabular}
\caption{Confusion matrix and accuracy by quadrant for the search task}
\label{fig:position_experiments_G0_search}
\end{longtable}
    
\end{subsection}

%Goal 2
\begin{subsection}{Does the position impact the retrieval performance for a specific date's value from time series data?}\label{goal1_retrieval}
Refer to Figure \ref{fig:position_experiments_retrieval} for bar plots that illustrate the accuracy across each quadrant. 

Once again, GPT achieves nearly perfect performance across all quadrants and time series types, suggesting no position bias in the retrieval task either. Similar to the findings in \ref{goal0_search}, Vicuna outperforms Llama2. Moreover, both Vicuna and Llama2 exhibit position bias in most datasets, including monotonic series both with and without noise, and sinusoidal series.

\begin{longtable}{c}
% \centering
% \begin{tabular}{c} % Two columns taking up equal space
% \toprule
\hline
\\
\textbf{GPT 3.5} \\
\begin{minipage}{4cm}
\includegraphics[scale = 0.25]{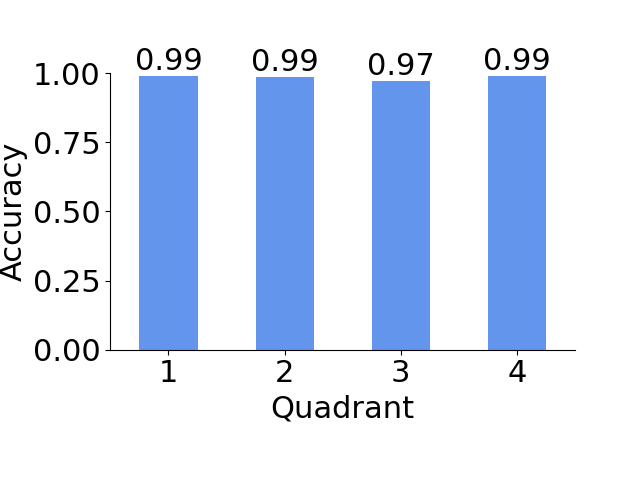}
\centering
\\
(a) Monotonic (no noise)
\end{minipage}
\begin{minipage}{4cm}
\includegraphics[scale = 0.25]
{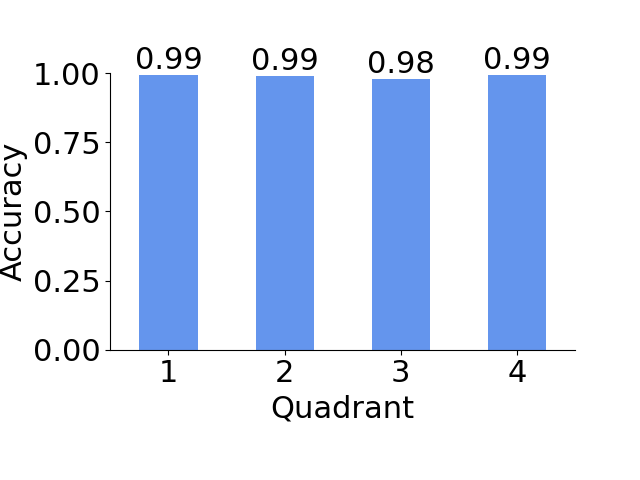}
\centering
(b) Monotonic with noise
\end{minipage}
\begin{minipage}{4cm}
\includegraphics[scale = 0.25]{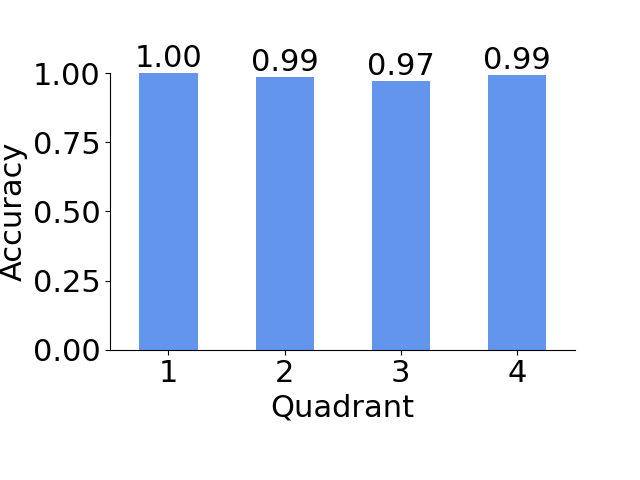}
\centering
(c) Spikes 
\end{minipage}
\begin{minipage}{4cm}
\includegraphics[scale = 0.25]{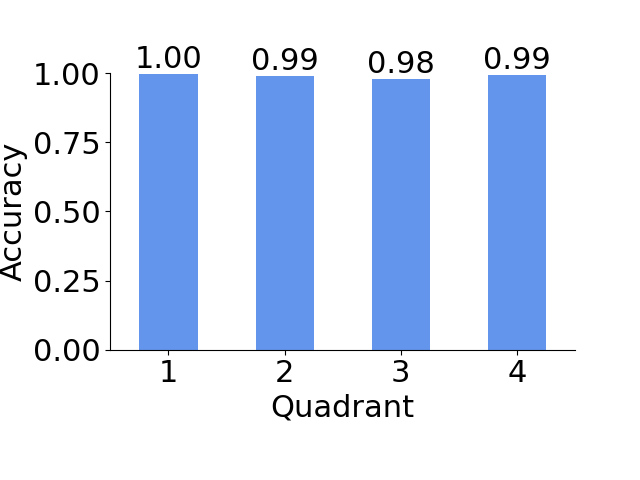}
\centering
(d) Brownian motion
\end{minipage}
\\ % End of the first row of images and labels
% \newline
% Text below last image in Block 1 \\
% Text below image in Block 2 \\
\midrule
\\
\clearpage
\textbf{Llama2} \\
\begin{minipage}{4cm}
\includegraphics[scale = 0.25]{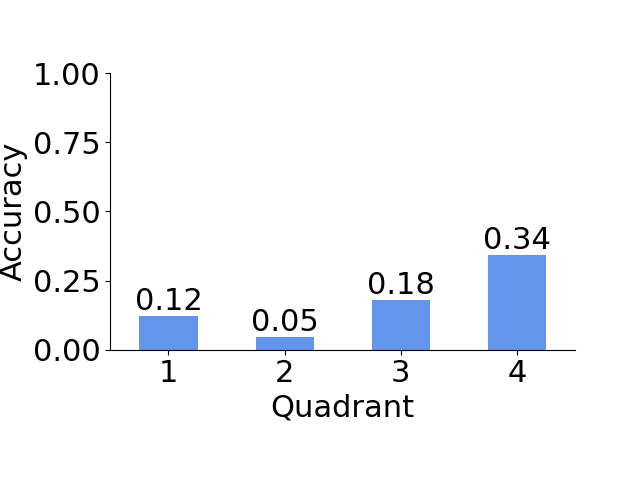}
\centering
(a) Monotonic (no noise)
\end{minipage}
\begin{minipage}{4cm}
\includegraphics[scale = 0.25]{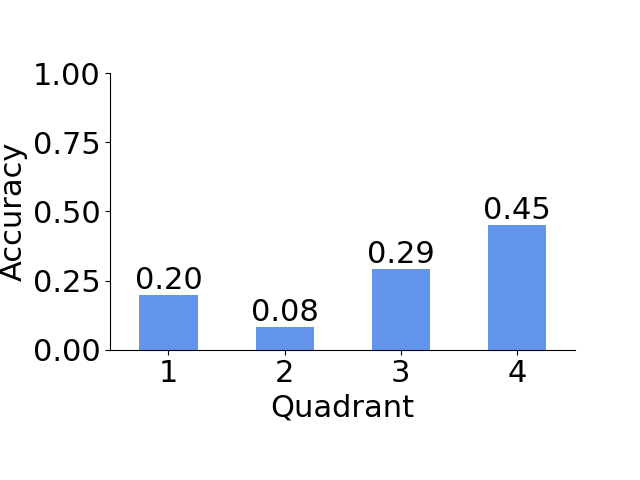}
\centering
(b) Monotonic with noise
\end{minipage}
\begin{minipage}{4cm}
\includegraphics[scale = 0.25]{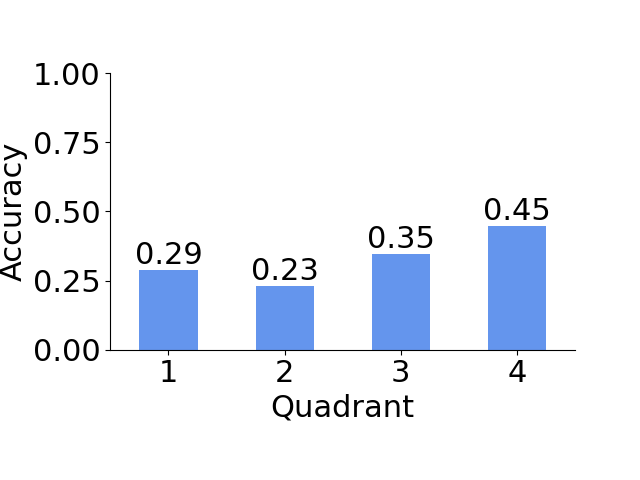}
\centering
(c) Spikes 
\end{minipage}
\begin{minipage}{4cm}
\includegraphics[scale = 0.25]{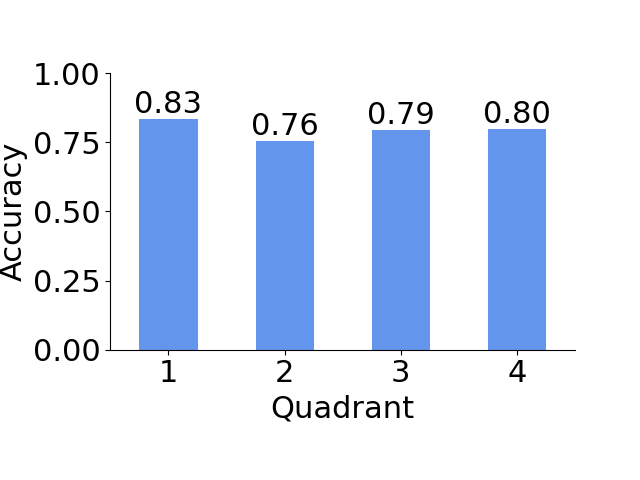}
\centering
(d) Brownian motion
\end{minipage}
\\
\midrule
\\
%%%%% G2 Vicuna is all remaining %%%%%%%%%%% 
\textbf{Vicuna} \\
\begin{minipage}{4cm}
\includegraphics[scale = 0.25]{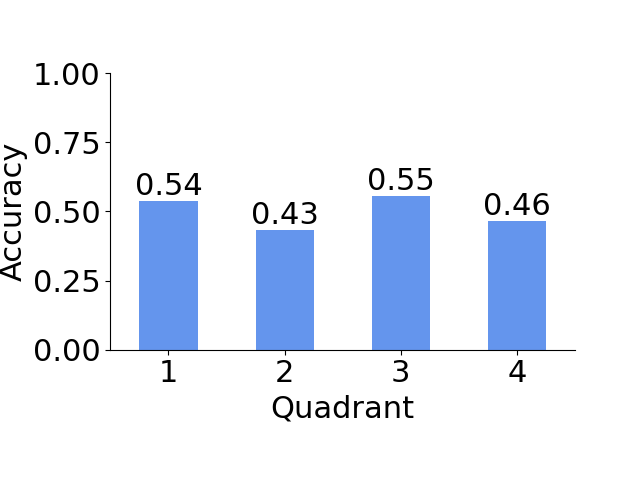}
\centering
(a) Monotonic (no noise)
\end{minipage}
\begin{minipage}{4cm}
\includegraphics[scale = 0.25]{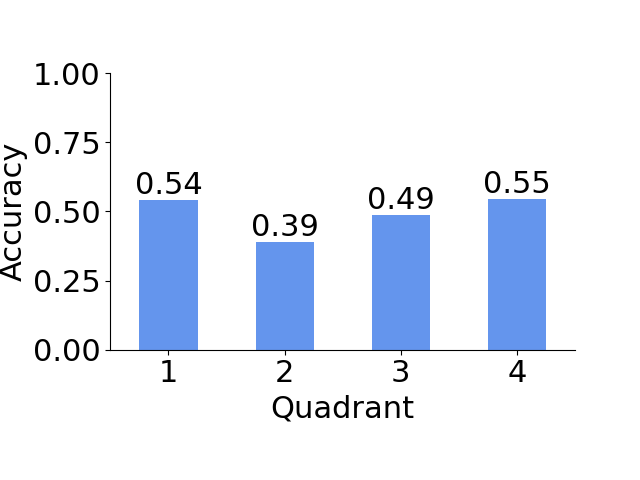}
\centering
(b) Monotonic with noise
\end{minipage}
\begin{minipage}{4cm}
\includegraphics[scale = 0.25]{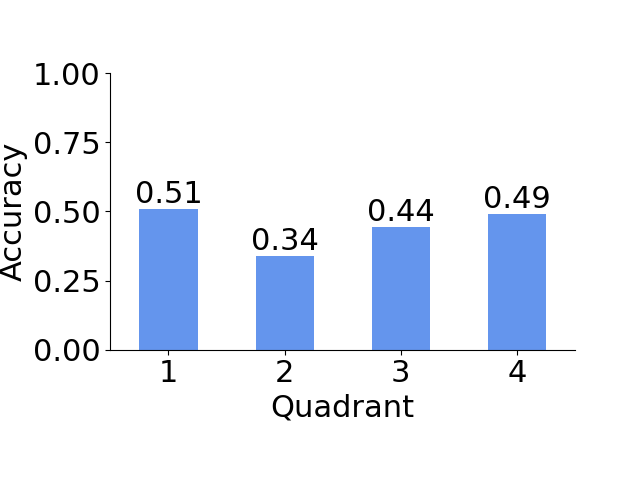}
\centering
(c) Spikes 
\end{minipage}
\begin{minipage}{4cm}
\includegraphics[scale = 0.25]{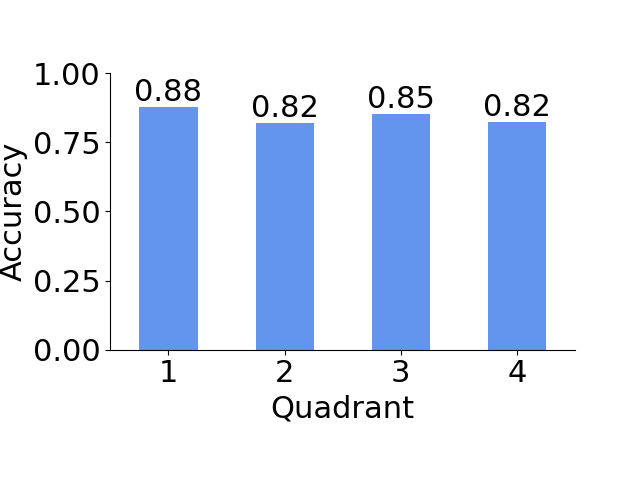}
\centering
(d) Brownian motion
\end{minipage}
\\ % End of the first row of images and labels
% \newline
% Text below last image in Block 1 \\
% Text below image in Block 2 \\
% \hline
\bottomrule
% \end{tabular}
\caption{Confusion matrix and accuracy by quadrant for the retrieval task}
\label{fig:position_experiments_retrieval}
\end{longtable}

\end{subsection}

%Goal 3
\begin{subsection}{Does the position impact the efficiency of identifying minimum and maximum values in different types of time series data?}\label{goal2_min_max}
Refer to Figure \ref{fig:position_experiments_min_max} for bar charts illustrating the accuracy distribution across quadrants. 

For the first time, GPT models show position bias in the spikes dataset, attributed to the increased complexity of the task, which involves arithmetic reasoning. Llama2 exhibits position bias in most datasets, notably in monotonic series with noise, spikes, and Brownian motion series. Vicuna also demonstrates position bias in most datasets, including monotonic series both with and without noise, as well as spikes series.

\begin{longtable}{c}
% \centering
% \begin{tabular}{c} % Two columns taking up equal space
% \toprule
\hline
\\
\textbf{GPT 3.5} \\
\begin{minipage}{4cm}
\includegraphics[scale = 0.25]{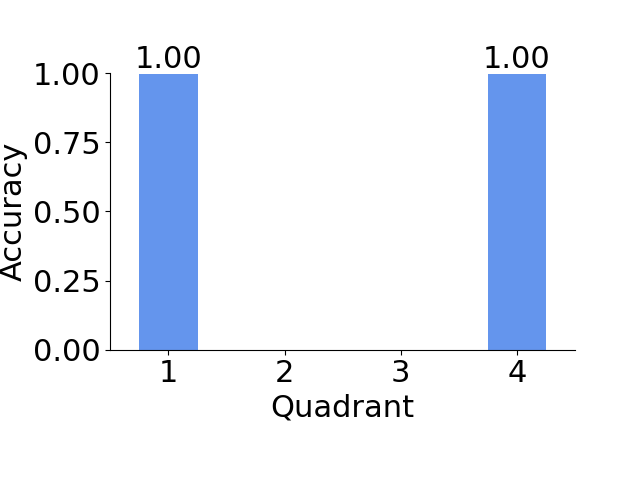}
\centering
\\
(a) Monotonic (no noise)
\end{minipage}
\begin{minipage}{4cm}
\includegraphics[scale = 0.25]
{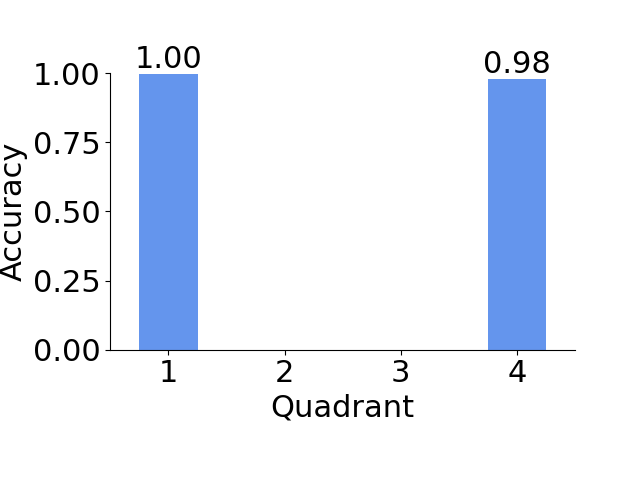}
\centering
(b) Monotonic with noise
\end{minipage}
\begin{minipage}{4cm}
\includegraphics[scale = 0.25]{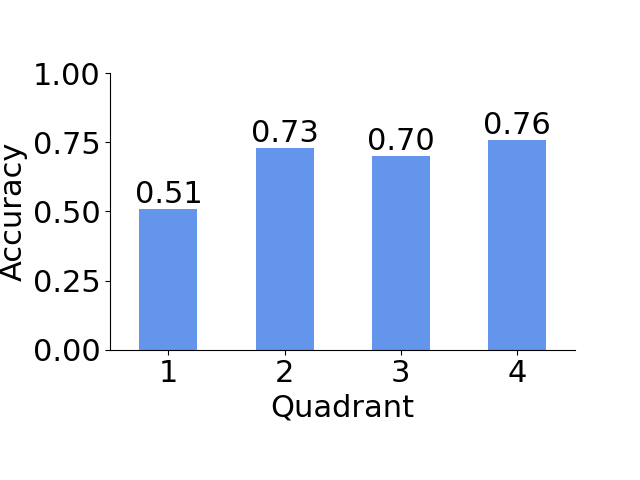}
\centering
(c) Spikes 
\end{minipage}
\begin{minipage}{4cm}
\includegraphics[scale = 0.25]{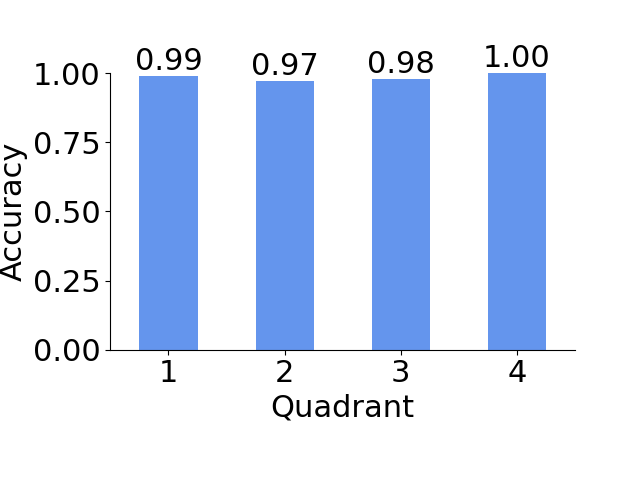}
\centering
(d) Brownian motion
\end{minipage}
\\ % End of the first row of images and labels
% \newline
% Text below last image in Block 1 \\
% Text below image in Block 2 \\
\midrule
\\
\textbf{Llama2} \\
\begin{minipage}{4cm}
\includegraphics[scale = 0.25]{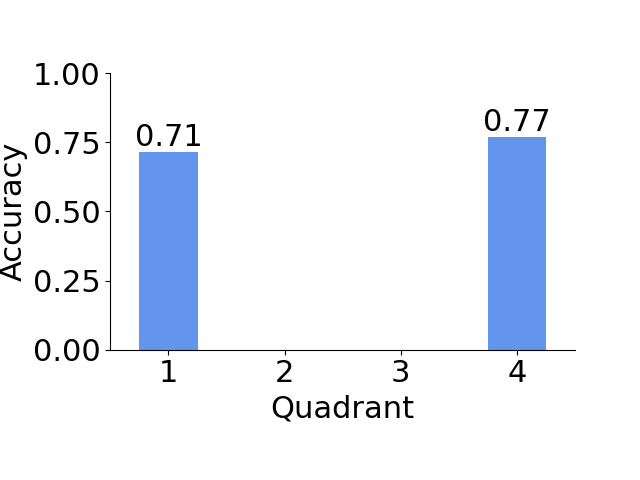}
\centering
(a) Monotonic (no noise)
\end{minipage}
\begin{minipage}{4cm}
\includegraphics[scale = 0.25]{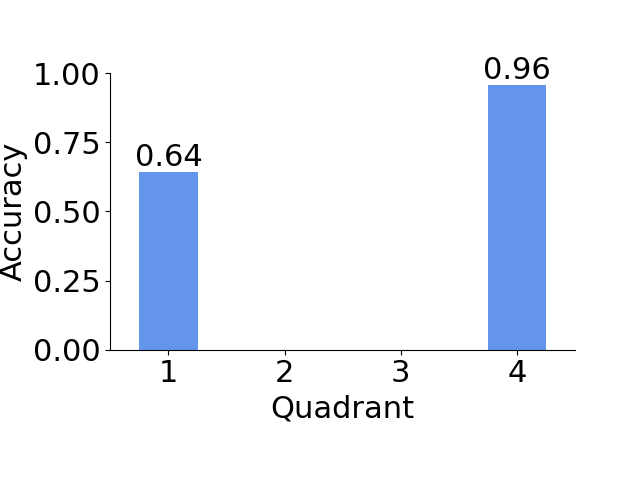}
\centering
(b) Monotonic with noise
\end{minipage}
\begin{minipage}{4cm}
\includegraphics[scale = 0.25]{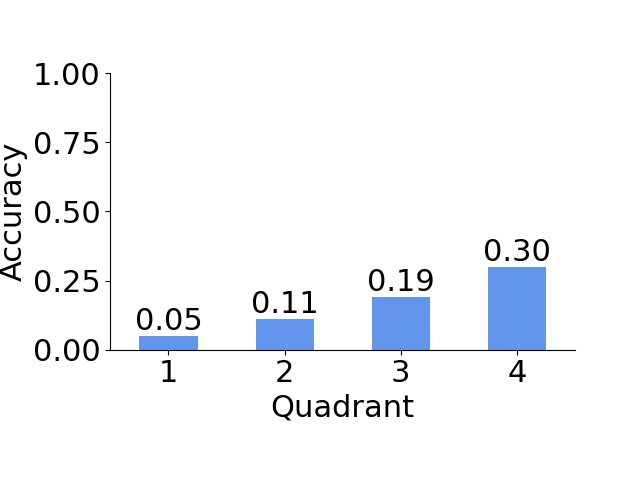}
\centering
(c) Spikes 
\end{minipage}
\begin{minipage}{4cm}
\includegraphics[scale = 0.25]{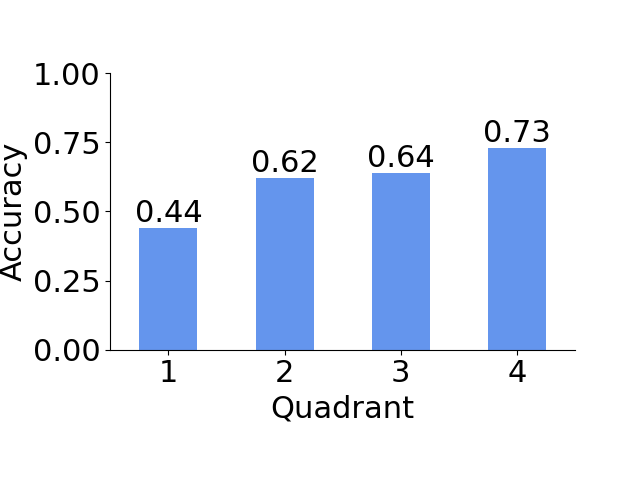}
\centering
(d) Brownian motion
\end{minipage}
\\
\midrule
\\
\textbf{Vicuna} \\
\begin{minipage}{4cm}
\includegraphics[scale = 0.25]{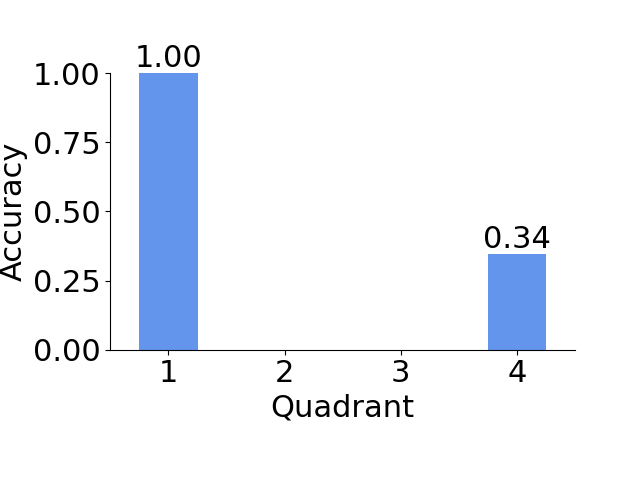}
\centering
(a) Monotonic (no noise)
\end{minipage}
\begin{minipage}{4cm}
\includegraphics[scale = 0.25]{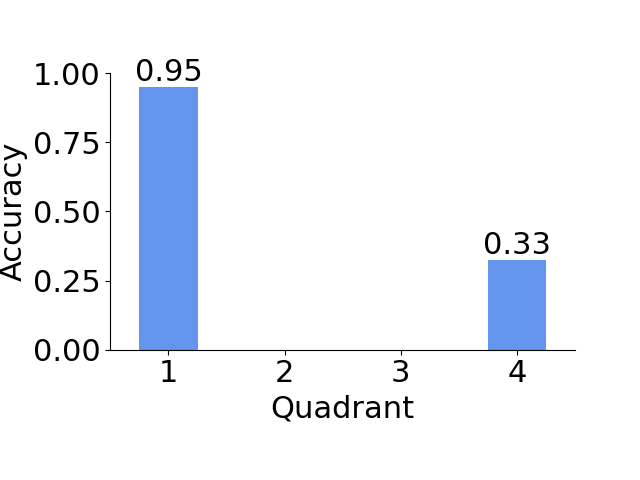}
\centering
(b) Monotonic with noise
\end{minipage}
\begin{minipage}{4cm}
\includegraphics[scale = 0.25]{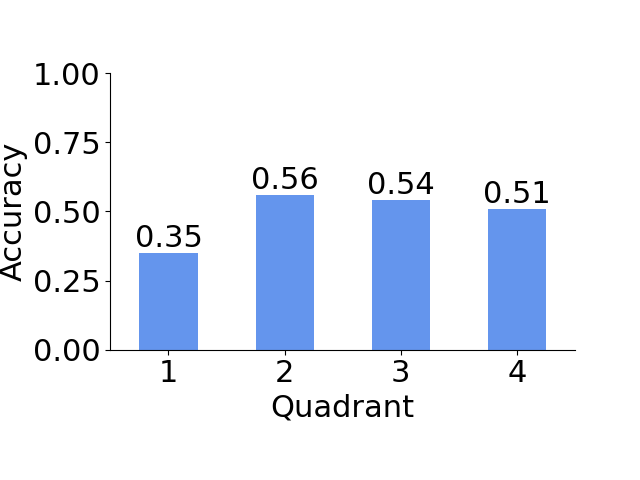}
\centering
(c) Spikes 
\end{minipage}
\begin{minipage}{4cm}
\includegraphics[scale = 0.25]{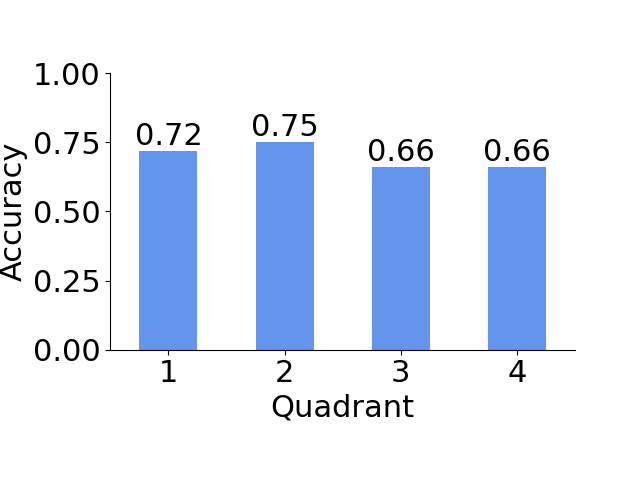}
\centering
(d) Brownian motion
\end{minipage}
\\ % End of the first row of images and labels
% \newline
% Text below last image in Block 1 \\
% Text below image in Block 2 \\
% \hline
\bottomrule
% \end{tabular}
\caption{Confusion matrix and accuracy by quadrant for the min-max extraction task. Note that monotonic series can have maximum or minimum values only in the first or fourth quadrant.}
\label{fig:position_experiments_min_max}
\end{longtable}

\end{subsection}

\end{section}

\section{Time Series formatting}
\label{sec:ts_format_extra}

\textbf{Custom}

\begin{lstlisting}[basicstyle=\footnotesize\ttfamily,columns=fullflexible]
"Date|Value\n2020-01-01|100\n2020-01-02|105\n2020-01-03|103\n2020-01-04|103\n"
\end{lstlisting}

\begin{lstlisting}[basicstyle=\ttfamily,columns=fullflexible]
Date|Value
2020-01-01|100
2020-01-02|105
2020-01-03|103
2020-01-04|103
\end{lstlisting}

% \begin{verbatim}
% Date|Value
% 2020-01-01|100
% 2020-01-02|105
% 2020-01-03|103
% 2020-01-04|103
% \end{verbatim}

\textbf{TSV}

\begin{lstlisting}[basicstyle=\footnotesize\ttfamily,columns=fullflexible]
"Date\tValue\n2020-01-01\t100\n2020-01-02\t105\n2020-01-03\t103\n2020-01-04\t103\n"
\end{lstlisting}

\begin{lstlisting}[basicstyle=\ttfamily,columns=fullflexible,keepspaces=true]
Date        Value
2020-01-01  100
2020-01-02  105
2020-01-03  103
2020-01-04  103
\end{lstlisting}

\textbf{Plain}

\begin{lstlisting}[basicstyle=\ttfamily, breaklines=true, tabsize=2, breakatwhitespace=false]
"Date: 2020-01-01, Value: 100\nDate: 2020-01-02, Value: 105\nDate: 2020-01-03, Value: 103\nDate: 2020-01-04, Value: 103"
\end{lstlisting}

\begin{lstlisting}[basicstyle=\ttfamily,columns=fullflexible,keepspaces=true]
Date: 2020-01-01, Value: 100
Date: 2020-01-02, Value: 105
Date: 2020-01-03, Value: 103
Date: 2020-01-04, Value: 103
\end{lstlisting}

\textbf{JSON}

\begin{lstlisting}[basicstyle=\ttfamily, breaklines=true, tabsize=2, breakatwhitespace=false]
{"Date":"2020-01-01","Value":100}\n{"Date":"2020-01-02","Value":105}\n{"Date":"2020-01-03","Value":103}\n{"Date":"2020-01-04","Value":103}\n
\end{lstlisting}

\begin{lstlisting}[basicstyle=\ttfamily,columns=fullflexible,keepspaces=true]
{"Date":"2020-01-01","Value":100}
{"Date":"2020-01-02","Value":105}
{"Date":"2020-01-03","Value":103}
{"Date":"2020-01-04","Value":103}
\end{lstlisting}

% '{"Date":"2020-01-01","Value":100}\n{"Date":"2020-01-02","Value":105}\n{"Date":"2020-01-03","Value":103}\n{"Date":"2020-01-04","Value":103}\n'

\textbf{Markdown}

\begin{lstlisting}[basicstyle=\ttfamily, breaklines=true, tabsize=2, breakatwhitespace=false]
"|Date|Value|\n|---|---|\n|2020-01-01|100|\n|2020-01-02|105|\n|2020-01-03|103|\n|2020-01-04|103|\n"
\end{lstlisting}

\begin{lstlisting}[basicstyle=\ttfamily,columns=fullflexible,keepspaces=true]
|Date|Value|
|---|---|
|2020-01-01|100|
|2020-01-02|105|
|2020-01-03|103|
|2020-01-04|103|
\end{lstlisting}

\textbf{Spaces}

\begin{lstlisting}[basicstyle=\ttfamily, breaklines=true, tabsize=2, breakatwhitespace=false]
"Date,Value\n2020-01-01,1 0 0\n2020-01-02,1 0 5\n2020-01-03,1 0 3\n2020-01-04,1 0 3\n"
\end{lstlisting}

\begin{lstlisting}[basicstyle=\ttfamily,columns=fullflexible,keepspaces=true]
Date,Value
2020-01-01,1 0 0
2020-01-02,1 0 5
2020-01-03,1 0 3
2020-01-04,1 0 3
\end{lstlisting}

\textbf{Context}

\begin{lstlisting}[basicstyle=\ttfamily, breaklines=true, tabsize=2, breakatwhitespace=false]
"Date,Value\n2020-01-01,[100]\n2020-01-02,[105]\n2020-01-03,[103]\n2020-01-04,[103]\n"
\end{lstlisting}

\begin{lstlisting}[basicstyle=\ttfamily,columns=fullflexible,keepspaces=true]
Date,Value
2020-01-01,[100]
2020-01-02,[105]
2020-01-03,[103]
2020-01-04,[103]
\end{lstlisting}

\textbf{Symbol}

\begin{lstlisting}[basicstyle=\ttfamily, breaklines=true, tabsize=2, breakatwhitespace=false, escapeinside={(*@}{@*)}]
"Date,Value,DirectionIndicator\n2020-01-01,100,(*@\rightarrowtext@*)\n2020-01-02,105,(*@\uparrowtext@*)\n2020-01-03,103,(*@\downarrowtext@*)\n2020-01-04,103,(*@\rightarrowtext@*)\n"
\end{lstlisting}

\begin{lstlisting}[basicstyle=\ttfamily, breaklines=true, escapeinside={(*@}{@*)}]
Date,Value,DirectionIndicator
2020-01-01,100,(*@\rightarrowtext@*)
2020-01-02,105,(*@\uparrowtext@*)
2020-01-03,103,(*@\downarrowtext@*)
2020-01-04,103,(*@\rightarrowtext@*)
\end{lstlisting}

\textbf{Base/csv}

\begin{lstlisting}[basicstyle=\ttfamily, breaklines=true, tabsize=2, breakatwhitespace=false]
"Date,Value\n2020-01-01,100\n2020-01-02,105\n2020-01-03,103\n2020-01-04,103\n"
\end{lstlisting}

\begin{lstlisting}[basicstyle=\ttfamily,columns=fullflexible,keepspaces=true]
Date,Value
2020-01-01,100
2020-01-02,105
2020-01-03,103
2020-01-04,103
\end{lstlisting}

\clearpage
\subsection{Additional results of time series formatting}

\begin{table}[ht]
\centering
\begin{subtable}{\linewidth}
\centering
\caption{GPT3.5}
%\label{tab:sub1}
\begin{tabular}{lccccccccc}
\toprule
 & csv & plain & tsv & custom & contextual & json & markdown & spaces & symbol \\
Trend det & 0.42 & 0.41 & 0.41 & 0.43 & \bfseries 0.44 & 0.41 & 0.41 & 0.42 & 0.42 \\
Trend class & 0.74 & 0.55 & 0.72 & 0.61 & 0.85 & 0.50 & 0.56 & 0.53 & \bfseries 0.92 \\
Season det & 0.61 & 0.77 & 0.69 & 0.60 & 0.58 & \bfseries 0.87 & 0.44 & 0.63 & 0.47 \\
Season class & \bfseries 0.27 & 0.19 & 0.21 & 0.16 & 0.23 & 0.22 & 0.09 & 0.17 & 0.18 \\
Outlier det & 0.55 & 0.52 & 0.50 & 0.49 & 0.46 & 0.49 & 0.48 & 0.52 & \bfseries 0.62 \\
Outlier class & \bfseries 0.17 & \bfseries 0.17 & \bfseries 0.17 & 0.16 & \bfseries 0.17 & \bfseries 0.17 & \bfseries 0.17 & \bfseries 0.17 & \bfseries 0.17 \\
\midrule
AvgRank & \bf 3.33 & 5.75 & 4.00 & 6.08 & 4.50 & 5.25 & 7.25 & 4.83 & 4.00 \\
\bottomrule
\end{tabular}
\bigskip % Adds some vertical space between the subtables
\end{subtable}
\begin{subtable}{\linewidth}
\centering
\caption{Llama2}
%\label{tab:sub2}
\begin{tabular}{lccccccccc}
\toprule
 & csv & plain & tsv & custom & contextual & json & markdown & spaces & symbol \\
Trend det & 0.51 & 0.44 & \bfseries 0.63 & 0.56 & 0.46 & 0.50 & 0.56 & 0.34 & 0.40 \\
Trend class & 0.41 & 0.48 & 0.40 & 0.43 & 0.45 & 0.42 & 0.36 & 0.43 & \bfseries 0.62 \\
Season det & 0.55 & 0.24 & 0.48 & 0.46 & \bfseries 0.59 & 0.38 & 0.45 & 0.40 & 0.50 \\
Season class & 0.11 & \bfseries 0.13 & 0.09 & 0.10 & 0.09 & 0.10 & 0.11 & 0.08 & 0.10 \\
Outlier det & 0.44 & 0.35 & 0.47 & 0.44 & 0.45 & 0.48 & \bfseries 0.51 & 0.41 & 0.47 \\
Outlier class & 0.13 & 0.14 & 0.10 & 0.14 & 0.17 & 0.18 & \bfseries 0.21 & 0.14 & 0.08 \\
\midrule
AvgRank & 4.83 & 5.50 & 5.33 & 4.33 & 4.33 & 4.83 & \bf 3.83 & 7.17 & 4.83 \\
\bottomrule
\end{tabular}
\bigskip % Adds some vertical space between the subtables
\end{subtable}
\begin{subtable}{\linewidth}
\centering
\caption{Vicuna}
%\label{tab:sub2}
\begin{tabular}{lccccccccc}
\toprule
 & csv & plain & tsv & custom & contextual & json & markdown & spaces & symbol \\
Trend det & 0.51 & 0.49 & 0.47 & 0.47 & \bfseries 0.55 & 0.44 & 0.51 & 0.54 & 0.45 \\
Trend class & 0.49 & 0.58 & 0.54 & 0.53 & 0.56 & 0.50 & 0.56 & 0.44 & \bfseries 0.64 \\
Season det & 0.47 & 0.47 & \bfseries 0.54 & 0.47 & 0.48 & 0.49 & 0.51 & 0.53 & \bfseries 0.54 \\
Season class & 0.14 & 0.14 & \bfseries 0.20 & \bfseries 0.20 & \bfseries 0.20 & 0.19 & 0.17 & 0.14 & 0.15 \\
Outlier det & 0.49 & 0.53 & \bfseries 0.54 & 0.52 & 0.47 & 0.50 & 0.52 & \bfseries 0.54 & 0.49 \\
Outlier class & 0.19 & 0.14 & 0.19 & 0.16 & \bfseries 0.22 & 0.16 & 0.13 & 0.14 & 0.08 \\
\midrule
AvgRank & 6.33 & 5.33 & \bf 3.00 & 5.33 & 3.83 & 5.83 & 4.83 & 5.17 & 5.33 \\
\bottomrule
\end{tabular}
\end{subtable}
\caption{Performance on Time Series Reasoning for different time series formatting.}
\label{tab:perf_format_reason_extra}
\end{table}

\begin{table}[ht]
\centering
\begin{subtable}{\linewidth}
\centering
\caption{GPT3.5}
\label{tab:sub1}
\begin{tabular}{lccccccccc}
\toprule
 & csv & plain & tsv & custom & contextual & json & markdown & spaces & symbol \\
\midrule
Min value & 0.98 & \bf 0.99 & 0.98 & 0.98 & 0.98 & 0.98 & 0.98 & 0.79 & 0.98 \\
Min date & 0.94 & \bf 0.95 & 0.94 & \bf 0.95 & 0.94 & 0.94 & 0.93 & 0.69 & 0.93 \\
Max value & 0.92 & 0.92 & 0.91 & 0.92 & 0.92 & 0.91 & 0.91 & 0.54 & \bf 0.94 \\
Max date & 0.88 & 0.88 & 0.88 & 0.88 & 0.88 & 0.86 & 0.86 & 0.51 & \bf 0.89 \\
Value on date & 0.94 & 0.94 & 0.94 & 0.94 & \bf 0.95 & 0.94 & 0.94 & 0.82 & 0.94 \\
\midrule
AvgRank & 4.80 & \bf 2.70 & 4.40 & 3.10 & 3.20 & 6.60 & 7.30 & 9.00 & 3.90 \\
\bottomrule
\end{tabular}
\bigskip % Adds some vertical space between the subtables
\end{subtable}
\begin{subtable}{\linewidth}
\centering
\caption{Llama2}
\label{tab:sub2}
\begin{tabular}{lccccccccc}
\toprule
 & csv & plain & tsv & custom & contextual & json & markdown & spaces & symbol \\
\midrule
Min value & 0.55 & \bf 0.58 & 0.54 & 0.54 & 0.56 & \bf 0.58 & 0.55 & 0.20 & \bf 0.58 \\
Min date & 0.28 & \bf 0.39 & 0.30 & 0.28 & 0.29 & 0.36 & 0.34 & 0.09 & 0.29 \\
Max value & 0.48 & \bf 0.56 & 0.49 & 0.48 & 0.50 & 0.55 & 0.54 & 0.05 & 0.52 \\
Max date & 0.34 & \bf 0.46 & 0.40 & 0.38 & 0.37 & 0.45 & 0.44 & 0.04 & 0.41 \\
Value on date & 0.39 & 0.38 & \bf 0.47 & 0.40 & 0.35 & 0.45 & 0.44 & 0.07 & 0.34 \\
\midrule
AvgRank & 6.80 & \bf 2.30 & 4.60 & 6.50 & 5.60 & 2.10 & 3.50 & 9.00 & 4.60 \\
\bottomrule
\end{tabular}
\bigskip % Adds some vertical space between the subtables
\end{subtable}
\begin{subtable}{\linewidth}
\centering
\caption{Vicuna}
\label{tab:sub3}
\begin{tabular}{lccccccccc}
\toprule
 & csv & plain & tsv & custom & contextual & json & markdown & spaces & symbol \\
\midrule
Min value & 0.63 & \bf 0.67 & 0.56 & 0.61 & 0.60 & 0.64 & 0.59 & 0.17 & 0.62 \\
Min date & 0.50 & \bf 0.55 & 0.47 & 0.49 & 0.53 & 0.52 & 0.51 & 0.13 & 0.49 \\
Max value & 0.49 & 0.46 & 0.45 & 0.44 & 0.48 & 0.47 & \bf 0.50 & 0.01 & \bf 0.50 \\
Max date & 0.38 & 0.42 & 0.41 & 0.39 & \bf 0.46 & 0.40 & 0.42 & 0.07 & 0.41 \\
Value on date & 0.36 & \bf 0.48 & 0.39 & 0.39 & 0.42 & 0.40 & 0.37 & 0.09 & 0.41 \\
\midrule
AvgRank & 5.40 & \bf 2.40 & 6.50 & 6.60 & 3.00 & 4.00 & 4.30 & 9.00 & 3.80 \\
\bottomrule
\end{tabular}
\end{subtable}
\caption{Accuracy for information retrieval and arithmetic reasoning tasks for different time series formatting.}
\label{tab:perf_format_ret_acc}
\end{table}

\begin{table}[ht]
\centering
\begin{subtable}{\linewidth}
\centering
\caption{GPT3.5}
\label{tab:sub1}
    \resizebox{\linewidth}{!}{
\begin{tabular}{lccccccccc}
\toprule
 & csv & plain & tsv & custom & contextual & json & markdown & spaces & symbol \\
\midrule
Min value & 0.04 & 0.04 & 0.05 & 0.04 & 0.04 & 0.06 & 0.07 & 0.32 & 0.04 \\
Max value & 0.06 & 0.07 & 0.07 & 0.07 & 0.07 & 0.10 & 0.09 & 1.01 & 0.10 \\
Value on date & 0.08 & 0.10 & 0.07 & 0.08 & 0.03 & 0.08 & 0.03 & 0.38 & 0.04 \\
\bottomrule
\end{tabular}}
\end{subtable}
\bigskip % Adds some vertical space between the subtables
\begin{subtable}{\linewidth}
\centering
\caption{Llama2}
\label{tab:sub2}
\resizebox{\linewidth}{!}{
\begin{tabular}{lccccccccc}
\toprule
 & csv & plain & tsv & custom & contextual & json & markdown & spaces & symbol \\
\midrule
Min value & 10.15 & 16.18 & 10.38 & 19.57 & 22.46 & 11.14 & 21.15 & 0.69 & 21.12 \\
Max value & 1.03 & 0.95 & 1.09 & 1.04 & 0.91 & 1.01 & 1.00 & 2.58 & 0.90 \\
Value on date & 0.81 & 0.65 & 0.40 & 0.73 & 0.61 & 0.48 & 0.44 & 0.96 & 0.90 \\
\bottomrule
\end{tabular}
}
\end{subtable}
\bigskip % Adds some vertical space between the subtables
\begin{subtable}{\linewidth}
\centering
\caption{Vicuna}
\label{tab:sub3}
\resizebox{\linewidth}{!}{
\begin{tabular}{lccccccccc}
\toprule
 & csv & plain & tsv & custom & contextual & json & markdown & spaces & symbol \\
\midrule
Min value & 12.79 & 12.24 & 29.45 & 13.89 & 12.06 & 26.62 & 25.54 & 0.96 & 22.50 \\
Max value & 0.85 & 0.74 & 1.01 & 1.14 & 0.94 & 0.67 & 0.98 & 2.51 & 0.59 \\
Value on date & 0.44 & 0.78 & 0.83 & 0.94 & 0.31 & 0.65 & 0.38 & 0.95 & 0.38 \\
\bottomrule
\end{tabular}
}
\end{subtable}
\caption{MAPE for information retrieval and arithmetic reasoning tasks for different time series formatting.}
\label{tab:perf_format_ret_mape}
\end{table}

\begin{figure*}
    \centering
    \includegraphics[width=0.33\textwidth]{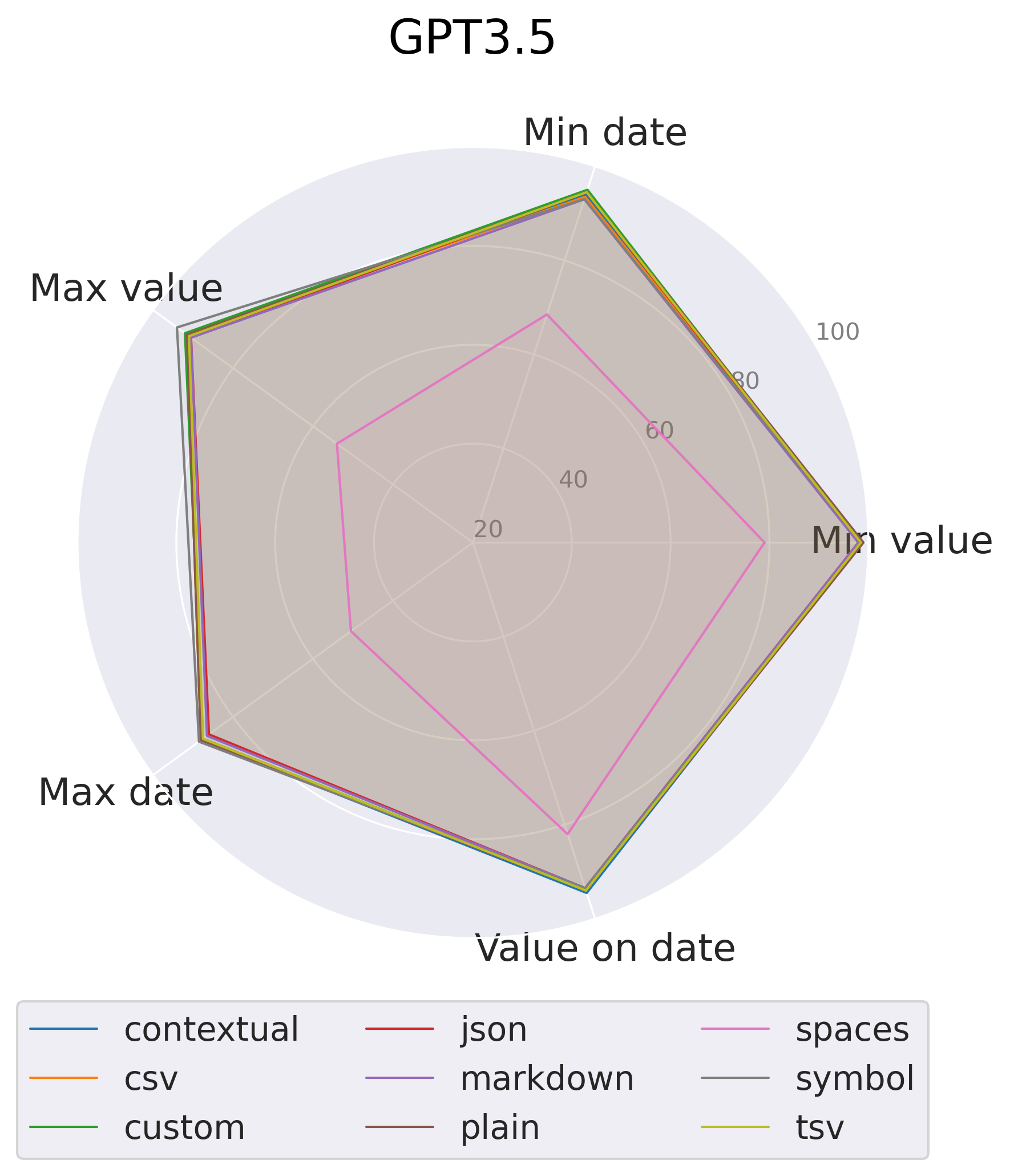}~\includegraphics[width=0.33\textwidth]{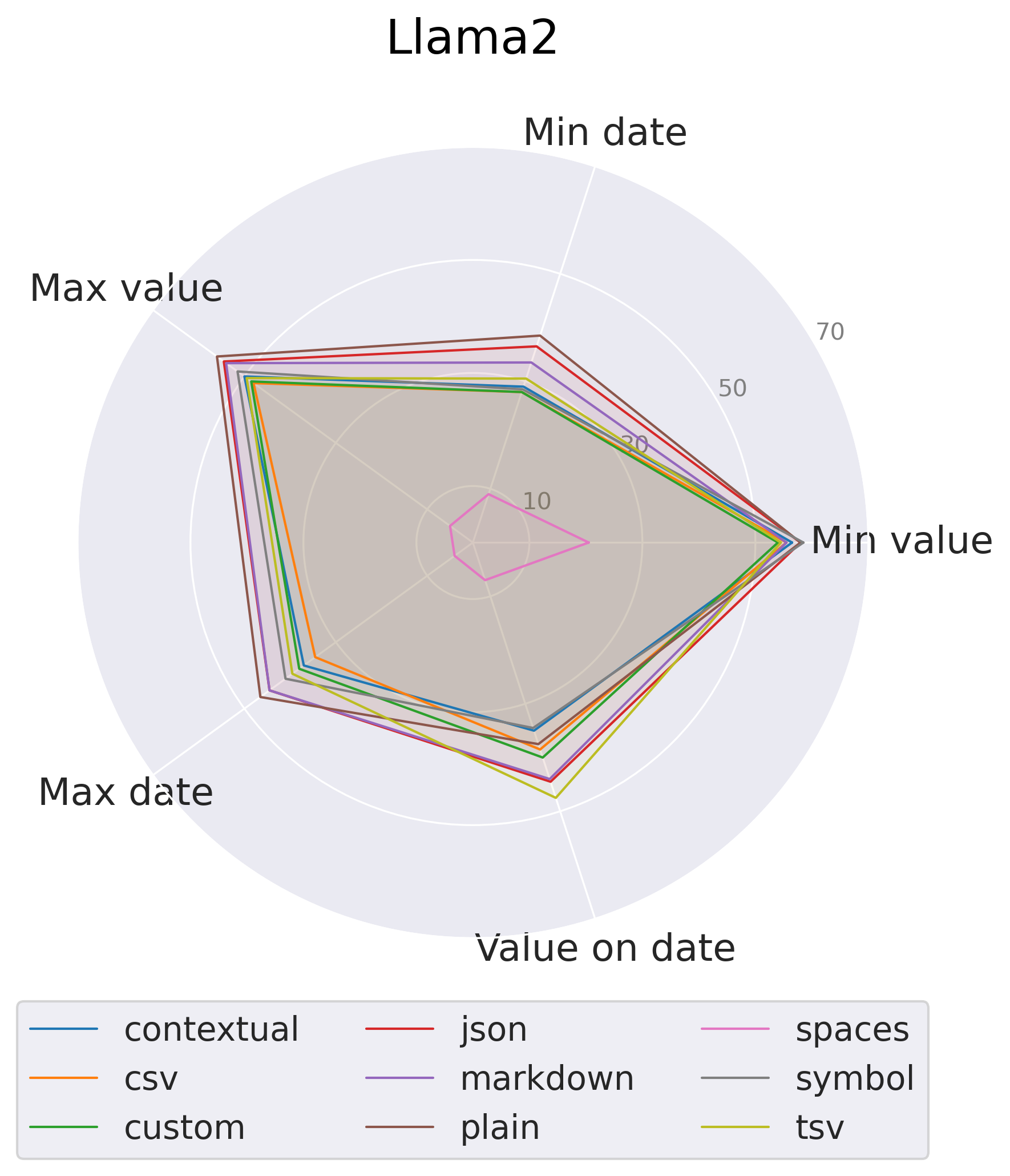}~\includegraphics[width=0.33\textwidth]{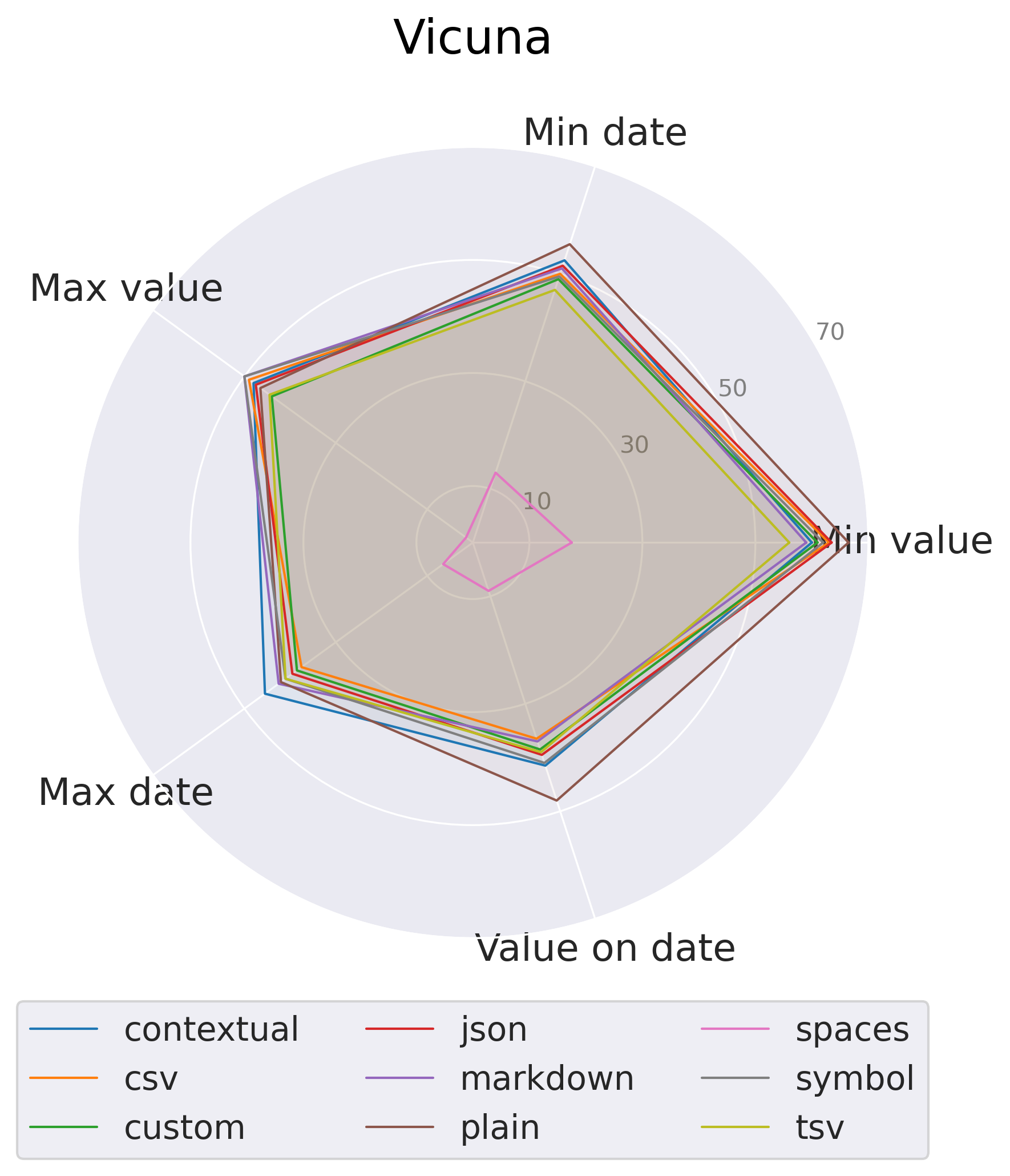}
    \caption{Accuracy for information retrieval and arithmetic reasoning tasks for different time series tokenization.}
    \label{fig:enter-label}
\end{figure*}

\clearpage
\section{Prompts}
\label{sec:app_prompts}

    \begin{tcolorbox}
    \textbf{Information retrieval and arithmetic reasoning prompts -- Zero-shot}
    \tcblower % Creates a visual separation within the box
    "Input:\texttt{<time series>}.\\
    Given the input time series, please answer the following questions and format your responses in a dictionary with the structure shown below: \\
        \texttt{\{'max\_value': \{'value':value, 'date':date\}, 
        'min\_value': \{'value':value, 'date':date\},  
        'value\_on\_date <date>': \{'value':value\}\}}.\\
        Only provide the numerical value and/or the date as the answer for each question. Format the reply as a dictionary following the instruction."
    \end{tcolorbox}

    \begin{tcolorbox}
    \textbf{Information retrieval and arithmetic reasoning prompts -- CoT}
    \tcblower % Creates a visual separation within the box
    "Input:\texttt{<time series>}.\\
    Given the input time series, please provide concise and precise answers to the following questions and format your responses in a dictionary: \\
        \texttt{\{'max\_value': \{'value':value, 'date':date\}, 
        'min\_value': \{'value':value, 'date':date\},  
        'value\_on\_date <date>': \{'value':value\}\}}.\\
        To ensure accuracy, let's follow these steps:\\
            1. Identify the maximum value and its date.\\
            2. Identify the minimum value and its date.\\
            3. Find the value on the specified date <date>.\\
        Note: Only provide the numerical value and/or the date as the answer for each question. Format the reply as a dictionary following the instruction.\\
        Let's think step by step."
    \end{tcolorbox}

    \begin{tcolorbox}
    \textbf{Trend Prompts -- Zero-shot}
    \tcblower % Creates a visual separation within the box
    "Input:\texttt{<time series>}."\\
    \colorbox{highlight}{Question 1: Detection}\\
    "Question: can you detect a general upward or downward trend in this time series? Answer yes or no only."\\
    \colorbox{highlight}{Question 2: Classification}\\
    "Select one of the following answers: (a) the time series has a positive trend, (b) the time series has a negative trend. Provide your answer as either (a) or (b)."
    \end{tcolorbox}

    \begin{tcolorbox}
    \textbf{Trend Prompts - CoT}
    \tcblower % Creates a visual separation within the box
    "Input:\texttt{<time series>}."\\
    \colorbox{highlight}{Question 1: Detection}\\
    "Question: Question: Can you detect a general upward or downward trend in this time series? Provide your reasoning and then answer 'Yes' or 'No'. \\
            Let's think step by step. First, observe the overall pattern of the data points. Do they generally increase or decrease over time? \\
            Consider the starting and ending points of the series. If the ending point is significantly higher or lower than the starting point, this might indicate a trend. \\
            Also, look at the intermediate points: do they show a consistent direction of movement, or are there major fluctuations that disrupt the trend? \\
            Now, based on these observations, determine if there is a consistent pattern indicating a trend. Finally, provide your answer as 'Yes' or 'No'.","\\
    \colorbox{highlight}{Question 2: Classification}\\
    ""Select one of the following answers: \\
            (a) The time series has a positive trend, (b) The time series has a negative trend. Provide your answer as either (a) or (b). \\
            Let's think step by step. First, identify the general direction of the data points. Do they appear to be moving upward or downward overall? \\
            Consider the slope of the line that could be drawn through the data points. A positive slope indicates an upward trend, while a negative slope indicates a downward trend. \\
            Check for consistency in the movement. Are most of the data points following this direction, or are there significant deviations? \\
            If the overall pattern is increasing, select (a). If it is decreasing, select (b)."
    \end{tcolorbox}

\begin{tcolorbox}
\textbf{Seasonality Prompts -- Zero-shot}
\tcblower % Creates a visual separation within the box
\colorbox{highlight}{Prompt 1: Detection}\\
"Input:\texttt{<time series>}.\\
Question: can you detect any cyclic or periodic patterns in this time series? Only answer 'Yes' or 'No'."

\colorbox{highlight}{Prompt 2: Classification}\\
"Given the following definitions:\\
            Fixed-period: Regular, predictable seasonal patterns occurring at fixed intervals (e.g., daily, weekly, monthly).\\
            Shifting Period: Seasonal patterns where the length of the period shifts over time.\\
            Multiple seasonality: Presence of multiple overlapping seasonal patterns (e.g., both weekly and monthly seasonality)\\
            Select one of the following answers: \\
            (a) The time series has fixed-period seasonality, (b) The time series has a shift in seasonal pattern, \
            (c) The time series has multiple seasonal patterns. \\
            Only answer (a), (b) or (c)."
\end{tcolorbox}

\begin{tcolorbox}
\textbf{Seasonality Prompts -- CoT}
\tcblower % Creates a visual separation within the box
\colorbox{highlight}{Prompt 1: Detection}\\
"Input:\texttt{<time series>}.\\
Question: Can you detect any cyclic or periodic patterns in this time series? Provide your reasoning and then answer 'Yes' or 'No'. \\
            Let's think step by step. First, observe the overall shape of the time series. Look for repeating patterns or cycles. \\
            Identify the peaks (high points) and troughs (low points) in the series. Are these peaks and troughs occurring at regular intervals? \\
            Measure the distance between these repeating points. If the intervals between them are consistent, it suggests a cyclic pattern. \\
            Also, consider the amplitude (height) of these peaks and troughs. Is the amplitude consistent or does it vary over time? \\
            Now, based on these observations, determine if there is a consistent cyclic or periodic pattern in the time series. Finally, provide your answer as 'Yes' or 'No'."
\colorbox{highlight}{Prompt 2: Classification}\\
"Given the following definitions:\\
            Fixed-Period: Seasonality with a constant, unchanging period (e.g., monthly seasonality).\\
            Shifting Period: Seasonality where the length of the period shifts over time (e.g., a seasonal pattern that shifts slightly each year).\\
            Multiple Seasonality: Presence of multiple overlapping seasonal patterns (e.g., both weekly and monthly seasonality).\\
            Select one of the following answers: \\
            (a) The time series has a fixed-period seasonality, (b) The time series has a shifting-period seasonality, \
            (c) The time series has multiple seasonality. \\
            Let's think step by step. First, identify if there is a repeating pattern at fixed intervals, which would indicate a fixed-period seasonality. 
            If the timing of the pattern shifts, it's a shifting-period seasonality. 
            Finally, if there are two or more overlapping seasonal patterns, identify it as multiple seasonality. 
            Compare the intervals and magnitudes of the peaks and troughs carefully to determine the correct pattern. Now, provide your final answer as either (a), (b), or (c)."
\end{tcolorbox}

\begin{tcolorbox}
\textbf{Anomaly Prompts -- Zero-shot}
\tcblower % Creates a visual separation within the box
"Input:\texttt{<time series>}.\\
\colorbox{highlight}{Prompt 1: Detection}\\
Question: can you detect any irregularities in this time series? Only answer 'Yes' or 'No'."\\
\colorbox{highlight}{Prompt 2: Classification}\\
"Given the following definitions:\\
        Spike: a sudden and brief deviation from the overall pattern of the data. \\
        Level shift: a sudden and lasting change in the average value of the series. \\
        Temporal disruption: an interval where data is missing or not recorded. \\
        Select one of the following answers that best describes the provided time series: \\
        (a) The time series has one or more spikes, (b) The time series has a level shift, 
        (c) The time series has a temporal disruption. \\
        Only answer (a), (b), or (c)."
\end{tcolorbox}

\begin{tcolorbox}
\textbf{Anomaly Prompts -- CoT}
\tcblower % Creates a visual separation within the box
"Input:\texttt{<time series>}.\\
\colorbox{highlight}{Prompt 1: Detection}\\
Question: Can you detect any irregularities in this time series? Provide your reasoning and then answer 'Yes' or 'No'. \\
            Let's think step by step. First, observe the overall pattern of the time series. Identify the general trend or pattern. \\
            Next, look for any points that deviate significantly from this overall pattern. These deviations could be much higher or lower than the rest of the data points. \\
            Consider the context of these deviations: are they isolated points, or do they occur in a sequence? \\
            Are there sudden jumps or drops that are not consistent with the trend? After examining these factors, determine if there are any significant irregularities. Finally, provide your answer as 'Yes' or 'No'."\\
\colorbox{highlight}{Prompt 2: Classification}\\
"Given the following definitions: \\
            Spike: a sudden and brief deviation from the overall pattern of the data. \\
            Level shift: a sudden and lasting change in the average value of the series. \\
            Temporal disruption: an interval where data is missing or not recorded. \\
            Select one of the following answers that best describes the provided time series: \\
            (a) The time series has one or more spikes, (b) The time series has a level shift, \\
            (c) The time series has a temporal disruption. \\
            Let's think step by step. First, identify if there are any points that stand out sharply from the rest of the data, which would indicate spikes. \\
            If there is a lasting change in the average value of the series, identify it as a level shift. \\
            If there are intervals where data appears to be missing or not recorded, classify it as a temporal disruption. \\
            Based on your observations, determine the type of irregularity present. Now, provide your final answer as either (a), (b), or (c)."
\end{tcolorbox}

\begin{tcolorbox}
\textbf{Volatility Prompts -- Zero-shot}
\tcblower % Creates a visual separation within the box
"Input:\texttt{<time series>}.\\
\colorbox{highlight}{Prompt 1: Detection}\\
Question: can you detect any volatility in this time series? Only answer 'Yes' or 'No'."\\
\colorbox{highlight}{Prompt 2: Classification}\\
"Given the following definitions: \\
        Constant Volatility: The degree of variation in the time series remains consistent and predictable over time.\\
        Trending Volatility: The level of variation in the time series shows a clear increasing or decreasing trend over time.\\
        Clustered Volatility: The time series exhibits periods where volatility is significantly higher or lower, with these periods tending to cluster together.\\
        Dynamic Volatility: The volatility of the time series changes over time in response to external factors (e.g., leverage effect where the 
        volatility of the time series tends to increase when the series experiences negative returns).\\
        Select one of the following answers: \\
        (a) The time series has constant volatility, (b) The time series has trending volatility, 
        (c) The time series has clustered volatility, (d) The time series has dynamic volatility. \\
        Only answer (a), (b), (c), or (d)."
\end{tcolorbox}

\begin{tcolorbox}
\textbf{Structural Break Prompts -- Zero-shot}
\tcblower % Creates a visual separation within the box
"Input:\texttt{<time series>}.\\
\colorbox{highlight}{Prompt 1: Detection}\\
Question: can you detect any regime switches or structural breaks in this time series? Only answer 'Yes' or 'No'."\\
\colorbox{highlight}{Prompt 2: Classification}\\
"Given the following definitions:\\
Regime Change: A shift in the time series data's statistical properties, such as mean, variance, or auto-correlation, that persists over time. This change is often gradual and represents a new phase or 'regime' in the data.\\
Structural Break: An abrupt change in the time series data that leads to a new level or trend. This change is typically sudden and can be linked to specific events or shifts in the underlying process.\\
Examine the provided time series data and select the correct option:\\
(a) The time series data exhibits a Regime Change. (b) The time series data exhibits a Structural Break.\\
Provide your answer as either (a) or (b)."
\end{tcolorbox}

\begin{tcolorbox}
\textbf{Fat tails Prompt -- Zero-shot}
\tcblower % Creates a visual separation within the box
"Input:\texttt{<time series>}.\\
\colorbox{highlight}{Prompt 1: Detection}\\
Question: Considering the data provided, does the time series exhibit fat tails? \
            Fat tails refer to a higher likelihood of extreme values compared to a normal distribution, \
            indicating a higher probability of observing significant positive or negative deviations. \
            Only answer 'Yes' or 'No'."\\
\end{tcolorbox}

\begin{tcolorbox}
\textbf{Stationarity Properties -- Zero-shot}
\tcblower % Creates a visual separation within the box
"Input:\texttt{<time series>}.\\
\colorbox{highlight}{Prompt 1: Detection}\\
Question: Considering the data provided, is the time series stationary? Only answer 'Yes' or 'No'."\\
\colorbox{highlight}{Prompt 2: Classification}\\
"Given the following definitions of non-stationary types in time series data: \\
            (a) Trend Change: The time series exhibits a significant shift in its underlying trend, indicating a change in the mean over time.\\
            (b) Variance Change: The time series shows a change in its variability or spread.\\
            (c) Seasonality: The time series displays regular and predictable patterns that repeat over a certain period.\\
            (d) Trend and Seasonality: The time series exhibits both a significant underlying trend and seasonal patterns. 
            This type combines elements of both trend changes and predictable seasonal fluctuations.\\
            Select one of the following answers based on your analysis of the time series: \\
            (a) The time series has a trend change, (b) The time series has a variance change, 
            (c) The time series has seasonality, (d) The time series has both trend and seasonality. \\
            Only answer (a), (b), (c) or (d)."
\end{tcolorbox}

\begin{tcolorbox}
\textbf{Correlation -- Zero-shot}
\tcblower % Creates a visual separation within the box
"Input:\texttt{<time series>}.\\
\colorbox{highlight}{Prompt 1: Detection}\\
Question: Considering the data provided, is there a correlation between the time series? Only answer 'Yes' or 'No'"\\
\colorbox{highlight}{Prompt 2: Classification}\\
"Select one of the following answers: \\
            (a) The time series are positively correlated or (b) The time series are negatively correlated. \\
            Provide your answer as either (a) or (b)."
\end{tcolorbox}

\begin{tcolorbox}
\textbf{Cross-Correlation -- Zero-shot}
\tcblower % Creates a visual separation within the box
"Input:\texttt{<time series>}.\\
\colorbox{highlight}{Prompt 1: Detection}\\
Question: Considering the data provided, is there a correlation (direct or lagged) between the two time series? Only answer 'Yes' or 'No'."\\
\colorbox{highlight}{Prompt 2: Classification}\\
"Given the following definitions:\\
            Direct Correlation: The two time series show a direct, immediate relationship between their values, \
            where changes in one series directly influence the other in a straightforward manner.\\
            Direct Lagged Correlation: The two time series demonstrate a delayed relationship, 
            where changes in one series influence the other after a certain lag period.\\
            Inverse Correlation: The two time series exhibit an inverse or negative relationship between their values, 
            where an increase in one series typically leads to a decrease in the other, and vice versa.\\
            Inverse Lagged Correlation: The two time series show a relationship where changes in one series negatively influence the other after a certain lag period, 
            suggesting that past increases in one series lead to future decreases in the other, and vice versa.\\
            Select one of the following answers that best describes the relationship between the two time series: \\
            (a) The two time series exhibit direct correlation, (b) The two time series exhibit direct lagged correlation, 
            (c) The two time series exhibit inverse correlation, (d) The two time series exhibit inverse lagged correlation. \\
            Only answer (a), (b), (c), or (d)."
\end{tcolorbox}

\clearpage
\section{Licenses}
Table \ref{tab:license} lists the licenses for the assets used in the paper.

\begin{table}[h!]
    \centering
    \begin{tabular}{c c}
    \toprule
        Asset & License \\
    \midrule
        Llama2  & \href{https://ai.meta.com/llama/license/}{Link} \\
        Vicuna1.5 & \href{https://huggingface.co/lmsys/vicuna-13b-v1.5}{Link} \\
        Phi3 & \href{https://huggingface.co/microsoft/Phi-3-medium-128k-instruct}{Link} \\
    \bottomrule
    \end{tabular}
    \caption{License of assets used.}
    \label{tab:license}
\end{table}

\clearpage
\section{Datasheet}

We provide a datasheet for evaluating large language models on time series feature understanding, following the framework in Gebru et al. (2021).

\begin{longtable}{|p{4cm}|p{10cm}|}
\caption{Datasheet for Time Series Feature Understanding} \\
\hline
\multicolumn{2}{|c|}{\textbf{Motivation}} \\
\hline
\textbf{For what purpose was the dataset created?} & The dataset was created to evaluate the capabilities of Large Language Models (LLMs) in understanding and captioning time series data, specifically in detecting, classifying, and reasoning about various time series features. \\
\hline
\textbf{Who created the dataset and on behalf of which entity?} & The dataset was created by the authors of this paper for the purposes of this research project. \\
\hline
\textbf{Who funded the creation of the dataset?} & The creation of the dataset was funded by the coauthors employers. \\
\hline
\textbf{Any other comment?} & The dataset is intended for evaluating the performance of LLMs on time series annotation and summarization tasks, highlighting both strengths and limitations. \\
\hline
\multicolumn{2}{|c|}{\textbf{Composition}} \\
\hline
\textbf{What do the instances that comprise the dataset represent?} & Instances are synthetic time series data points, representing various time series features such as trends, seasonality, anomalies, and more. \\
\hline
\textbf{How many instances are there in total?} & The dataset comprises 10 synthetic datasets with 5000 samples in the train split, 2000 samples in the validation split and 200 time series samples in the test set. \\
\hline
\textbf{Does the dataset contain all possible instances or is it a sample (not necessarily random) of instances from a larger set?} & The dataset is a curated sample representing a wide range of time series features and complexities. \\
\hline
\textbf{What data does each instance consist of?} & Each instance is a time series data point with associated features, metadata, and annotations for trend, seasonality, anomalies, etc. \\
\hline
\textbf{Is there a label or target associated with each instance?} & No. The dataset is primarily for evaluation of time series description and understanding tasks performed by LLMs. \\
\hline
\textbf{Is any information missing from individual instances?} & No. \\
\hline
\textbf{Are relationships between individual instances made explicit?} & No. Each instance is considered independently for the purpose of this benchmark. \\
\hline
\textbf{Are there recommended data splits?} & Yes, the dataset includes splits for training, validation, and test to ensure consistent evaluation metrics. \\
\hline
\textbf{Are there any errors, sources of noise, or redundancies in the dataset?} & We make efforts to remove errors and noise, but due to the complex nature of isolating time series features, there may be some redundancies. \\
\hline
\textbf{Is the dataset self-contained, or does it link to or otherwise rely on external resources?} & The dataset is self-contained. \\
\hline
\textbf{Does the dataset contain data that might be considered confidential?} & No. All data used in the dataset is synthetically generated. \\
\hline
\multicolumn{2}{|c|}{\textbf{Collection Process}} \\
\hline
\textbf{How was the data associated with each instance acquired?} & The synthetic data was generated using predefined rules for each feature. \\
\hline
\textbf{Was the data directly obtained from the individuals, or was it provided by third parties or obtained from publicly available sources?} & The data was synthesized using algorithmic generation methods. \\
\hline
\textbf{Were the individuals in question notified about the data collection?} & Not applicable. The dataset does not contain individual personal data. \\
\hline
\textbf{Did the individuals in question consent to the collection and use of their data?} & Not applicable. The dataset does not contain individual personal data. \\
\hline
\textbf{If consent was obtained, were the consenting individuals provided with any mechanism to revoke their consent in the future or for certain uses?} & Not applicable. The dataset does not contain individual personal data. \\
\hline
\textbf{Has an analysis of the potential impact of the dataset and its use on data subjects been conducted?} & Not applicable. The dataset does not contain individual personal data. \\
\hline
\multicolumn{2}{|c|}{\textbf{Preprocessing/Cleaning/Labeling}} \\
\hline
\textbf{What preprocessing/cleaning was done?} & Synthetic data was generated with controlled features. \\
\hline
\textbf{Was the “raw” data saved in addition to the preprocessed/cleaned/labeled data?} & Yes, both raw and preprocessed data are saved for transparency and reproducibility. \\
\hline
\textbf{Is the software used to preprocess/clean/label the instances available?} & Not at the moment, preprocessing scripts and tools might be made available in a project repository. \\
\hline
\multicolumn{2}{|c|}{\textbf{Uses}} \\
\hline
\textbf{Has the dataset been used for any tasks already?} & Yes, the dataset has been used for evaluating LLMs on time series feature detection, classification, and arithmetic reasoning tasks. \\
\hline
\textbf{Is there a repository that links to any or all papers or systems that use the dataset?} & Not at the moment. \\
\hline
\textbf{What (other) tasks could the dataset be used for?} & The dataset could be used for further time series analysis, forecasting, anomaly detection, and other machine learning tasks involving time series data. \\
\hline
\textbf{Is there anything about the composition of the dataset or the way it was collected and preprocessed/cleaned/labeled that might impact future uses?} & The synthetic nature of some datasets might limit their applicability to real-world scenarios, but they are useful for controlled benchmarking. \\
\hline
\textbf{Are there tasks for which the dataset should not be used?} & The dataset is not suitable for tasks requiring personal data or highly sensitive financial predictions without further analysis. \\
\hline
\multicolumn{2}{|c|}{\textbf{Distribution}} \\
\hline
\textbf{Will the dataset be distributed to third parties outside of the entity on behalf of which the dataset was created?} & Yes, the dataset will be publicly available for research purposes. \\
\hline
\textbf{How will the dataset be distributed?} & The dataset will be distributed via an online repository with appropriate licensing. \\
\hline
\textbf{When will the dataset be distributed?} & The dataset will be available for distribution after the publication of the paper. \\
\hline
\textbf{Will the dataset be distributed under a copyright or other intellectual property license, and/or under applicable terms of use?} & Yes. \\
\hline
\textbf{Have any third parties imposed IP-based or other restrictions on the data associated with the instances?} & No. \\
\hline
\textbf{Do any export controls or other regulatory restrictions apply to the dataset or to individual instances?} & No. \\
\hline
\multicolumn{2}{|c|}{\textbf{Maintenance}} \\
\hline
\textbf{Who is supporting/hosting/maintaining the dataset?} & The dataset is maintained by the research team and contributors. \\
\hline
\textbf{How can the owner/curator/manager of the dataset be contacted?} & Contact details will be provided in the dataset repository. \\
\hline
\textbf{Is there an erratum?} & Not yet, but any updates or errors will be documented in the repository. \\
\hline
\textbf{Will the dataset be updated?} & Yes, future updates will be made to improve and expand the dataset. \\
\hline
\textbf{If the dataset relates to people, are there applicable limits on the retention of the data associated with the instances?} & Not applicable. \\
\hline
\textbf{Will older versions of the dataset continue to be supported/hosted/maintained?} & Yes, previous versions will remain available for reference. \\
\hline
\textbf{If others want to extend/augment/build on/contribute to the dataset, is there a mechanism for them to do so?} & Yes, contributions are welcomed via the dataset repository, and code for expanding the dataset will be provided upon request. \\
\hline
\multicolumn{2}{|c|}{\textbf{Ethical Considerations}} \\
\hline
\textbf{Were any ethical review processes conducted (e.g., by an institutional review board)?} & No formal ethical review was conducted as the dataset does not contain sensitive personal information. \\
\hline
\textbf{Does the dataset contain data that, if viewed directly, might be offensive, insulting, threatening, or might otherwise cause anxiety?} & No. The dataset contains time series data without any sensitive or potentially offensive content. \\
\hline
\textbf{Does the dataset relate to people?} & No. \\
\hline
\textbf{Does the dataset identify any subpopulations (e.g., by age, gender)?} & No. \\
\hline
\textbf{Is it possible to identify individuals (i.e., one or more people) from the dataset?} & No. \\
\hline
\textbf{Does the dataset contain data that might be considered sensitive in any way (e.g., data that reveals racial or ethnic origins, sexual orientations, religious beliefs, political opinions or affiliations, health data)?} & No. \\
\hline
\textbf{Are there any known risks to individuals that are represented in the dataset?} & No. \\
\hline
\textbf{Does the dataset contain data that might be subject to GDPR or other data protection laws?} & No. \\
\hline
\end{longtable}

\end{document}